%% file: ms.tex
\documentclass[12pt,a4paper,english,greek,twoside]{biomed-thesis}
\usepackage[utf8]{inputenc}
\usepackage[caption=false,font=footnotesize]{subfig}
\usepackage[greek]{babel}
\usepackage[sectionbib]{chapterbib}
\usepackage{adjustbox}
\usepackage{algorithmic}
\usepackage{algorithm}
\usepackage{alphabeta}
\usepackage{amssymb}
\usepackage{bm}
\usepackage{booktabs}
\usepackage{caption}
\usepackage{datatool}
\usepackage{diagbox}
\usepackage{dsfont}
\usepackage{indentfirst}
\usepackage{makecell}
\usepackage{multirow}
\usepackage{rotating}
\usepackage{siunitx}
\usepackage{tikz}
\usepackage{tikz-3dplot}
\usepackage{titlesec}
\usepackage{url}

\usetikzlibrary{positioning}
\usetikzlibrary{shapes}

\addto\captionsenglish{\renewcommand{\contentsname}{Περιεχόμενα}}
\addto\captionsenglish{\renewcommand{\listfigurename}{Λίστα εικόνων}}
\addto\captionsenglish{\renewcommand{\listtablename}{Λίστα πινάκων}}
\addto\captionsenglish{}
\abstractgreek{Περίληψη}
\acknowledgementsgreek{Ευχαριστίες}

\BeforeBeginEnvironment{tabular}{\begin{adjustbox}{max width=\textwidth}}
\AfterEndEnvironment{tabular}{\end{adjustbox}}

\titleformat{\chapter}[hang]{\bfseries\huge}{\thechapter.}{2pc}{}
\titlelabel{\thetitle.\quad}

\graphicspath{{./images/}}

\title{Δίκτυα Αραιής Ενεργοποίησης:\\ Μια νέα μέθοδος αποσύνθεσης και συμπίεσης δεδομένων}
\author{Μπιζόπουλος Πασχάλης}
\authorbsc{Διπλωματούχος Ηλεκτρολόγος Μηχανικός και Μηχανικός Υπολογιστών Α.Π.Θ.}
\authoralt{ΜΠΙΖΟΠΟΥΛΟΥ ΠΑΣΧΑΛΗ}
\documenttype{Διδακτορική Διατριβή}
\copyrightdeclaration{Με επιφύλαξη παντός δικαιώματος.\\
Απαγορεύεται η αντιγραφή, αποθήκευση και διανομή της παρούσας εργασίας, εξ ολοκλήρου ή τμήματος αυτής, για εμπορικό σκοπό.
Επιτρέπεται η ανατύπωση, αποθήκευση και διανομή για σκοπό μη κερδοσκοπικό, εκπαιδευτικής ή ερευνητικής φύσης, υπό την προϋπόθεση να αναφέρεται η πηγή προέλευσης και να διατηρείται το παρόν μήνυμα.
Ερωτήματα που αφορούν τη χρήση της εργασίας για κερδοσκοπικό σκοπό πρέπει να απευθύνονται προς τον συγγραφέα.

Οι απόψεις και τα συμπεράσματα που περιέχονται σε αυτό το έγγραφο εκφράζουν τον συγγραφέα και δεν πρέπει να ερμηνευτεί ότι αντιπροσωπεύουν τις επίσημες θέσεις του Εθνικού Μετσόβιου Πολυτεχνείου.}
\placetime{Αθήνα, Οκτώβριος 2019}
\supervisor{Επιβλέπων : \=Δημήτριος Κουτσούρης\\ \>Καθηγητής Ε.Μ.Π.}
\institution{Εθνικό Μετσόβιο Πολυτεχνείο}
\department{Σχολή Ηλεκτρολόγων Μηχανικών και Μηχανικών Υπολογιστών}
\division{Τομέας Συστημάτων Μετάδοσης Πληροφορίας\\ και Τεχνολογίας Υλικών}
\lab{Εργαστήριο Βιοϊατρικής Τεχνολογίας}

\approval{
	Εγκρίθηκε από την επταμελή εξεταστική επιτροπή την 30η Οκτωβρίου 2019.
	\vspace{1cm}
	\begin{tabbing}
		aaaaaaaaaaaaaaaaaaaaaaaaaa\=aaaaaaaaaaaaaaaaaaaaaaaaaa\=\kill
		\ldots\ldots\ldots\ldots\ldots\ldots\>  \ldots\ldots\ldots\ldots\ldots\ldots\>  \ldots\ldots\ldots\ldots\ldots\ldots\\
		Δημήτριος Κουτσούρης\>             Δημήτριος Φωτιάδης\>        Κωνσταντίνα Νικήτα\\
		Καθηγητής Ε.Μ.Π.\>     Καθηγητής Παν. Ιωαννίνων\>      Καθηγήτρια Ε.Μ.Π.\\\\\\\\
		\ldots\ldots\ldots\ldots\ldots\ldots\>  \ldots\ldots\ldots\ldots\ldots\ldots\>  \ldots\ldots\ldots\ldots\ldots\ldots\\
		Εμμανουήλ Βαβουρανάκης\>         Γεώργιος Ματσόπουλος\>    Θεόδωρος Παπαϊωάννου\\
		Καθηγητής Ε.Κ.Π.Α.\>     Καθηγητής Ε.Μ.Π.\>      Αν.Καθηγητής Ε.Κ.Π.Α.\\\\\\\\
		\ldots\ldots\ldots\ldots\ldots\ldots\\
		Ανδριάνα Πρέντζα\\
		Καθηγήτρια Παν. Πειραιώς\\\\\\\\
	\end{tabbing}
}

\begin{document}
\maketitle
\selectlanguage{english}
\frontmatter
\mainmatter{}
\include{acknowledgements}
\include{abstract}
\include{abstracteng}
\tableofcontents
\listoffigures
\cleardoublepage{}
\listoftables
\cleardoublepage{}
\include{chapter1}
\include{chapter2}
\include{chapter3}
\include{chapter4}
\include{chapter5}
\include{chapter6}
\include{chapter7}
\include{acronyms}

\backmatter{}

\include{publications}
\end{document}

%% file: acknowledgements.tex
\begin{acknowledgements}
	Θα ήθελα να ευχαριστήσω τον επιβλέπων καθηγητή κ. Κουτσούρη για την δυνατότητα που με έδωσε να ξεκινήσω διδακτορικές σπουδές, την ελευθερία στην ερευνητική διαδικασία αλλά και για τις πολύτιμες επιστημονικές συμβουλές του κατά τη διάρκεια του διδακτορικού.
	Θέλω επίσης να ευχαριστήσω τον κ. Φωτιάδη για την πολύτιμη βοήθειά του και τις κατατοπιστικές ερευνητικές συμβουλές, ιδίως στα πρώτα χρόνια του διδακτορικού.
	Θα ήθελα να ευχαριστήσω τα μέλη της επταμελούς επιτροπής και όλα τα μέλη του Εργαστηρίου Βιοϊατρικής Τεχνολογίας για την συμπαράσταση και βοήθεια που προσέφεραν.
	Τέλος θα ήθελα να ευχαριστήσω την οικογένεια μου στην οποία οφείλω τα πάντα.
\end{acknowledgements}

%% file: abstract.tex
\begin{abstract}
	Η πρόσφατη βιβλιογραφία σχετικά με τη μη-επιβλεπώμενη μάθηση επικεντρώθηκε στο σχεδιασμό δομών με στόχο την μάθηση χαρακτηριστικών.
	Αυτό όμως γινόταν χωρίς να ληφθεί υπόψη το μήκος περιγραφής των αναπαραστάσεων, το οποίο είναι ένα άμεσο και αμερόληπτο μέτρο της πολυπλοκότητας του μοντέλου.
	Στο πλαίσιο της διδακτορικής διατριβής προτείνουμε ένα μέτρο $\varphi$ το οποίο αξιολογεί μη-επιβλεπώμενα μοντέλα με βάση την ακρίβεια ανακατασκευής και το βαθμό συμπίεσης των εσωτερικών αναπαραστάσεων.
	Έπειτα παρουσιάζουμε και ορίζουμε δύο συναρτήσεις ενεργοποίησης (Ταυτότητα, ReLU) ως βάσεις αναφοράς και τρεις αραιές συναρτήσεις ενεργοποίησης (Απόλυτα κ-μέγιστα, Δείκτες συγκέντρωσης ακρότατων, Ακρότατα) ως υποψήφιες δομές για την ελαχιστοποίηση του προηγουμένως ορισμένου μέτρου $\varphi$.
	Τέλος προτείνουμε μια νέα αρχιτεκτονική Νευρωνικών Δικτύων, τα \textbf{Δίκτυα Αραιής Ενεργοποίησης} (SANs), τα οποία αποτελούνται από πυρήνες με κοινά βάρη που κατά την κωδικοποίηση συνελλίσονται με την είσοδο και στη συνέχεια διέρχονται μέσω μιας συνάρτησης αραιής ενεργοποίησης.
	Κατά τη διάρκεια της αποκωδικοποίησης, τα ίδια βάρη συνελλίσονται με τον χάρτη αραιής ενεργοποίησης και έπειτα οι μερικές ανακατασκευές από κάθε βάρος αθροίζονται για να ανακατασκευάσουν την είσοδο.
	Συγκρίνουμε τα SANs χρησιμοποιώντας τις προηγουμένως ορισμένες συναρτήσεις ενεργοποίησης σε ένα σύνολο από βάσεις δεδομένων (15 βάσεις δεδομένων από την Physionet και EEG από την UCI για ταξινόμηση επιληπτικών κρίσεων) και δείχνουμε ότι τα SANs που επιλέγονται με χρήση του $\varphi$ έχουν αναπαραστάσεις με μικρό μήκος περιγραφής και περιέχουν ερμηνεύσιμους πυρήνες.

	\section*{Λέξεις Κλειδιά}
	νευρωνικά δίκτυα, αυτοκωδικοποιητές, αραιότητα, συμπίεση
\end{abstract}

%% file: abstracteng.tex
\begin{abstracteng}
	Recent literature on unsupervised learning focused on designing structural priors with the aim of learning meaningful features, but without considering the description length of the representations.
	In this thesis, first we introduce the $\varphi$ metric that evaluates unsupervised models based on their reconstruction accuracy and the degree of compression of their internal representations.
	We then present and define two activation functions (Identity, ReLU) as base of reference and three sparse activation functions (top-k absolutes, Extrema-Pool indices, Extrema) as candidate structures that minimize the previously defined metric $\varphi$.
	We lastly present Sparsely Activated Networks (SANs) that consist of kernels with shared weights that, during encoding, are convolved with the input and then passed through a sparse activation function.
	During decoding, the same weights are convolved with the sparse activation map and subsequently the partial reconstructions from each weight are summed to reconstruct the input.
	We compare SANs using the five previously defined activation functions on a variety of datasets (Physionet, UCI-epilepsy, MNIST, FMNIST) and show that models that are selected using $\varphi$ have small description representation length and consist of interpretable kernels.

	\section*{Keywords}
	\textlatin{neural networks, autoencoders, sparsity, compression}
\end{abstracteng}

%% file: chapter1.tex
\chapter{Εισαγωγή}
\label{chapter1}
Οι γιατροί κάνουν διαγνώσεις με βάση το ιατρικό ιστορικό, βιοδείκτες και τις εξετάσεις μεμονωμένων ασθενών, τα οποία ερμηνεύουν σύμφωνα με την προσωπική τους κλινική εμπειρία.
Στη συνέχεια, αντιστοιχούν τον κάθε ασθενή στην παραδοσιακή ταξινόμηση των ιατρικών νοσημάτων/παθήσεων σύμφωνα με μια υποκειμενική ερμηνεία της ιατρικής βιβλιογραφίας.
Αυτή η διαδικασία είναι όλο και περισσότερο επιρρεπής σε σφάλματα.
Επιπλέον, οι τεχνολογίες απεικονίσεων αυξάνουν συνεχώς την ικανότητά τους να παράγουν μεγάλες ποσότητες δεδομένων, τα οποία οι ειδικοί αδυνατούν να καταλάβουν και να χρησιμοποιήσουν αποτελεσματικά, καθιστώντας το έργο τους δυσκολότερο.
Επομένως, απαιτείται η αυτοματοποίηση των ιατρικών διαδικασιών για την αύξηση της ποιότητας της υγείας των ασθενών και τη μείωση του κόστους των συστημάτων υγειονομικής περίθαλψης.

Η ανάγκη για αυτοματοποίηση των ιατρικών διαδικασιών κυμαίνεται από τη διάγνωση έως τη θεραπεία και ιδίως σε περιπτώσεις/περιοχές όπου υπάρχει έλλειψη υγειονομικής περίθαλψης.
Προηγούμενες προσπάθειες αυτοματοποίησης περιλαμβάνουν συστήματα εμπειρογνωμόνων βασισμένα σε κανόνες, τα οποία έχουν σχεδιαστεί να μιμούνται τη διαδικασία που ακολουθούν οι ιατρικοί εμπειρογνώμονες όταν επιλύουν ιατρικά προβλήματα.
Αυτά τα συστήματα έχουν αποδειχθεί αναποτελεσματικά επειδή απαιτούν χειροκίνητη δημιουργία χαρακτηριστικών και γνώση ειδικών πάνω στο πρόβλημα για να επιτευχθεί επαρκής ακρίβεια και επίσης είναι δύσκολο να εμφανίσουν γραμμική βελτίωση με την παρουσία νέων δεδομένων.

Η βαθιά μάθηση έχει αναδειχθεί ως μια πιο ακριβής και αποτελεσματική τεχνολογία σε ένα ευρύ φάσμα ιατρικών προβλημάτων όπως η διάγνωση, η πρόβλεψη και η παρέμβαση.
Είναι μια μέθοδος μάθησης αναπαραστάσεων που αποτελείται από επίπεδα που μετασχηματίζουν τα δεδομένα με μη-γραμμικό τρόπο, αποκαλύπτοντας έτσι ιεραρχικές σχέσεις και δομές.
Παρόλο που η βαθιά μάθηση έχει εφαρμοστεί με επιτυχία σε πολλούς τομείς, αυτό έρχεται με κόστος της ερμηνευσιμότητας των αναπαραστάσεων, καθώς αυτές αποτελούνται από εκατομμύρια παραμέτρους.

Σκοπός της διδακτορικής διατριβής είναι η δημιουργία μιας νέας αρχιτεκτονικής νευρωνικών δικτύων, τα \textbf{Δίκτυα Αραιής Ενεργοποίησης} (Sparsely Activated Networks, SANs), τα οποία αποσυνθέτουν τις εισόδους τους σε ένα σύνολο από αραιά επαναλαμβανόμενα πρότυπα διαφορετικού πλάτους και σε συνδυασμό με ένα προτεινόμενο μέτρο $\varphi$ μαθαίνουν αναπαραστάσεις με ελάχιστα μήκη περιγραφής, αυξάνοντας έτσι την ερμηνευσιμότητα του μοντέλου.

\section{Οργάνωση τόμου}
Στο Κεφάλαιο~\ref{chapter2}, παρουσιάζουμε τις θεμελιώδεις έννοιες των νευρωνικών δικτύων και της βαθιάς μάθησης καθώς επίσης και τις γενικές ιδιότητες των πιο συχνά χρησιμοποιούμενων αρχιτεκτονικών.

Στα Κεφάλαια~\ref{chapter3} και~\ref{chapter4} παρουσιάζουμε μια βιβλιογραφική επισκόπηση των αρχιτεκτονικών βαθιάς μάθησης σε δομημένα δεδομένα, σήματα και απεικονίσεις που έχουν χρησιμοποιηθεί στην ιατρική και ιδιαίτερα στην καρδιολογία.
Συζητάμε τα πλεονεκτήματα και τους περιορισμούς των εφαρμογών της βαθιάς μάθησης στην ιατρική, ενώ προτείνουμε ορισμένες κατευθύνσεις ως τις πιο βιώσιμες για κλινική χρήση.
Πιο συγκεκριμένα στις ενότητες~\ref{sec3:structured} και~\ref{sec3:signals} παρουσιάζουμε τις εφαρμογές βαθιάς μάθησης με χρήση δομημένων δεδομένων και μορφές σημάτων, ενώ στην ενότητα~\ref{sec4:discussion} παρουσιάζουμε πλεονεκτήματα και περιορισμούς των εφαρμογών βαθιάς μάθησης στην καρδιολογία και προτείνουμε κάποιες κατευθύνσεις για την υλοποίηση μοντέλων βαθιάς μάθησης που μπορούν να εφαρμοστούν κλινικά.

Στο Κεφάλαιο~\ref{chapter5} προτείνουμε τα Signal2Image (S2Is) ως εκπαιδεύσιμα επίπεδα προθέματος νευρωνικών δικτύων, τα οποία μετατρέπουν σήματα, όπως το ηλεκτροεγκεφαλογράφημα (Electroencephalogram, EEG), σε αναπαραστάσεις εικόνων, καθιστώντας τα κατάλληλα για την εκπαίδευση βαθιών νευρωνικών δικτύων βασισμένα σε εικόνες, τα οποία ορίζονται ως `μοντέλα βάσης'.
Συγκρίνουμε την ακρίβεια και τις επιδόσεις τεσσάρων S2Is (`σήμα ως εικόνα', φασματογράφημα, CNN ενός και δύο επιπέδων) σε συνδυασμό με ένα σύνολο `μοντέλων βάσης' μαζί με μοντέλα διαφορετικού βάθους και 1D παραλλαγές των τελευταίων.
Παρέχουμε επίσης εμπειρικές αποδείξεις ότι το CNN S2I ενός επιπέδου αποδίδει καλύτερα σε 11 από τα 15 μοντέλα που δοκιμάστηκαν σε σύγκριση με τα μη-εκπαιδεύσιμα S2Is για την ταξινόμηση σημάτων EEG και οπτικοποιούμε τις εξόδους κάποιων από τα S2Is.

Στο Κεφάλαιο~\ref{chapter6}, ενότητα~\ref{sec6:flithos} ορίζουμε το μέτρο $\varphi$, έπειτα στην ενότητα~\ref{sec6:sans} ορίζουμε τις πέντε συναρτήσεις ενεργοποίησης που θα συγκριθούν καθώς επίσης και την αρχιτεκτονική και διαδικασία εκπαίδευσης των SAN, στην ενότητα~\ref{sec6:experiments} δοκιμάζουμε τα SANs στις βάσεις δεδομένων Physionet, UCI-epilepsy και MNIST και οπτικοποιούμε τις ενδιάμεσες αναπαραστάσεις και αποτελέσματα.
Στην ενότητα~\ref{sec6:discussion} συζητάμε τα ευρήματα των πειραμάτων και τους περιορισμούς των SAN\@.

Τέλος στο Κεφάλαιο~\ref{chapter7} παρουσιάζουμε τα τελικά συμπεράσματα και προτείνουμε πιθανές μελλοντικές κατευθύνσεις.

%% file: chapter2.tex
\chapter{Νευρωνικά δίκτυα}
\label{chapter2}
\graphicspath{{./images/deep-learning-in-cardiology/}}

\newcommand{\networkLayer}[6]{
	\def\a{#1}
	\def\b{0.02}
	\def\c{#2}
	\def\t{#3}
	\def\d{#4}
	\draw[line width=0.3mm] (\c+\t,0,\d) -- (\c+\t,\a,\d) -- (\t,\a,\d);                                                      
	\draw[line width=0.3mm] (\t,0,\a+\d) -- (\c+\t,0,\a+\d) node[midway,below] {#6} -- (\c+\t,\a,\a+\d) -- (\t,\a,\a+\d) -- (\t,0,\a+\d); 
	\draw[line width=0.3mm] (\c+\t,0,\d) -- (\c+\t,0,\a+\d);
	\draw[line width=0.3mm] (\c+\t,\a,\d) -- (\c+\t,\a,\a+\d);
	\draw[line width=0.3mm] (\t,\a,\d) -- (\t,\a,\a+\d);
	\filldraw[#5] (\t+\b,\b,\a+\d) -- (\c+\t-\b,\b,\a+\d) -- (\c+\t-\b,\a-\b,\a+\d) -- (\t+\b,\a-\b,\a+\d) -- (\t+\b,\b,\a+\d); 
	\filldraw[#5] (\t+\b,\a,\a-\b+\d) -- (\c+\t-\b,\a,\a-\b+\d) -- (\c+\t-\b,\a,\b+\d) -- (\t+\b,\a,\b+\d);
	\ifthenelse {\equal{#5} {}}
	{}
	{\filldraw[#5] (\c+\t,\b,\a-\b+\d) -- (\c+\t,\b,\b+\d) -- (\c+\t,\a-\b,\b+\d) -- (\c+\t,\a-\b,\a-\b+\d);} 
}

\section{Εισαγωγή}
Η μηχανική μάθηση είναι ένα σύνολο μεθόδων Τεχνητής Νοημοσύνης (Artificial Intelligence, AI) που επιτρέπει στους υπολογιστές να μαθαίνουν μια διαδικασία χρησιμοποιώντας δεδομένα, αντί να προγραμματιστούν ρητά.
Έχει αναδειχθεί ως ένας αποτελεσματικός τρόπος χρήσης και συνδυασμού βιολογικών δεικτών, απεικόνισης, συσσωρευμένης κλινικής έρευνας από τη βιβλιογραφία και τους Ηλεκτρονικούς Φακέλους Υγείας (Electronic Health Record, EHR) για την αύξηση της ακρίβειας λύσεων ενός ευρύ φάσματος ιατρικών προβλημάτων.
Οι ιατρικές διαδικασίες που χρησιμοποιούν την μηχανική μάθηση εξελίσσονται από τέχνη σε επιστήμη με γνώμονα τα δεδομένα, προσφέροντας διόραση από τα πληθυσμιακά δεδομένα στην προσωποποιημένη ιατρική.

Η βαθιά μάθηση, και η εφαρμογή της στα νευρωνικά δίκτυα τα Βαθιά Νευρωνικά Δίκτυα (Deep Neural Networks, DNN), είναι ένα σύνολο μεθόδων μηχανικής μάθησης που αποτελούνται από πολλαπλά στοιβαγμένα επίπεδα και χρησιμοποιούν δεδομένα για να μάθουν ιεραρχικά επίπεδα.
Η βαθιά μάθηση προέκυψε λόγω της αύξησης της υπολογιστικής ισχύος των μονάδων επεξεργασίας γραφικών και της διαθεσιμότητας δεδομένων μεγάλου όγκου και έχει αποδειχθεί ότι είναι μια ισχυρή λύση για προβλήματα όπως ταξινόμηση εικόνων~\cite{krizhevsky2012imagenet}, κατάτμηση εικόνας~\cite{ronneberger2015u}, επεξεργασία φυσικής γλώσσας~\cite{collobert2008unified}, αναγνώριση ομιλίας~\cite{graves2013speech} και γονιδιωματική~\cite{alipanahi2015predicting}.

Πλεονεκτήματα των DNN έναντι των παραδοσιακών τεχνικών μηχανικής μάθησης περιλαμβάνουν ότι απαιτούν λιγότερη εξειδικευμένη γνώση για το πρόβλημα που προσπαθούν να λύσουν και επίσης είναι ευκολότερο να αυξήσουν την ακρίβεια τους είτε με την αύξηση του συνόλου δεδομένων εκπαίδευσης ή την αύξηση της χωρητικότητας του δικτύου.
Τα ρηχά μοντέλα μηχανικής μάθησης, όπως τα δέντρα αποφάσεων και οι Μηχανές Διανυσμάτων Υποστήριξης (Support Vector Machines, SVM), είναι `ανεπαρκή'; το οποίο σημαίνει ότι απαιτούν μεγάλο αριθμό υπολογισμών κατά τη διάρκεια της εκπαίδευσης/συμπερασμού, μεγάλο αριθμό παρατηρήσεων για την επίτευξη γενίκευσης και σημαντική ανθρώπινη εργασία για τον προσδιορισμό της πρότερης γνώσης του μοντέλου~\cite{bengio2007scaling}.

\section{Επισκόπηση θεωρίας}
Τα νευρωνικά δίκτυα είναι ένα σύνολο τεχνικών μηχανικής μάθησης εμπνευσμένων από τον εγκέφαλο αλλά χωρίς πρωταρχικό στόχο να τον προσομοιώνουν.
Πρόκειται για μεθόδους προσέγγισης συναρτήσεων όπου η είσοδος $\mathbf{x}$ μπορεί να είναι κείμενο, εικόνα, ήχος, σήμα, τρισδιάστατος όγκος, βίντεο (ή συνδυασμός αυτών) και η έξοδος $\mathbf{y}$ είναι από το ίδιο σετ με το $\mathbf{x}$ αλλά αυξημένου ενημερωτικού περιεχομένου.
Με μαθηματικούς όρους, ο στόχος ενός νευρωνικού δικτύου είναι να βρει το σύνολο παραμέτρων $\theta$ (το οποίο αποτελείται από βάρη $\mathbf{w}$ και πολώσεις $\mathbf{b}$):
\begin{equation}
	\centering
	f(\mathbf{x};\mathbf{\theta}) = \mathbf{\hat{y}}
\end{equation}

\noindent
όπου $f$ είναι μια προκαθορισμένη συνάρτηση και $\mathbf{\hat{y}}$ είναι η πρόβλεψη.
Ο περιορισμός για το $\theta$ είναι να έχουμε ένα όσο το δυνατόν χαμηλότερη τιμή για μια συνάρτηση κόστους $J(\theta)$ μεταξύ του $\mathbf{y}$ και του $\mathbf{\hat{y}}$.

\begin{figure}[!t]
	\centering
	\input{chapter2-nn.tex}
	\caption[Αντιλήπτωρ]{Αντιλήπτωρ.
	Περιέχει ένα σύνολο συνδέσεων $x_{1, \ldots, n}$ ως εισόδους, τα βάρη $w_{1, \ldots, n}$, τη πόλωση $b$, τη συνάρτηση ενεργοποίησης $\phi$ και την έξοδο $\alpha$.
	}
	\label{fig:perceptron}
\end{figure}
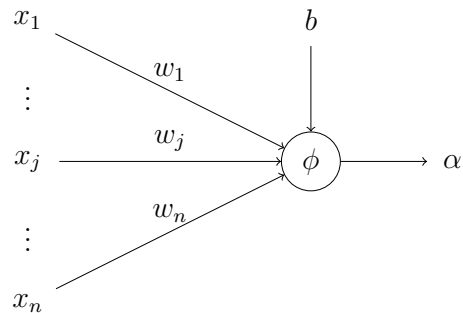

Η βασική μονάδα των νευρωνικών δικτύων είναι ο αντιλήπτωρ που απεικονίζεται στην Εικ.\ref{fig:perceptron}, ο οποίος δημοσιεύθηκε για πρώτη φορά από τον Rosenblatt~\cite{rosenblatt1958perceptron} το 1958.
Αποτελείται από ένα σύνολο εισόδων που συμβολίζεται με το διάνυσμα $\mathbf{x}=[x_1, \ldots, x_j, \ldots, x_n]$, ένα σύνολο βαρών για την κάθε είσοδο, που συμβολίζεται με το διάνυσμα $\mathbf{w}=[w_1, \ldots, w_j, \ldots, w_n]$ και η πόλωση $b$.
Η απόφαση του κόμβου να πυροδοτήσει ένα σήμα $\alpha$ στον επόμενο νευρώνα ή στην έξοδο καθορίζεται από τη συνάρτηση ενεργοποίησης $\phi$, το σταθμισμένο άθροισμα $\mathbf{w}$ και τη πόλωση $b$, με τον ακόλουθο τρόπο:
\begin{equation}
	\centering
	\alpha = \phi(\sum\limits_{j} w_{j}x_{j} + b)
\end{equation}

Οι τιμές του διανύσματος βαρών $\mathbf{w}$ και των πολώσεων $\mathbf{b}$ στα DNN ρυθμίζονται επαναληπτικά χρησιμοποιώντας αλγόριθμους βελτιστοποίησης που βασίζονται στην κλίση καθόδου και την αντίστροφη διάδοση σφάλματος (backpropagation)~\cite{rumelhart1986learning} ο οποίος υπολογίζει την κλίση της συνάρτησης κόστους σε σχέση με τις παραμέτρους $\nabla_{\theta}J(\theta)$~\cite{goodfellow2016deep}.
Η αξιολόγηση της γενίκευσης ενός νευρωνικού δικτύου απαιτεί τη διάσπαση του αρχικού συνόλου δεδομένων $D=(\mathbf{x}, \mathbf{y})$ σε τρία μη-αλληλεπικαλυπτόμενα σύνολα δεδομένων:
\begin{equation}
	\centering
	D_{train}=(\mathbf{x}_{train}, \mathbf{y}_{train})
\end{equation}

\begin{equation}
	\centering
	D_{validation}=(\mathbf{x}_{validation}, \mathbf{y}_{validation})
\end{equation}

\begin{equation}
	\centering
	D_{test}=(\mathbf{x}_{test}, \mathbf{y}_{test})
\end{equation}

Το $D_{train}$ χρησιμοποιείται για την μάθηση των $\mathbf{w}$ και $\mathbf{b}$, των οποίων οι τιμές κάνουν το δίκτυο να ελαχιστοποιεί την επιλεγμένη συνάρτηση κόστους, ενώ το $D_{validation}$ χρησιμοποιείται για την επιλογή των υπερπαραμέτρων του δικτύου.
Το $D_{test}$ χρησιμοποιείται για την αξιολόγηση της γενίκευσης του δικτύου και θα πρέπει ιδανικά να προέρχεται από διαφορετικές μηχανές/ασθενείς/οργανισμούς ανάλογα με το ερευνητικό ερώτημα.

Η απόδοση των DNN έχει βελτιωθεί σημαντικά με τη χρήση της ανορθωμένης γραμμικής μονάδας (Rectified Linear Unit, ReLU) ως συνάρτηση ενεργοποίησης σε σύγκριση με την σιγμοειδής και την υπερβολική εφαπτομένη~\cite{glorot2011deep}.
Η συνάρτηση ενεργοποίησης του τελευταίου επιπέδου επιλέγεται με βάση τη φύση του ερευνητικού ερωτήματος που πρέπει να απαντηθεί (π.χ.\ softmax για ταξινόμηση, σιγμοειδής για παλινδρόμηση).

Οι συναρτήσεις κόστους που χρησιμοποιούνται στα νευρωνικά δίκτυα εξαρτώνται επίσης από το πρόβλημα που πρέπει να επιλυθεί.
Η διασταυρωμένη εντροπία (cross-entropy) ποσοτικοποιεί τη διαφορά μεταξύ της πραγματικής και της προβλεπόμενης κατανομής πιθανοτήτων και επιλέγεται συνήθως για προβλήματα ανίχνευσης και ταξινόμησης.
Το εμβαδόν της περιοχής κάτω από την καμπύλη λειτουργίας δέκτη (Area Under Curve, AUC) αντιπροσωπεύει την πιθανότητα ένα τυχαίο ζεύγος κανονικών και μη κανονικών εικονοστοιχείων/σημάτων/εικόνων να ταξινομηθεί σωστά~\cite{hanley1982meaning} και χρησιμοποιείται σε προβλήματα δυαδικής τμηματοποίησης.
Ο συντελεστής Dice~\cite{dice1945measures} είναι ένα μέτρο ομοιότητας που χρησιμοποιείται στα προβλήματα τμηματοποίησης και οι τιμές του κυμαίνονται μεταξύ μηδέν (συνολική αναντιστοιχία) και μονάδας (τέλεια αντιστοίχιση).

\section{Επισκόπηση αρχιτεκτονικών}
Τα Πλήρως Συνδεδεμένα Δίκτυα (Fully Connected Networks, FNN) είναι δίκτυα που αποτελούνται από πολλαπλούς αντιλήπτωρες στοιβαγμένους σε πλάτος και βάθος, που σημαίνει ότι κάθε αντιλήπτωρ σε κάθε επίπεδο συνδέεται με κάθε αντιλήπτωρ στα επίπεδα αμέσως πριν και μετά.
Αν και έχει αποδειχθεί~\cite{hornik1989multilayer} ότι τα FNN ενός επιπέδου με επαρκή αριθμό κρυφών μονάδων είναι καθολικοί προσεγγιστές συναρτήσεων, δεν είναι υπολογιστικά αποδοτικά για την προσέγγιση σύνθετων συναρτήσεων.
Τα Δίκτυα Βαθιάς Πίστης (Deep Belief Networks, DBN)~\cite{hinton2006fast} είναι στοιβαγμένες Περιορισμένες Μηχανές Boltzmann (Restricted Boltzmann Machines, RBMs) όπου κάθε επίπεδο κωδικοποιεί τις στατιστικές εξαρτήσεις μεταξύ των μονάδων στο προηγούμενο επίπεδο; εκπαιδεύονται για να μεγιστοποιήσουν την πιθανοφάνεια των δεδομένων εκπαίδευσης.

\begin{figure}[!t]
	\centering
	\input{chapter2-cnn.tex}
	\caption[Συνελικτικό νευρωνικό δίκτυο]{Ένα συνελικτικό νευρωνικό δίκτυο που υπολογίζει την εμβαδόν της αριστερής καρδιακής κοιλίας ($\hat{y}$) από μια εικόνα MRI ($x$).
	Η πυραμιδοειδής δομή στο πάνω μέρος της εικόνας υποδηλώνει τη ροή των υπολογισμών κατά τη διάρκεια της προς-τα-εμπρός διάδοσης, ξεκινώντας από την εικόνα εισόδου μέσω του συνόλου των χαρτών ενεργοποιήσεων που απεικονίζονται ως τρισδιάστατα ορθογώνια στην έξοδο $\hat{y}$.
	Το ύψος και το πλάτος των ορθογωνίων είναι ανάλογο του ύψους και του πλάτους των χαρτών ενεργοποιήσεων, ενώ το βάθος είναι ανάλογο του αριθμού των χαρτών ενεργοποιήσεων.
	Τα βέλη στο κάτω μέρος υποδηλώνουν τη ροή της αντίστροφης διάδοσης του σφάλματος (backpropagation) ξεκινώντας από τον υπολογισμό του σφάλματος χρησιμοποιώντας τη συνάρτηση κόστους $J$, την πραγματική έξοδο $y$ και την πρόβλεψη $\hat{y}$.
	Αυτή η απώλεια διαδίδεται προς τα πίσω μέσω των φίλτρων του δικτύου, τα οποία προσαρμόζουν τις τιμές τους.
	Οι διακεκομμένες γραμμές με παύλες υποδηλώνουν ένα 2D συνελικτικό επίπεδο με ReLU και Extrema-Pooling (το οποίο μειώνει το ύψος και το πλάτος των χαρτών ενεργοποιήσεων), η διακεκομμένη γραμμή με τελείες δηλώνει το πλήρως συνδεδεμένο επίπεδο και τέλος οι διακεκομμένες γραμμές με παύλες-τελείες υποδηλώνουν σιγμοειδές επίπεδο.
	Για ευκολία απεικόνισης εμφανίζονται μόνο μερικοί από τους χάρτες ενεργοποιήσεων και τα φίλτρα και δεν εμφανίζονται σε κλίμακα.}
	\label{fig:cnn}
\end{figure}
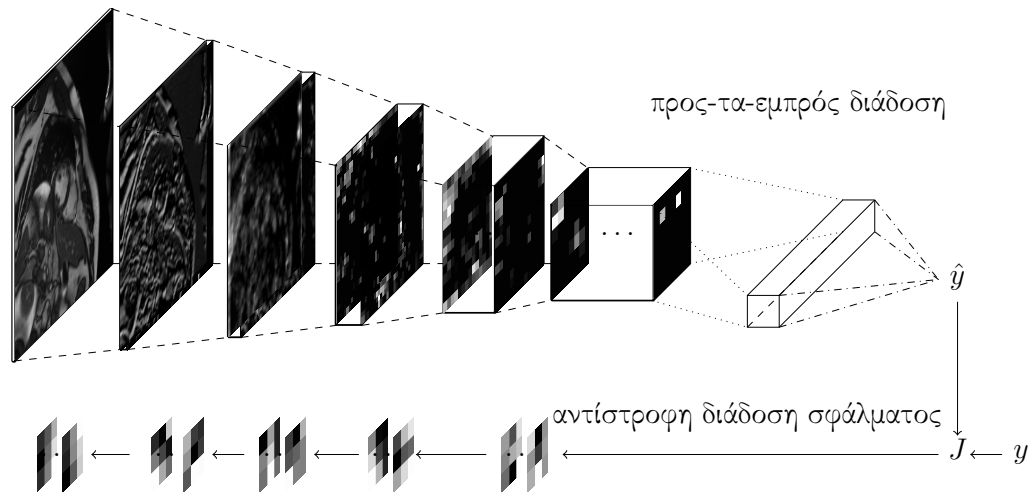

Τα Συνελικτικά Νευρωνικά Δίκτυα (Convolutional Neural Networks, CNN), όπως φαίνεται στο Σχήμα~\ref{fig:cnn}, αποτελούνται από ένα συνελικτικό μέρος στο οποίο γίνεται ιεραρχική εξαγωγή χαρακτηριστικών (χαρακτηριστικά χαμηλού επιπέδου όπως ακμές και γωνίες και χαρακτηριστικά υψηλού επιπέδου, όπως τμήματα αντικειμένων) και ένα πλήρως συνδεδεμένο μέρος για ταξινόμηση ή παλινδρόμηση, ανάλογα με τη φύση της εξόδου $y$.
Τα συνελικτικά επίπεδα είναι πολύ καλοί βελτιστοποιητές χαρακτηριστικών που χρησιμοποιούν τις τοπικές σχέσεις των δεδομένων, ενώ τα πλήρως συνδεδεμένα επίπεδα είναι καλοί ταξινομητές, επομένως χρησιμοποιούνται ως τα τελευταία επίπεδα ενός CNN\@.
Επιπλέον, τα συνελικτικά επίπεδα δημιουργούν χάρτες ενεργοποιήσεων χρησιμοποιώντας κοινόχρηστα βάρη που έχουν σταθερό αριθμό παραμέτρων σε αντίθεση με τα πλήρως συνδεδεμένα επίπεδα, καθιστώντας τα έτσι πολύ ταχύτερα.
Το VGGnet~\cite{simonyan2014very} είναι μια απλή CNN αρχιτεκτονική που χρησιμοποιεί μικρά συνελικτικά φίλτρα ($3\times 3$) και η απόδοση του αυξάνεται απλά αυξάνοντας το βάθος του δικτύου.
Το GoogleNet~\cite{szegedy2015going} είναι μια άλλη CNN αρχιτεκτονική η οποία χρησιμοποιεί την μονάδα inception.
Η μονάδα inception χρησιμοποιεί παράλληλα πολλαπλά επίπεδα περιελίξεων από τα οποία το αποτέλεσμα συσσωρεύεται σειριακά, επιτρέποντας έτσι στο δίκτυο να μαθαίνει χαρακτηριστικά πολλαπλών επιπέδων.
Το ResNet~\cite{he2016deep} είναι μια CNN αρχιτεκτονική η οποία διαμορφώνει τα επίπεδα ως συναρτήσεις μάθησης υπολειμμάτων με βάση τις εισόδους των επιπέδων, επιτρέποντας την δημιουργία πολύ πιο βαθιών δικτύων, σε σχέση με τα προηγουμένως αναφερθέντα.

Οι Αυτο-κωδικοποιητές (Autoencoders, AE) είναι νευρωνικά δίκτυα που εκπαιδεύονται με στόχο να αντιγράψουν την είσοδο στην έξοδο με τέτοιο τρόπο ώστε να κωδικοποιούν χρήσιμες ιδιότητες των δεδομένων.
Συνήθως αποτελείται από ένα τμήμα κωδικοποίησης που υποδειγματολειπτεί την είσοδο μέχρι αυτό να γίνει γραμμικό χαρακτηριστικό και ένα τμήμα αποκωδικοποίησης που υπερδειγματολειπτεί προς τις αρχικές διαστάσεις.
Μια κοινή αρχιτεκτονική AE είναι ο Στοιβαγμένος Αυτο-κωδικοποιητής Αποθορυβοποίησης (Stacked Denoised Autoencoder, SDAE) που έχει ως στόχο την ανακατασκευή της εισόδου από μια τεχνητά αλλοιωμένη έκδοση της~\cite{vincent2010stacked}, η οποία εμποδίζει το μοντέλο να μάθει προφανείς λύσεις.
Μια άλλη αρχιτεκτονική τύπου AE είναι το u-net~\cite{ronneberger2015u}, το οποίο παρουσιάζει ιδιαίτερο ενδιαφέρον για τη βιοϊατρική κοινότητα καθώς αρχικά εφαρμόστηκε για κατάτμηση βιοϊατρικών εικόνων.
Το u-net εισήγαγε τις συνδέσεις παράλειψης (skip connections) που συνδέουν τα επίπεδα του κωδικοποιητή με τα αντίστοιχα επίπεδα του αποκωδικοποιητή.

Τα επαναλαμβανόμενα νευρωνικά δίκτυα (Recurrent Neural Network, RNN) είναι δίκτυα που αποτελούνται από βρόχους ανατροφοδότησης και σε αντίθεση με τις προηγούμενες καθορισμένες αρχιτεκτονικές μπορούν να χρησιμοποιήσουν την εσωτερική τους κατάσταση για να επεξεργαστούν την είσοδο.
Τα απλά RNN έχουν το πρόβλημα της εξαφάνισης των κλίσεων (vanishing gradients) και για το λόγο αυτό προτάθηκε η Μακρο-Βραχυ Πρόθεσμη Μονάδα Μνήμης (Long-Short Term Memory, LSTM) ως λύση για την αποθήκευση πληροφοριών για παρατεταμένο χρόνο.
Η Κεκλεισμένη Επαναληπτική Μονάδα (Gated Reccurrent Unit, GRU) προτάθηκε αργότερα ως μια απλούστερη εναλλακτική λύση έναντι του LSTM\@.

\clearpage
\bibliography{chapter3.bib}
\bibliographystyle{unsrt}

%% file: chapter2-nn.tex
\begin{tikzpicture}[]
	\node[circle, draw=white] (x1) at (0, 4) {$x_1$};
	\node at (0, 3) {$\vdots$};
	\node[circle, draw=white] (xj) at (0, 2) {$x_j$};
	\node at (0, 1) {$\vdots$};
	\node[circle, draw=white] (xn) at (0, 0) {$x_n$};
	\node[circle, draw=white] (b) at (4, 4) {$b$};
	\node[circle, draw=black] (sigma) at (4, 2) {$\phi$};
	\node[circle, draw=white] (output) at (6, 2) {$\alpha$};
	\draw[->] (x1) to node[above]{$w_1$} (sigma);
	\draw[->] (xj) to node[above]{$w_j$} (sigma);
	\draw[->] (xn) to node[above]{$w_n$} (sigma);
	\draw[->] (b) -- (sigma);
	\draw[->] (sigma) -- (output);
\end{tikzpicture}

%% file: chapter2-cnn.tex
\begin{tikzpicture}[scale=0.9]
	\draw[dashed] (0.03, 0) -- (1.5, 0);
	\draw[dashed] (1.61, 0) -- (3.01, 0);
	\draw[dashed] (3.2, 0) -- (4.5, 0);
	\draw[dashed] (4.9, 0) -- (6, 0);
	\draw[dashed] (6.8, 0) -- (7.5, 0);
	\networkLayer{4.0}{0.03}{0}{0.0}{color=white}{}
	\networkLayer{3.5}{0.1}{1.5}{0.0}{color=white}{}
	\networkLayer{3.0}{0.2}{3.0}{0.0}{color=white}{}
	\networkLayer{2.5}{0.4}{4.5}{0.0}{color=white}{}
	\networkLayer{2.0}{0.8}{6.0}{0.0}{color=white}{}
	\networkLayer{1.5}{1.6}{7.5}{0.0}{color=white}{}
	\draw[dashed] (10, -1) -- (11.5, 0.5);
	\draw[-] (10.5, -1) -- (12, 0.5);
	\draw[-] (10.5, -0.5) -- (12, 1);
	\draw[-] (10, -0.5) -- (11.5, 1);
	\draw[-] (12, 1) -- (11.5, 1);
	\draw[-] (12, 1) -- (12, 0.5);
	\draw[-] (10, -1) rectangle ++ (0.5,0.5) node{};
	\node[canvas is zy plane at x=0, scale=0.9] at (-0.735, 1.235){\includegraphics[scale=0.491]{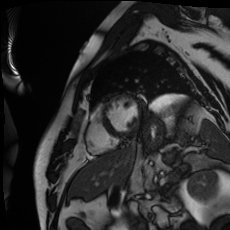}};
	\node[canvas is zy plane at x=0, scale=0.9] at (0.93, 1.07){\includegraphics[scale=0.89]{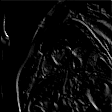}};
	\node[canvas is zy plane at x=0, scale=0.9] at (0.83, 1.07){\includegraphics[scale=0.89]{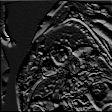}};
	\node[canvas is zy plane at x=0, scale=0.9] at (2.63, 0.93){\includegraphics[scale=1.52]{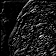}};
	\node[canvas is zy plane at x=0, scale=0.9] at (2.42, 0.93){\includegraphics[scale=1.52]{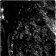}};
	\node[canvas is zy plane at x=0, scale=0.9] at (4.42, 0.78){\includegraphics[scale=2.53]{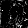}};
	\node[canvas is zy plane at x=0, scale=0.9] at (4.03, 0.78){\includegraphics[scale=2.53]{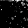}};
	\node[canvas is zy plane at x=0, scale=0.9] at (6.42, 0.611){\includegraphics[scale=4.05]{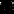}};
	\node[canvas is zy plane at x=0, scale=0.9] at (5.6, 0.611){\includegraphics[scale=4.05]{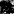}};
	\node[canvas is zy plane at x=0, scale=0.9] at (8.815, 0.472){\includegraphics[scale=6.05]{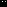}};
	\node[canvas is zy plane at x=0, scale=0.9] at (7.22, 0.472){\includegraphics[scale=6.05]{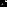}};
	\node at (8, 0.472) {$\dots$};
	\node at (6, 0.472) {$\dots$};
	\node[canvas is zy plane at x=0] at (-1, -3){\includegraphics[scale=10]{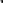}};
	\node at (-0.8, -3) {$\dots$};
	\node[canvas is zy plane at x=0] at (-0.6, -3){\includegraphics[scale=10]{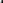}};
	\node[canvas is zy plane at x=0] at (0.8, -3){\includegraphics[scale=10]{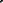}};
	\node at (1, -3) {$\dots$};
	\node[canvas is zy plane at x=0] at (1.3, -3){\includegraphics[scale=10]{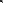}};
	\node[canvas is zy plane at x=0] at (2.5, -3){\includegraphics[scale=10]{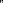}};
	\node at (2.7, -3) {$\dots$};
	\node[canvas is zy plane at x=0] at (2.9, -3){\includegraphics[scale=10]{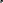}};
	\node[canvas is zy plane at x=0] at (4.2, -3){\includegraphics[scale=10]{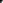}};
	\node at (4.4, -3) {$\dots$};
	\node[canvas is zy plane at x=0] at (4.6, -3){\includegraphics[scale=10]{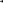}};
	\node[canvas is zy plane at x=0] at (6.3, -3){\includegraphics[scale=10]{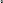}};
	\node at (6.5, -3) {$\dots$};
	\node[canvas is zy plane at x=0] at (6.7, -3){\includegraphics[scale=10]{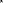}};
	\draw[<-] (-0.3, -3) -- (0.3, -3);
	\draw[<-] (1.6, -3) -- (2.1, -3);
	\draw[<-] (3.2, -3) -- (3.9, -3);
	\draw[<-] (4.8, -3) -- (5.9, -3);
	\draw[<-] (7.1, -3) -- (13.0, -3);
	\draw[dashed] (-1.51, -1.53) -- (0.15, -1.35);
	\draw[dashed] (0.27, -1.35) -- (1.85, -1.16);
	\draw[dashed] (2.05, -1.16) -- (3.55, -0.97);
	\draw[dashed] (3.95, -0.97) -- (5.25, -0.77);
	\draw[dashed] (6.02, -0.78) -- (6.94, -0.57);
	\draw[dashed] (-1.51, 2.45) -- (0.15, 2.15);
	\draw[dashed] (0.27, 2.15) -- (1.85, 1.85);
	\draw[dashed] (2.05, 1.85) -- (3.55, 1.54);
	\draw[dashed] (3.95, 1.54) -- (5.25, 1.23);
	\draw[dashed] (6.02, 1.22) -- (6.94, 0.91);
	\draw[dashed] (0.03, 4) -- (1.5, 3.51);
	\draw[dashed] (1.61, 3.51) -- (3.01, 3.01);
	\draw[dashed] (3.2, 3) -- (4.5, 2.5);
	\draw[dashed] (4.9, 2.51) -- (6, 2);
	\draw[dashed] (6.8, 2) -- (7.5, 1.5);
	\draw[dotted] (9.1, 1.5) --(11.5, 1);
	\draw[dotted] (9.1, 0) -- (11.5, 0.5);
	\draw[dotted] (8.5, 0.92) -- (10, -0.5);
	\draw[dotted] (8.5, -0.58) --  (10, -1);
	\draw[dashdotted] (10.5, -0.5) -- (13, -0.25);
	\draw[dashdotted] (10.5, -1) -- (13, -0.25);
	\draw[dashdotted] (12, 1) -- (13, -0.25);
	\draw[dashdotted] (12, 0.5) -- (13, -0.25);
	\node[align=left] at (13.3,-0.2) {$\hat{y}$};
	\draw[->] (13.3, -0.6) -- (13.3, -2.7) node [pos=1.1] {$J$};
	\draw[<-] (13.5, -3) -- (14, -3) node [pos=1.6] {$y$};
	\node[align=left] at (10.0, -2.4) {αντίστροφη διάδοση σφάλματος};
	\node[align=left] at (10.8, 2.5) {προς-τα-εμπρός διάδοση};
\end{tikzpicture}

%% file: chapter3.tex
\chapter{Βαθιά μάθηση με δομημένα δεδομένα και σήματα}
\label{chapter3}
\graphicspath{{./images/deep-learning-in-cardiology/}}

\section{Εισαγωγή}
Οι καρδιαγγειακές παθήσεις (Cardiovascular Diseases, CVDs) είναι η κύρια αιτία θανάτου παγκοσμίως αντιπροσωπεύοντας το 30\% των θανάτων το 2014 στις Ηνωμένες Πολιτείες~\cite{benjamin2017heart}, το 45\% των θανάτων στην Ευρώπη και εκτιμάται ότι κοστίζουν 210 δισ. Ευρώ ετησίως μόνο για την Ευρωπαϊκή Ένωση~\cite{wilkins2017european}.

Στα επόμενα δύο κεφάλαια θα παρουσιάσουμε μια συστηματική βιβλιογραφική ανασκόπηση, που έγινε στο πλαίσιο της παρούσας διδακτορικής διατριβής, των εφαρμογών της βαθιάς μάθησης σε δομημένα δεδομένα, σήματα και απεικόνιση από την καρδιολογία, που σχετίζονται με καρδιακές δομές και αγγεία.
Η φράση αναζήτησης της βιβλιογραφίας είναι η συνδυασμός καθενός από τους όρους καρδιολογίας και βαθιάς μάθησης που αναφέρονται στα Ακρωνύμια, χρησιμοποιώντας το Google Scholar\footnote{\url{https://scholar.google.com}}, το Pubmed\footnote{\url{https://ncbi.nlm.nih.gov/pubmed/}} και το Scopus\footnote{\url{https://www.scopus.com/search/form.uri?=display=basic}}.
Έπειτα τα αποτελέσματα επιλέγονται έτσι ώστε να ταιριάζουν με τα κριτήρια επιλογής της ανασκόπησης τα οποία συνοψίζονται σε δύο κύριους άξονες: την αρχιτεκτονική του νευρωνικού δικτύου και ο τύπος δεδομένων που χρησιμοποιήθηκε για εκπαίδευση/επικύρωση/δοκιμή.
Οι αξιολογήσεις αναφέρονται για περιοχές που χρησιμοποίησαν ένα συνεκτικό σύνολο μετρήσεων με την ίδια μη-τροποποιημένη βάση δεδομένων και το ίδιο ερευνητικό ερώτημα.
Δημοσιεύσεις που δεν παρέχουν πληροφορίες σχετικά με την αρχιτεκτονική του νευρωνικού δικτύου ή δημοσιεύσεις που απλά αναπαράγουν μεθόδους προηγούμενων δημοσιεύσεων ή προκαταρκτικές δημοσιεύσεις αποκλείονται από αυτήν την ανασκόπηση.
Όταν αναφέρεται η λέξη \textit{πολλαπλά} στη στήλη `Αποτελέσματα', αυτά αναφέρονται στο κυρίως κείμενο όπου είναι κατάλληλο και ειδικά για τις αρχιτεκτονικές συνελικτικών νευρωνικών δικτύων, η χρήση του όρου \textit{επίπεδο} συνεπάγεται `συνελικτικό επίπεδο' για χάρη συντομίας.
Οι πίνακες~\ref{table:cardiologypublicdatabases1},~\ref{table:cardiologypublicdatabases2} και~\ref{table:cardiologypublicdatabases3} παρέχουν μια επισκόπηση των δημόσια διαθέσιμων βάσεων δεδομένων καρδιολογίας που έχουν χρησιμοποιηθεί για εκπαίδευση μοντέλων βαθιάς μάθησης.

\begin{sidewaystable}
	\centering
	\caption{Δημόσιες καρδιολογικές βάσεις δεδομένων, δομημένων δεδομένων και σημάτων}
	\label{table:cardiologypublicdatabases1}
	\begin{tabular}{l c r l}
		\toprule
		\thead{Βάση δεδομένων}                                & \thead{Ακρωνύμιο} & \thead{Ασθενείς}                                                               & \thead{Πρόβλημα}                                                      \\
		\midrule
		\multicolumn{4}{l}{\thead{Δομημένες βάσεις δεδομένων}}                                                                                                                                                                                                                                                           \\
		\midrule
		Medical Information Mart for Intensive Care III~\cite{johnson2016mimic}                                                              & MIMIC             & 38597                                                                          & 53423 νοσοκομειακές εισαγωγές για πρόβλεψη ICD-9     \\
		KNHANES-VI~\cite{kweon2014data}                                                                                                      & KNH               & 8108                                                                           & δημογραφικά, τεστ αίματος και τρόπος ζωής \\
		\midrule
		\multicolumn{4}{l}{\thead{Βάσεις δεδομένων σημάτων (όλα ECG εκτός του~\cite{karlen2013multiparameter})}}                                                                                                                                                                                                          \\
		\midrule
		IEEE-TBME PPG Respiratory Rate Benchmark Dataset~\cite{karlen2013multiparameter}                                                     & PPGDB             & 42                                                                             & εκτίμηση ρυθμού αναπνοής με χρήση PPG                                 \\
		Creighton University Ventricular Tachyarrhythmia~\cite{nolle1986crei}                                                                & CREI              & 35                                                                             & εντοπισμός κοιλιακής ταχυαρρυθμίας                                    \\
		MIT-BIH Atrial Fibrillation Database~\cite{moody1983new}                                                                             & AFDB              & 25                                                                             & πρόβλεψη HF                                                           \\
		BIH Deaconess Medical Center CHF Database~\cite{baim1986survival}                                                                    & CHFDB             & 15                                                                             & ταξινόμηση CHF                                                        \\
		St.Petersburg Institute of Cardiological Technics~\cite{goldberger2000physiobank}                                                    & INDB              & 32                                                                             & εντοπισμός QRS και ταξινόμηση χτύπων ECG                              \\
		Long-Term Atrial Fibrillation Database~\cite{petrutiu2007abrupt}                                                                     & LTAFDB            & 84                                                                             & εντοπισμός QRS και ταξινόμηση χτύπων ECG                              \\
		Long-Term ST Database~\cite{jager2003long}                                                                                           & LTSTDB            & 80                                                                             & ανίχνευση χτύπων ST και ταξινόμηση                                    \\
		MIT-BIH Arrhythmia Database~\cite{moody2001impact}                                                                                   & MITDB             & 47                                                                             & ανίχνευση αρρυθμιών                                                   \\
		MIT-BIH Noise Stress Test Database~\cite{moody1984noise}                                                                             & NSTDB             & 12                                                                             & χρήση για δοκιμή αντοχής θορύβου μοντέλων                             \\
		MIT-BIH Normal Sinus Rhythm Database~\cite{goldberger2000physiobank}                                                                 & NSRDB             & 18                                                                             & ανίχνευση αρρυθμιών                                                   \\
		MIT-BIH Normal Sinus Rhythm RR Interval Database~\cite{goldsmith1992comparison}                                                      & NSR2DB            & 54                                                                             & ταξινόμηση χτύπων ECG                                                 \\
		Fantasia Database~\cite{iyengar1996age}                                                                                              & FAN               & 40                                                                             & ταξινόμηση χτύπων ECG                                                 \\
		AF Classification short single lead ECG Physionet 2017~\cite{moody2001impact}                                                        & PHY17             & ---\footnote{Ο αριθμός των ασθενών δεν αναφέρεται.} & ταξινόμηση χτύπων ECG (12186 σήματα μονού καναλιού)                   \\
		Physionet 2016 Challenge~\cite{liu2016open}                                                                                          & PHY16             & 1297                                                                           & ταξινόμηση ήχων καρδιάς με χρήση PCG (3126 σήματα)                    \\
		Physikalisch-Technische Bundesanstalt ECG Database~\cite{bousseljot1995nutzung}                                                      & PTBDB             & 268                                                                            & διάγνωση καρδιακών ασθενειών                                          \\
		QT Database~\cite{laguna1997database}                                                                                                & QTDB              & 105                                                                            & ανίχνευση χτύπων QT και ταξινόμηση                                    \\
		MIT-BIH Supraventricular Arrhythmia Database~\cite{greenwald1990improved}                                                            & SVDB              & 78                                                                             & ανίχνευση υπερκοιλιακών αρρυθμιών                                     \\
		Non-Invasive Fetal ECG Physionet Challenge Dataset~\cite{silva2013noninvasive}                                                       & PHY13             & 447                                                                            & μέτρηση εμβρυϊκού HR, διαστήματος RR και QT                           \\
		DeepQ Arrhythmia Database~\cite{wu2017deepq} & DeepQ             & 299                                                                            & ταξινόμηση χτύπων ECG (897 σήματα)                                    \\
		\bottomrule
	\end{tabular}
\end{sidewaystable}

\section{Βαθιά μάθηση με δομημένα δεδομένα}
\label{sec3:structured}
Τα δομημένα δεδομένα περιλαμβάνουν κυρίως EHRs και συνήθως διατηρούνται σε σχεσιακές βάσεις δεδομένων.
Μια σύνοψη των εφαρμογών βαθιάς μάθησης που χρησιμοποιούν δομημένα δεδομένα παρουσιάζεται στον Πίνακα~\ref{table:structured}.

\begin{sidewaystable}
	\centering
	\caption{Εφαρμογές βαθιάς μάθησης με χρήση δομημένων δεδομένων}
	\label{table:structured}
	\begin{tabular}{l c l l}
		\toprule
		\thead{Αναφορά}                                    & \thead{Μέθοδος} & \thead{Εφαρμογή/Σημειώσεις\footnote{Σε παρένθεση οι βάσεις δεδομένων που χρησιμοποιήθηκαν.}}                        & \thead{Αποτέλεσμα\footnote{\label{structuredlabel}Υπάρχει μεγάλη μεταβλητότητα στην αναφορά αποτελεσμάτων. Όλα τα αποτελέσματα ειναι ακρίβειες εκτός από το~\cite{choi2016using} που αναφέρει AUC και το~\cite{hsiao2016deep} που είναι στατιστική μελέτη.}} \\
		\midrule
		Gopalswamy 2017~\cite{gopalswamy2017deep}           & LSTM            & πρόβλεψη BP και διάρκεια περίθαλψης με χρήση πολλαπλών βιοδεικτών (μη-δημόσια)                                      & 73.1\%                                                                                                                                                                                                                                                       \\
		Choi 2016~\cite{choi2016using}                      & GRU             & διάγνωση του HF με χρήση GRU και παράθυρου παρατήρησης (μη-δημόσια)                               & 0.883\textsuperscript{b}                                                                                                                                                                                                                                        \\
		Purushotham 2018~\cite{purushotham2018benchmarking} & FNN, GRU        & προβλήματα πρόβλεψης της MIMIC με χρήση ενός FNN και ενός μοντέλου GRU (MIMIC)                                      & \textit{πολλαπλά}                                                                                                                                                                                                                                            \\
		Kim 2017~\cite{kim2017highrisk}                     & GRU, CNN        & πρόβλεψη υψηλού ρίσκου αγγειακής ασθένειας με ένα Bi-GRU και ενός 1D CNN (μη-δημόσια) & \textit{πολλαπλά}                                                                                                                                                                                                                                            \\
		Hsiao 2016~\cite{hsiao2016deep}                     & AE              & ανάλυση ρίσκου CVD με AE και softmax σε ασθενείς, μετεωρολογικά δεδομένα (μη-δημόσια)          & \_\_\textsuperscript{b}                                                                                                                                                                                                                                              \\
		Kim 2017~\cite{kim2017statistics}                   & DBN             & πρόβλεψη καρδιαγγειακού ρίσκου με χρήση DBN (KNH)                                                                   & 83.9\%                                                                                                                                                                                                                                                       \\
		Huang 2018~\cite{huang2018regularized}              & SDAE            & πρόβλεψη ACS με SDAE με δύο περιορισμούς συστηματοποίησης και softmax (μη-δημόσια)          & 73.0\%                                                                                                                                                                                                                                                       \\
		\bottomrule
	\end{tabular}
\end{sidewaystable}

Τα RNNs έχουν χρησιμοποιηθεί για τη διάγνωση καρδιαγγειακών νοσημάτων με χρήση δομημένων δεδομένων.
Στο~\cite{gopalswamy2017deep} οι συγγραφείς προβλέπουν την πίεση του αίματος (Blood Pressure, BP) κατά τη διάρκεια χειρουργικής επέμβασης και μετά, χρησιμοποιώντας LSTM\@.
Πραγματοποίησαν πειράματα σε ένα σύνολο 12036 χειρουργικών επεμβάσεων που περιέχουν πληροφορίες για ενδοεγχειρητικά σήματα (θερμοκρασία σώματος, αναπνευστικός ρυθμός, καρδιακός ρυθμός, διαστολική BP (Diastolic Blood Pressure, DBP), συστολική BP (Systolic Blood Pressure, SBP), κλάσμα εισπνεόμενου $O_2$), επιτυγχάνοντας καλύτερα αποτελέσματα από τα KNN και SVM\@.
Οι Choi et al.~\cite{choi2016using} εκπαίδευσαν ένα GRU με διαχρονικά δεδομένα EHR, ανιχνεύοντας σχέσεις μεταξύ χρονικών συμβάντων (διάγνωση ασθενειών, εντολές φαρμάκων κ.λπ.), χρησιμοποιώντας ένα παράθυρο παρατήρησης.
Κάνουν διάγνωση καρδιακής ανεπάρκειας (Heart Failure, HF) με AUC 0.777 για παράθυρο 12 μηνών και 0.883 για παράθυρο 18 μηνών, καλύτερα από τα MLP, SVM και KNN\@.
Οι Purushotham et al.~\cite{purushotham2018benchmarking} σύγκριναν τον super-learner (σύνολο ρηχών αλγορίθμων μηχανικής μάθησης)~\cite{polley2010super} με FNN, RNN και ένα πολυτροπικό μοντέλο βαθιάς μάθησης που προτάθηκε από τους συγγραφείς στη βάση δεδομένων MIMIC\@.
Το προτεινόμενο πλαίσιο χρησιμοποιεί FNN και GRU για το χειρισμό των μη-χρονικών και χρονικών χαρακτηριστικών αντίστοιχα, μαθαίνοντας έτσι τις κοινές λανθάνουσες αναπαραστάσεις τους για πρόβλεψη.
Τα αποτελέσματα δείχνουν ότι οι μέθοδοι βαθιάς μάθησης υπερβαίνουν κατά πολύ τον super-learner στην πλειονότητα των προβλημάτων πρόβλεψης της MIMIC (πρόβλεψη θνησιμότητας εντός νοσοκομείου με AUC 0.873, πρόβλεψη βραχυπρόθεσμης θνησιμότητας με AUC 0.871, πρόβλεψη μακροπρόθεσμης θνησιμότητας με AUC 0.87 και πρόβλεψη κωδικού ICD-9 με AUC 0.777).
Οι Kim et al.~\cite{kim2017highrisk} δημιούργησαν δύο μοντέλα πρόγνωσης ιατρικού ιστορικού χρησιμοποιώντας δίκτυα προσοχής και τα αξιολόγησαν σε 50000 ασθενείς με υπέρταση.
Έδειξαν ότι η χρήση ενός GRU διπλής κατεύθυνσης παρέχει καλύτερη διακριτική ικανότητα από ένα αντίστοιχο συνελικτικό δίκτυο, το οποίο όμως έχει μικρότερο χρόνο εκπαίδευσης με ανταγωνιστική ακρίβεια.

Τα ΑΕ χρησιμοποιήθηκαν για τη διάγνωση καρδιαγγειακών νοσημάτων με δομημένα δεδομένα.
Οι Hsiao et al.~\cite{hsiao2016deep} εκπαίδευσαν ένα ΑΕ και ένα επίπεδο softmax για την ανάλυση κινδύνου τεσσάρων κατηγοριών καρδιαγγειακών παθήσεων.
Η είσοδος περιελάμβανε δημογραφικά στοιχεία, κωδικούς ICD-9 από αρχεία εξωτερικών ασθενών, ρύπους συγκέντρωσης και μετεωρολογικές παραμέτρους από περιβαλλοντικά αρχεία.
Οι Huang et al.~\cite{huang2018regularized} εκπαίδευσαν ένα SDAE χρησιμοποιώντας ένα σύνολο δεδομένων EHR 3464 ασθενών για να προβλέψουν Οξύ Στεφανιαίο Σύνδρομο (Acute Coronary Syndrome, ACS).
Το SDAE έχει δύο περιορισμούς συστηματοποίησης (regularization) που κάνουν τις ανακατασκευασμένες αναπαραστάσεις χαρακτηριστικών να περιέχουν περισσότερες πληροφορίες σχετικές με το επίπεδο κινδύνου, καταγράφοντας έτσι τα χαρακτηριστικά των ασθενών σε παρόμοια επίπεδα κινδύνου και διατηρώντας τις διακριτικές πληροφορίες σε διαφορετικά επίπεδα κινδύνου.
Στη συνέχεια, τοποθέτησαν ένα επίπεδο softmax, το οποίο προσαρμόζεται στο πρόβλημα της κλινικής πρόβλεψης κινδύνου.

Τα DBN έχουν επίσης χρησιμοποιηθεί σε συνδυασμό με δομημένα δεδομένα εκτός από τα RNNs και AEs.
Στο~\cite{kim2017statistics} οι συγγραφείς πρώτα εφήρμοσαν μια στατιστική μελέτη ενός συνόλου δεδομένων 4244 εγγραφών για την εύρεση μεταβλητών που σχετίζονται με καρδιαγγειακές παθήσεις (ηλικία, φύλο, χοληστερόλη, λιποπρωτεΐνη υψηλής πυκνότητας, SBP, DBP, κάπνισμα, διαβήτης).
Στη συνέχεια, ανέπτυξαν ένα μοντέλο DBN για την πρόβλεψη καρδιαγγειακών παθήσεων (υπέρταση, υπερλιπιδαιμία, μυοκαρδιακό έμφραγμα (Myocardial Infarction, MI), στηθάγχη).
Σύγκριναν το μοντέλο τους με Naive Bayes, λογιστική παλινδρόμηση, SVM, RF και ένα βασικό DBN πετυχαίνοντας καλύτερα αποτελέσματα.

Σύμφωνα με τη βιβλιογραφία, τα RNN χρησιμοποιούνται ευρέως σε δομημένα δεδομένα καρδιολογίας επειδή είναι σε θέση να βρουν χρονικά χαρακτηριστικά καλύτερα από άλλες μεθόδους βαθιάς/μηχανικής μάθησης.
Από την άλλη πλευρά, οι εφαρμογές σε αυτόν τον τομέα είναι σχετικά λίγες και αυτό συμβαίνει κυρίως επειδή υπάρχει μικρός αριθμός δημόσιων βάσεων δεδομένων, κάτι το οποίο εμποδίζει την περαιτέρω αξιολόγηση και σύγκριση διαφορετικών αρχιτεκτονικών.
Επιπλέον, οι δομημένες βάσεις δεδομένων λόγω σχεδιασμού τους περιέχουν λιγότερες πληροφορίες για έναν μεμονωμένο ασθενή και επικεντρώνονται περισσότερο σε ομάδες ασθενών, καθιστώντας αυτές πιο κατάλληλες για επιδημιολογικές μελέτες παρά για τον τομέα της καρδιολογίας.

\section{Βαθιά μάθηση με σήματα}
\label{sec3:signals}
Τα σήματα περιλαμβάνουν χρονοσειρές όπως ηλεκτροκαρδιογράφημα (Electrocardiogram, ECG), φωνοκαρδιογράφημα (Phonocardiogram, PCG), παλμομετρικά δεδομένα και δεδομένα από φορητά εξαρτήματα.
Ένας λόγος που η παραδοσιακή μηχανική μάθηση έχει δουλέψει αρκετά καλά σε αυτόν τον τομέα τα προηγούμενα χρόνια οφείλεται στη χρήση χειροποίητων και προσεκτικά σχεδιασμένων χαρακτηριστικών από εμπειρογνώμονες, όπως τα στατιστικά μέτρα από το ECG beats και το διάστημα RR~\cite{faziludeen2013ecg}.
Η βαθιά μάθηση μπορεί να βελτιώσει τα μοντέλα όταν οι επισημάνσεις των ειδικών είναι χαμηλής ποιότητας ή όταν είναι δύσκολο να δημιουργηθεί ένα μοντέλο με χρήση χειροποίητων χαρακτηριστικών.
Μια σύνοψη των εφαρμογών βαθιάς μάθησης που χρησιμοποιούν σήματα εμφανίζεται στους πίνακες~\ref{table:signals1},~\ref{table:signals2},~\ref{table:signals3} και~\ref{table:signals4}.

\begin{sidewaystable}
	\centering
	\caption{Εφαρμογές βαθιάς μάθησης για ανίχνευση αρρυθμιών με την MITDB}
	\label{table:signals1}
	\begin{tabular}{l c l l}
		\toprule
		\thead{Αναφορά}                                & \thead{Μέθοδος} & \thead{Εφαρμογή/Σημειώσεις\footnote{Σε παρένθεση οι βάσεις δεδομένων που χρησιμοποιήθηκαν.}}          & \thead{Ακρίβεια\footnote{Υπάρχει μεγάλη μεταβλητότητα στην αναφορά αποτελεσμάτων. Τα αποτελέσματα του~\cite{kiranyaz2016real} είναι για κοιλιακούς/υπερκοιλιακούς εκτοπικούς χτύπους, το~\cite{isin2017cardiac} είναι για τρεις τύπους αρρυθμίας, το~\cite{wu2016novel} είναι για πέντε τύπους αρρυθμίας.}} \\
		\midrule
		Zubair 2016~\cite{zubair2016automated}          & CNN             & μη-γραμμικός μετασχηματισμός για ανίχνευση R-peak και 1D CNN με μεταβλητό ρυθμό μάθησης           & 92.7\%                                                                                                                                                                                                                                                                                                                                                                                                                                                                                                                                                                                                                                                                                                                                                                                                                                                                           \\
		Li 2017~\cite{li2017classification}             & CNN             & WT για αποθορυβοποίηση και ανίχνευση των R-peak και 1D CNN δύο \textit{επιπέδων}                      & 97.5\%                                                                                                                                                                                                                                                                                                                                                                                                                                                                                                                                                                                                                                                                                                                                                                                                                                                                           \\
		Kiranyaz 2016~\cite{kiranyaz2016real}           & CNN             & CNN ειδικό για κάθε ασθενή με χρήση προσαρμόσιμων 1D συνελικτικών επιπέδων                            & 99\%,97.6\%\textsuperscript{b}                                                                                                                                                                                                                                                                                                                                                                                                                                                                                                                                                                                                                                                                                                                                                                                                                                              \\
		Isin 2017~\cite{isin2017cardiac}                & CNN             & φίλτρα αποθορυβοποίησης, Pan-Tomkins, AlexNet για χαρακτηριστικά και PCA για ταξινόμηση      & 92.0\%\textsuperscript{b}                                                                                                                                                                                                                                                                                                                                                                                                                                                                                                                                                                                                                                                                                                                                                                                                                                                   \\
		Luo 2017~\cite{luo2017patient}                  & SDAE            & φίλτρα αποθορυβοποίησης, ανίχνευση R-peak με χρήση παραγώγων, WT, SDAE και softmax                    & 97.5\%                                                                                                                                                                                                                                                                                                                                                                                                                                                                                                                                                                                                                                                                                                                                                                                                                                                                           \\
		Jiang 2017~\cite{jiang2017heartbeat}            & SDAE            & φίλτρα αποθορυβοποίησης, Pan-Tomkins, SDAE και FNN                                                    & 97.99\%                                                                                                                                                                                                                                                                                                                                                                                                                                                                                                                                                                                                                                                                                                                                                                                                                                                                          \\
		Yang 2017~\cite{yang2017novel}                  & SSAE            & κανονικοποίηση του ECG, SSAE                                                                          & 99.45\%                                                                                                                                                                                                                                                                                                                                                                                                                                                                                                                                                                                                                                                                                                                                                                                                                                                                          \\
		Wu 2016~\cite{wu2016novel}                      & DBN             & φίλτρα αποθορυβοποίησης, ecgpuwave, δύο τύποι RBMs                                                    & 99.5\%\textsuperscript{b}                                                                                                                                                                                                                                                                                                                                                                                                                                                                                                                                                                                                                                                                                                                                                                                                                                                   \\
		\bottomrule
	\end{tabular}
\end{sidewaystable}

\subsection{Ηλεκτροκαρδιογράφημα}
Το ECG είναι η μέθοδος μέτρησης των ηλεκτρικών δυναμικών της καρδιάς για τη διάγνωση καρδιακών προβλημάτων~\cite{badnjevic2017inspection}.
Είναι μη-επεμβατική, εύκολη στην απόκτηση και παρέχει χρήσιμες πληροφορίες για τη διάγνωση ασθενειών.
Έχει χρησιμοποιηθεί κυρίως για την ανίχνευση αρρυθμιών χρησιμοποιώντας τον μεγάλο αριθμό δημόσιων βάσεων δεδομένων ECG όπως φαίνεται στον Πίνακα~\ref{table:cardiologypublicdatabases1}.

\subsubsection{Ανίχνευση αρρυθμιών με την MITDB}
Τα CNN έχουν χρησιμοποιηθεί για την ανίχνευση αρρυθμιών με την MITDB\@.
Οι Zubair et al.~\cite{zubair2016automated} ανίχνευσαν τις κορυφές R χρησιμοποιώντας ένα μη-γραμμικό μετασχηματισμό και τμηματοποιούν χτύπους γύρω από αυτό.
Στη συνέχεια, χρησιμοποίησαν τα τμήματα για να εκπαιδεύσουν ένα 1D CNN τριών \textit{επιπέδων} με μεταβλητό ρυθμό μάθησης, βασισμένο στο μέσο τετραγωνικό σφάλμα πετυχαίνοντας καλύτερα αποτελέσματα από τις προηγούμενες καλύτερες μεθόδους.
Οι Li et al.~\cite{li2017classification} χρησιμοποίησαν Μετασχηματισμό Κυματιδίων (Wavelet Transform, WT) για την απομάκρυνση του θορύβου υψηλής συχνότητας και της μετατόπισης της γραμμής βάσης και διορθογωνικού spline για την ανίχνευση των κορυφών R.
Έπειτα, δημιούργησαν τμήματα γύρω από την κορυφή R τα οποία τροφοδότησαν σε ένα 1D CNN δυο \textit{επιπέδων}.
Στο άρθρο τους οι Kiranyaz et al.~\cite{kiranyaz2016real} εκπαίδευσαν CNNs που μπορούν να χρησιμοποιηθούν για την ταξινόμηση δεδομένων ECG μακράς διάρκειας και για την παρακολούθηση ECG σε πραγματικό χρόνο, ως μέρος συστήματος έγκαιρης προειδοποίησης σε φορητή συσκευή.
Το CNN αποτελούνταν από τρία προσαρμόσιμα 1D συνελικτικά \textit{επίπεδα}.
Επιτεύχθηκε 99\% και 97.6\% για την ταξινόμηση κοιλιακών και υπερκοιλιακών εκτοπικών παλμών αντίστοιχα.
Στο~\cite{isin2017cardiac} οι συγγραφείς χρησιμοποίησαν αφαίρεση μέσου όρου για απομάκρυνση της βάσης αναφοράς, κινητό φίλτρο μέσου όρου για απομάκρυνση των υψηλών συχνοτήτων, διαφορικό φίλτρο για απομάκρυνση της γραμμής βάσης και φίλτρο χτένας (comb) για την απομάκρυνση του θορύβου γραμμής ισχύος.
Ανίχνευσαν τα QRS με τον αλγόριθμο Pan-Tompkins~\cite{pan1985real}, εξήγαγαν τμήματα χρησιμοποιώντας δείγματα μετά την κορυφή R τα οποία μετέτρεψαν σε $256\times 256\times 3$ δυαδικές εικόνες.
Οι εικόνες έπειτα τροφοδοτήθηκαν σε ένα εξαγωγέα χαρακτηριστικών AlexNet προεκπαιδευμένο στην βάση δεδομένων ImageNet και έπειτα στην Ανάλυση Κύριων Συνιστωσών (Principal Component Analysis, PCA).
Πέτυχαν υψηλή ακρίβεια στην ταξινόμηση τριών τύπων αρρυθμιών της MITDB\@.

Τα ΑΕ έχουν επίσης χρησιμοποιηθεί για ανίχνευση αρρυθμιών με MITDB\@.
Στο άρθρο τους οι Luo et al.~\cite{luo2017patient} χρησιμοποίησαν αξιολόγηση ποιότητας για την αφαίρεση των καρδιακών παλμών χαμηλής ποιότητας, δύο φίλτρα μέσης τιμής για την αφαίρεση του θορύβου της γραμμής ισχύος, του θορύβου υψηλής συχνότητας και της μετατόπισης της γραμμής βάσης.
Στη συνέχεια, χρησιμοποίησαν έναν διαφορικό αλγόριθμο για να ανιχνεύσουν τις κορυφές R και τα χρονικά παράθυρα για την κατάτμηση κάθε καρδιακού παλμού.
Χρησιμοποίησαν επίσης WT για τον υπολογισμό του φάσματος κάθε καρδιακού παλμού και ενός SDAE για την εξαγωγή χαρακτηριστικών από το φάσμα.
Έπειτα, δημιούργησαν έναν ταξινομητή τεσσάρων αρρυθμιών από τον κωδικοποιητή του SDAE και ένα softmax, επιτυγχάνοντας ακρίβεια 97.5\%.
Στο~\cite{jiang2017heartbeat} οι συγγραφείς αποθορυβοποίησαν τα σήματα με ένα χαμηλοπερατό, ένα ζωνοπερατό και ένα φίλτρο διάμεσου.
Εντόπισαν κορυφές R χρησιμοποιώντας τον αλγόριθμο Pan-Tomkins και κατάτμησαν/ανακατασκεύασαν τους καρδιακούς παλμούς.
Χαρακτηριστικά εξήχθησαν από το σήμα της καρδιάς χρησιμοποιώντας ένα SDAE και χρησιμοποιήθηκε ένα FNN για την ταξινόμηση καρδιακών παλμών από 16 τύπους αρρυθμίας.
Παρατηρήθηκε συγκρίσιμη απόδοση σε σχέση με προηγούμενες μεθόδους βασισμένες στην χειροκίνητη εξαγωγή χαρακτηριστικών.
Οι Yang et al.~\cite{yang2017novel} κανονικοποίησαν το ECG και στη συνέχεια το τροφοδότησαν σε ένα Στοιβαγμένο Αραιό AE (Stacked Sparse Autoencoder, SSAE).
Ταξινομούν έξι τύπους αρρυθμιών επιτυγχάνοντας ακρίβεια 99.5\% ενώ ταυτόχρονα αποδεικνύουν την ανθεκτικότητα του έναντι στο θόρυβο με χρήση τεχνητά προστιθέμενου θορύβου.

Τα DBN έχουν επίσης χρησιμοποιηθεί για αυτό το πρόβλημα εκτός από CNN και AE\@.
Οι Wu et al.~\cite{wu2016novel} χρησιμοποίησαν φίλτρα διάμεσου για την απομάκρυνση της γραμμής βάσης, ένα χαμηλοπερατό φίλτρο για την απομάκρυνση του θορύβου γραμμής ισχύος και του θορύβου υψηλής συχνότητας.
Εντόπισαν κορυφές R χρησιμοποιώντας το λογισμικό ecgpuwave από την Physionet και κατάτμησαν τους ECG χτύπους.
Δύο τύποι RBMs, εκπαιδεύτηκαν για την εξαγωγή χαρακτηριστικών από το ECG για ανίχνευση αρρυθμιών.
Επιτεύχθηκε ακρίβεια 99.5\% σε πέντε κατηγορίες της MITDB\@.

\subsubsection{Ανίχνευση αρρυθμιών με άλλες βάσεις δεδομένων}
Τα CNN χρησιμοποιήθηκαν για ανίχνευση αρρυθμιών χρησιμοποιώντας άλλες βάσεις δεδομένων εκτός της MITDB\@.
Στο~\cite{wu2018personalizing} οι συγγραφείς δημιούργησαν ένα CNN δύο \textit{επιπέδων} χρησιμοποιώντας την DeepQ και την MITDB για να ταξινομήσουν τέσσερις τύπους αρρυθμιών.
Τα σήματα υπόκεινται σε έντονη προεπεξεργασία με φίλτρα απομάκρυνσης (μεσαία, υψηλή, και χαμηλή διέλευση, αφαίρεση ακραίων τιμών) και κατατμήσονται σε 0.6 δευτερόλεπτα γύρω από την κορυφή R.
Στη συνέχεια, τροφοδοτούνται στο CNN μαζί με το διάστημα RR για εκπαίδευση.
Οι συγγραφείς χρησιμοποιούν επίσης μια ενεργή μέθοδο μάθησης για την επίτευξη εξατομικευμένων αποτελεσμάτων και βελτιωμένης ακρίβειας, επιτυγχάνοντας υψηλή ευαισθησία και θετική προβλεψιμότητα και στα δύο σύνολα δεδομένων.
Οι Hannun et al.~\cite{hannun2019cardiologist} δημιούργησαν ένα σύνολο δεδομένων με ECG από φορητές συσκευές το οποίο περιέχει τον μεγαλύτερο αριθμό ασθενών (30000) σε σχέση με προηγούμενες βάσεις δεδομένων και το χρησιμοποίησαν για να εκπαιδεύσουν ένα Residual-CNN με 34 \textit{επίπεδα}.
Το μοντέλο τους ανιχνεύει ένα ευρύ φάσμα αρρυθμιών συνολικά 14 κατηγορίες, ξεπερνώντας τον μέσο καρδιολόγο σε ακρίβεια.
Στο άρθρο τους, οι Acharya et al.~\cite{acharya2017automateda} εκπαίδευσαν ένα CNN τεσσάρων \textit{επιπέδων} στην AFDB, MITDB και CREI, για την ταξινόμηση μεταξύ φυσιολογικού, AF, κολπικού πτερυγισμού και κοιλιακής μαρμαρυγής.
Χωρίς την ανίχνευση του QRS, πέτυχαν συγκρίσιμες επιδόσεις με προηγούμενες μεθόδους που βασίστηκαν στην ανίχνευση κορυφών R και στην χειροκίνητη εξαγωγή χαρακτηριστικών.
Οι ίδιοι συγγραφείς επίσης εκπαίδευσαν την προηγούμενη CNN αρχιτεκτονική για τον εντοπισμό των απινιδωτικών και μη-απινιδωτικών κοιλιακών αρρυθμιών~\cite{acharya2018automated}, εντοπισμό CAD ασθενών με τις FAN και INDB~\cite{acharya2017automatedb}, ταξινόμηση CHF με τις CHFDB, NSTDB, FAN~\cite{acharya2018deep} και την δοκιμασία της αντοχής τους στο θόρυβο με αποθορυβοποίηση WT~\cite{acharya2017application}.

\begin{sidewaystable}
	\centering
	\caption{Εφαρμογές βαθιάς μάθησης με χρήση ECG για ανίχνευση αρρυθμιών και AF}
	\label{table:signals2}
	\begin{tabular}{l c l l}
		\toprule
		\thead{Αναφορά}                                & \thead{Μέθοδος} & \thead{Εφαρμογή/Σημειώσεις\footnote{Σε παρένθεση οι βάσεις δεδομένων που χρησιμοποιήθηκαν.}}          & \thead{Ακρίβεια\footnote{Υπάρχει μεγάλη μεταβλητότητα στην αναφορά αποτελεσμάτων. Τα αποτελέσματα του~\cite{hannun2019cardiologist} είναι precision, το~\cite{xiong2015denoising} αναφέρει SNR και πολλαπλά αποτελέσματα ανάλογα με τον προστιθέμενο θόρυβο, το αποτέλεσμα του~\cite{taji2017false} αφορά μελέτη αντοχής θορύβου.}} \\
		\midrule
		\multicolumn{4}{l}{\thead{Ανίχνευση αρρυθμιών}}                                                                                                                                                                                                                                                                                                                                                                                                                                                                                                                                                                                                                                                                                                                                                                                                                                                                                                                                                                                                             \\
		\midrule
		Wu 2018~\cite{wu2018personalizing}              & CNN             & ενεργή μάθηση και CNN δύο \textit{επιπέδων} με ECG και διαστήματα RR (MITDB, DeepQ) & \textit{πολλαπλά}                                                                                                                                                                                                                                                                                                                                                                                                                                                                                                                                                                                                                                                                                                                                                                                                                                                                \\
		Hannun 2019~\cite{hannun2019cardiologist} & CNN             & CNN 34-\textit{επιπέδων} (μη-δημόσια)                                                                 & 80\%\textsuperscript{b}                                                                                                                                                                                                                                                                                                                                                                                                                                                                                                                                                                                                                                                                                                                                                                                                                                                     \\
		Acharya 2017~\cite{acharya2017automateda}       & CNN             & CNN τεσσάρων \textit{επιπέδων} (AFDB, MITDB, Creighton)                                               & 92.5\%                                                                                                                                                                                                                                                                                                                                                                                                                                                                                                                                                                                                                                                                                                                                                                                                                                                                           \\
		Schwab 2017~\cite{schwab2017beat}               & RNN             & ensemble από RNNs με μηχανισμό προσοχής (PHY17)                                                       & 79\%                                                                                                                                                                                                                                                                                                                                                                                                                                                                                                                                                                                                                                                                                                                                                                                                                                                                             \\
		\midrule
		\multicolumn{4}{l}{\thead{Ανίχνευση AF}}                                                                                                                                                                                                                                                                                                                                                                                                                                                                                                                                                                                                                                                                                                                                                                                                                                                                                                                                                                                                                    \\
		\midrule
		Yao 2017~\cite{yao2017atrial}                   & CNN             & πολυ-κλιμακωτό CNN (AFDB, LTAFDB, μη-δημόσια)                                                         & 98.18\%                                                                                                                                                                                                                                                                                                                                                                                                                                                                                                                                                                                                                                                                                                                                                                                                                                                                          \\
		Xia 2018~\cite{xia2018detecting}                & CNN             & CNN με φασματογράφημα από STFT ή στατικού WT (AFDB)                                                   & 98.29\%                                                                                                                                                                                                                                                                                                                                                                                                                                                                                                                                                                                                                                                                                                                                                                                                                                                                          \\
		Andersen 2018~\cite{andersen2018deep}           & CNN, LSTM       & διαστήματα RR με CNN-LSTM (MITDB, AFDB, NSRDB)                                                        & 87.40\%                                                                                                                                                                                                                                                                                                                                                                                                                                                                                                                                                                                                                                                                                                                                                                                                                                                                          \\
		Xiong 2015~\cite{xiong2015denoising}            & AE              & κατωφλίωση προσαρμοσμένης κλίμακας WT και AE αποθορυβοποίησης (MITDB, NSTDB)                          & $\sim$18.7\textsuperscript{b}                                                                                                                                                                                                                                                                                                                                                                                                                                                                                                                                                                                                                                                                                                                                                                                                                                               \\
		Taji 2017~\cite{taji2017false}                  & DBN             & μείωση false alarm κατά τη διάρκεια ανίχνευσης AF σε θορυβώδη ECG σήματα (AFDB, NSTDB)                & 87\%\textsuperscript{b}                                                                                                                                                                                                                                                                                                                                                                                                                                                                                                                                                                                                                                                                                                                                                                                                                                                     \\
		\bottomrule
	\end{tabular}
\end{sidewaystable}

Μια εφαρμογή των RNN σε αυτή την περιοχή είναι από τους Schwab et al.~\cite{schwab2017beat} που δημιούργησαν ένα ensemble από RNN που διακρίνει μεταξύ φυσιολογικών φλεβοκομβικών ρυθμών, AF, άλλων τύπων αρρυθμίας και θορυβώδους σήματος.
Εισήγαγαν μια μορφοποίηση του προβλήματος κατά την οποία τμηματοποιούν το ECG σε καρδιακούς παλμούς για να μειώσουν τον αριθμό των χρονικών βημάτων ανά ακολουθία.
Επέκτειναν επίσης τα RNNs με έναν μηχανισμό προσοχής που τους επιτρέπει να εκτιμήσουν σε ποιους καρδιακούς παλμούς επικεντρώνεται το RNN για να λάβει τις αποφάσεις του και να επιτύχει συγκρίσιμη αποτελεσματικότητα με άλλες μεθόδους χρησιμοποιώντας λιγότερες παραμέτρους.

\subsubsection{Ανίχνευση AF}
Τα CNN χρησιμοποιήθηκαν για την ανίχνευση AF\@.
Οι Yao et al.~\cite{yao2017atrial} εξήγαγαν την ακολουθία του στιγμιαίου καρδιακού ρυθμού, η οποία τροφοδοτείται σε ένα CNN πολλαπλής κλίμακας που εξάγει το αποτέλεσμα ανίχνευσης AF, επιτυγχάνοντας καλύτερα αποτελέσματα από τις προηγούμενες μεθόδους όσον αφορά την ακρίβεια.
Οι Xia et al.~\cite{xia2018detecting} σύγκριναν δύο CNN, με τρία και δύο \textit{επίπεδα}, τα οποία τροφοδοτήθηκαν με φάσματα σημάτων από την AFDB χρησιμοποιώντας βραχυπρόθεσμο μετασχηματισμό Fourier (Short-Time Fourier Transform, STFT) και στατικό WT αντίστοιχα.
Τα πειράματά τους κατέληξαν στο συμπέρασμα ότι η χρήση του στατικού WT επιτυγχάνει μια ελαφρώς καλύτερη ακρίβεια για αυτό το πρόβλημα.

Εκτός από τα CNN έχουν χρησιμοποιηθεί και άλλες αρχιτεκτονικές για την ανίχνευση AF\@.
Οι Andersen et al.~\cite{andersen2018deep} μετέτρεψαν τα σήματα ECG από την AFDB σε διαστήματα RR για να τα ταξινομήσουν για ανίχνευση AF\@.
Στη συνέχεια, κατάτμησαν τα διαστήματα RR σε 30 δείγματα το καθένα και τα τροφοδότησαν σε ένα δίκτυο με δύο \textit{επίπεδα} ακολουθούμενα από ένα επίπεδο συγκέντρωσης (pooling) και ένα επίπεδο LSTM με 100 νευρώνες.
Η μέθοδος επικυρώθηκε στην MITDB και την NSRDB επιτυγχάνοντας ακρίβεια που δείχνει ότι μπορεί να γενικεύσει.
Στο~\cite{xiong2015denoising} οι συγγραφείς πρόσθεσαν σήματα θορύβου από την NSTDB στην MITDB και στη συνέχεια χρησιμοποίησαν κατωφλίωση προσαρμοσμένης κλίμακας WT, για να απομακρύνουν το μεγαλύτερο μέρος του θορύβου και ένα SDAE για να αφαιρεθεί ο υπολειπόμενος θόρυβος.
Τα πειράματά τους έδειξαν ότι η αύξηση του αριθμού των δεδομένων εκπαίδευσης σε 1000, αυξάνει δραματικά το λόγο σήματος προς θόρυβο μετά την αποθορυβοποίηση.
Οι Taji et al.~\cite{taji2017false} εκπαίδευσαν ένα DBN για να ταξινομήσουν αποδεκτά από μη-αποδεκτά τμήματα ECG, έτσι ώστε να μειώσουν το ποσοστό ψευδούς συναγερμού που προκαλείται από κακής ποιότητας ECG κατά την ανίχνευση AF\@.
Οκτώ διαφορετικά επίπεδα ποιότητας ECG παρέχονται, προσθέτοντας στο ECG θόρυβο κίνησης από την NSTDB\@.
Με SNR $-20dB$ στο ECG, η μέθοδος τους πέτυχε αύξηση 22\% της ακρίβειας σε σύγκριση με ένα μοντέλο βάσης αναφοράς.

\subsubsection{Άλλες εφαρμογές με δημόσιες βάσεις δεδομένων}
Ταξινόμηση χτύπων του ECG πραγματοποιήθηκε επίσης από μια σειρά μελετών που χρησιμοποίησαν δημόσιες βάσεις δεδομένων.
Στο~\cite{xiao2018monitoring} οι συγγραφείς μετεκπαίδευσαν ένα Inception-v3 προεκπαιδευμένο στην ImageNet, χρησιμοποιώντας σήματα από την LTSTDB για την ταξινόμηση συμβάντων ST\@.
Τα δείγματα εκπαίδευσης ήταν πάνω από 500000 τμήματα ST και μη-ST ECG σημάτων με διάρκεια δέκα δευτερολέπτων που έπειτα μετατράπηκαν σε εικόνες.
Επιτυγχάνουν συγκρίσιμες επιδόσεις με προηγούμενες πολύπλοκες μεθόδους βασισμένες σε χειροκίνητο ορισμό κανόνων.
Οι Rahhal et al.~\cite{al2016deep} εκπαίδευσαν ένα SDAEs με περιορισμό αραιότητας και ένα softmax για την ταξινόμηση των ECG χτύπων.
Σε κάθε επανάληψη, ο εμπειρογνώμονας επισημάνει τους πιο αβέβαιους χτύπους ECG στο σετ δοκιμών, τα οποία στη συνέχεια χρησιμοποιούνται για εκπαίδευση, ενώ η έξοδος του δικτύου εκχωρεί τα μέτρα εμπιστοσύνης σε κάθε χτύπο.
Πειράματα που εκτελούνται στις MITDB, INDB, SVDB δείχνουν την αξιοπιστία και την υπολογιστική αποτελεσματικότητα της μεθόδου.
Στο~\cite{abrishami2018p} οι συγγραφείς εκπαίδευσαν τρεις ξεχωριστές αρχιτεκτονικές για να προσδιορίσουν τα κύματα P-QRS-T στο ECG με την QTDB\@.
Σύγκριναν ένα FNN δύο \textit{επιπέδων}, ένα CNN δύο \textit{επιπέδων} και ένα CNN δύο \textit{επιπέδων} με dropout, με το δεύτερο να πετυχαίνει τα καλύτερα αποτελέσματα.

\begin{sidewaystable}
	\centering
	\caption{Εφαρμογές βαθιάς μάθησης με χρήση ECG σε άλλες εφαρμογές}
	\label{table:signals3}
	\begin{tabular}{l c l l}
		\toprule
		\thead{Αναφορά}                       & \thead{Μέθοδος} & \thead{Εφαρμογή/Σημειώσεις\footnote{Σε παρένθεση οι βάσεις δεδομένων που χρησιμοποιήθηκαν.}} & \thead{Ακρίβεια\footnote{Υπάρχει μεγάλη μεταβλητότητα στην αναφορά αποτελεσμάτων. Τα αποτελέσματα του~\cite{xiao2018monitoring} αναφέρει AUC, το~\cite{al2016deep} αναφέρει πολλαπλές ακρίβειες για υπερκοιλιακούς/κοιλιακούς εκτοπικούς χτύπους, το~\cite{wu2016myocardial} αναφέρει ευαισθησία και εξειδίκευση (specificity), το~\cite{hwang2018deep} αναφέρει αποτελέσματα για δύο περιπτώσεις.}} \\
		\midrule
		\multicolumn{4}{l}{\thead{Άλλες εφαρμογές}}                                                                                                                                                                                                                                                                                                                                                                                                                                                                                                                                                                                                                                                                                                                                                                                                                                                                                                                                                                                               \\
		\midrule
		Xiao 2018~\cite{xiao2018monitoring}    & CNN             & ταξινόμηση γεγονότων ST από ECG με μεταφοράς μάθησης στο Inception-v3 (LTSTDB)               & 0.867\textsuperscript{b}                                                                                                                                                                                                                                                                                                                                                                                                                                                                                                                                                                                                                                                                                                                                                                                                                                                    \\
		Rahhal 2016~\cite{al2016deep}          & SDAE            & SDAE με περιορισμό αραιότητας και softmax (MITDB, INDB, SVDB)                                 & \textgreater{99\%}\textsuperscript{b}                                                                                                                                                                                                                                                                                                                                                                                                                                                                                                                                                                                                                                                                                                                                                                                                                                       \\
		Abrishami 2018~\cite{abrishami2018p}   & Multiple        & σύγκριναν ένα FNN, CNN και CNN με dropout για εντοπισμό κυμάτων ECG (QTDB)                & 96.2\%                                                                                                                                                                                                                                                                                                                                                                                                                                                                                                                                                                                                                                                                                                                                                                                                                                                                           \\
		Wu 2016~\cite{wu2016myocardial}        & SAE             & εντοπισμός και ταξινόμηση MI με SAE και πολυ-κλιμακωτό διακριτό WT (PTBDB)                       & $\sim$99\%\textsuperscript{b}                                                                                                                                                                                                                                                                                                                                                                                                                                                                                                                                                                                                                                                                                                                                                                                                                                               \\
		Reasat 2017~\cite{reasat2017detection} & Inception       & ανίχνευση MI με Inception μπλοκ για κάθε κανάλι ECG (PTBDB)                                    & 84.54\%                                                                                                                                                                                                                                                                                                                                                                                                                                                                                                                                                                                                                                                                                                                                                                                                                                                                          \\
		Zhong 2018~\cite{zhong2018deep}        & CNN             & CNN τριών \textit{επιπέδων} για την ταξινόμηση εμβρυϊκών ECG (PHY13)                         & 77.85\%                                                                                                                                                                                                                                                                                                                                                                                                                                                                                                                                                                                                                                                                                                                                                                                                                                                                          \\
		\midrule
		\multicolumn{4}{l}{\thead{Άλλες εφαρμογές (μη-δημόσιες βάσεις)}}                                                                                                                                                                                                                                                                                                                                                                                                                                                                                                                                                                                                                                                                                                                                                                                                                                                                                                                                                                          \\
		\midrule
		Ripoll 2016~\cite{ripoll2016ecg}       & RBM             & ανίχνευση μη-κανονικών ECG με προεκπαιδευμένα RBMs                                           & 85.52\%                                                                                                                                                                                                                                                                                                                                                                                                                                                                                                                                                                                                                                                                                                                                                                                                                                                                          \\
		Jin 2017~\cite{jin2017classification}  & CNN             & ανίχνευση μη-κανονικών ECG με lead-CNN και κανόνα συμπερασμού                                & 86.22\%                                                                                                                                                                                                                                                                                                                                                                                                                                                                                                                                                                                                                                                                                                                                                                                                                                                                          \\
		Liu 2018~\cite{liu2018detecting}       & Multiple        & συνέκριναν το Inception και ένα 1D CNN για πρόωρη κοιλιακή συστολή με ECG                    & 88.5\%                                                                                                                                                                                                                                                                                                                                                                                                                                                                                                                                                                                                                                                                                                                                                                                                                                                                           \\
		Hwang 2018~\cite{hwang2018deep}        & CNN, RNN        & ανίχνευση στρες με ένα CNN  ενός \textit{επιπέδου} με dropout και δύο RNNs                   & 87.39\%\textsuperscript{b}                                                                                                                                                                                                                                                                                                                                                                                                                                                                                                                                                                                                                                                                                                                                                                                                                                                  \\
		\bottomrule
	\end{tabular}
\end{sidewaystable}

Το ECG έχει επίσης χρησιμοποιηθεί για τον εντοπισμό και την ταξινόμηση του ΜΙ.
Στο άρθρο τους οι Wu et al.~\cite{wu2016myocardial} εντόπισαν και ταξινόμησαν MI στην PTBDB\@.
Χρησιμοποίησαν διακριτό WT πολλαπλών κλιμάκων, για να διευκολύνουν την εξαγωγή χαρακτηριστικών για ΜΙ σε συγκεκριμένες αναλυτικότητες συχνοτήτων και ένα επίπεδο παλινδρόμησης softmax για να δημιουργήσουν έναν ταξινομητή πολλαπλών κατηγοριών.
Τα πειράματα επικύρωσης δείχνουν ότι η μέθοδος τους απέδωσε καλύτερα από τις προηγούμενες μεθόδους, όσον αφορά την ευαισθησία και την εξειδίκευση.
Η PTBDB χρησιμοποιήθηκε επίσης από τους Reasat et al.~\cite{reasat2017detection} για να εκπαιδεύσουν ένα μοντέλο CNN βασισμένο στο inception.
Κάθε ηλεκτρόδιο του ECG τροφοδοτείται σε ένα inception μπλοκ, ακολουθούμενο από επίπεδα συγκόλλησης (concatenation), παγκόσμιας μέσης συγκέντρωσης (global average pooling) και ένα softmax.
Οι συγγραφείς συνέκριναν τη μέθοδο τους με μια προηγούμενη σύγχρονη μέθοδο που χρησιμοποιεί SWT, πετυχαίνοντας καλύτερα αποτελέσματα.

Συμπλέγματα εμβρυϊκού QRS ταυτοποιήθηκαν με ένα CNN τριών \textit{στρωμάτων} με dropout από τους Zhong et al.~\cite{zhong2018deep} με την PHY13.
Αρχικά, τα σήματα κακής ποιότητας απορρίπτονται χρησιμοποιώντας την εντροπία του δείγματος και στη συνέχεια κανονικοποιημένα τμήματα διάρκειας 100ms τροφοδοτούνται στο CNN για εκπαίδευση.
Οι συγγραφείς συνέκριναν τη μέθοδο τους με τα KNN, Naive Bayes και SVM επιτυγχάνοντας καλύτερα αποτελέσματα.

\subsubsection{Άλλες εφαρμογές με μη-δημόσιες βάσεις δεδομένων}
Η ανίχνευση μη-φυσιολογικών ECG μελετήθηκε από μια σειρά δημοσιεύσεων.
Οι Ripoll et al.~\cite{ripoll2016ecg} χρησιμοποίησαν προεκπαιδευμένα μοντέλα με ECG από 1390 ασθενείς, για να αξιολογήσουν εάν ένας ασθενής στο ασθενοφόρο ή στα επείγοντα πρέπει να παραπεμφθεί σε μια καρδιολογική υπηρεσία.
Σύγκριναν το μοντέλο τους με τα KNN, SVM, μηχανές ακραίας μάθησης (Extreme Learning Machines, ELMs) και με ένα σύστημα εμπειρογνωμόνων πετυχαίνοντας καλύτερα αποτελέσματα στην ακρίβεια και την εξειδίκευση.
Στο~\cite{jin2017classification} οι συγγραφείς εκπαίδευσαν ένα μοντέλο το οποίο ταξινομεί φυσιολογικούς και μη-φυσιολογικούς ασθενείς με 193690 αρχεία ECG 10 έως 20 δευτερολέπτων.
Το μοντέλο τους αποτελείται από δύο παράλληλα μέρη; την στατιστική μάθηση και ένα κανόνα συμπεράσματος.
Κατά τη διάρκεια της μάθησης, τα ECG υπόκεινται σε προεπεξεργασία με χρήση χαμηλοπερατών και ζωνοπερατών φίλτρων, στη συνέχεια τροφοδοτούνται σε δύο παράλληλα lead-CNNs και τέλος χρησιμοποιείται Bayesian σύντηξη για τον συνδυασμό των εξόδων πιθανότητας.
Κατά τη διάρκεια του συμπερασμού, ανιχνεύονται οι θέσεις κορυφής R στο αρχείο ECG και χρησιμοποιούνται τέσσερις κανόνες ασθένειας για την ανάλυση.
Τέλος, χρησιμοποιούν το μέσο όρο bias για τον προσδιορισμό του αποτελέσματος.

Άλλες εφαρμογές περιλαμβάνουν την ταξινόμηση της πρόωρης κοιλιακής συστολής και την ανίχνευση του στρες.
Οι Liu et al.~\cite{liu2018detecting} χρησιμοποίησαν ένα σύνολο δεδομένων μονού καναλιού με 2400 φυσιολογικά και πρόωρης κοιλιακής συστολής ECG από το Νοσοκομείο Παίδων της Σαγκάης για εκπαίδευση.
Δύο διαφορετικά μοντέλα εκπαιδεύτηκαν χρησιμοποιώντας τις εικόνες των κυματομορφών.
Το πρώτο ήταν ένα CNN δύο \textit{επιπέδων} με dropout και το δεύτερο ένα Inception-v3 εκπαιδευμένο στην Imagenet.
Άλλα τρία μοντέλα εκπαιδεύτηκαν χρησιμοποιώντας τα 1D σήματα.
Το πρώτο μοντέλο ήταν ένα FNN με dropout, το δεύτερο ένα 1D CNN τριών \textit{επιπέδων} και το τρίτο ένα 2D CNN το ίδιο με το πρώτο αλλά εκπαιδευμένο με μια στοιβαγμένη έκδοση του σήματος (με επαύξηση δεδομένων).
Τα πειράματα τους έδειξαν ότι το 1D CNN με τα τρία \textit{επίπεδα} είχαν καλύτερα και πιο σταθερά αποτελέσματα.
Στο~\cite{hwang2018deep} οι συγγραφείς εκπαίδευσαν ένα δίκτυο με ένα συνελικτικό επίπεδο με dropout, ακολουθούμενο από δύο RNNs για τον εντοπισμό του άγχους χρησιμοποιώντας βραχυπρόθεσμα δεδομένα ECG\@.
Έδειξαν ότι το δίκτυό τους πέτυχε τα καλύτερα αποτελέσματα σε σύγκριση με παραδοσιακές μεθόδους μάθησης μηχανών και DNN βάσης αναφοράς.

\subsubsection{Συνολική άποψη για την χρήση της βαθιάς μάθησης στο ECG}
Πολλές μέθοδοι βαθιάς μάθησης έχουν χρησιμοποιήσει ECG για την εκπαίδευση μοντέλων χρησιμοποιώντας τον μεγάλο αριθμό διαθέσιμων βάσεων δεδομένων.
Είναι προφανές από τη βιβλιογραφία ότι οι περισσότερες μέθοδοι βαθιάς μάθησης (κυρίως CNN και SDAE) σε αυτή την περιοχή αποτελούνται από τρία μέρη: φιλτράρισμα για αποθορυβοποίηση, ανίχνευση κορυφών R για κατάτμηση χτύπων και ένα νευρωνικό δίκτυο για την εξαγωγή χαρακτηριστικών.
Ένα άλλο δημοφιλές σύνολο μεθόδων είναι η μετατροπή των ECG σε εικόνες, για την αξιοποίηση ενός ευρύ φάσματος των αρχιτεκτονικών και των προεκπαιδευμένων μοντέλων που έχουν ήδη κατασκευαστεί για τις μορφές απεικόνισης.
Αυτό έγινε χρησιμοποιώντας τεχνικές ανάλυσης φάσματος~\cite{luo2017patient, xia2018detecting} και μετατροπών σε δυαδική εικόνα~\cite{xiao2018monitoring, liu2018detecting, isin2017cardiac}.

\subsection{Φωνοκαρδιογράφημα με χρήση της Physionet 2016}
Ο διαγωνισμός της Physionet/Computing στην Καρδιολογία (Cinc) 2016 (PHY16) αφορούσε την ταξινόμηση των φυσιολογικών/παθολογικών καρδιακών ηχογραφήσεων.
Το δεδομένα εκπαίδευσης αποτελούνται από πέντε βάσεις δεδομένων (Α έως Ε) που περιέχουν 3126 φωνοκαρδιογραφήματα (PCGs), τα οποία διαρκούν από 5 δευτερόλεπτα έως 120 δευτερόλεπτα.

Οι περισσότερες από τις μεθόδους μετατρέπουν τα PCG σε εικόνες χρησιμοποιώντας τεχνικές φασματογραφίας.
Οι Rubin et al.~\cite{rubin2017recognizing} χρησιμοποίησαν ένα κρυφό ημι-Markov μοντέλο για την κατάτμηση της έναρξης κάθε παλμού της καρδιάς, το οποίο στη συνέχεια μετατράπηκε σε φασματογράφημα με τη χρήση μετρητών συχνότητας Mel-Frequency Cepstral (MFCCs).
Κάθε φασματογράφημα ταξινομήθηκε σε φυσιολογικό ή μη-φυσιολογικό χρησιμοποιώντας ένα CNN δύο \textit{επιπέδων} με τροποποιημένη συνάρτηση απώλειας που μεγιστοποιεί την ευαισθησία και την εξειδίκευση, μαζί με μια παράμετρο συστηματοποίησης.
Η τελική ταξινόμηση του σήματος ήταν η μέση πιθανότητα όλων των πιθανοτήτων των τμημάτων.
Πέτυχαν συνολική ακρίβεια 83.99\% τοποθετώντας την μέθοδο όγδοη κατά τη διάρκεια του διαγωνισμού στο PHY16.
Οι Kucharski et al.~\cite{kucharski2017deep} χρησιμοποίησαν ένα φασματογράφημα οκτώ δευτερολέπτων για τα τμήματα, πριν τροφοδοτηθούν σε ένα CNN πέντε \textit{επιπέδων} με dropout.
Η μέθοδος τους πέτυχε ευαισθησία 99.1\% και ειδικότητα 91.6\% οι οποίες είναι συγκρίσιμες με αποτελέσματα προηγούμενων μεθόδων.
Οι Dominguez et al.~\cite{dominguez2018deep} ταξινόμησαν τα σήματα και τα προεπεξεργάστηκαν χρησιμοποιώντας τον νευρομορφικό ακουστικό αισθητήρα~\cite{jimenez2017binaural} για να αποσυνθέσουν τις πληροφορίες ήχου σε ζώνες συχνοτήτων.
Στη συνέχεια, υπολογίζουν τα φασματογραφήματα που τροφοδοτούνται σε μια τροποποιημένη έκδοση του δικτύου AlexNet.
Το μοντέλο τους πέτυχε ακρίβεια 94.16\%, η οποία είναι μια σημαντική βελτίωση σε σύγκριση με το μοντέλο που κέρδισε το PHY16.
Στο~\cite{potes2016ensemble} οι συγγραφείς χρησιμοποίησαν το Adaboost το οποίο τροφοδοτήθηκε με χαρακτηριστικά φασματογραφίας από PCG και ένα CNN το οποίο εκπαιδεύτηκε χρησιμοποιώντας καρδιακούς κύκλους αποσυνθεμένους σε τέσσερις ζώνες συχνοτήτων.
Τέλος, οι έξοδοι του Adaboost και του CNN συνδυάστηκαν για να παράξουν το τελικό αποτέλεσμα ταξινόμησης χρησιμοποιώντας έναν απλό κανόνα απόφασης.
Η συνολική ακρίβεια ήταν 89\%, τοποθετώντας τη μέθοδο πρώτη στο διαγωνισμό του PHY16.

Τα μοντέλα που δεν μετατρέπουν τα PCG σε φασματογραφήματα φαίνεται να έχουν μικρότερη απόδοση.
Οι Ryu et al.~\cite{ryu2016classification} εφήρμοσαν ένα φίλτρο με παράθυρο Hamming Window-sinc για αποθορυβοποίηση, έπειτα κλιμάκωσαν το σήμα και χρησιμοποίησαν ένα σταθερό παράθυρο για κατάτμηση.
Εκπαίδευσαν ένα 1D CNN τεσσάρων \textit{επιπέδων} χρησιμοποιώντας τα τμήματα ενώ η τελική ταξινόμηση ήταν ο μέσος όρος όλων των πιθανοτήτων των τμημάτων.
Επιτεύχθηκε συνολική ακρίβεια 79.5\% στην επίσημη φάση του PHY16.

Τα φωνοκαρδιογραφήματα έχουν επίσης χρησιμοποιηθεί για προβλήματα όπως η αναγνώριση ήχου καρδιάς S1 και S2 από τον Chen et al.~\cite{chen2017s1}.
Μετασχημάτισαν τα ηχητικά σήματα της καρδιάς σε μια ακολουθία MFCCs και έπειτα εφάρμοσαν Κ-μέσους για να συσσωρεύσουν τα χαρακτηριστικά MFCC σε δύο ομάδες για να βελτιώσουν την εκπροσώπηση και τη διακριτική τους ικανότητα.
Τα χαρακτηριστικά τροφοδοτούνται στη συνέχεια σε ένα DBN για την εκτέλεση ταξινόμησης S1 και S2.
Οι συγγραφείς συνέκριναν τη μέθοδο τους με τα μοντέλα KNN, Gaussian mixture, λογιστική παλινδρόμηση και SVM επιτυγχάνοντας καλύτερα αποτελέσματα.

Σύμφωνα με τη βιβλιογραφία, τα CNN αποτελούν την πλειονότητα των αρχιτεκτονικών νευρωνικών δικτύων που χρησιμοποιήθηκαν για την επίλυση προβλημάτων με PCG\@.
Επιπλέον, όπως και στο ECG, πολλές μέθοδοι βαθιάς μάθησης μετασχηματίζουν τα σήματα σε εικόνες χρησιμοποιώντας τεχνικές φασματογραφίας~\cite{potes2016ensemble, rubin2017recognizing, kucharski2017deep, dominguez2018deep, pan2017variation, shashikumar2017deep}.

\subsection{Άλλα σήματα}
\subsubsection{Παλμομετρικά δεδομένα}
Τα παλμομετρικά δεδομένα χρησιμοποιούνται για την εκτίμηση της SBP και της DBP που είναι οι αιμοδυναμικές πιέσεις που ασκούνται στο αρτηριακό σύστημα κατά τη διάρκεια της συστολής και της διαστολής αντίστοιχα~\cite{everly2012clinical}.

Τα DBN έχουν χρησιμοποιηθεί για την εκτίμηση των SBP και DBP\@.
Στο άρθρο τους οι Lee et al.~\cite{lee2017deepa} χρησιμοποίησαν bootstrap-aggregation για να δημιουργήσουν ensemble παραμέτρους και χρησιμοποίησαν τον Adaboost για την εκτίμηση των SBP και DBP\@.
Στη συνέχεια, χρησιμοποίησαν bootstrap και Monte-Carlo για να προσδιορίσουν τα διαστήματα εμπιστοσύνης βασισμένα στο BP, τα οποία εκτιμήθηκαν χρησιμοποιώντας τον ensemble εκτιμητή παλινδρόμησης του DBN\@.
Αυτή η τροποποίηση βελτίωσε σημαντικά την εκτίμηση του BP σε σχέση με το βασικό μοντέλο DBN\@.
Παρόμοια κατεύθυνση έχει ακολουθηθεί για το ίδιο πρόβλημα από τους ίδιους συγγραφείς στα~\cite{lee2017oscillometric, lee2017deepc, lee2017deepd}.

Παλμομετρικά δεδομένα έχουν επίσης χρησιμοποιηθεί από τους Pan et al.~\cite{pan2017variation} για την εκτίμηση της μεταβολής των ήχων Korotkoff.
Οι χτύποι χρησιμοποιήθηκαν για να δημιουργήσουν παράθυρα κεντραρισμένα στις κορυφές των παλμών, οι οποίες μετά εξήχθησαν.
Ανάλυση φάσματος λήφθηκε από κάθε χτύπο και όλοι οι χτύποι μεταξύ των χειροκίνητα επισημασμένων SBPs και DBPs επισημάνθηκαν ως Korotkoff.
Στη συνέχεια χρησιμοποιήθηκε ένα CNN τριών \textit{επιπέδων}, για την ανάλυση της συνέπειας στα ηχητικά μοτίβα που συσχετίστηκαν με τους ήχους Korotkoff.
Σύμφωνα με τους συγγραφείς, αυτή ήταν η πρώτη μελέτη που διεξήχθη για τέτοιου είδους πρόβλημα, αποδεικνύοντας ότι είναι δύσκολο να προσδιοριστούν οι ήχοι Korotkoff στη συστολή και διαστολή.

\subsubsection{Δεδομένα από φορητές συσκευές}
Οι φορητές συσκευές, οι οποίες επιβάλλουν περιορισμούς στο μέγεθος, την ισχύ και την κατανάλωση μνήμης για τα μοντέλα, έχουν επίσης χρησιμοποιηθεί για τη συλλογή δεδομένων καρδιολογίας για εκπαίδευση μοντέλων βαθιάς μάθησης ανίχνευσης AF\@.

Οι Shashikumar et al.~\cite{shashikumar2017deep} πήραν ECG, φωτοπληθυσμογραφία (PPG) και δεδομένα από επιταχυνσιόμετρο από 98 άτομα χρησιμοποιώντας μια φορητή συσκευή καρπού και παρήξαν την ανάλυση φάσματος χρησιμοποιώντας συνεχή WT\@.
Εκπαίδευσαν ένα CNN πέντε \textit{επιπέδων} σε μια σειρά βραχέων παραθύρων με θόρυβο κίνησης και η έξοδός τους συνδυάστηκε με χαρακτηριστικά που υπολογίστηκαν βάσει της μεταβλητότητας του χτύπου-προς-χτύπο και του δείκτη ποιότητας του σήματος.
Η μέθοδος πέτυχε ακρίβεια 91.8\% στην ανίχνευση AF και σε συνδυασμό με την υπολογιστική αποτελεσματικότητά της είναι πολλά υποσχόμενη για κλινική εφαρμογή σύμφωνα με τους συγγραφείς.
Οι Gotlibovych et al.~\cite{gotlibovych2018end} εκπαίδευσαν ένα δίκτυο CNN ενός \textit{επιπέδου} ακολουθούμενο από ένα LSTM, χρησιμοποιώντας 180 ώρες PPG από δεδομένα φορητών συσκευών για την ανίχνευση AF\@.
Η χρήση του επιπέδου LSTM επιτρέπει στο δίκτυο να μαθαίνει συσχετίσεις μεταβλητού μήκους σε αντίθεση με το σταθερό μήκος του συνελικτικού επιπέδου.
Οι Poh et al.~\cite{poh2018diagnostic} δημιούργησαν μια μεγάλη βάση δεδομένων PPG (πάνω από 180000 σήματα από 3373 ασθενείς), συμπεριλαμβανομένων των δεδομένων από την MIMIC για την ταξινόμηση τεσσάρων ρυθμών.
Ένα πυκνά συνδεδεμένο (densely) CNN με έξι \textit{επίπεδα} χρησιμοποιήθηκε για ταξινόμηση, το οποίο τροφοδοτήθηκε με τμήματα 17 δευτερολέπτων που έχουν αποθορυβοποιηθεί χρησιμοποιώντας ένα ζωνοπερατό φίλτρο.
Τα αποτελέσματα ελήφθησαν χρησιμοποιώντας ένα ανεξάρτητο σύνολο δεδομένων 3039 PPG πετυχαίνοντας καλύτερα αποτελέσματα από τις προηγούμενες μεθόδους που βασίστηκαν σε χειροποίητα χαρακτηριστικά.

Εκτός από την ανίχνευση AF, δεδομένα από φορητές συσκευές χρησιμοποιήθηκαν και για την αναζήτηση καλύτερων προγνωστικών καρδιοαγγειακών παθήσεων.
Στο~\cite{ballinger2018deepheart} οι συγγραφείς εκπαίδευσαν ένα ημι-επιβλεπώμενο, διπλής κατεύθυνσης LSTM σε δεδομένα από 14011 χρήστες της εφαρμογής Cardiogram για την ανίχνευση του διαβήτη, της υψηλής χοληστερόλης, της υψηλής BP και της άπνοιας.
Τα αποτελέσματά τους δείχνουν ότι η ανταπόκριση της καρδιάς στη σωματική δραστηριότητα είναι ένας σημαντικός βιοδείκτης για την πρόβλεψη της εμφάνισης μιας νόσου και μπορεί να εντοπιστεί χρησιμοποιώντας βαθιά μάθηση.

\begin{sidewaystable}
	\centering
	\caption{Εφαρμογές βαθιάς μάθησης με χρήση PCG και άλλων σημάτων}
	\label{table:signals4}
	\begin{tabular}{l c l l}
		\toprule
		\thead{Αναφορά}                             & \thead{Μέθοδος} & \thead{Εφαρμογή/Σημειώσεις\footnote{Σε παρένθεση οι βάσεις δεδομένων που χρησιμοποιήθηκαν. Στον υποπίνακα `Διαγωνισμός PCG/Physionet 2016' όλες οι δημοσιεύσεις χρησιμοποιούν την PHY εκτός του~\cite{chen2017s1}, και στον υποπίνακα `Άλλα Σήματα' όλες οι δημοσιεύσεις χρησιμοποιούν μη-δημόσιες βάσεις δεδομένων εκτός του~\cite{poh2018diagnostic}.}} & \thead{Ακρίβεια\footnote{Υπάρχει μεγάλη μεταβλητότητα στην αναφορά αποτελεσμάτων. Το~\cite{kucharski2017deep} αναφέρουν specificity, το~\cite{pan2017variation} αναφέρει αποτελέσματα για το SBP και το DBP, το~\cite{gotlibovych2018end} αναφέρει ευαισθησία, ειδικότητα, το~\cite{poh2018diagnostic} αναφέρει την τιμή της θετικής προβλεψιμότητας,το~\cite{ballinger2018deepheart} αναφέρει AUC για το diabetest, αποτελέσματα επίσης αναφέρονται για υψηλή χοληστερόλη, άπνοια και υψηλό BP.}} \\
		\midrule
		\multicolumn{4}{l}{\thead{Διαγωνισμός PCG/Physionet 2016}}                                                                                                                                                                                                                                                                                                                                                                                                                                                                                                                                                                                                                                                                                                                                                                                                                                                                                           \\
		\midrule
		Rubin 2017~\cite{rubin2017recognizing}       & CNN             & λογιστική παλινδρόμηση με ένα κρυφό ημι-Markov μοντέλο, MFCCs και ένα CNN δύο \textit{επιπέδων}                                                                                                                                                                                                                                                         & 83.99\%                                                                                                                                                                                                                                                                                                                                                                                                                                                                                                                    \\
		Kucharski 2017~\cite{kucharski2017deep}      & CNN             & ανάλυση φάσματος και ένα CNN πέντε~\textit{επιπέδων} CNN με dropout                                                                                                                                                                                                                                                                                      & 91.6\%\textsuperscript{b}                                                                                                                                                                                                                                                                                                                                                                                                                                                                                             \\
		Dominguez 2018~\cite{dominguez2018deep}      & CNN             & ανάλυση φάσματος και τροποποιημένο AlexNet                                                                                                                                                                                                                                                                                                              & 94.16\%                                                                                                                                                                                                                                                                                                                                                                                                                                                                                                                    \\
		Potes 2016~\cite{potes2016ensemble}          & CNN             & ensemble από Adaboost και CNN, η έξοδος συνδυάζεται με ένα κανόνα απόφασης                                                                                                                                                                                                                                                                               & 86.02\%                                                                                                                                                                                                                                                                                                                                                                                                                                                                                                                    \\
		Ryu 2016~\cite{ryu2016classification}        & CNN             & φίλτρα αποθορυβοποίησης και ένα CNN τεσσάρων \textit{επιπέδων}                                                                                                                                                                                                                                                                                          & 79.5\%                                                                                                                                                                                                                                                                                                                                                                                                                                                                                                                     \\
		Chen 2017~\cite{chen2017s1}                  & DBN             & αναγνώριση S1 και S2 ήχων καρδιάς με χρήση MFCCs, K-means και DBN (μη-δημόσια)                                                                                                                                                                                                                                                                          & 91\%                                                                                                                                                                                                                                                                                                                                                                                                                                                                                                                       \\
		\midrule
		\multicolumn{4}{l}{\thead{Άλλα σήματα}}                                                                                                                                                                                                                                                                                                                                                                                                                                                                                                                                                                                                                                                                                                                                                                                                                                                                                                            \\
		\midrule
		Lee 2017~\cite{lee2017deepa}                 & DBN             & εκτίμηση BP με χρήση bootstrap-aggregation, Monte-Carlo και DBN με παλμομετρικά δεδομένα                                                                                                                                                                                                                                                                & \textit{πολλαπλά}                                                                                                                                                                                                                                                                                                                                                                                                                                                                                                          \\
		Pan 2017~\cite{pan2017variation}             & CNN             & εύρεση ήχων Korotkoff με χρήση ενός CNN τριών \textit{επιπέδων} με παλμομετρικά δεδομένα                                                                                                                                                                                                                                                                & \textit{πολλαπλά}\textsuperscript{b}                                                                                                                                                                                                                                                                                                                                                                                                                                                                                  \\
		Shashikumar 2017~\cite{shashikumar2017deep}  & CNN             & ανίχνευση AF με χρήση ECG, φωτοπληθυσμογραφία, και επιταχυνσιόμετρο με WT και ένα CNN                                                                                                                                                                                                                                                                   & 91.8\%                                                                                                                                                                                                                                                                                                                                                                                                                                                                                                                     \\
		Gotlibovych 2018~\cite{gotlibovych2018end}   & CNN, LSTM       & ανίχνευση AF με χρήση PPG από φορητή συσκευή και ένα LSTM-CNN                                                                                                                                                                                                                                                                                           & \textgreater{99\%}\textsuperscript{b}                                                                                                                                                                                                                                                                                                                                                                                                                                                                                 \\
		Poh 2018~\cite{poh2018diagnostic}            & CNN             & ανίχνευση τεσσάρων ρυθμών με PPG και ένα densely CNN (MIMIC, VORTAL, PPGDB)                                                                                                                                                                                                                                                                             & 87.5\%\textsuperscript{b}                                                                                                                                                                                                                                                                                                                                                                                                                                                                                             \\
		Ballinger 2018~\cite{ballinger2018deepheart} & LSTM            & πρόβλεψη διαβήτη, υψηλής χοληστερόλης, BP και άπνοιας από αισθητήρα                                                                                                                                                                                                                                                     & 0.845\textsuperscript{b}                                                                                                                                                                                                                                                                                                                                                                                                                                                                                                         \\
		\bottomrule
	\end{tabular}
\end{sidewaystable}

\clearpage
\bibliography{chapter3.bib}
\bibliographystyle{unsrt}

%% file: chapter4.tex
\chapter{Βαθιά μάθηση με απεικονίσεις}
\label{chapter4}
\graphicspath{{./images/deep-learning-in-cardiology/}}

\section{Εισαγωγή}
Οι μέθοδοι απεικόνισης που έχουν βρει χρήση στην καρδιολογία περιλαμβάνουν τη τομογραφία μαγνητικού συντονισμού (Magnetic Resonance Imaging, MRI), την Fundus, την ηλεκτρονική τομογραφία (Computerized Tomography, CT), το ηχοκαρδιογράφημα, την τομογραφία οπτικής συνοχής (Optical Coherence Tomography, OCT), το ενδοαγγειακό υπερηχογράφημα (Intravascular Ultrasound, IVUS) και άλλες.
Η βαθιά μάθηση υπήρξε ως επί το πλείστον επιτυχής σε αυτόν τον τομέα, κυρίως λόγω αρχιτεκτονικών που χρησιμοποιούν μεγάλο αριθμό συνελικτικών επιπέδων.
Μια σύνοψη των εφαρμογών βαθιάς μάθησης που χρησιμοποιούν απεικονίσεις παρουσιάζονται στους Πίνακες~\ref{table:imaging1},~\ref{table:imaging2},~\ref{table:imaging3},~\ref{table:imaging4},~\ref{table:imaging5},~\ref{table:imaging6},~\ref{table:imaging7} και~\ref{table:imaging8}.

\begin{sidewaystable}
	\caption{Δημόσιες καρδιολογικές βάσεις δεδομένων, MRI}
	\label{table:cardiologypublicdatabases2}
	\centering
	\begin{tabular}{l c r l}
		\toprule
		\thead{Βάση Δεδομένων} & \thead{Ακρωνύμιο} & \thead{Ασθενείς}    & \thead{Πρόβλημα}                                  \\
		\midrule
		MICCAI 2009 Sunnybrook~\cite{radau2009evaluation}                                                          & SUN09             & 45                  & κατάτμηση LV                                      \\
		MICCAI 2011 Left Ventricle Segmentation STACOM~\cite{fonseca2011cardiac}                                   & STA11             & 200                 & κατάτμηση LV                                      \\
		MICCAI 2012 Right Ventricle Segmentation Challenge~\cite{petitjean2015right}                               & RV12              & 48                  & κατάτμηση RV                                      \\
		MICCAI 2013 SATA~\cite{asman2013miccai}                                                                    & SAT13             & ---\footnote{Ο αριθμός των ασθενών δεν αναφέρεται.} & κατάτμηση LV                                      \\
		MICCAI 2016 HVSMR~\cite{pace2015interactive}                                                               & HVS16             & 20                  & κατάτμηση καρδιάς                                 \\
		MICCAI 2017 ACDC~\cite{bernard2018deep}                                                                    & AC17              & 150                 & κατάτμηση LV/RV                                   \\
		York University Database~\cite{andreopoulos2008efficient}                                                  & YUDB              & 33                  & κατάτμηση LV                                      \\
		Data Science Bowl Cardiac Challenge Data~\cite{dsbcdc2016}                                                 & DS16              & 1140                & εκτίμηση όγκου LV μετά την συστολή και διαστολή \\
		\bottomrule
	\end{tabular}
\end{sidewaystable}

\begin{sidewaystable}
	\caption{Δημόσιες καρδιολογικές βάσεις δεδομένων, Fundus και άλλες}
	\label{table:cardiologypublicdatabases3}
	\centering
	\begin{tabular}{l c r l}
		\toprule
		\thead{Βάση Δεδομένων} & \thead{Ακρωνύμιο} & \thead{Ασθενείς}    & \thead{Πρόβλημα}                                                \\
		\midrule
		\multicolumn{4}{l}{\thead{Βάσεις δεδομένων αμφιβληστροειδούς (όλα Fundus εκτός του~\cite{zhang2016robust})}}                                                                                                           \\
		\midrule
		Digital Retinal Images for Vessel Extraction~\cite{staal2004ridge}                                         & DRIVE             & 40                  & κατάτμηση αγγείων                                               \\
		Structured Analysis of the Retina~\cite{hoover2000locating}                                                & STARE             & 20                  & κατάτμηση αγγείων                                               \\
		Child Heart and Health Study in England Database~\cite{owen2009measuring}                                  & CHDB              & 14                  & κατάτμηση αγγείων                                               \\
		High Resolution Fundus~\cite{odstrcilik2013retinal}                                                        & HRF               & 45                  & κατάτμηση αγγείων                 \\
		Kaggle Retinopathy Detection Challenge 2015~\cite{graham2015kaggle}                                        & KR15              & \_\_\textsuperscript{b} & ταξινόμηση διαβητικής αμφιβληστροειδοπάθειας                    \\
		TeleOptha~\cite{decenciere2013teleophta}                                                                   & e-optha           & 381                 & MA και ανίχνευση αιμορραγίας                                    \\
		Messidor~\cite{decenciere2014feedback}                                                                     & Messidor          & 1200                & διάγνωση διαβητικής αμφιβληστροειδοπάθειας                      \\
		Messidor2~\cite{decenciere2014feedback}                                                                    & Messidor2         & 874                 & διάγνωση διαβητικής αμφιβληστροειδοπάθειας                      \\
		Diaretdb1~\cite{kauppi2013constructing}                                                                    & DIA               & 89                  & MA και ανίχνευση αιμορραγίας                                    \\
		Retinopathy Online Challenge~\cite{niemeijer2010retinopathy}                                               & ROC               & 100                 & ανίχνευση MA                                                    \\
		IOSTAR~\cite{zhang2016robust}                                                                              & IOSTAR            & 30                  & κατάτμηση αγγείων με χρήση SLO                                  \\
		RC-SLO~\cite{zhang2016robust}                                                                              & RC-SLO            & 40                  & κατάτμηση αγγείων με χρήση SLO                                  \\
		\midrule
		\multicolumn{4}{l}{\thead{Άλλες βάσεις δεδομένων απεικόνισης}}                                                                                                                                                        \\
		\midrule
		MICCAI 2011 Lumen+External Elastic Laminae~\cite{balocco2014standardized}                                  & IV11              & 32                  & κατάτμηση περιγράμματος lumen, external IVUS \\
		UK Biobank~\cite{sudlow2015uk}                                                                             & UKBDB             & ---\footnote{Ο αριθμός των ασθενών δεν αναφέρεται.} & πολλαπλές βάσεις δεδομένων απεικόνισης                          \\
		Coronary Artery Stenoses Detection and Quantification~\cite{kiricsli2013standardized}                      & CASDQ             & 48                  & αγγειογραφία CT για στένωση κορωνιαίων αρτηριών        \\
		\midrule
		\multicolumn{4}{l}{\thead{Πολυτροπικές βάσεις δεδομένων}}                                                                                                                                                                      \\
		\midrule
		VORTAL~\cite{charlton2016assessment}                                                                       & VORTAL            & 45                  & εκτίμηση ρυθμού αναπνοής με χρήση ECG και PCG                   \\
		Left Atrium Segmentation Challenge STACOM 2013~\cite{tobon2015benchmark}                                   & STA13             & 30                  & κατάτμηση αριστερού καρδιακού κόλπου με MRI, CT                 \\
		MICCAI MMWHS 2017~\cite{zhuang2016multi}                                                                   & MM17              & 60                  & 120 εικόνες για κατάτμηση της καρδιάς με MRI, CT                \\
		\bottomrule
	\end{tabular}
\end{sidewaystable}

\section{Τομογραφία μαγνητικού συντονισμού}
Η τομογραφία μαγνητικού συντονισμού (MRI) βασίζεται στην αλληλεπίδραση μεταξύ ενός συστήματος ατομικών πυρήνων και ενός εξωτερικού μαγνητικού πεδίου παρέχοντας μια εικόνα του εσωτερικού ενός φυσικού αντικειμένου~\cite{sebastiani1991mathematical}.
Οι κύριες χρήσεις της MRI περιλαμβάνουν την κατάτμηση της αριστερής κοιλίας (Left Ventricle, LV), της δεξιάς κοιλίας (Right Ventricle, RV) και ολόκληρης της καρδιάς.

\subsection{Κατάτμηση LV}
Τα CNNs χρησιμοποιήθηκαν για κατάτμηση LV με MRI\@.
Οι Tan et al.~\cite{tan2016cardiac} χρησιμοποίησαν ένα CNN για να εντοπίσουν το ενδοκάρδιο του LV και ένα επιπλέον CNN για να προσδιορίσουν την ακτίνα του ενδοκαρδίου, χρησιμοποιώντας τις STA11 και SUN09 για εκπαίδευση και αξιολόγηση αντίστοιχα.
Χωρίς φιλτράρισμα των εικόνων που απεικονίζουν τα apical ή την χρήση παραμορφώσιμων μοντέλων επιτυγχάνουν συγκρίσιμες επιδόσεις με προηγούμενες μεθόδους.
Στο~\cite{romaguera2017left} οι συγγραφείς εκπαίδευσαν ένα CNN πέντε \textit{επιπέδων} χρησιμοποιώντας MRI από την SUN09.
Εκπαίδευσαν το μοντέλο τους χρησιμοποιώντας SGD και RMSprop με το πρώτο να φτάνει Dice 92\%.

\begin{sidewaystable}
	\caption{Εφαρμογές βαθιάς μάθησης με χρήση MRI, για κατάτμηση LV}
	\label{table:imaging1}
	\centering
	\begin{tabular}{l c l l}
		\toprule
		\thead{Αναφορά}                            & \thead{Μέθοδος} & \thead{Εφαρμογή/Σημειώσεις\footnote{Σε παρένθεση οι βάσεις δεδομένων που χρησιμοποιήθηκαν.}}               & \thead{Dice\footnote{το ($\S$) υποδηλώνει `για κάθε βάση', το ($*$) υποδηλώνει μέσο τετραγωνικό σφάλμα για EF το ($+$) υποδηλώνει `για ενδοκαρδιακά και επικαρδιακά', το ($-$) υποδηλώνει ακρίβεια, και το ($\#$) υποδηλώνει `για CT και MRI'}} \\
		\midrule
		Tan 2016~\cite{tan2016cardiac}              & CNN             & CNN για εύρεση τοποθεσίας και CNN για ευθυγράμμιση ενδοκαρδιακών συνόρων (SUN09, STA11)                & 88\%                                                                                                                                                                                                                                            \\
		Romaguera 2017~\cite{romaguera2017left}     & CNN             & CNN πέντε \textit{επιπέδων} με SGD και RMSprop (SUN09)                                                     & 92\%                                                                                                                                                                                                                                            \\
		Poudel 2016~\cite{poudel2016recurrent}      & u-net, RNN      & συνδυασμός u-net και RNN (SUN09, μη-δημόσια)                                                               & \textit{πολλαπλά}                                                                                                                                                                                                                               \\
		Rupprecht 2016~\cite{rupprecht2016deep}     & CNN             & συνδυασμός ενός CNN τεσσάρων \textit{επιπέδων} με Sobolev (STA11, non-medical)                             & 85\%                                                                                                                                                                                                                                            \\
		Ngo 2014~\cite{anh2014fully}                & DBN             & συνδυασμός DBN με level set (SUN09)                                                                        & 88\%                                                                                                                                                                                                                                            \\
		Avendi 2016~\cite{avendi2016combined}       & CNN, AE         & CNN για εντοπισμό καρδιακού θαλάμου, AEs για συμπερασμό σχήματος (SUN09) & 96.69\%                                                                                                                                                                                                                                         \\
		Yang 2016~\cite{yang2016deep}               & CNN             & δίκτυο εξαγωγής χαρακτηριστικών και ένα μη-τοπικό δίκτυο σύντηξης επισημάνσεων (SAT13)                     & 81.6\%                                                                                                                                                                                                                                          \\
		Luo 2016~\cite{luo2016cardiac}              & CNN             & άτλας αντιστοίχισης του LV και ένα CNN τριών \textit{επιπέδων} (DS16)                                      & 4.98\%$^*$                                                                                                                                                                                                                                      \\
		Yang 2017~\cite{yang2017deep}               & CNN, u-net      & εντοπισμός με CNN παλινδρόμησης και κατάτμηση με u-net (YUDB, SUN09)                                       & 91\%, 93\%$^\S$                                                                                                                                                                                                                                 \\
		Tan 2017~\cite{tan2017convolutional}        & CNN             & CNN παλινδρόμησης (STA11, DS16)                                                                            & \textit{πολλαπλά}                                                                                                                                                                                                                               \\
		Curiale 2017~\cite{curiale2017automatic}    & u-net           & residual u-net (SUN09)                                                                                     & 90\%                                                                                                                                                                                                                                            \\
		Liao 2017~\cite{liao2017estimation}         & CNN             & τοπικό δυαδικό μοτίβο για εντοπισμό και FCN για κατάτμηση (DS16)                                           & 4.69\%$^*$                                                                                                                                                                                                                                      \\
		Emad 2015~\cite{emad2015automatic}          & CNN             & εντοπισμός LV με χρήση CNN και πυραμίδες κλιμάκων (YUDB)                                                   & 98.66\%$^-$                                                                                                                                                                                                                                     \\
		\bottomrule
	\end{tabular}
\end{sidewaystable}

Χρησιμοποιήθηκαν επίσης συνδυασμοί CNN με RNNs.
Στο~\cite{poudel2016recurrent} οι συγγραφείς δημιούργησαν ένα recurrent u-net που μαθαίνει αναπαραστάσεις από μια στοίβα από 2D και έχει την ικανότητα να αξιοποιεί τις χωρικές εξαρτήσεις μεταξύ των τμημάτων μέσω εσωτερικών μονάδων μνήμης.
Συνδυάζει ανίχνευση ανατομίας και κατάτμηση σε μια ενιαία αρχιτεκτονική από-άκρο-σε-άκρο, επιτυγχάνοντας συγκρίσιμα αποτελέσματα με άλλες μεθόδους, ξεπερνώντας τις βάσεις αναφοράς για το DBN, τα recurrent DBN και FCN όσον αφορά το Dice.

Άλλες δημοσιεύσεις συνδυάζουν μεθόδους βαθιάς μάθησης με το level-set για την κατάτμηση της LV\@.
Οι Rupprecht et al.~\cite{rupprecht2016deep} εκπαίδευσαν ένα CNN τεσσάρων \textit{επιπέδων}, που προβλέπει ένα διάνυσμα που υποδεικνύει το σημείο στο εξελισσόμενο περίγραμμα προς το πλησιέστερο σημείο στο όριο του αντικειμένου ενδιαφέροντος.
Αυτές οι προβλέψεις σχημάτισαν ένα διανυσματικό πεδίο το οποίο στη συνέχεια χρησιμοποιήθηκε για την εξέλιξη του περιγράμματος, χρησιμοποιώντας το πλαίσιο ενεργού περιγράμματος Sobolev.
Οι Anh et al.~\cite{anh2014fully} δημιούργησαν μια μέθοδο μη-άκαμπτης κατάτμησης βασισμένη στη level-set μέθοδο ρυθμισμένη ανάλογα με την απόσταση, που αρχικοποιήθηκε από τα αποτελέσματα μιας δομημένης εξαγωγής από ένα DBN\@.
Οι Avendi et al.~\cite{avendi2016combined} χρησιμοποίησαν ένα CNN για να ανιχνεύσουν τον LV θάλαμο και στη συνέχεια χρησιμοποίησαν στοιβαγμένα AE για να συμπεράνουν το σχήμα της LV\@.
Στη συνέχεια το αποτέλεσμα ενσωματώθηκε σε παραμορφώσιμα μοντέλα για να βελτιωθεί η ακρίβεια και η ευρωστία της κατάτμησης.

Μέθοδοι που βασίζονται σε άτλαντες έχουν επίσης χρησιμοποιηθεί για την επίλυση αυτού του προβλήματος.
Οι Yang et al.~\cite{yang2016deep} δημιούργησαν ένα δίκτυο βαθιάς σύντηξης συνδυάζοντας ένα δίκτυο εξαγωγής χαρακτηριστικών και ένα μη-τοπικό δίκτυο σύντηξης επισημάνσεων βασισμένο σε patch.
Τα χαρακτηριστικά που δημιουργούνται κατά τη διάρκεια της μάθησης χρησιμοποιούνται περαιτέρω για τον ορισμό ενός μέτρου ομοιότητας για την επιλογή άτλα MRI\@.
Σύγκριναν τη μέθοδος τους με την ψηφοφορία με πλειοψηφία, τη σύντηξη ετικετών με βάση τα patch, την αντιστοίχιση patch πολλαπλών ατλάντων και το SVM με επαυξημένα χαρακτηριστικά επιτυγχάνοντας καλύτερα αποτελέσματα όσον αφορά την ακρίβεια.
Οι Luo et al.~\cite{luo2016cardiac} υιοθέτησαν μια μέθοδο χαρτογράφησης άτλα LV για να επιτευχθεί ακριβής εντοπισμός με χρήση δεδομένων MRI από το DS16.
Στη συνέχεια, ένα CNN τριών \textit{επιπέδων} εκπαιδεύτηκε για την πρόβλεψη του όγκου LV, πετυχαίνοντας συγκρίσιμα αποτελέσματα με τους νικητές του διαγωνισμού ACDC 2017 με βάση αναφοράς την μέση τετραγωνική ρίζα της τελικής διαστολής και των τελικών συστολικών όγκων.

Μέθοδοι παλινδρόμησης έχουν χρησιμοποιηθεί για τον εντοπισμό της LV πριν την κατάτμηση της.
Οι Yang et al.~\cite{yang2017deep} πρώτα εντόπισαν το LV σε ολόκληρη την εικόνα χρησιμοποιώντας CNN παλινδρόμησης και στη συνέχεια το ταξινόμησαν μέσα στην περιοχή ενδιαφέροντος (Region of Interest, ROI), χρησιμοποιώντας μια αρχιτεκτονική που βασίζεται στο u-net.
Το μοντέλο τους επιτυγχάνει υψηλή ακρίβεια με καλές υπολογιστικές επιδόσεις κατά τη διάρκεια των συμπερασμών.

Επίσης χρησιμοποιήθηκαν διάφορες άλλες μέθοδοι για την κατάτμηση του LV\@.
Οι Tan et al.~\cite{tan2017convolutional} παραμετροποίησαν όλες τις εικόνες του προβλήματος της τμηματοποίησης του LV με βάση τις ακτινικές αποστάσεις μεταξύ του κεντρικού σημείου LV και των ενδοκαρδιακών και επικαρδιακών περιγραμμάτων στις πολικές συντεταγμένες.
Στη συνέχεια, εκπαίδευσαν ένα CNN παλινδρόμησης στην STA11 για να συμπεραίνουν αυτές τις παραμέτρους και δοκίμασαν τη γενικευσιμότητα της μεθόδου στην DS16 παρουσιάζοντας καλά αποτελέσματα.
Στο~\cite{curiale2017automatic} οι συγγραφείς χρησιμοποίησαν την απόσταση Jaccard ως τη αντικειμενική συνάρτηση βελτιστοποίησης, ενσωματώνοντας μια residual στρατηγική μάθησης και εισάγοντας ένα επίπεδο κανονικοποίησης παρτίδας (batch normalization) για την εκπαίδευση ενός u-net.
Αποδείχθηκε ότι αυτή η διαμόρφωση είχε καλύτερα αποτελέσματα από άλλα απλά u-nets σε σχέση με το δείκτη Dice.
Στο άρθρο τους οι Liao et al.~\cite{liao2017estimation} ανίχνευσαν το ROI που περιείχε LV και έπειτα χρησιμοποίησαν FCN για να ταξινομήσουν τα LV μέσα στο ROI\@.
Τα αποτελέσματα της 2D τμηματοποίησης ενσωματώθηκαν μεταξύ διαφορετικών εικόνων για την εκτίμηση του όγκου.
Το μοντέλο εκπαιδεύτηκε εναλλάξ στην κατάτμηση του LV και στην εκτίμηση του όγκου, τοποθετώντας το τέταρτο στον διαγωνισμό του DS16.
Οι Emad et al.~\cite{emad2015automatic} εντόπισαν το LV χρησιμοποιώντας ένα CNN και μια ανάλυση πυραμίδας κλιμάκων, για να λάβουν υπόψη διαφορετικά μεγέθη της καρδιάς με την YUDB\@.
Πέτυχαν καλά αποτελέσματα, αλλά με σημαντικό κόστος υπολογισμού (10 δευτερόλεπτα ανά εικόνα κατά τη διάρκεια συμπερασμών).

\subsection{Κατάτμηση LV/RV}
Μια βάση δεδομένων που χρησιμοποιήθηκε για την κατάτμηση LV/RV ήταν η MICCAI 2017 ACDC Challenge (AC17), που περιέχει εικόνες MRI από 150 ασθενείς χωρισμένες σε πέντε ομάδες (φυσιολογικοί, προηγούμενο MI, διαστολή της καρδιομυοπάθειας, υπερτροφική καρδιομυοπάθεια, μη-φυσιολογική RV).
Οι Zotti et al.~\cite{zotti2017gridnet} χρησιμοποίησαν ένα μοντέλο που περιλαμβάνει μια καρδιακή μονάδα παλινδρόμησης κέντρου μάζας, που επιτρέπει την καταγραφή του σχήματος εκ των προτέρων και μια συνάρτηση απώλειας προσαρμοσμένη στην καρδιακή ανατομία.
Τα χαρακτηριστικά δημιουργούνται με μια αρχιτεκτονική μετατροπής πολλαπλών αναλυτικοτήτων conv-deconv `πλέγματος', η οποία αποτελεί επέκταση του u-net.
Αυτό το μοντέλο σε σύγκριση με το απλό conv-deconv και το u-net, εμφανίζει καλύτερα αποτελέσματα κατά μέσο όρο 5\% όσον αφορά τον δείκτη Dice.
Οι Patravali et al.~\cite{patravali20172d} εκπαίδευσαν ένα μοντέλο βασισμένο στο u-net, χρησιμοποιώντας το Dice σε συνδυασμό με το cross-entropy ως μέτρηση για την τμηματοποίηση των LV/RV και του μυοκαρδίου.
Το μοντέλο σχεδιάστηκε για να δέχεται μια στοίβα εικόνων ως κανάλια εισόδου, ενώ η έξοδος προβλέπει τη μεσαία εικόνα.
Με βάση τα πειράματα που διεξήγαγαν, βγήκε το συμπέρασμα ότι τρεις εικόνες ήταν βέλτιστες ως είσοδος για το μοντέλο αντί για μια ή πέντε.
Οι Isensee et al.~\cite{isensee2017automatic} χρησιμοποίησαν ένα σύνολο 2D και 3D u-net για κατάτμηση των LV/RV και του μυοκαρδίου του LV σε κάθε επανάληψη του καρδιακού κύκλου.
Πληροφορίες εξήχθησαν από την κατατμημένη χρονοσειρά με τη μορφή χαρακτηριστικών, που αντικατοπτρίζουν τις διαγνωστικές κλινικές διαδικασίες του σκοπού της ταξινόμησης.
Με βάση αυτά τα χαρακτηριστικά, εκπαίδευσαν ένα ensemble perceptrons πολλαπλών~\textit{επιπέδων} και έναν ταξινομητή RF για την πρόβλεψη της παθολογικής κατηγορίας.
Το μοντέλο τους κατέλαβε την πρώτη θέση στον διαγωνισμό του ACDC\@.

\begin{sidewaystable}
	\caption{Εφαρμογές βαθιάς μάθησης με χρήση MRI, για κατάτμηση LV/RV}
	\label{table:imaging2}
	\centering
	\begin{tabular}{l c l l}
		\toprule
		\thead{Αναφορά}                            & \thead{Μέθοδος} & \thead{Εφαρμογή/Σημειώσεις\footnote{Σε παρένθεση οι βάσεις δεδομένων που χρησιμοποιήθηκαν.}}               & \thead{Dice\footnote{το ($\S$) υποδηλώνει `για κάθε βάση', το ($*$) υποδηλώνει μέσο τετραγωνικό σφάλμα για EF το ($+$) υποδηλώνει `για ενδοκαρδιακά και επικαρδιακά', το ($-$) υποδηλώνει ακρίβεια, και το ($\#$) υποδηλώνει `για CT και MRI'}} \\
		\midrule
		Zotti 2017~\cite{zotti2017gridnet}          & u-net           & παραλλαγή του u-net με πολυ-κλιμακωτό conv-deconv αρχιτεκτονική πλέγματος (AC17)                           & 90\%                                                                                                                                                                                                                                            \\
		Patravali 2017~\cite{patravali20172d}       & u-net           & 2D/3D u-net (AC17)                                                                                         & \textit{πολλαπλά}                                                                                                                                                                                                                               \\
		Isensee 2017~\cite{isensee2017automatic}    & u-net           & ensemble u-net, με συστηματοποίηση πολυ-επιπέδων perceptron και ταξινομητή RF (AC17)                 & \textit{πολλαπλά}                                                                                                                                                                                                                               \\
		Tran 2016~\cite{tran2016fully}              & CNN             & FCN τεσσάρων \textit{επιπέδων} (SUN09, STA11)                                                              & 92\%, 96\%$^+$                                                                                                                                                                                                                                  \\
		Bai 2017~\cite{bai2017semi}                 & CNN             & VGGnet16 και DeepLab με χρήση CRF και πιο λεπτομερή αποτελέσματα (UKBDB)                                  & 90.3\%                                                                                                                                                                                                                                          \\
		Lieman 2017~\cite{lieman2017fastventricle}  & u-net           & επέκταση του ENet~\cite{paszke2016enet} με skip-connections (μη-δημόσια)                                    & \textit{πολλαπλά}                                                                                                                                                                                                                               \\
		Winther 2017~\cite{winther2017nu}           & u-net           & $\nu$-net παραλλαγή του u-net (DS16, SUN09, RV12, μη-δημόσια)                                              & \textit{πολλαπλά}                                                                                                                                                                                                                               \\
		Du 2018~\cite{du2018deep}                   & DBN             & χαρακτηριστικά DAISY και DBN παλινδρόμησης με χρήση 2900 εικόνων (μη-δημόσια)                              & 91.6\%, 94.1\%$^+$                                                                                                                                                                                                                              \\
		Giannakidis 2016~\cite{giannakidis2016fast} & CNN             & κατάτμηση RV με χρήση 3D πολυ-κλιμακωτών CNN με δύο διαδρομές (μη-δημόσια)                                 & 82.81\%                                                                                                                                                                                                                                         \\
		\bottomrule
	\end{tabular}
\end{sidewaystable}

Διάφορες άλλες βάσεις δεδομένων έχουν επίσης χρησιμοποιηθεί για την επίλυση της κατάτμησης LV/RV με CNNs.
Στο~\cite{bai2017semi} οι συγγραφείς δημιούργησαν μια μέθοδο ημι-επιβλεπώμενης μάθησης, στην οποία το δίκτυο κατάτμησης για το LV/RV και το μυοκάρδιο εκπαιδεύτηκε από τα επισημασμένα και μη-επισημασμένα δεδομένα.
Η αρχιτεκτονική του δικτύου βασίστηκε στο VGGnet16, παρόμοια με την αρχιτεκτονική του DeepLab~\cite{chen2018deeplab}, ενώ η τελική κατάτμηση βελτιώθηκε με τη χρήση ενός υπό όρους τυχαίου πεδίου (Conditional Random Field, CRF).
Οι συγγραφείς καταδεικνύουν ότι η εισαγωγή μη-επισημασμένων δεδομένων, βελτιώνει την απόδοση τμηματοποίησης όταν τα δεδομένα εκπαίδευσης είναι λίγα.
Στο~\cite{giannakidis2016fast} οι συγγραφείς υιοθετούν ένα 3D πολλαπλών κλιμάκων CNN για τον εντοπισμό των pixel που ανήκουν στον RV\@.
Το δίκτυο έχει δύο συνελικτικές διαδρομές με τις εισόδους του κεντροθετημένες στην ίδια θέση εικόνας, ενώ το δεύτερο τμήμα εξάγεται από μια εκδοχή της εικόνας που έχει υποβληθεί σε δειγματοληψία.
Τα αποτελέσματα που προέκυψαν ήταν καλύτερα από προηγούμενες μεθόδους, παρόλο που τα τελευταία βασίστηκαν σε χειροποίητα χαρακτηριστικά και εκπαιδεύτηκαν σε λιγότερο μεταβλητές βάσεις δεδομένων.

Τα FCNs έχουν επίσης χρησιμοποιηθεί για κατάτμηση LV/RV\@.
Στο άρθρο τους, οι Tran et al.~\cite{tran2016fully} εκπαίδευσαν ένα μοντέλο FCN τεσσάρων \textit{επιπέδων} για την κατάτμηση των LV/RV με τις SUN09, STA11.
Σύγκριναν προηγούμενες μεθόδους μαζί με δύο αρχικοποιήσεις του μοντέλου τους: μία fine-tuned έκδοση του μοντέλου τους χρησιμοποιώντας το STA11 και μια αρχικοποιημένη με Xavier, με το τελευταίο να έχει τις καλύτερες επιδόσεις σε σχεδόν όλα τα προβλήματα.

Τα FCN με skip-connections και u-net έχουν επίσης χρησιμοποιηθεί για την επίλυση αυτού του προβλήματος.
Οι Lieman et al.~\cite{lieman2017fastventricle} δημιούργησαν μια αρχιτεκτονική FCN με skip-connections με το όνομα FastVentricle βασισμένη στο ENet~\cite{paszke2016enet}, η οποία είναι ταχύτερη και λειτουργεί με λιγότερη μνήμη από τις προηγούμενες αρχιτεκτονικές κοιλιακής κατάτμησης επιτυγχάνοντας υψηλή κλινική ακρίβεια.
Στο~\cite{winther2017nu} οι συγγραφείς εισάγουν το $\nu$-net το οποίο είναι μια παραλλαγή u-net για την κατάτμηση του ενδοκαρδίου και του επικαρδίου LV/RV με τη χρήση των DS16, SUN09 και RV12.
Αυτή η μέθοδος απέδωσε καλύτερα από τον ειδικό καρδιολόγο σε αυτή τη μελέτη, ειδικά για την κατάτμηση του RV\@.

Ορισμένες άλλες μέθοδοι βασίστηκαν σε μοντέλα παλινδρόμησης.
Στο άρθρο τους οι Du et al.~\cite{du2018deep} δημιούργησαν ένα πλαίσιο κατάτμησης παλινδρόμησης για να οριοθετήσουν το LV/RV\@.
Πρώτον, εξάγονται χαρακτηριστικά DAISY και στη συνέχεια χρησιμοποιήθηκε μια μέθοδος αναπαράστασης βάσει σημείων, για την απεικόνιση των ορίων.
Τέλος, τα χαρακτηριστικά DAISY χρησιμοποιήθηκαν ως είσοδος και τα σημεία ορίων ως ετικέτες για την εκπαίδευση του μοντέλου παλινδρόμησης με βάση το DBN\@.
Η απόδοση του μοντέλου αξιολογείται χρησιμοποιώντας διαφορετικά χαρακτηριστικά από το DAISY (GIST, ιστόγραμμα πυραμίδων προσανατολισμένων διαβαθμίσεων) και επίσης συγκρίνεται με την παλινδρόμηση διανυσμάτων υποστήριξης (Support Vector Regression, SVR) και άλλες παραδοσιακές μεθόδους (γραφήματα, ενεργά περιγράμματα, level set), επιτυγχάνοντας καλύτερα αποτελέσματα.

\subsection{Κατάτμηση της καρδιάς}
Το MICCAI 2016 HVSMR (HVS16) χρησιμοποιήθηκε για την κατάτμηση της καρδιάς η οποία περιέχει εικόνες MRI από 20 ασθενείς.
Οι Wolterink et al.~\cite{wolterink2016dilated} εκπαίδευσαν ένα CNN δέκα \textit{επιπέδων} με αυξανόμενα επίπεδα διαστολής για την κατάτμηση του μυοκαρδίου και του αίματος στις αξονικές, σαγματοειδής και στεφανιαίες εικόνες.
Επίσης, χρησιμοποιούν βαθιά επίβλεψη~\cite{lee2015deeply} για να επιλύσουν το πρόβλημα του vanishing gradients και να βελτιώσουν την αποτελεσματικότητα της εκπαίδευσης του δικτύου τους χρησιμοποιώντας ένα μικρό σύνολο δεδομένων.
Τα πειράματα που πραγματοποίησαν με και χωρίς διαστολές (dilations) σε αυτήν την αρχιτεκτονική έδειξαν τη χρησιμότητα της.
Στο άρθρο τους οι Li et al.~\cite{li2016automatic} ξεκίνησαν με ένα 3D FCN με επισήμανση των voxel και στη συνέχεια εισήγαγαν διαστελλόμενα συνελικτικά επίπεδα στο βασικό μοντέλο για να επεκτείνουν το δεκτικό πεδίο.
Έπειτα χρησιμοποιούν μονοπάτια βαθιά επίβλεψης, για την επιταχύνουν την εκπαίδευση και την αξιοποίηση πληροφοριών πολλαπλών κλιμάκων.
Σύμφωνα με τους συγγραφείς το μοντέλο παρουσιάζει καλή ακρίβεια κατάτμησης, σε συνδυασμό με χαμηλό υπολογιστικό κόστος.
Οι Yu et al.~\cite{yu20163d} δημιούργησαν ένα φράκταλ δίκτυο 3D FCN για κατάτμηση της καρδιάς και των μεγάλων αγγείων.
Εφαρμόζοντας αναδρομικά έναν κανόνα απλής επέκτασης, κατασκευάζουν το φράκταλ δίκτυο συνδυάζοντας ιεραρχικές ενδείξεις για ακριβή κατάτμηση.
Επιτυγχάνουν επίσης καλά αποτελέσματα με χαμηλό υπολογιστικό κόστος (12 δευτερόλεπτα ανά όγκο).

Μια άλλη βάση δεδομένων που χρησιμοποιήθηκε για την κατάτμηση της καρδιάς ήταν η MM17 η οποία περιέχει 120 πολυτροπικές εικόνες από καρδιακή MRI/CT\@.
Η μέθοδος των Payer et al.~\cite{payer2017multi} βασίζεται σε δύο FCN για τον εντοπισμό και την κατάτμηση της καρδιάς.
Αρχικά, το CNN εντοπισμού βρίσκει το κέντρο του πλαισίου οριοθέτησης γύρω από όλες τις δομές της καρδιάς, έτσι ώστε το CNN κατάτμησης να μπορεί να επικεντρωθεί σε αυτήν την περιοχή.
Έπειτα, το CNN κατάτμησης μετατρέπει τις προβλέψεις ενδιάμεσων ετικετών σε θέσεις άλλων ετικετών.
Επομένως το δίκτυο μαθαίνει από τις σχετικές θέσεις μεταξύ των επισημάνσεων και επικεντρώνεται στις ανατομικά εφικτές διαμορφώσεις.
Το μοντέλο συγκρίθηκε με το u-net επιτυγχάνοντας καλύτερα αποτελέσματα, ειδικά στο σύνολο δεδομένων του MRI\@.
Οι Mortazi et al.~\cite{mortazi2017multi} εκπαίδευσαν ένα πολυεπίπεδο CNN με μια προσαρμοστική στρατηγική σύντηξης, για την κατάτμηση επτά περιοχών της καρδιάς.
Σχεδίασαν τρία CNN (ένα για κάθε κάθετο επίπεδο) με την ίδια αρχιτεκτονική και τα εκπαίδευσαν για επισήμανση των voxel.
Από τα πειράματά τους καταλήγουν στο συμπέρασμα ότι το μοντέλο τους οριοθετεί τις καρδιακές δομές με υψηλή ακρίβεια και αποτελεσματικά.
Στο~\cite{yang2017hybrid} οι συγγραφείς χρησιμοποίησαν ένα FCN, το οποίο συνδύασαν με 3D τελεστές, μεταφορά μάθησης και έναν μηχανισμό βαθιάς επίβλεψης για την απόσταξη 3D συμφραζομένων πληροφοριών και την επίλυση πιθανών δυσκολιών στην εκπαίδευση.
Χρησιμοποιήθηκε υβριδική απώλεια που καθοδηγεί τη διαδικασία εκπαίδευσης για την εξισορρόπηση των κατηγοριών και διατηρεί τις λεπτομέρειες των ορίων.
Σύμφωνα με τα πειράματά τους, η χρήση της υβριδικής απώλειας επιτυγχάνει καλύτερα αποτελέσματα από το Dice μέτρο.

\subsection{Άλλες εφαρμογές}
Μέθοδοι βαθιάς μάθησης έχουν επίσης χρησιμοποιηθεί και για ανίχνευση άλλων καρδιακών δομών με MRI\@.
Οι Yang et al.~\cite{yang2017segmenting} δημιούργησαν μια μέθοδο διάδοσης πολλαπλών ατλάντων, για να ενσωματώσουν την ανατομική δομή του μυοκαρδίου του αριστερού κόλπου και των πνευμονικών φλεβών.
Αυτό ακολουθήθηκε από ένα μη-επιτηρούμενο εκπαιδευμένο SSAE με ένα softmax για την κατάτμηση της κολπικής ίνωσης, χρησιμοποιώντας 20 εικόνες από ασθενείς με AF\@.
Στο άρθρο τους οι Zhang et al.~\cite{zhang2016automated} προσπάθησαν να ανιχνεύσουν εικόνες έλλειψης apical και basal.
Ελέγχουν την παρουσία τυπικών basal και apical μοτίβων στις τελευταίες και πρώτες εικόνες της βάσης δεδομένων και εκπαιδεύουν δύο CNN για να κατασκευάσουν ένα σύνολο διακριτικών χαρακτηριστικών.
Τα πειράματά τους έδειξαν ότι το μοντέλο με τέσσερα \textit{επίπεδα}, έχει καλύτερη απόδοση από τις SAE και τις Βαθιές Μηχανές Boltzmann.

\begin{sidewaystable}
	\caption{Εφαρμογές βαθιάς μάθησης με χρήση MRI, για κατάτμηση της καρδιάς και άλλα}
	\label{table:imaging3}
	\centering
	\begin{tabular}{l c l l}
		\toprule
		\thead{Αναφορά}                           & \thead{Μέθοδος} & \thead{Εφαρμογή/Σημειώσεις\footnote{Σε παρένθεση οι βάσεις δεδομένων που χρησιμοποιήθηκαν.}}    & \thead{Dice\footnote{το ($\S$) υποδηλώνει `για κάθε βάση', το ($*$) υποδηλώνει μέσο τετραγωνικό σφάλμα για EF το ($+$) υποδηλώνει `για ενδοκαρδιακά και επικαρδιακά', το ($-$) υποδηλώνει ακρίβεια, και το ($\#$) υποδηλώνει `για CT και MRI'}} \\
		\midrule
		\multicolumn{4}{l}{\thead{Τμηματοποίηση όλης της καρδιάς}}                                                                                                                                                                                                                                                                                                                                                      \\
		\midrule
		Wolterink 2016~\cite{wolterink2016dilated} & CNN             & διαστελλόμενα CNN με ορθογώνια patches (HVS16)                                                  & 80\%, 93\%                                                                                                                                                                                                                                      \\
		Li 2016~\cite{li2016automatic}             & CNN             & βαθιά επιβλεπώμενο 3D FCN με διαστολές (HVS16)                                                  & 69.5\%                                                                                                                                                                                                                                          \\
		Yu 2017~\cite{yu20163d}                    & CNN             & βαθιά επιβλεπώμενο 3D FCN κατασκευασμένο με αυτο-επιβλεπώμενο φρακταλ τρόπο (HVS16)             & \textit{πολλαπλά}                                                                                                                                                                                                                               \\
		Payer~\cite{payer2017multi}                & CNN             & δύο ξεχωριστά FCN για εντοπισμό και κατάτμηση (MM17)                                            & 90.7\%, 87\%$^\#$                                                                                                                                                                                                                               \\
		Mortazi 2017~\cite{mortazi2017multi}       & CNN             & πολυ-επίπεδο FCN (MM17)                                                                         & 90\%, 85\%$^\#$                                                                                                                                                                                                                                 \\
		Yang~\cite{yang2017hybrid}                 & CNN             & βαθιά επιβλεπώμενο 3D FCN εκπαιδευμένο με μεταφορά μάθησης (MM17)                               & 84.3\%, 77.8\%$^\#$                                                                                                                                                                                                                             \\
		\midrule
		\multicolumn{4}{l}{\thead{Άλλες εφαρμογές}}                                                                                                                                                                                                                                                                                                                                                                     \\
		\midrule
		Yang 2017~\cite{yang2017segmenting}        & SSAE            & κατάτμηση κολπικής ίνωσης με διάδοση πολλαπλών ατλάντων, SSAE, softmax (μη-δημόσια)      & 82\%                                                                                                                                                                                                                                            \\
		Zhang 2016~\cite{zhang2016automated}       & CNN             & ανίχνευση έλλειψης apical και basal με δύο CNNs τεσσάρων \textit{επιπέδων} (UKBDB)              & \textit{πολλαπλά}                                                                                                                                                                                                                               \\
		Kong 2016~\cite{kong2016recognizing}       & CNN, RNN        & CNN για χωρική πληροφορία και RNN για χρονική πληροφορία                                        & \textit{πολλαπλά}                                                                                                                                                                                                                               \\
		Yang 2017~\cite{yang2017convolutional}     & CNN             & CNN για την ανίχνευση του τέλους της διαστολής και συστολής απο το LV (STA11, μη-δημόσια)       & 76.5\%$^-$                                                                                                                                                                                                                                      \\
		Xu 2017~\cite{xu2017direct}                & Multiple        & ανίχνευση MI με χρήση Fast R-CNN για εντοπισμό της καρδιάς, LSTM και SAE (μη-δημόσια)           & 94.3\%$^-$                                                                                                                                                                                                                                      \\
		Xue 2018~\cite{xue2018full}                & CNN, LSTM       & CNN, δύο παράλληλα LSTMs και Bayesian μοντέλο για ποσοτικοποίηση του LV (μη-δημόσια) & \textit{πολλαπλά}                                                                                                                                                                                                                               \\
		Zhen 2016~\cite{zhen2016multi}             & RBM             & πολυ-κλιμακωτό συνελικτικό RBM και RF για δι-κολπική εκτίμηση του όγκου (μη-δημόσια)            & 3.87\%$^*$                                                                                                                                                                                                                                      \\
		Biffi 2016~\cite{biffi2018learning}        & CNN             & ανίχνευση υπερτροφικής καρδιομυοπάθειας με χρήση μεταβολικών AE (AC17, μη-δημόσια)              & 90\%$^-$                                                                                                                                                                                                                                        \\
		Oktay 2016~\cite{oktay2016multi}           & CNN             & αύξηση αναλυτικότητας εικόνας με residual CNN (μη-δημόσια)                                      & \textit{πολλαπλά}                                                                                                                                                                                                                               \\
		\bottomrule
	\end{tabular}
\end{sidewaystable}

Άλλα ιατρικά προβλήματα με MRI μελετήθηκαν επίσης, όπως η ανίχνευση εικόνων ακραίας συστολής και διαστολής.
Οι Kong et al.~\cite{kong2016recognizing} δημιούργησαν ένα χρονικό δίκτυο παλινδρόμησης που είχε προεκπαιδευτεί στο ImageNet με την ενσωμάτωση ενός CNN με ένα RNN, για να προσδιορίσει τις εικόνες της τελικής διαστολής και της τελικής συστολής από τις ακολουθίες MRI\@.
Το CNN κωδικοποιεί τις χωρικές πληροφορίες μίας καρδιακής αλληλουχίας ενώ το RNN αποκωδικοποιεί τις χρονικές.
Επίσης σχεδίασαν μια συνάρτηση απώλειας για να περιορίσουν τη δομή των προβλεπόμενων επισημάνσεων.
Το μοντέλο τους επιτυγχάνει καλύτερη μέση διαφορά εικόνων από τις προηγούμενες μεθόδους.
Στο άρθρο τους οι Yang et al.~\cite{yang2017convolutional} χρησιμοποίησαν ένα CNN για να ανιχνεύσουν τις εικόνες της τελικής διαστολής και της τελικής συστολής από την LV, επιτυγχάνοντας ακρίβεια 76.5\%.

Δημιουργήθηκαν επίσης μέθοδοι ποσοτικοποίησης διαφόρων καρδιαγγειακών χαρακτηριστικών.
Στο~\cite{xu2017direct} οι συγγραφείς εντοπίζουν τη θέση και το σχήμα του ΜΙ χρησιμοποιώντας ένα μοντέλο που αποτελείται από τρία επίπεδα; πρώτον, το επίπεδο εντοπισμού καρδιάς είναι ένα Fast R-CNN το οποίο απομονώνει τις ακολουθίες ROI συμπεριλαμβανομένου του LV\@; δεύτερον, τα στατιστικά επίπεδα κίνησης, τα οποία κατασκευάζουν μια αρχιτεκτονική χρονοσειράς για να καταγράψουν τα χαρακτηριστικά τοπικής κίνησης που παράγονται από το LSTM-RNN και τα χαρακτηριστικά κίνησης που παράγονται από βαθιές οπτικές ροές από την ακολουθία ROI\@; τρίτον, τα FNN διάκρισης, τα οποία χρησιμοποιούν το SAE για να μάθουν περαιτέρω τα χαρακτηριστικά από το προηγούμενο επίπεδο και τέλος έναν ταξινομητή softmax\@.
Οι Xue et al.~\cite{xue2018full} εκπαίδευσαν ένα δίκτυο βαθιάς μάθησης πολλαπλών προβλημάτων, σε MRI από 145 άτομα με 20 εικόνες ο καθένας για πλήρη ποσοτικοποίηση του LV\@.
Αποτελείται από ένα CNN τριών \textit{επιπέδων} που εξάγει τις καρδιακές αναπαραστάσεις, και στη συνέχεια δύο παράλληλα LSTM-RNN για τη μοντελοποίηση της χρονικής δυναμικής των καρδιακών ακολουθιών.
Τέλος, τοποθετείται ένα Bayesian πλαίσιο ικανό να μάθει τις σχέσεις μεταξύ των προβλημάτων και ένας ταξινομητής softmax.
Εκτεταμένες συγκρίσεις με προηγούμενες μεθόδους δείχνουν την αποτελεσματικότητα αυτής της μεθόδου όσον αφορά το μέσο απόλυτο σφάλμα.
Στο~\cite{zhen2016multi} οι συγγραφείς δημιούργησαν μια μέθοδο μη-επιβλεπώμενης μάθησης καρδιακής απεικόνισης χρησιμοποιώντας πολυεπίπεδο συνελικτικό RBM και μια άμεση εκτίμηση όγκου των δύο κοιλοτήτων χρησιμοποιώντας RF\@.
Σύγκριναν το μοντέλο τους με ένα Bayesian μοντέλο, ένα μοντέλο βασισμένο σε χειροποίητα χαρακτηριστικά, τα level-set και την περικοπή γραφήματος, επιτυγχάνοντας καλύτερα αποτελέσματα από πλευράς συντελεστή συσχέτισης για όγκους LV/RV και εκτίμηση σφάλματος του EF\@.

Άλλες μέθοδοι χρησιμοποιήθηκαν επίσης για την ανίχνευση υπερτροφικής καρδιομυοκαρδιοπάθειας ή για την αύξηση της αναλυτικότητας των MRI\@.
Οι Biffi et al.~\cite{biffi2018learning} εκπαίδευσαν ένα VΑΕ για την ταυτοποίηση ασθενών με υπερτροφική μυοκαρδιοπάθεια χρησιμοποιώντας ένα ισορροπημένο σύνολο δεδομένων 1365 ασθενών και την AC17.
Δείχνουν επίσης ότι το δίκτυο είναι σε θέση να απεικονίσει και να ποσοτικοποιήσει τα πρότυπα αναδιαμόρφωσης που είναι σχετικά με την παθολογία στον αρχικό χώρο εισόδου των εικόνων, αυξάνοντας έτσι την ερμηνευσιμότητα του μοντέλου.
Στο~\cite{oktay2016multi} οι συγγραφείς δημιούργησαν μια μέθοδο επαύξησης της αναλυτικότητας της εικόνας βασισμένη σε residual CNN η οποία επιτρέπει τη χρήση δεδομένων εισόδου που αποκτήθηκαν από διαφορετικά επίπεδα προβολής για βελτιωμένη απόδοση.
Σύγκριναν με άλλες μεθόδους παρεμβολής (γραμμική, spline, ταίριασμα patch πολλαπλών ατλάντων, ρηχό CNN, CNN), επιτυγχάνοντας καλύτερα αποτελέσματα όσον αφορά το PSNR\@.
Συγγραφείς από την ίδια ομάδα πρότειναν μια στρατηγική εκπαίδευσης~\cite{oktay2018anatomically} που ενσωματώνει ανατομική προηγούμενη γνώση σε CNNs μέσω ενός μοντέλου συστηματοποίησης, ενθαρρύνοντας την να ακολουθήσει την ανατομία μέσω μη-γραμμικών παραστάσεων του σχήματος.

\subsection{Συμπεράσματα της χρήσης βαθιάς μάθησης με MRI}
Υπάρχει ένα ευρύ φάσμα αρχιτεκτονικών που έχουν εφαρμοστεί στα MRI\@.
Οι περισσότερες είναι CNNs ή u-net οι οποίες είτε χρησιμοποιούνται αποκλειστικά είτε σε συνδυασμό με RNNs, AEs ή ensemble.
Το πρόβλημα είναι ότι οι περισσότερες από αυτές δεν εκπαιδεύονται από-άκρο-σε-άκρο. Βασίζονται σε προεπεξεργασία, χρήση χειροποίητων χαρακτηριστικών, ενεργά περιγράμματα, level-set και σε άλλες μη-διαφοροποιήσιμες μεθόδους, χάνοντας έτσι μερικώς τη δυνατότητα κλιμάκωσης στην παρουσία νέων δεδομένων.
Κύριος στόχος αυτού του τομέα πρέπει να είναι η δημιουργία μοντέλων από-άκρο-σε-άκρο, ακόμη και αν αυτό σημαίνει μικρότερη ακρίβεια βραχυπρόθεσμα; πιο αποδοτικές αρχιτεκτονικές θα μπορούσαν να καλύψουν το χάσμα στο μέλλον.

Ένα ενδιαφέρον εύρημα σχετικά με την κατάτμηση της καρδιάς έγινε στο~\cite{konukoglu2018exploration} όπου οι συγγραφείς διερεύνησαν την καταλληλότητα των προηγούμενων 2D, 3D CNN αρχιτεκτονικών και των τροποποιήσεών τους.
Διαπίστωσαν ότι η επεξεργασία ανά εικόνα χρησιμοποιώντας δίκτυα 2D ήταν καλύτερη λόγω του μεγάλου πάχους του τμήματος.
Ωστόσο, η επιλογή της αρχιτεκτονικής δικτύου διαδραματίζει μικρό ρόλο.

\section{Fundus}
Η απεικόνιση Fundus είναι ένα κλινικό εργαλείο για την αξιολόγηση της αμφιβληστροειδοπάθειας σε ασθενείς στην οποία η ένταση αντιπροσωπεύει την ποσότητα του ανακλώμενου φωτός συγκεκριμένης ζώνης κυμάτων~\cite{abramoff2010retinal}.
Μία από τις πιο ευρέως διαδεδομένες βάσεις δεδομένων στο Fundus είναι η DRIVE, η οποία περιέχει 40 εικόνες και τις αντίστοιχες επισημάνσεις της μάσκας των αγγείων.

\subsection{Κατάτμηση αγγείων}
Τα CNN χρησιμοποιήθηκαν για την κατάτμηση αγγείων σε εικόνες Fundus.
Στο~\cite{wang2015hierarchical} οι συγγραφείς αρχικά χρησιμοποίησαν ισορροπία ιστογράμματος και φιλτράρισμα Gauss για τη μείωση του θορύβου.
Στη συνέχεια χρησιμοποιήθηκε ένα CNN τριών \textit{επιπέδων} ως εξαγωγέας χαρακτηριστικών και ένα RF ως ταξινομητής.
Σύμφωνα με τα πειράματά τους, η καλύτερη απόδοση επιτεύχθηκε από ένα ensemble νικητής-τα-παίρνει-όλα, σε σύγκριση με ένα μέσο, σταθμισμένο και διάμεσο ensemble.
Οι Zhou et al.~\cite{zhou2017improving} εφάρμοσαν προεπεξεργασία εικόνας για την εξάλειψη των ισχυρών άκρων γύρω από το οπτικό πεδίο και κανονικοποίησαν την φωτεινότητα και την αντίθεση μέσα σε αυτό.
Στη συνέχεια, εκπαίδευσαν ένα CNN για να παράξουν χαρακτηριστικά για γραμμικά μοντέλα και εφάρμοσαν φίλτρα για την ενίσχυση των λεπτών αγγείων, μειώνοντας τη διαφορά έντασης μεταξύ λεπτών και ευρέων αγγείων.
Στη συνέχεια ένας πυκνός CRF προσαρμόστηκε για να επιτευχθεί η τελική κατάτμηση του αγγείου, λαμβάνοντας τα διακριτικά χαρακτηριστικά για μοναδιαία δυναμικά και την εικόνα με τα ενισχυμένα λεπτά αγγεία.
Μεταξύ των αποτελεσμάτων τους, στα οποία παρουσιάζουν μεγαλύτερη ακρίβεια από τις περισσότερες μεθόδους τελευταίας τεχνολογίας, παρέχουν επίσης στοιχεία υπέρ της χρήσης των πληροφοριών RGB του Fundus αντί για χρήση μόνο του πράσινου καναλιού.
Οι Chen et al.~\cite{chen2017labeling} σχεδίασαν ένα σύνολο κανόνων για τη δημιουργία τεχνητών δειγμάτων εκπαίδευσης με πρότερη γνώση και χωρίς χειροκίνητη επισήμανση.
Εκπαίδευσαν ένα FCN με ένα skip-connection που επιτρέπει την υψηλού επιπέδου πληροφορία να καθοδηγεί την εργασία σε χαμηλότερα επίπεδα.
Αξιολογούν το μοντέλο τους στις DRIVE και STARE, επιτυγχάνοντας συγκρίσιμα αποτελέσματα με άλλες μεθόδους που χρησιμοποιούν πραγματική επισήμανση.
Στο~\cite{maji2016ensemble} οι συγγραφείς εκπαίδευσαν ένα ensemble 12 CNNs με τρία \textit{επίπεδα} το καθένα στη DRIVE, όπου κατά τη διάρκεια του συμπερασμού οι απαντήσεις των CNNs υπολογίζονται κατά μέσο όρο για να σχηματίσουν την τελική κατάτμηση.
Δείχνουν ότι το μοντέλο τους επιτυγχάνει υψηλότερη μέση ακρίβεια από τις προηγούμενες μεθόδους.
Οι Fu et al.~\cite{fu2016retinal} εκπαιδεύουν ένα CNN στις DRIVE και STARE δημιουργώντας ένα χάρτη πιθανοτήτων και έπειτα χρησιμοποιούν έναν πλήρως συνδεδεμένο CRF για να συνδυάσουν τους χάρτες πιθανότητας και τις αλληλεπιδράσεις μεγάλης εμβέλειας μεταξύ των εικονοστοιχείων.
Στο~\cite{wu2016deep} οι συγγραφείς χρησιμοποίησαν ένα CNN για να μάθουν τα χαρακτηριστικά και μια αναζήτηση πλησιέστερων γειτόνων βασισμένη στο PCA που χρησιμοποιήθηκε για την εκτίμηση της τοπικής κατανομής δομών.
Εκτός από την παρουσίαση καλών αποτελεσμάτων υποστηρίζουν ότι είναι σημαντικό για το CNN να ενσωματώσει πληροφορίες σχετικά με τη δομή δέντρων όσον αφορά την ακρίβεια.

\begin{sidewaystable}
	\caption{Εφαρμογές βαθιάς μάθησης με χρήση Fundus, για κατάτμηση αγγείων}
	\label{table:imaging4}
	\centering
	\begin{tabular}{l c l l}
		\toprule
		\thead{Αναφορά}                                  & \thead{Μέθοδος} & \thead{Εφαρμογή/Σημειώσεις\footnote{Σε παρένθεση οι βάσεις δεδομένων που χρησιμοποιήθηκαν.}}                  & \thead{AUC\footnote{Το ($*$) υποδηλώνει ακρίβεια.}} \\
		\midrule
		Wang 2015~\cite{wang2015hierarchical}             & CNN, RF         & CNN τριών \textit{επιπέδων} συνδυασμένα με ensemble RF (DRIVE, STARE)                                         & 0.9475                                              \\
		Zhou 2017~\cite{zhou2017improving}                & CNN, CRF        & CNN για εξαγωγή χαρακτηριστικών και CRF για τελικό αποτέλεσμα (DRIVE, STARE, CHDB)                     & 0.7942                                              \\
		Chen 2017~\cite{chen2017labeling}                 & CNN             & τεχνητά δεδομένα, FCN (DRIVE, STARE)                                                                          & 0.9516                                              \\
		Maji 2016~\cite{maji2016ensemble}                 & CNN             & 12 CNNs ensemble με τρία \textit{επίπεδα} (DRIVE)                                                             & 0.9283                                              \\
		Fu 2016~\cite{fu2016retinal}                      & CNN, CRF        & CNN και CRF (DRIVE, STARE)                                                                                    & 94.70\%$^*$                                         \\
		Wu 2016~\cite{wu2016deep}                         & CNN             & τμηματοποίηση αγγείου και ανίχνευση βρόχου με CNN και PCA (DRIVE)                                             & 0.9701                                              \\
		Li 2016~\cite{li2016cross}                        & SDAE            & FNN και SDAE (DRIVE, STARE, CHDB)                                                                             & 0.9738                                              \\
		Lahiri 2016~\cite{lahiri2016deep}                 & SDAE            & ensemble δύο επιπέδων SDAE (DRIVE)                                                        & 95.30\%$^*$                                         \\
		Oliveira 2017~\cite{oliveira2017augmenting}       & u-net           & επαύξηση δεδομένων και u-net (DRIVE)                                                                          & 0.9768                                              \\
		Leopold 2017~\cite{leopold2017use}                & CNN             & CNN ως ένα πολυ-καναλικός ταξινομητής και φίλτρα Gabor (DRIVE)                                                & 94.78\%$^*$                                         \\
		Leopold 2017~\cite{leopold2017pixelbnn}           & AE              & πλήρως residual AE με περιφραγμένο ρεύμα βασισμένο στο u-net (DRIVE, STARE, CHDB)                             & 0.8268                                              \\
		Mo 2017~\cite{mo2017multi}                        & CNN             & επαυξημένοι ταξινομητές και μεταφορά μάθησης (DRIVE, STARE, CHDB)                                             & 0.9782                                              \\
		Melinscak 2015~\cite{melinvsvcak2015retinal}      & CNN             & CNN τεσσάρων \textit{επιπέδων} (DRIVE)                                                                        & 0.9749                                              \\
		Sengur 2017~\cite{sengur2017retinal}              & CNN             & CNN δύο \textit{επιπέδων} με dropout (DRIVE)                                                                  & 0.9674                                              \\
		Meyer 2017~\cite{meyer2017deep}                   & u-net           & τμηματοποίηση αγγείου με χρήση u-net στην SLO (IOSTAR, RC-SLO)                                                & 0.9771                                              \\
		\bottomrule
	\end{tabular}
\end{sidewaystable}

Τα ΑΕ έχουν επίσης χρησιμοποιηθεί για την κατάτμηση των αγγείων.
Οι Li et al.~\cite{li2016cross} εκπαίδευσαν ένα FNN και ένα AE αποθορυβοποίησης με τις DRIVE, STARE και CHDB\@.
Υποστηρίζουν ότι τα χαρακτηριστικά του μοντέλου τους είναι πιο ανθεκτικά στο θόρυβο και τις διαφορετικές συνθήκες απεικόνισης, επειδή η διαδικασία μάθησης εκμεταλλεύεται τα χαρακτηριστικά των αγγείων σε όλες τις εικόνες εκπαίδευσης.
Στο~\cite{lahiri2016deep} οι συγγραφείς χρησιμοποίησαν μη-επιβλεπώμενα ιεραρχικά χαρακτηριστικά χρησιμοποιώντας ένα ensemble δύο \textit{επιπέδων} SDAE\@.
Το επίπεδο εκπαίδευσης εξασφαλίζει αποσύνδεση και το επίπεδο του ensemble εξασφαλίζει την αρχιτεκτονική αναθεώρηση.
Δείχνουν επίσης ότι η εκπαίδευση του ensemble των ΑΕ ενισχύει την ποικιλομορφία στο λεξικό των χαρακτηριστικών που έχουν μαθευτεί για την κατάτμηση των αγγείων.
Ο ταξινομητής Softmax χρησιμοποιήθηκε στη συνέχεια για το fine-tuning κάθε ΑΕ και διερευνήθηκαν στρατηγικές για συγχώνευση δύο επιπέδων μελών του ensemble.

Άλλες αρχιτεκτονικές χρησιμοποιήθηκαν επίσης για την κατάτμηση των αγγείων.
Στο άρθρο τους οι Oliveira et al.~\cite{oliveira2017augmenting} εκπαίδευσαν ένα u-net με την DRIVE, παρουσιάζοντας καλά αποτελέσματα και ενδείξεις των πλεονεκτημάτων της επαύξησης των δεδομένων εκπαίδευσης με χρήση ελαστικών μετασχηματισμών.
Οι Leopold et al.~\cite{leopold2017use} διερεύνησαν τη χρήση ενός CNN ως ταξινομητή πολλαπλών καναλιών και τη χρήση φίλτρων Gabor για να ενισχύσουν την ακρίβεια της μεθόδου που περιγράφεται στο~\cite{leopold2017segmentation}.
Εφάρμοσαν το μέσο μιας σειράς φίλτρων Gabor με διάφορες συχνότητες και τιμές σίγμα στην έξοδο του δικτύου για να καθορίσουν εάν ένα εικονοστοιχείο αντιπροσωπεύει ένα αγγείο ή όχι.
Εκτός από τη διαπίστωση ότι τα βέλτιστα φίλτρα διαφέρουν μεταξύ των καναλιών, οι συγγραφείς δηλώνουν επίσης την `ανάγκη' να επιβάλλουν στα δίκτυα να ευθυγραμμίζονται με την ανθρώπινη αντίληψη, στο πλαίσιο της χειρωνακτικής επισήμανσης, ακόμη και αν απαιτεί πληροφορίες υποδειγματοληψίας, οι οποίες διαφορετικά θα μείωναν το υπολογιστικό κόστος.
Οι ίδιοι συγγραφείς~\cite{leopold2017pixelbnn} δημιούργησαν το PixelBNN, το οποίο είναι ένα πλήρως residual AE\@.
Είναι πάνω από οκτώ φορές χρονικά αποδοτικό από τις προηγούμενες μεθόδους κατά την διάρκεια του συμπερασμού με καλά αποτελέσματα, λαμβάνοντας υπόψη τη σημαντική μείωση των πληροφοριών από την αλλαγή μεγέθους των εικόνων κατά την προεπεξεργασία.
Στο άρθρο τους οι Mo et al.~\cite{mo2017multi} χρησιμοποίησαν βαθιά επίβλεψη με βοηθητικούς ταξινομητές στα ενδιάμεσα επίπεδα του δικτύου, για να βελτιώσουν τη διακριτική ικανότητα των χαρακτηριστικών στα χαμηλότερα επίπεδα του δικτύου και να καθοδηγήσουν το backpropagation να ξεπεράσει τα vanishing gradients.
Επιπλέον, μεταφορά μάθησης χρησιμοποιήθηκε για να ξεπεραστεί το ζήτημα των ανεπαρκών δεδομένων εκπαίδευσης.

\subsection{Ανίχνευση μικροανευρυσμάτων και αιμορραγίας}
Οι Haloi et al.~\cite{haloi2015improved} εκπαίδευσαν ένα CNN τριών \textit{επιπέδων} με dropout και maxout για ανίχνευση ΜΑ.
Τα πειράματα στις ROC και DIA έδειξαν θετικά αποτελέσματα.
Στο~\cite{giancardo2017representation} οι συγγραφείς δημιούργησαν ένα μοντέλο που μαθαίνει έναν γενικό περιγραφέα της μορφολογίας των αγγείων χρησιμοποιώντας την εσωτερική αναπαράσταση ενός variational u-net.
Έπειτα, εξέτασαν τις αγγειακές ενσωματώσεις (embeddings) σε ένα παρόμοιο πρόβλημα ανάκτησης εικόνων σύμφωνα με το αγγειακό σύστημα και σε ένα πρόβλημα ταξινόμησης της διαβητικής αμφιβληστροειδοπάθειας, στο οποίο δείχνουν πως τα embeddings των αγγείων μπορούν να βελτιώσουν την ταξινόμηση μιας μεθόδου η οποία βασίζεται στην ανίχνευση ΜΑ.
Στο~\cite{orlando2018ensemble} οι συγγραφείς συνδυάζουν ενισχυμένα χαρακτηριστικά από ένα CNN με χειροποίητα χαρακτηριστικά.
Αυτό το ensemble διάνυσμα χαρακτηριστικών χρησιμοποιήθηκε στη συνέχεια για την αναγνώριση των υποψηφίων αιμορραγίας και των ΜΑ, χρησιμοποιώντας έναν ταξινομητή RF\@.
Η ανάλυσή τους με χρήση t-SNE καταδεικνύει ότι τα χαρακτηριστικά από το CNN έχουν μεγάλη λεπτομέρεια όπως ένδειξη του προσανατολισμού της βλάβης ενώ τα χειροποίητα χαρακτηριστικά είναι σε θέση να διακρίνουν βλάβες χαμηλής αντίθεσης όπως αιμορραγίες.
Στο~\cite{van2016fast} οι συγγραφείς εκπαίδευσαν ένα CNN πέντε \textit{επιπέδων}, για ανίχνευση αιμορραγίας χρησιμοποιώντας 6679 εικόνες από τις βάσεις δεδομένων DS16 και Messidor.
Εφάρμοσαν επιλεκτική δειγματοληψία σε ένα CNN, κάτι το οποίο αύξησε την ταχύτητα της εκπαίδευσης με δυναμική επιλογή λανθασμένα ταξινομημένων δειγμάτων κατά τη διάρκεια της εκπαίδευσης.
Τα βάρη αποδίδονται στα δείγματα εκπαίδευσης και τα ενημερωμένα δείγματα περιλαμβάνονται στην επόμενη επανάληψη εκπαίδευσης.

\subsection{Άλλες εφαρμογές}
Το Fundus έχει χρησιμοποιηθεί και για ταξινόμηση αρτηριών/φλεβών.
Στο άρθρο τους οι Girard et al.~\cite{girard2017artery} εκπαίδευσαν ένα CNN τεσσάρων \textit{επιπέδων}, που ταξινομεί τα εικονοστοιχεία των αγγείων σε αρτηρίες/φλέβες χρησιμοποιώντας επαύξηση των δεδομένων εκπαίδευσης με περιστροφικό μετασχηματισμό.
Στη συνέχεια κατασκευάστηκε ένα γράφημα από το αγγειακό δίκτυο του αμφιβληστροειδούς, όπου οι κόμβοι ορίζονται ως οι κλάδοι των αγγείων και κάθε άκρη συνδέεται με ένα κόστος που εκτιμά εάν οι δύο κλάδοι θα πρέπει να έχουν την ίδια ετικέτα.
Η ταξινόμηση του CNN διαδόθηκε μέσω του ελάχιστου δένδρου του γραφήματος.
Τα πειράματα κατέδειξαν την αποτελεσματικότητα της μεθόδου, ιδίως στην παρουσία εμφραγμάτων.
Οι Welikala et al.~\cite{welikala2017automated} εκπαίδευσαν και αξιολόγησαν ένα CNN τριών \textit{επιπέδων}, χρησιμοποιώντας εικονοστοιχεία της κεντρικής γραμμής που προέρχονται από εικόνες αμφιβληστροειδούς.
Από τα πειράματά τους διαπίστωσαν ότι η επαύξηση των δεδομένων εκπαίδευσης με χρήση περιστροφικών και κλιμακωτών μετασχηματισμών, δεν βοήθησε στην αύξηση της ακρίβειας αποδίδοντάς το στην παρεμβολή μεταξύ των εντάσεων των εικονοστοιχείων, η οποία είναι προβληματική λόγω της ευαισθησίας του CNN στην κατανομή των εικονοστοιχείων.

\begin{sidewaystable}
	\caption{Εφαρμογές βαθιάς μάθησης με χρήση Fundus, εκτός της κατάτμησης αγγείων}
	\label{table:imaging5}
	\centering
	\begin{tabular}{l c l l}
		\toprule
		\thead{Αναφορά}                                  & \thead{Μέθοδος} & \thead{Εφαρμογή/Σημειώσεις\footnote{Σε παρένθεση οι βάσεις δεδομένων που χρησιμοποιήθηκαν.}}                  & \thead{AUC\footnote{Το ($*$) υποδηλώνει ακρίβεια.}} \\
		\midrule
		\multicolumn{4}{l}{\thead{Μικροανεύρισμα και ανίχνευση αιμορραγίας}}                                                                                                                                                                     \\
		\midrule
		Haloi 2015~\cite{haloi2015improved}               & CNN             & ανίχνευση MA με χρήση CNN με dropout και maxout (ROC, Messidor, DIA)                                          & 0.98                                                \\
		Giancardo 2017~\cite{giancardo2017representation} & u-net           & ανίχνευση MA με χρήση εσωτερικών αναπαραστάσεων με εκπαιδευμένα u-net (DRIVE, Messidor)                       & \textit{πολλαπλά}                                   \\
		Orlando 2018~\cite{orlando2018ensemble}           & CNN             & ανίχνευση MA και αιμορραγιών με χρήση χαρακτηριστικών και CNN (DIA, e-optha, Messidor)       & \textit{πολλαπλά}                                   \\
		van Grinsven 2017~\cite{van2016fast}              & CNN             & ανίχνευση αιμορραγιών με δειγματοληψία και χρήση CNN πέντε \textit{επιπέδων} (KR15, Messidor) & \textit{πολλαπλά}                                   \\
		\midrule
		\multicolumn{4}{l}{\thead{Άλλες εφαρμογές}}                                                                                                                                                                                              \\
		\midrule
		Girard 2017~\cite{girard2017artery}               & CNN             & ταξινόμηση αρτηριών/φλεβών με χρήση CNN και διάδοση πιθανοφάνειας (DRIVE, Messidor)                           & \textit{πολλαπλά}                                   \\
		Welikala 2017~\cite{welikala2017automated}        & CNN             & ταξινόμηση αρτηριών/φλεβών με χρήση ενός CNN τριών \textit{επιπέδων} (UKBDB)                                  & 82.26\%$^*$                                         \\
		Pratt 2017~\cite{pratt2017automatica}             & ResNet          & ταξινόμηση διασταυρώσεων με χρήση ενός ResNet18 (DRIVE, IOSTAR)                                               & \textit{πολλαπλά}                                   \\
		Poplin 2017~\cite{poplin2017predicting}           & Inception       & πρόβλεψη παραγόντων καρδιαγγειακών κινδύνου (UKBDB, μη-δημόσια)                                               & \textit{πολλαπλά}                                   \\
		\bottomrule
	\end{tabular}
\end{sidewaystable}

Υπάρχουν επίσης και άλλες εφαρμογές του Fundus όπως η αναγνώριση διακλάδωσης/διέλευσης.
Οι Pratt et al.~\cite{pratt2017automatica} εκπαίδευσαν ένα ResNet 18 \textit{επιπέδων}, για να εντοπίσουν μικρά patches που περιλαμβάνουν είτε διακλάδωση είτε διέλευση.
Ένα άλλο ResNet18 εκπαιδεύτηκε σε patches που έχουν ταξινομηθεί ώστε να έχουν διακλαδώσεις και διελεύσεις έτσι ώστε να διακρίνουν σε ποια κατηγορία ανήκει.
Παρόμοια επίλυση σε αυτό το πρόβλημα έχει γίνει από τους ίδιους συγγραφείς~\cite{pratt2017automaticb} χρησιμοποιώντας ένα CNN\@.

Ένα σημαντικό αποτέλεσμα στον τομέα της Καρδιολογίας χρησιμοποιώντας τη Fundus είναι από τους Poplin et al.~\cite{poplin2017predicting} που χρησιμοποίησαν ένα Inception-v3 για να προβλέψουν παράγοντες καρδιαγγειακού κινδύνου (ηλικία, φύλο, κατάσταση καπνίσματος, HbA1c, SBP) και μείζον καρδιακά επεισόδια.
Τα μοντέλα τους χρησιμοποίησαν ξεχωριστές πτυχές της ανατομίας για τη δημιουργία κάθε πρόβλεψης, όπως ο οπτικός δίσκος ή τα αιμοφόρα αγγεία, όπως αποδείχθηκε χρησιμοποιώντας την τεχνική soft-attention.
Τα περισσότερα αποτελέσματα ήταν σημαντικά καλύτερα από ότι θεωρούνταν προηγουμένως δυνατό με τη Fundus (\textgreater{70\%} AUC).

\subsection{Συμπεράσματα της χρήσης βαθιάς μάθησης με Fundus}
Όσον αφορά τη χρήση των αρχιτεκτονικών, υπάρχει σαφής προτίμηση στα CNN ειδικά στον τομέα της κατάτμησης των αγγείων, ενώ μια ενδιαφέρουσα προσέγγιση από ορισμένες δημοσιεύσεις είναι η χρήση των CRF μετά την επεξεργασία για τη περαιτέρω βελτίωση της κατάτμησης των αγγείων.
Το γεγονός ότι υπάρχουν πολλές βάσεις δεδομένων που είναι διαθέσιμες και ότι η βάση δεδομένων DRIVE χρησιμοποιείται κατά κύριο λόγο στην πλειοψηφία της βιβλιογραφίας, καθιστά τον τομέα αυτό ευκολότερο στη σύγκριση και επικύρωση νέων αρχιτεκτονικών.
Επιπλέον, η μη-επεμβατική φύση του Fundus και η πρόσφατη χρήση του ως εργαλείου για την εκτίμηση καρδιαγγειακού κινδύνου, το καθιστά μια πολλά υποσχόμενη απεικονιστική αυξημένης χρησιμότητας στον τομέα της καρδιολογίας.

\section{Ηλεκτρονική τομογραφία}
Η ηλεκτρονική τομογραφία (CT) είναι μια μη-επεμβατική μέθοδος για την ανίχνευση της αποφρακτικής αρτηριακής νόσου.
Ορισμένες από τις περιοχές στις οποίες εφαρμόστηκε η βαθιά μάθηση στην CT, περιλαμβάνουν την αξιολόγηση της βαθμολογίας ασβεστίου της στεφανιαίας αρτηρίας, τον εντοπισμό και τον κατάτμηση των καρδιακών περιοχών.

\begin{sidewaystable}
	\caption{Εφαρμογές βαθιάς μάθησης με χρήση CT}
	\label{table:imaging6}
	\centering
	\begin{tabular}{l c l}
		\toprule
		\thead{Αναφορά}                              & \thead{Μέθοδος} & \thead{Εφαρμογή/Σημειώσεις\footnote{Αποτελέσματα από αυτές τις απεικονιστικές τεχνικές δεν δημοσιεύονται σε αυτήν την βιβλιογραφική αναφορά καθώς υπάρχει υψηλή μεταβλητότητα σε σχέση με το ερευνητικό ερώτημα προς προς απάντηση και στην χρήση των μετρήσεων. Επιπλέον όλες οι δημοσιεύσεις χρησιμοποιούν μη-δημόσιες βάσεις δεδομένων εκτός της~\cite{liu2017left}.}} \\
		\midrule
		Lessman 2016~\cite{lessmann2016deep}          & CNN             & ανίχνευση στεφανιαίου ασβεστίου με χρήση τριών ανεξάρτητα εκπαιδευμένων CNNs                                                                                                                                                                                                                                                                                                                \\
		Shadmi 2018~\cite{shadmi2018fully}            & DenseNet        & συνέκριναν το DenseNet και το u-net για τον εντοπισμό στεφανιαίου ασβεστίου                                                                                                                                                                                                                                                                                                                 \\
		Cano 2018~\cite{cano2018automated}            & CNN             & 3D CNN παλινδρόμησης για τον υπολογισμό της βαθμολογίας Agatston                                                                                                                                                                                                                                                                                                                            \\
		Wolterink 2016~\cite{wolterink2016automatic}  & CNN             & ανίχνευση στεφανιαίου ασβεστίου με χρήση τριών CNNs για τον εντοπισμό και δύο CNN για την ανίχνευση                                                                                                                                                                                                                                                                                         \\
		Santini 2017~\cite{santini2017automatic}      & CNN             & ανίχνευση στεφανιαίου ασβεστίου με χρήση ενός CNN επτά \textit{επιπέδων} σε patches εικόνας                                                                                                                                                                                                                                                                                                 \\
		Lopez 2017~\cite{lopez2017dcnn}               & CNN             & χαρακτηρισμός όγκου θρόμβου με χρήση ενός 2D CNN και μετα-επεξεργασία                                                                                                                                                                                                                                                                                                                       \\
		Hong 2016~\cite{hong2016automatic}            & DBN             & ανίχνευση, τμηματοποίηση, ταξινόμηση αορτικού ανευρύσματος με χρήση DBN και patches εικόνων                                                                                                                                                                                                                                                                                                 \\
		Liu 2017~\cite{liu2017left}                   & CNN             & τμηματοποίηση αριστερού κόλπου καρδιάς με CNN δώδεκα \textit{επιπέδων} και μοντέλα ενεργού σχήματος (STA13)                                                                                                                                                                                                                                                                           \\
		de Vos 2016~\cite{de20162d}                   & CNN             & 3D εντοπισμός των ανατομικών δομών με χρήση τριών CNNs, ένα για κάθε ορθογώνιο επίπεδο                                                                                                                                                                                                                                                                                                      \\
		Moradi 2016~\cite{moradi2016hybrid}           & CNN             & ανίχνευση θέσης στο CT με προεκπαιδευμένο VGGnet, χειροποίητα χαρακτηριστικά και SVM                                                                                                                                                                                                                                                            \\
		Zheng 2015~\cite{zheng20153d}                 & Multiple        & ανίχνευση διακλάδωσης της καρωτίδας με χρήση πολυ-επίπεδων perceptrons και πιθανοτικά boosting-trees                                                                                                                                                                                                                                                                             \\
		Montoya 2018~\cite{montoya2018deep}           & ResNet          & 3D ανακατασκευή αγγειογραφείας με χρήση ενός ResNet-30 \textit{επιπέδων}                                                                                                                                                                                                                                                                                                                    \\
		Zreik 2018~\cite{zreik2018deep}               & CNN, AE         & ανίχνευση στεφανιαίας αρτηριακής στένωσης με CNN για LV τμηματοποίηση και AE, SVM για ταξινόμηση                                                                                                                                                                                                                                                                              \\
		Commandeur 2018~\cite{commandeur2018deep}     & CNN             & ποσοτικοποιήση του επικάρδιου και θωρακικού λιπώδες ιστού από χωρίς-αντίθεση CT                                                                                                                                                                                                                                                                                                             \\
		Gulsun 2016~\cite{gulsun2016coronary}         & CNN             & εξαγωγή στεφανιαίων κεντρικών γραμμών με χρήση βέλτιστου μονοπατιού από πεδίο ροών και ένα CNN                                                                                                                                                                                                                                                                                              \\
		\bottomrule
	\end{tabular}
\end{sidewaystable}

Η μέθοδος των Lessman et al.~\cite{lessmann2016deep} για τη βαθμολόγηση του στεφανιαίου ασβεστίου χρησιμοποιεί τρία ανεξάρτητα εκπαιδευμένα CNNs για να εκτιμήσει ένα πλαίσιο οριοθέτησης γύρω από την καρδιά, στο οποίο τα συνδεδεμένα συστατικά (connected components) πάνω από ένα όριο μονάδας Hounsfield θεωρούνται υποψήφια για CACs.
Η ταξινόμηση των voxels πραγματοποιήθηκε τροφοδοτώντας δισδιάστατα patches από τρία ορθογώνια επίπεδα ταυτόχρονα σε τρία CNNs, για να διαχωριστούν από άλλες περιοχές υψηλής έντασης.
Οι ασθενείς ταξινομήθηκαν σε μία από τις πέντε πρότυπες κατηγορίες καρδιαγγειακού κινδύνου βάσει της βαθμολογίας Agatston.
Συγγραφείς από την ίδια ομάδα δημιούργησαν μια μέθοδο~\cite{lessmann2017automatic} για την ανίχνευση ασβεστοποιήσεων σε θωρακικό CT χαμηλής δόσης χρησιμοποιώντας ένα CNN για ανατομική θέση και ένα άλλο CNN για την ανίχνευση της ασβεστοποίησης.
Στο~\cite{shadmi2018fully} οι συγγραφείς σύγκριναν ένα u-net και DenseNet για τον υπολογισμό της βαθμολογίας Agatston χρησιμοποιώντας πάνω από 1000 εικόνες θωρακικού CT\@.
Οι συγγραφείς επεξεργάστηκαν έντονα τις εικόνες χρησιμοποιώντας κατωφλίωση, ανάλυση connected components και μορφολογικές λειτουργίες για την ανίχνευση των πνευμόνων, της τραχείας και της τροπίδας.
Τα πειράματά τους έδειξαν ότι το DenseNet είχε καλύτερη απόδοση όσον αφορά την ακρίβεια.
Οι Cano et al.~\cite{cano2018automated} εκπαίδευσαν ένα 3D CNN παλινδρόμησης που υπολόγισε την βαθμολογία Agatston χρησιμοποιώντας 5973 εικόνες CT χωρίς ECG περίφραξης επιτυγχάνοντας συσχέτιση Pearson 0.932.
Στο~\cite{wolterink2016automatic} οι συγγραφείς δημιούργησαν μια μέθοδο για την ανίχνευση και τον ποσοτικό προσδιορισμό της CAC, χωρίς την εξαγωγή της στεφανιαίας αρτηρίας.
Η ανίχνευση του πλαισίου οριοθέτησης γύρω από την καρδιά χρησιμοποιεί τρία CNNs, όπου το καθένα ανιχνεύει την καρδιά στο αξονικό, ισχαιμικό και στεφανιαίο επίπεδο.
Ένα άλλο ζεύγος από CNNs χρησιμοποιήθηκε για την ανίχνευση CAC\@.
Το πρώτο CNN αναγνωρίζει κύστεις τύπου CAC, απορρίπτοντας έτσι την πλειονότητα των voxels που δεν είναι υποψήφια CAC όπως ο πνεύμονας και ο λιπώδης ιστός.
Τα αναγνωρισμένα voxels τύπου CAC ταξινομούνται περαιτέρω από το δεύτερο CNN, το οποίο διακρίνει τα CAC\@.
Αν και τα CNNs μοιράζονται την αρχιτεκτονική, δεδομένου ότι έχουν διαφορετικό έργο, δεν μοιράζονται βάρη.
Επιτυγχάνουν μια συσχέτιση Pearson 0.95, συγκρίσιμη με προηγούμενες καλύτερες τεχνικές.
Οι Santini et al.~\cite{santini2017automatic} εκπαίδευσαν ένα CNN επτά \textit{επιπέδων} χρησιμοποιώντας patches, για την κατάτμηση και ταξινόμηση των στεφανιαίων περιοχών στις εικόνες CT\@.
Εκπαίδευσαν, επικύρωσαν και δοκίμασαν το δίκτυό τους σε 45, 18 και 56 CT όγκους αντιστοίχως, επιτυγχάνοντας μια συσχέτιση Pearson 0.983.

Το CT έχει χρησιμοποιηθεί επίσης και για την κατάτμηση διαφόρων καρδιακών περιοχών.
Οι Lopez et al.~\cite{lopez2017dcnn} εκπαίδευσαν ένα 2D CNN για την εκτίμηση του όγκου αορτής θρόμβου από προεγχειρητικές και μετεγχειρητικές κατατμήσεις, χρησιμοποιώντας περιστροφικές και κατοπτρικές επαυξήσεις δεδομένων.
Μεταγενέστερη επεξεργασία περιλαμβάνει Gaussian φιλτράρισμα και ομαδοποίηση κ-μέσων.
Στο άρθρο τους, οι Hong et al.~\cite{hong2016automatic} εκπαίδευσαν ένα DBN χρησιμοποιώντας patches εικόνας για την ανίχνευση, τμηματοποίηση και ταξινόμηση της σοβαρότητας του κοιλιακού αορτικού ανευρύσματος σε CT εικόνες.
Οι Liu et al.~\cite{liu2017left} χρησιμοποίησαν ένα FCN με δώδεκα \textit{επίπεδα} για την κατάτμηση του αριστερού κόλπου σε 3D όγκους CT και στη συνέχεια βελτίωσαν τα αποτελέσματα τμηματοποίησης του FCN με ένα μοντέλο ενεργού σχήματος επιτυγχάνοντας Dice 93\%.

Το CT έχει επίσης χρησιμοποιηθεί για τον εντοπισμό των καρδιακών περιοχών.
Στο~\cite{de20162d} οι συγγραφείς δημιούργησαν μια μέθοδο για την ανίχνευση ανατομικών ROI (καρδιά, αορτική αψίδα και φθίνουσα αορτή) σε εικόνες από θωρακικό CT προκειμένου να εντοπιστούν σε 3D.
Κάθε ROI εντοπίστηκε χρησιμοποιώντας έναν συνδυασμό τριών CNNs, κάθε ένα αναλύοντας ένα ορθογώνιο επίπεδο εικόνας.
Ενώ ένα CNN προέβλεψε την παρουσία ενός συγκεκριμένου ROI στο δεδομένο επίπεδο, ο συνδυασμός των αποτελεσμάτων τους παρείχε ένα 3D πλαίσιο οριοθέτησης γύρω του.
Στο άρθρο τους οι Moradi et al.~\cite{moradi2016hybrid} αντιμετωπίζουν το πρόβλημα της ανίχνευσης κατακόρυφης θέσης για μια δεδομένη καρδιακή εικόνα CT\@.
Διαχωρίζουν την περιοχή του σώματος που απεικονίζεται στο θωρακικό CT σε εννέα σημασιολογικές κατηγορίες, οι οποίες κάθε μια αντιπροσωπεύουν μία περιοχή που είναι σχετική με τη μελέτη αντίστοιχων νόσων.
Χρησιμοποιώντας ένα σύνολο χειροποίητων χαρακτηριστικών εικόνας μαζί με τα χαρακτηριστικά που προέρχονται από ένα προεκπαιδευμένο VGGnet με πέντε \textit{επίπεδα}, χτίζουν ένα σχήμα ταξινόμησης για να αντιστοιχίσουν μια δεδομένη εικόνα CT στο σχετικό επίπεδο.
Κάθε ομάδα χαρακτηριστικών χρησιμοποιήθηκε για την εκπαίδευση ενός ξεχωριστού ταξινομητή SVM και οι προβλεπόμενες ετικέτες στη συνέχεια συνδυάζονται σε ένα γραμμικό μοντέλο, το οποίο επίσης αντλήθηκε από τα δεδομένα εκπαίδευσης.

Η βαθιά μάθηση χρησιμοποιήθηκε σε συνδυασμό με το CT και για άλλες περιοχές εκτός από την καρδιάς.
Οι Zheng et al.~\cite{zheng20153d} δημιούργησαν μια μέθοδο για 3D ανίχνευση σε ογκομετρικά δεδομένα, τα οποία αξιολογήθηκαν ποσοτικά για την ανίχνευση της διακλάδωσης της καρωτιδικής αρτηρίας σε CT\@.
Χρησιμοποιήθηκε ένα δίκτυο κρυφών επιπέδων για τον αρχικό έλεγχο όλων των voxels για να αποκτηθεί ένας μικρός αριθμός υποψηφίων, ακολουθούμενος από μια ακριβέστερη ταξινόμηση με ένα βαθύ δίκτυο.
Τα χαρακτηριστικά από το δίκτυο συνδυάζονται περαιτέρω με τα χαρακτηριστικά κυματιδίων Haar, για την αύξηση της ακρίβειας ανίχνευσης.
Οι Montoya et al.~\cite{montoya2018deep} εκπαίδευσαν ένα ResNet με 30 επίπεδα, για να δημιουργήσουν 3D αγγειογραφήματα χρησιμοποιώντας τρεις τύπους ιστών (αγγειακό σύστημα, οστό και μαλακό ιστό).
Δημιούργησαν τις επισημάνσεις χρησιμοποιώντας κατωφλίωση και connected components σε 3D, έχοντας ένα συνδυασμένο σύνολο 13790 εικόνων.

Το CT έχει επίσης χρησιμοποιηθεί για την επίλυση και άλλων προβλημάτων.
Οι Zreik et al.~\cite{zreik2018deep} δημιούργησαν μια μέθοδο για την ταυτοποίηση ασθενών με στένωση στεφανιαίας αρτηρίας από το μυοκάρδιο του LV από CT\@.
Χρησιμοποίησαν ένα CNN πολλαπλών κλιμάκων για να τμηματοποιήσουν το μυοκάρδιο του LV και έπειτα το κωδικοποίησαν χρησιμοποιώντας ένα μη-επιβλεπώμενο συνελικτικό ΑΕ.
Η τελική ταξινόμηση έγινε με έναν ταξινομητή SVM με βάση τα εξαγόμενα και ομαδοποιημένα clustering.
Παρόμοια δουλειά έχει γίνει από τους ίδιους συγγραφείς~\cite{zreik2016automatic}, οι οποίοι χρησιμοποίησαν τρία CNNs για να ανιχνεύσουν ένα πλαίσιο οριοθέτησης γύρω από την LV και εφάρμοσαν ταξινόμηση voxel LV μέσα στο κιβώτιο οριοθέτησης.
Οι Commandeur et al.~\cite{commandeur2018deep} χρησιμοποίησαν ένα συνδυασμό δύο βαθιών δικτύων για τον ποσοτικό προσδιορισμό του επικαρδιακού και θωρακικού λιπώδες ιστού, με CT από 250 ασθενείς με 55 εικόνες ανά ασθενή κατά μέσο όρο.
Το πρώτο δίκτυο είναι ένα CNN έξι \textit{επιπέδων}, που ανιχνεύει την περιοχή που βρίσκεται μέσα στα όρια της καρδιάς και διαχωρίζει τις θωρακικές και επικαρδιακές-παρακαρδιακές μάσκες.
Το δεύτερο δίκτυο είναι ένα CNN πέντε \textit{επιπέδων}, που ανιχνεύει τη γραμμή του περικαρδίου από την αξονική τομογραφία σε κυλινδρικές συντεταγμένες.
Στη συνέχεια ένα στατιστικό μοντέλο συστηματοποίησης μαζί με εφαρμογή κατωφλίωσης και φιλτράρισμα διαμέσου, παρέχουν τις τελικές κατατμήσεις.
Οι Gulsun et al.~\cite{gulsun2016coronary} δημιούργησαν μια μέθοδο για την εξαγωγή των κεντρικών γραμμών των αγγείων στην CT\@.
Πρώτον, οι βέλτιστες διαδρομές ανιχνεύονται σε ένα υπολογισμένο πεδίο ροής και στη συνέχεια χρησιμοποιείται ένας ταξινομητής CNN για την αφαίρεση των εξωτερικών διαδρομών στις ανιχνευμένες κεντρικές γραμμές.
Η μέθοδος ενισχύθηκε χρησιμοποιώντας μια ανίχνευση βασισμένη στο μοντέλο των συγκεκριμένων στεφανιαίων περιοχών και των κύριων κλάδων για να περιορίσουν τον χώρο αναζήτησης.

\section{Ηχοκαρδιογράφημα}
Το ηχοκαρδιογράφημα είναι μια μέθοδος απεικόνισης της περιοχής της καρδιάς χρησιμοποιώντας υπερηχητικά κύματα.
Χρήσεις της βαθιάς μάθησης στο ηχοκαρδιογράφημα περιλαμβάνουν κυρίως την κατάτμηση του LV και την αξιολόγηση της ποιότητας της βαθμολογίας της εικόνας, μεταξύ άλλων.

\begin{sidewaystable}
	\caption{Εφαρμογές βαθιάς μάθησης με χρήση Ηχοκαρδιογραφήματος}
	\label{table:imaging7}
	\centering
	\begin{tabular}{l c l}
		\toprule
		\thead{Αναφορά}                              & \thead{Μέθοδος} & \thead{Εφαρμογή/Σημειώσεις\footnote{Αποτελέσματα από αυτές τις απεικονιστικές τεχνικές δεν δημοσιεύονται σε αυτήν την βιβλιογραφική αναφορά καθώς υπάρχει υψηλή μεταβλητότητα σε σχέση με το ερευνητικό ερώτημα προς απάντηση και στην χρήση των μέτρων. Επιπλέον όλες οι δημοσιεύσεις χρησιμοποιούν μη-δημόσιες βάσεις δεδομένων.}} \\
		\midrule
		Carneiro 2012~\cite{carneiro2012segmentation} & DBN             & τμηματοποίηση LV με αποσύνδεση στοιβαρών και μη-στοιβαρών ανιχνεύσεων με χρήση DBN σε 480 εικόνες                                                                                                                                                                                                                                                                                           \\
		Nascimento 2016~\cite{nascimento2016multi}    & DBN             & τμηματοποίηση LV με μάθηση επιφανειών και ένα DBN                                                                                                                                                                                                                                                                                                                                     \\
		Chen 2016~\cite{chen2016iterative}            & CNN             & τμηματοποίηση LV με συστηματοποιημένο FCN πολλαπλών τομέων και μεταφορά μάθησης                                                                                                                                                                                                                                                                                                      \\
		Madani 2018~\cite{madani2018fast}             & CNN             & ταξινόμηση όψης διαθωρακικού ηχοκαρδιογραφήματος με CNN έξι \textit{επιπέδων} \\
		Silva 2018~\cite{silva2018ejection}           & CNN             & ταξινόμηση κλάσματος εξώθησης με ένα residual 3D CNN και διαθωρακικά ηχοκαρδιογραφήματα                                                                                                                                                                                                                                                                                               \\
		Gao 2017~\cite{gao2017fused}                  & CNN             & ταξινόμηση όψης με σύντηξη δύο CNNs με επτά \textit{επίπεδα} το καθένα                                                                                                                                                                                                                                                                                                                      \\
		Abdi 2017~\cite{abdi2017quality}              & CNN, LSTM       & αξιολόγηση ποιότητας βαθμολογίας με συνελικτικά και επαναληπτικά επιπέδα                                                                                                                                                                                                                                                                                                           \\
		Ghesu 2016~\cite{ghesu2016marginal}           & CNN             & τμηματοποίηση αορτικής βαλβίδας με χρήση 2891 3D διαεισοφαγικών ηχοκαρδιογραφικών εικόνων                                                                                                                                                                                                                                                                                                   \\
		Perrin 2017~\cite{perrin2017application}      & CNN             & ταξινόμηση εκ-γενετής καρδιακής ασθένειας με CNN                                                                                                                                                                                                                                                                                                                                      \\
		Moradi 2016~\cite{moradi2016cross}            & VGGnet, doc2vec & δημιουργία σημαντικών περιγραφέων για εικόνες                                                                                                                                                                                                                                                                                                                                               \\
		\bottomrule
	\end{tabular}
\end{sidewaystable}

Τα DBN έχουν χρησιμοποιηθεί για την κατάτμηση της LV στα ηχοκαρδιογραφήματα.
Στο~\cite{carneiro2012segmentation} οι συγγραφείς δημιούργησαν μια μέθοδο που αποσυνδέει τις άκαμπτες και μη-άκαμπτες ανιχνεύσεις, με ένα DBN που μοντελοποιεί το LV δείχνοντας ότι είναι πιο ισχυρό από τα level-sets και τα παραμορφώσιμα μοντέλα.
Οι Nascimento et al.~\cite{nascimento2016multi} χρησιμοποιούν μάθηση του πολυειδούς που χωρίζει τα δεδομένα σε patches όπου το κάθε ένα προτείνει μια κατάτμηση της LV\@.
Η σύντηξη των patches πραγματοποιήθηκε από έναν πολλαπλό ταξινομητή DBN ο οποίος αποδίδει ένα βάρος σε κάθε patch.
Με τον τρόπο αυτό, η μέθοδος δεν βασίζεται σε μια μόνο κατάτμηση και η διαδικασία εκπαίδευσης παράγει ισχυρά μοντέλα χωρίς την ανάγκη μεγάλων βάσεων δεδομένων εκπαίδευσης.
Στο~\cite{chen2016iterative} οι συγγραφείς χρησιμοποίησαν ένα συστηματοποιημένο FCN και μεταφορά μάθησης.
Συγκρίνουν τη μέθοδο τους με απλούστερες αρχιτεκτονικές FCN και μια προηγούμενη μέθοδο παρουσιάζοντας καλύτερα αποτελέσματα.

Το ηχοκαρδιογράφημα έχει επίσης χρησιμοποιηθεί για την ταξινόμηση των όψεων της απεικόνισης.
Οι Madani et al.~\cite{madani2018fast} εκπαίδευσαν CNN έξι \textit{επιπέδων} για να ταξινομήσουν 15 προβολές (12 βίντεο και 3 ακίνητα) με διαθωρακικό υπερηχογράφημα, επιτυγχάνοντας καλύτερα αποτελέσματα από τους ειδικούς ηχοκαρδιογραφήματος.
Στο~\cite{silva2018ejection} οι συγγραφείς δημιούργησαν ένα residual 3D CNN για ταξινόμηση κλάσματος εξώθησης από εικόνες διαθωρακικού ηχοκαρδιογραφήματος.
Χρησιμοποίησαν 8715 εξετάσεις κάθε μια με 30 διαδοχικές εικόνες του apical θαλάμου για να εκπαιδεύσουν και να δοκιμάσουν τη μέθοδο τους επιτυγχάνοντας ικανοποιητικά αποτελέσματα.
Οι Gao et al.~\cite{gao2017fused} ενσωμάτωσαν χωρική και χρονική πληροφορία από τις εικόνες του βίντεο της κινούμενης καρδιάς, με τη σύντηξη δύο CNNs με επτά \textit{επίπεδα} το καθένα.
Η μέτρηση της επιτάχυνσης σε κάθε σημείο υπολογίστηκε με τη χρήση μεθόδου πυκνής οπτικής ροής, για την απεικόνιση πληροφοριών χρονικής κίνησης.
Στη συνέχεια η σύντηξη των CNN πραγματοποιήθηκε χρησιμοποιώντας γραμμικές ενσωματώσεις των διανυσμάτων των εξόδων τους.
Συγκρίσεις έγιναν με προηγούμενες προσεγγίσεις βασισμένες σε χειροκίνητη δημιουργία χαρακτηριστικών, καταδεικνύοντας καλύτερα αποτελέσματα.

Τέλος, η αξιολόγηση της ποιότητας των εικόνων και άλλα προβλήματα επιλύθηκαν με τη χρήση του ηχοκαρδιογραφήματος.
Στο~\cite{abdi2017quality} οι συγγραφείς δημιούργησαν μια μέθοδο για τη μείωση της μεταβλητότητας των δεδομένων κατα τη διάρκεια της επισήμανσης από τον χειριστή, υπολογίζοντας μια βαθμολογία ποιότητας που ανατροφοδοτείται σε πραγματικό χρόνο.
Το μοντέλο συνίσταται από συνελικτικά επίπεδα για την εξαγωγή χαρακτηριστικών από την είσοδο και επαναλαμβανόμενα επίπεδα για χρήση των χρονικών πληροφοριών.
Η μέθοδος των Ghesu et al.~\cite{ghesu2016marginal} ανίχνευσης αντικειμένων και κατάτμησης στο πλαίσιο της ογκομετρικής ανάλυσης εικόνων, γίνεται με την επίλυση της ανατομικής εκτίμησης της θέσης και της οριοθέτησης.
Για το σκοπό αυτό εισάγουν βαθιά μάθηση της οριοθέτησης, η οποία παρέχει υψηλής απόδοσης χρόνο εκτέλεσης, μαθαίνοντας ταξινομητές σε περιοχές υψηλής πιθανότητας και σε χώρους σταδιακά αυξανόμενων διαστάσεων.
Δεδομένου του εντοπισμού των αντικειμένων, προτείνουν ένα συνδυασμένο μοντέλο ενεργού σχήματος βαθιάς μάθησης για την εκτίμηση του μη-άκαμπτου ορίου αντικειμένου.
Στο άρθρο τους οι Perrin et al.~\cite{perrin2017application} εκπαίδευσαν και αξιολόγησαν το AlexNet με 59151 εικόνες, για να ταξινομήσουν μεταξύ πέντε παιδιατρικών πληθυσμών με συγγενή καρδιακή νόσο.
Οι Moradi et al.~\cite{moradi2016cross} δημιούργησαν μια μέθοδο που βασίζεται σε VGGnet και doc2vec για να παράξουν σημασιολογικούς περιγραφείς εικόνων, οι οποίοι μπορούν να χρησιμοποιηθούν ως ασθενώς επισημασμένες περιπτώσεις ή να διορθωθούν από ιατρικούς εμπειρογνώμονες.
Το μοντέλο τους ήταν σε θέση να αναγνωρίσει το 91\% των ασθενειών και 77\% των σοβαρών νόσων από εικόνες Doppler καρδιακών βαλβίδων.

\section{Τομογραφία οπτικής συνοχής}
Η Τομογραφία Οπτικής Συνοχής (Optical Coherence Tomography, OCT) είναι μια ενδοαγγειακή απεικόνιση που παρέχει εικόνες αρτηριών υψηλής ανάλυσης και ποσοτικές μετρήσεις της στεφανιαίας γεωμετρίας στο κλινικό περιβάλλον~\cite{kubo2013oct}.

Στο άρθρο τους οι Roy et al.~\cite{roy2016multiscale} χαρακτηρίζουν τον ιστό στο OCT, μαθαίνοντας την πολυκλιμακωτή κατανομή των δεδομένων με ένα AE\@.
Ο κανόνας μάθησης του δικτύου εισάγει μια παράμετρο κλίμακας που συνδέεται με το backpropagation.
Σε σύγκριση με τρία προεκπαιδευμένα ΑΕ βάσης αναφοράς επιτυγχάνεται καλύτερη απόδοση όσον αφορά την ακρίβεια στην ανίχνευση πλακών/κανονικών εικονοστοιχείων.
Οι Yong et al.~\cite{yong2017linear} δημιούργησαν ένα CNN γραμμικής παλινδρόμησης με τέσσερα \textit{επίπεδα}, για να χωρίσουν τον αυλό του αγγείου παραμετροποιημένο σε όρους ακτινικών αποστάσεων από το κέντρο του καθετήρα στις πολικές συντεταγμένες.
Η υψηλή ακρίβεια αυτής της μεθόδου μαζί με την υπολογιστική αποτελεσματικότητά της (40.6ms/εικόνα), υποδεικνύουν τη δυνατότητα χρήσης σε πραγματικό κλινικό περιβάλλον.
Στο~\cite{xu2017fibroatheroma} οι συγγραφείς σύγκριναν την διακριτική ικανότητα των βαθιών χαρακτηριστικών που εξάγονται από τα AlexNet, GoogleNet, VGGnet16 και VGGnet19 για την ταυτοποίηση του ινωαθερώματος.
Η επαύξηση δεδομένων εφαρμόστηκε σε μια βάση δεδομένων OCT για κάθε σχήμα ταξινόμησης και επίσης εφαρμόστηκε γραμμικό SVM για την ταξινόμηση των εικόνων ως κανονικών ή ινωαθερωματικών.
Τα αποτελέσματα υποδεικνύουν ότι το VGGnet19 είναι καλύτερο για τον εντοπισμό εικόνων που περιέχουν ινωαθέρωμα.
Οι Abdolmanafi et al.~\cite{abdolmanafi2017deep} ταξινομούν τον ιστό στο OCT χρησιμοποιώντας ένα προεκπαιδευμένο AlexNet ως εξαγωγέα χαρακτηριστικών και συγκρίνουν τις προβλέψεις τριών ταξινομητών, CNN, RF και SVM, με το πρώτο να επιτυγχάνει τα καλύτερα αποτελέσματα.

\section{Άλλες απεικονιστικές τεχνικές}
Το ενδοαγγειακό υπερηχογράφημα (Intravascular Ultrasound, IVUS) χρησιμοποιεί μετασχηματιστές τοποθετημένους σε τροποποιημένους ενδοστεφανιαίους καθετήρες για την παροχή ακτινικής ανατομικής απεικόνισης ενδοστεφανιαίας ασβεστοποίησης και σχηματισμού πλάκας~\cite{parrillo2013critical}.
Οι Lekadir et al.~\cite{lekadir2017convolutional} χρησιμοποίησαν ένα βασισμένο σε patches CNN τεσσάρων \textit{επιπέδων}, για τον χαρακτηρισμό της σύνθεσης της πλάκας σε καρωτιδικές εικόνες υπερήχων.
Τα πειράματα που έγιναν από τους συγγραφείς έδειξαν ότι το μοντέλο επιτυγχάνει καλύτερη ακρίβεια από SVM μονών και πολλαπλών κλιμάκων.
Στο~\cite{tajbakhsh2017automatic} οι συγγραφείς αυτοματοποίησαν ολόκληρη τη διαδικασία της ερμηνείας του πάχους του καρωτιδικού intima-media.
Εκπαίδευσαν ένα CNN δύο \textit{επιπέδων} με δύο εξόδους για την επιλογή της εικόνας και ένα CNN δύο \textit{επιπέδων} με τρεις εξόδους για τον εντοπισμό του ROI και τις μετρήσεις πάχους intima-media.
Το μοντέλο αυτό αποδίδει καλύτερα από προηγούμενη μέθοδο από τους ίδιους συγγραφείς, το οποίο δικαιολογεί την ικανότητα του CNN να μαθαίνει την εμφάνιση του QRS και του ROI, αντί να στηρίζεται σε εφαρμογή κατωφλίωσης με κατώτατο πλάτος και καμπυλότητα.
Οι Tom et al.~\cite{tom2018simulating} δημιούργησαν μια μέθοδο βασισμένη στα Δίκτυα Γεννήτριας-Διάκρισης (Generative Adversarial Networks, GANs) για χρονικά αποδοτική προσομοίωση ρεαλιστικού IVUS\@.
Η προσομοίωση στο πρώτο στάδιο αφορούσε τον προσομοιωτή ψευδότροπου IVUS και την χαρτογράφηση του στίγματος της ψηφιακά καθορισμένης εικόνας.
Στο δεύτερο στάδιο βελτιώθηκαν οι αντιστοιχίσεις για τη διατήρηση των συγκεκριμένων εντάσεων στίγματος με ιστούς χρησιμοποιώντας ένα GAN με τέσσερα residual \textit{επίπεδα}, ενώ στο τελευταίο στάδιο το GAN παρήξε εικόνες υψηλής αναλυτικότητας με παθορεαλιστικά προφίλ στίγματος.

\begin{sidewaystable}
	\caption{Εφαρμογές βαθιάς μάθησης με χρήση OCT και άλλων απεικονιστικών τεχνικών}
	\label{table:imaging8}
	\centering
	\begin{tabular}{l c l}
		\toprule
		\thead{Αναφορά}                             & \thead{Μέθοδος} & \thead{Εφαρμογή/Σημειώσεις\footnote{Αποτελέσματα από αυτές τις απεικονιστικές τεχνικές δεν δημοσιεύονται σε αυτήν την βιβλιογραφική αναφορά καθώς υπάρχει υψηλή μεταβλητότητα σε σχέση με το ερευνητικό ερώτημα προς προς απάντηση και στην χρήση των μετρήσεων. Επιπλέον όλες οι δημοσιεύσεις χρησιμοποιούν μη-δημόσιες βάσεις δεδομένων εκτός της~\cite{tom2018simulating}.}} \\
		\midrule
		\multicolumn{3}{l}{\thead{OCT}}                                                                                                                                                                                                                                                                                                                                                                                                                             \\
		\midrule
		Roy 2016~\cite{roy2016multiscale}            & AE              & χαρακτηρισμός ιστού με χρήση ενός AE διατήρησης της κατανομής                                                                                                                                                                                                                                                                                                                               \\
		Yong 2017~\cite{yong2017linear}              & CNN             & τμηματοποίηση του lumen με χρήση ενός CNN γραμμικής-παλινδρόμησης με τέσσερα \textit{επίπεδα}                                                                                                                                                                                                                                                                                               \\
		Xu 2017~\cite{xu2017fibroatheroma}           & CNN             & παρουσία του ινωαθερώματος με χρήση χαρακτηριστικών εξαγόμενων από προηγούμενες αρχιτεκτονικές και ένα SVM                                                                                                                                                                                                                                                                                   \\
		Abdolmanafi 2017~\cite{abdolmanafi2017deep}  & CNN             & τμηματοποίηση του intima-media με χρήση ενός προεκπαιδευμένου AlexNet και σύγκριση διαφόρων ταξινομητών                                                                                                                                                                                                                                                                                     \\
		\midrule
		\multicolumn{3}{l}{\thead{Άλλες απεικονιστικές τεχνικές}}                                                                                                                                                                                                                                                                                                                                                                                                   \\
		\midrule
		Lekadir 2017~\cite{lekadir2017convolutional} & CNN             & χαρακτηρισμός καρωτιδικής πλάκας με χρήση CNN τεσσάρων \textit{επιπέδων} με υπερήχους                                                                                                                                                                                                                                                                                                       \\
		Tajbakhsh 2017~\cite{tajbakhsh2017automatic} & CNN             & ερμηνεία video του πάχους των καρωτιδικών intima-media με χρήση δύο CNNs δύο \textit{επίπεδα} με υπερήχους                                                                                                                                                                                                                                                                                 \\
		Tom 2017~\cite{tom2018simulating}            & GAN             & δημιουργία IVUS εικόνων με χρήση δύο GANs (IV11)                                                                                                                                                                                                                                                                                                                                            \\
		Wang 2017~\cite{wang2017detecting}           & CNN             & ασβεστοποίηση αρτηριών μαστού με χρήση CNN δέκα \textit{επιπέδων} σε μαστογραφίες                                                                                                                                                                                                                                                                                                           \\
		Liu 2017~\cite{liu2017coronary}              & CNN             & ανίχνευση CAC με χρήση CNNs σε 1768 X-Rays                                                                                                                                                                                                                                                                                                                                                  \\
		Pavoni 2017~\cite{pavoni2017image}           & CNN             & αποθορυβοποίηση του διαδερμικού αυλού στεφανιαίων εικόνων αγγειοπλαστικής με ένα CNN τεσσάρων \textit{επιπέδων}                                                                                                                                                                                                                                                                             \\
		Nirschl 2018~\cite{nirschl2018deep}          & CNN             & CNN έξι \textit{επιπέδων} βασισμένο σε patch για ανίχνευση καρδιακής ανεπάρκειας σε ενδομυοκαρδιακές εικόνες βιοψίας                                                                                                                                                                                                                                                                        \\
		Betancur 2018~\cite{betancur2018deep}        & CNN             & CNN τριών \textit{επιπέδων} για προβλεψη CAD από απεικονίσεις του μυοκαρδίου                                                                                                                                                                                                                                                                                                                \\
		\bottomrule
	\end{tabular}
\end{sidewaystable}

Άλλες εφαρμογές καρδιολογίας με χρήση μεθόδων βαθιάς μάθησης, περιλαμβάνουν μαστογραφίες, ακτίνες Χ, διαδερμική διαφραγματική αγγειοπλαστική, εικόνες βιοψίας και απεικόνιση διάχυσης του μυοκαρδίου.
Στο άρθρο τους οι Wang et al.~\cite{wang2017detecting} εφάρμοσαν μια διαδικασία βασισμένη σε εικονοστοιχεία και patches για την ανίχνευση ασβεστοποιητικού αρτηριακού μαστού σε μαστογραφίες, χρησιμοποιώντας ένα CNN δέκα \textit{επιπέδων} και μορφολογικές λειτουργίες για μετα-επεξεργασία.
Οι συγγραφείς χρησιμοποίησαν 840 εικόνες και τα πειράματά τους οδήγησαν σε ένα μοντέλο που πέτυχε συντελεστή προσδιορισμού 96.2\%.
Οι Liu et al.~\cite{liu2017coronary} εκπαίδευσαν ένα CNN χρησιμοποιώντας 1768 εικόνες ακτίνων Χ με αντίστοιχες επισημάνσεις διαγνώσεων.
Η μέση διαγνωστική ακρίβεια των μοντέλων έφτασε στο 0.89 όταν το βάθος του δικτύου ήταν οκτώ επίπεδα; μετά από αυτό η αύξηση της ακρίβειας ήταν περιορισμένη.
Στο~\cite{pavoni2017image} οι συγγραφείς δημιούργησαν μια μέθοδο για την αποθορυβοποίηση εικόνων διαδερμικής αγγειοπλαστικής στεφανιαίας.
Δοκίμασαν το μέσο τετραγωνικό σφάλμα και τη δομική ομοιότητα ως συναρτήσεις απωλειών σε δύο patch CNN με τέσσερα \textit{επίπεδα} και τα σύγκριναν με διαφορετικούς τύπους και επίπεδα θορύβου.
Οι Nirschl et al.~\cite{nirschl2018deep} χρησιμοποίησαν εικόνες ενδομυοκαρδιακής βιοψίας από 209 ασθενείς για να εκπαιδεύσουν και να δοκιμάσουν ένα patch CNN έξι \textit{επιπέδων}, για τον εντοπισμό της καρδιακής ανεπάρκειας.
Επίσης εφάρμοσαν περιστροφική επαύξηση δεδομένων, ενώ υπολογίστηκε ο μέσος όρος των εξόδων του CNN σε κάθε patch για να ληφθεί η πιθανότητα για κάθε εικόνα.
Αυτό το μοντέλο έδειξε καλύτερα αποτελέσματα από τα AlexNet, Inception και ResNet50.
Στο~\cite{betancur2018deep} οι συγγραφείς εκπαίδευσαν ένα CNN τριών \textit{επιπέδων} για την πρόβλεψη της αποφρακτικής αιμάτωσης μυοκαρδίου του CAD από 1638 ασθενείς και το συνέκριναν με το συνολικό έλλειμμα διάχυσης.
Το μοντέλο υπολογίζει την πιθανότητα ανά αγγείο κατά τη διάρκεια της εκπαίδευσης, ενώ κατά τη διάρκεια της δοκιμής χρησιμοποιούνται οι μέγιστες πιθανότητες ανά αρτηρία ανά βαθμολογία ασθενούς.
Τα αποτελέσματα δείχνουν ότι αυτή η μέθοδος υπερέχει του συνολικού ελλείμματος διάχυσης στο πρόβλημα της πρόβλεψης των αγγείων και των ασθενών.

\section{Συμπεράσματα της χρήσης βαθιάς μάθησης με CT, ηχοκαρδιογράφημα, OCT και άλλων απεικονιστικών τεχνικών}
Δημοσιεύσεις που χρησιμοποίησαν αυτούς τους τρόπους απεικόνισης παρουσίασαν υψηλή μεταβλητότητα από την πλευρά του ερευνητικού τους ερωτήματος και ήταν κατά πλειοψηφία ασυμβίβαστα σε σχέση με τη χρήση μέτρων αξιολόγησης για τα αποτελέσματα που ανέφεραν.
Αυτές οι μέθοδοι απεικόνισης επίσης δεν έχουν διαθέσιμες δημόσιες βάσεις δεδομένων, περιορίζοντας έτσι τις ευκαιρίες για δημιουργία νέων αρχιτεκτονικών από ερευνητικές ομάδες που δεν έχουν κλινικούς συνεργάτες.
Από την άλλη πλευρά, υπάρχει σχετικά υψηλή ομοιομορφία όσον αφορά τη χρήση αρχιτεκτονικών με το πιο διαδεδομένο το CNN, ειδικά τις προεκπαιδευμένες αρχιτεκτονικές του διαγωνισμού ImageNet (AlexNet, VGGnet, GoogleNet, ResNet).

\section{Συζήτηση και μελλοντικές κατευθύνσεις}
\label{sec4:discussion}

\begin{sidewaystable}
	\caption{Βιβλιογραφικές αναφορές βαθιάς μάθησης με εφαρμογή στην καρδιολογία}
	\label{table:reviews}
	\centering
	\begin{tabular}{l l}
		\toprule
		\thead{Αναφορά}                                    & \thead{Εφαρμογή/Σημειώσεις}                                                                                     \\
		\midrule
		Mayer 2015~\cite{mayer2015big}                      & Big data στην καρδιολογία αλλάζει τον τρόπο με τον οποίο δημιουργούνται καινούργιες ιδέες                       \\
		Austin 2016~\cite{austin2016application}            & σφαιρική εικόνα των big data, τα πλεονεκτήματα, πιθανά μειονεκτήματα και μελλοντική συνεισφορά στην καρδιολογία \\
		Greenspan 2016~\cite{greenspan2016guest}            & ανίχνευση ιστών, τμηματοποίηση και μοντελοποίηση ενεργού σχήματος                                               \\
		Miotto 2017~\cite{miotto2017deep}                   & απεικονιστικές, EHR, γονιδιωματική, δεδομένα από φορητές συσκευές και η ανάγκη για αύξηση της ερμηνευσιμότητας  \\
		Krittanawong 2017~\cite{krittanawong2017rise}       & μελέτες πάνω στις τεχνολογίες αναγνώρισης εικόνας που προβλέπουν καλύτερα από τους ειδικούς                     \\
		Litjens 2017~\cite{litjens2017survey}               & ταξινόμηση εικόνων, εντοπισμός αντικειμένου, τμηματοποίηση και εγγραφή (registration)                           \\
		Qayyum 2017~\cite{qayyum2017medical}                & μέθοδοι βασισμένοι στο CNN στην τμηματοποίηση εικόνας, ταξινόμηση, διάγνωση και ανάκτηση εικόνας                \\
		Hengling 2017~\cite{henglin2017machine}             & επίπτωση που θα έχει η μηχανική μάθηση στο μέλλον της καρδιαγγειακής απεικόνισης                                \\
		Blair 2017~\cite{blair2017advanced}                 & πρόοδος στη νευροεπιστήμη με MRI στην νόσο των μικρών αγγείων                                                   \\
		Slomka 2017~\cite{slomka2017cardiac}                & πυρηνική καρδιολογία, CT αγγειογραφία Ηχοκαρδιογράφημα, MRI                                                      \\
		Carneiro 2017~\cite{carneiro2017review}             & μαστογραφία, καρδιαγγειακή και μικροσκοπική απεικόνιση                                                          \\
		Johnson 2018~\cite{johnson2018artificial}           & AI στην καρδιολογία, έννοιες προβλεπτικών μοντέλων, κοινοί αλγόριθμοι και χρήση της βαθιάς μάθησης               \\
		Jiang 2017~\cite{jiang2017artificial}               & εφαρμογές του AI στην αναγνώριση της καρδιακής προσβολής, διάγνωση, θεραπεία, πρόβλεψη και εκτίμηση πρόγνωσης   \\
		Lee 2017~\cite{lee2017deepb}                        & AI στην απεικονιστική καρδιακής προσβολής εστιασμένο στις τεχνικές αρχές και κλινικές εφαρμογές                 \\
		Loh 2017~\cite{loh2017deep}                         & διάγνωση καρδιακής ασθένειας και διαχείριση μέσα στο πλαίσιο της υγειονομικής περίθαλψης                        \\
		Krittanawong 2017~\cite{krittanawong2017artificial} & καρδιαγγειακή κλινική περίθαλψη και ο ρόλος της ακριβής καρδιαγγειακής ιατρικής                                 \\
		Gomez 2018~\cite{gomez2018new}                      & πρόσφατη πρόοδος στην αυτοματοποίηση και ποσοτική ανάλυση στην πυρηνική καρδιολογία                             \\
		Shameer 2018~\cite{shameer2018machine}              & υποσχέσεις και περιορισμοί της υλοποίησης της μηχανικής μάθησης στην καρδιαγγειακή ιατρική                      \\
		Shrestha 2018~\cite{shrestha2018machine}            & εφαρμογές μηχανικής μάθησης στην πυρηνική καρδιολογία                                                           \\
		Kikuchi 2018~\cite{kikuchi2018future}               & εφαρμογές του AI στην πυρηνική καρδιολογία και το πρόβλημα των περιορισμένων αριθμών δεδομένων                  \\
		Awan 2018~\cite{awan2018machine}                    & εφαρμογές μηχανικής μάθησης στην διάγνωση καρδιακής ανεπάρκειας, ταξινόμηση και πρόβλεψη επανεισδοχής           \\
		Faust 2018~\cite{faust2018deep}                     & εφαρμογές βαθιάς μάθησης στα φυσιολογικά δεδομένα συμπεραλαμβανομένου του ECG                                   \\
		\bottomrule
	\end{tabular}
\end{sidewaystable}

Είναι προφανές από τη βιβλιογραφία ότι οι μέθοδοι βαθιάς μάθησης θα αντικαταστήσουν τα συστήματα που βασίζονται σε χειροποίητους κανόνες και την παραδοσιακή μηχανική μάθηση.
Στο~\cite{awan2018machine} οι συγγραφείς υποστηρίζουν ότι η βαθιά μάθηση είναι καλύτερη στην απεικόνιση πολύπλοκων μοτίβων κρυμμένων σε ιατρικά δεδομένα υψηλής διαστάσεων.
Οι Krittanawong et al.~\cite{krittanawong2017artificial} υποστηρίζουν ότι η αυξανόμενη διαθεσιμότητα αυτοματοποιημένων εργαλείων AI πραγματικού χρόνου στα EHRs, θα μειώσει την ανάγκη για συστήματα βαθμολόγησης όπως το σκορ Framingham.
Στο~\cite{krittanawong2017rise} οι συγγραφείς υποστηρίζουν ότι οι προγνωστικές αναλύσεις του AI και η εξατομικευμένη κλινική υποστήριξη για την αναγνώριση του ιατρικού κινδύνου, είναι ανώτερες από τις ανθρώπινες γνωσιακές ικανότητες.
Επιπλέον, το AI μπορεί να διευκολύνει την επικοινωνία μεταξύ ιατρών και ασθενών, μειώνοντας τους χρόνους επεξεργασίας και αυξάνοντας έτσι την ποιότητα της περίθαλψης των ασθενών.
Οι Loh et al.~\cite{loh2017deep} υποστηρίζουν ότι οι τεχνολογίες βαθιάς μάθησης και κινητής τηλεφωνίας θα επιταχύνουν τον πολλαπλασιασμό των υπηρεσιών υγειονομικής περίθαλψης σε εκείνους που βρίσκονται σε φτωχές περιοχές, γεγονός που με τη σειρά του οδηγεί σε περαιτέρω μείωση των ποσοστών ασθενειών.
Οι Mayer et al.~\cite{mayer2015big} δηλώνουν ότι τα big data υπόσχονται να αλλάξουν την καρδιολογία μέσω της αύξησης των δεδομένων που συλλέγονται, αλλά ο αντίκτυπος τους υπερβαίνει τη βελτίωση των υπαρχουσών μεθόδων όπως η αλλαγή του τρόπου με τον οποίο δημιουργούνται νέες ιδέες.

Η βαθιά μάθηση απαιτεί δεδομένα εκπαίδευσης μεγάλου όγκου για την επίτευξη αποτελεσμάτων υψηλής ποιότητας~\cite{krizhevsky2012imagenet}.
Αυτό είναι ιδιαίτερα δύσκολο με τα ιατρικά δεδομένα, λόγω του ότι η διαδικασία επισήμανσης των ιατρικών δεδομένων είναι δαπανηρή; απαιτεί χειρωνακτική εργασία από ιατρικούς εμπειρογνώμονες.
Επιπλέον, τα περισσότερα ιατρικά δεδομένα ανήκουν στις κανονικές περιπτώσεις αντί στις μη-κανονικές, καθιστώντας τα εξαιρετικά μη-ισορροπημένα.
Άλλες προκλήσεις της εφαρμογής της βαθιάς μάθησης στην ιατρική που έχει εντοπίσει η βιβλιογραφία είναι τα ζητήματα τυποποίησης/διαθεσιμότητας/διαστάσεων/όγκου/ποιότητας, δυσκολία στην απόκτηση των αντίστοιχων επισημάνσεων και θόρυβος στις επισημάνσεις~\cite{litjens2017survey, greenspan2016guest, miotto2017deep, slomka2017cardiac}.
Πιο συγκεκριμένα, στο~\cite{blair2017advanced} οι συγγραφείς σημειώνουν ότι οι εφαρμογές βαθιάς μάθησης στην ασθένεια μικρών αγγείων, έχουν αναπτυχθεί χρησιμοποιώντας μόνο λίγα αντιπροσωπευτικά σύνολα δεδομένων και πρέπει να αξιολογηθούν σε μεγάλα σύνολα δεδομένων πολλαπλών κέντρων.
Οι Kikuchi et al.~\cite{kikuchi2018future} αναφέρουν ότι σε σύγκριση με τα CT και MRI, οι απεικονιστικές της πυρηνικής καρδιολογίας έχουν περιορισμένο αριθμό εικόνων ανά ασθενή και απεικονίζεται μόνο συγκεκριμένος αριθμός οργάνων.
Ο Liebeskind~\cite{liebeskind2018artificial} αναφέρει ότι οι μέθοδοι μηχανικής μάθησης δοκιμάζονται σε επίλεκτα και ομοιογενή κλινικά δεδομένα, αλλά η γενικευσιμότητα θα προέκυπτε χρησιμοποιώντας ετερογενή και πολύπλοκα δεδομένα.
Το ισχαιμικό αγγειακό εγκεφαλικό επεισόδιο αναφέρεται ως παράδειγμα μιας ετερογενούς και σύνθετης ασθένειας, όπου η απόφραξη της μεσαίας εγκεφαλικής αρτηρίας μπορεί να οδηγήσει σε αποκλίνουσες μορφές απεικόνισης.
Στο~\cite{gomez2018new} οι συγγραφείς καταλήγουν στο συμπέρασμα ότι απαιτούνται πρόσθετα δεδομένα που επικυρώνουν αυτές τις εφαρμογές σε μη-ελεγχόμενες κλινικές ρυθμίσεις πολλαπλών κέντρων, πριν από την κλινική εφαρμογή τους.
Πρέπει επίσης να διερευνηθεί ο αντίκτυπος αυτών των εργαλείων στην λήψη αποφάσεων, χρήση πόρων, και κόστος.
Επιπλέον, η παρούσα βιβλιογραφία έδειξε ότι υπάρχει μια άνιση κατανομή των διαθέσιμων στο κοινό βάσεων δεδομένων ανάμεσα στις διάφορες απεικονιστικές τεχνικές στην καρδιολογία (π.χ. δεν υπάρχει διαθέσιμη δημόσια βάση δεδομένων για τα OCT σε αντίθεση με τα MRI).

Η προηγούμενη βιβλιογραφία αναφέρει ότι τα προβλήματα που σχετίζονται με τα δεδομένα μπορούν να λυθούν με τεχνικές επαύξησης δεδομένων, την ανοικτή συνεργασία μεταξύ των ερευνητικών οργανισμών και την αύξηση της χρηματοδότησης.
Οι Hengling et al.~\cite{henglin2017machine} υποστηρίζουν ότι θα χρειαστούν σημαντικές επενδύσεις για τη δημιουργία επισημασμένων βάσεων δεδομένων υψηλής ποιότητας, οι οποίες είναι απαραίτητες για την επιτυχία των επιβλεπώμενων μεθόδων βαθιάς μάθησης.
Στο~\cite{austin2016application} οι συγγραφείς υποστηρίζουν ότι η επιτυχία αυτού του τομέα εξαρτάται από τις τεχνολογικές εξελίξεις στην πληροφορική και την αρχιτεκτονική των υπολογιστών, καθώς και τη συνεργασία και την ανοικτή ανταλλαγή δεδομένων μεταξύ των γιατρών και άλλων ενδιαφερομένων.
Οι Lee et al.~\cite{lee2017deepb} καταλήγουν στο συμπέρασμα ότι απαιτείται διεθνής συνεργασία για την κατασκευή ενός υψηλής ποιότητας πολυτροπικού συνόλου δεδομένων για απεικόνιση εγκεφαλικών επεισοδίων.
Μια άλλη λύση για την καλύτερη αξιοποίηση των δεδομένων μεγάλου όγκου στην καρδιολογία είναι η εφαρμογή μη-επιβλεπώμενων μεθόδων μάθησης, οι οποίες δεν απαιτούν επισημάνσεις.
Η παρούσα βιβλιογραφική αναφορά έδειξε ότι η μη-επιβλεπώμενη μάθηση δεν χρησιμοποιείται ευρέως, καθώς η πλειονότητα των μεθόδων σε όλες τις απεικονιστικές τεχνικές είναι επιβλεπώμενες.

Όσον αφορά το πρόβλημα της έλλειψης ερμηνευσιμότητας όπως υποδεικνύεται από τον Hinton~\cite{hinton2018deep}, είναι γενικά ανέφικτο να ερμηνευτούν τα μη-γραμμικά χαρακτηριστικά των βαθιών δικτύων, επειδή το νόημά τους εξαρτάται από πολύπλοκες αλληλεπιδράσεις με μη-ερμηνεύσιμα χαρακτηριστικά από άλλα επίπεδα.
Επιπλέον, αυτά τα μοντέλα είναι στοχαστικά, που σημαίνει ότι κάθε φορά που ένα δίκτυο εκπαιδεύεται με τα ίδια δεδομένα αλλά με διαφορετικά αρχικά βάρη, μαθαίνει διαφορετικά χαρακτηριστικά.
Πιο συγκεκριμένα σε μια εκτενή ανασκόπηση~\cite{betancur2018deep} του αν επιλύεται το πρόβλημα της τμηματοποίησης LV/RV, οι συγγραφείς δηλώνουν ότι αν και η πτυχή ταξινόμησης του προβλήματος επιτυγχάνει σχεδόν τέλεια αποτελέσματα, η χρήση ενός `διαγνωστικού μαύρου κουτιού' δεν μπορεί να ενσωματωθεί στην κλινική πρακτική.
Οι Miotto et al.~\cite{miotto2017deep} αναφέρουν την ερμηνευσιμότητα ως μία από τις κύριες προκλήσεις που αντιμετωπίζει η κλινική εφαρμογή της βαθιάς μάθησης στην υγειονομική περίθαλψη.
Στο~\cite{lee2017deepb} οι συγγραφείς σημειώνουν ότι η ιδιότητα του μαύρου κουτιού που έχουν μερικές ΑΙ μέθοδοι, όπως η βαθιά μάθηση, είναι αντίθετη με την έννοια της τεκμηριωμένης ιατρικής και εγείρει νομικά και ηθικά ζητήματα στη χρήση τους στην κλινική πρακτική.
Αυτή η έλλειψη ερμηνευσιμότητας είναι ο κύριος λόγος για τον οποίο οι ιατρικοί εμπειρογνώμονες αντιστέκονται στη χρήση αυτών των μοντέλων και υπάρχουν επίσης νομικοί περιορισμοί όσον αφορά την ιατρική χρήση των μη-ερμηνευόμενων εφαρμογών~\cite{slomka2017cardiac}.
Από την άλλη πλευρά, κάθε μοντέλο μπορεί να τοποθετηθεί σε έναν άξονα~\cite{beam2018big} `ανθρώπου-μηχανής', συμπεριλαμβανομένων στατιστικών που οι ιατρικοί εμπειρογνώμονες βασίζονται στην καθημερινή λήψη κλινικών αποφάσεων.
Για παράδειγμα, οι ανθρώπινες αποφάσεις όπως η επιλογή των μεταβλητών που πρέπει να συμπεριληφθούν στο μοντέλο, η σχέση εξαρτημένων και ανεξάρτητων μεταβλητών και μεταβλητών μετασχηματισμών, μετακινούν τον αλγόριθμο προς τον ανθρώπινο άξονα αποφάσεων, καθιστώντας τον πιο ερμηνεύσιμο αλλά ταυτόχρονα περισσότερο πιο επιρρεπή σε σφάλματα.

Όσον αφορά την επίλυση του προβλήματος ερμηνευσιμότητας, όταν νέες μέθοδοι είναι απαραίτητες οι ερευνητές είναι προτιμητέο να δημιουργούν απλούστερες μεθόδους βαθιάς μάθησης (από-άκρο-σε-άκρο και μη-ensembles) για να αυξήσουν την πιθανότητα κλινικής εφαρμογής τους, ακόμα κι αν αυτό σημαίνει μειωμένη αναφερόμενη ακρίβεια.
Υπάρχουν επίσης επιχειρήματα κατά της δημιουργίας νέων μεθόδων, επικεντρώνοντας στην επικύρωση των ήδη υφιστάμενων.
Στο~\cite{damen2016prediction} οι συγγραφείς καταλήγουν στο συμπέρασμα ότι υπάρχει πληθώρα μοντέλων που προβλέπουν περιστατικά CVD στο γενικό πληθυσμό.
Η χρησιμότητα των περισσότερων μοντέλων είναι ασαφής λόγω σφαλμάτων στη μεθοδολογία και έλλειψης εξωτερικών μελετών επικύρωσης.
Αντί να αναπτυχθούν νέα μοντέλα πρόβλεψης κινδύνου CVD, η μελλοντική έρευνα πρέπει να επικεντρωθεί στην επικύρωση και σύγκριση υφιστάμενων μοντέλων και να διερευνήσει εάν μπορούν να βελτιωθούν περαιτέρω.

Μια δημοφιλής μέθοδος που χρησιμοποιείται για ερμηνεύσιμα μοντέλα είναι τα δίκτυα προσοχής~\cite{bahdanau2014neural}.
Τα δίκτυα προσοχής είχαν ως αρχική έμπνευση την ικανότητα εστίασης της ανθρώπινης όρασης σε ένα συγκεκριμένο σημείο με υψηλή ανάλυση και την αντίληψη του περιβάλλοντος με χαμηλή ανάλυση.
Έχουν χρησιμοποιηθεί από αρκετές δημοσιεύσεις στην καρδιολογία στην πρόβλεψη του ιατρικού ιστορικού~\cite{kim2017highrisk}, ταξινόμηση χτύπων ECG~\cite{schwab2017beat} και την πρόβλεψη CVD με χρήση Fundus~\cite{poplin2017predicting}.
Ένα άλλο απλούστερο εργαλείο για την ερμηνεία είναι οι χάρτες αξιοπιστίας (saliency maps)~\cite{simonyan2013deep} που χρησιμοποιούν την κλίση της εξόδου σε σχέση με την είσοδο, η οποία δείχνει διαισθητικά τις περιοχές που συνεισφέρουν περισσότερο στην έξοδο.

Εκτός από την επίλυση των προβλημάτων των δεδομένων και της ερμηνείας, οι ερευνητές στην καρδιολογία θα μπορούσαν να χρησιμοποιήσουν τις ήδη καθιερωμένες αρχιτεκτονικές βαθιάς μάθησης που δεν έχουν εφαρμοστεί ευρέως στην καρδιολογία, όπως τα δίκτυα καψουλών (Capsule Networks, CapsNets).
Τα CapsNet~\cite{sabour2017dynamic} είναι βαθιά νευρωνικά δίκτυα που χρειάζονται λιγότερα δεδομένα εκπαίδευσης από τα CNN και τα επίπεδά τους αποτυπώνουν τον `προσανατολισμό' των χαρακτηριστικών, καθιστώντας έτσι τις εσωτερικές τους λειτουργίες πιο ερμηνεύσιμες και πιο κοντά στον ανθρώπινο τρόπο αντίληψης.
Εντούτοις, ένα σημαντικό μειονέκτημα που έχουν, το οποίο τους περιορίζει από την επίτευξη ευρύτερης χρήσης, είναι το υψηλό υπολογιστικό κόστος σε σύγκριση με τα CNN λόγω του αλγορίθμου `δρομολόγησης με συμφωνία'.
Μεταξύ των πρόσφατων χρήσεών τους στην ιατρική περιλαμβάνονται η ταξινόμηση όγκων στον εγκέφαλο~\cite{afshar2018brain} και η ταξινόμηση του καρκίνου του μαστού~\cite{iesmantas2018convolutional}.
Τα CapsNet δεν έχουν χρησιμοποιηθεί ακόμη σε καρδιολογικά δεδομένα.

Μια άλλη λιγότερο χρησιμοποιούμενη αρχιτεκτονική βαθιάς μάθησης στην καρδιολογία είναι τα GANs~\cite{goodfellow2014generative}, τα οποία αποτελούνται από ένα δίκτυο γεννήτρια που δημιουργεί ψεύτικες εικόνες από θόρυβο και ένα δίκτυο διάκρισης που είναι υπεύθυνο για τη διαφοροποίηση μεταξύ πλαστών εικόνων από τη γεννήτρια και πραγματικών εικόνων.
Και τα δύο δίκτυα προσπαθούν να βελτιστοποιήσουν μια απώλεια σε ένα παιχνίδι με μηδενικό άθροισμα, με αποτέλεσμα μια γεννήτρια που παράγει ρεαλιστικές εικόνες.
Τα GAN έχουν χρησιμοποιηθεί μόνο για την προσομοίωση παθο-ρεαλιστικών εικόνων IVUS~\cite{tom2018simulating} και ο τομέας της καρδιολογίας έχει πολλά να κερδίσει από τη χρήση αυτού του είδους των μοντέλων, ειδικά λόγω της έλλειψης επισημασμένων δεδομένων υψηλής ποιότητας.

Οι ερευνητές θα μπορούσαν επίσης να χρησιμοποιήσουν CRF, τα οποία είναι γραφικά μοντέλα που συλλαμβάνουν πληροφορίες από γειτονικές περιοχές και είναι σε θέση να ενσωματώσουν στατιστικά στοιχεία υψηλότερης τάξης, τα οποία οι παραδοσιακές μέθοδοι βαθιάς μάθησης δεν είναι σε θέση να κάνουν.
Εκπαιδευμένα CRFs από κοινού με CNNs έχουν χρησιμοποιηθεί σε εκτίμηση του βάθους στην ενδοσκόπηση~\cite{mahmood2018deep} και την κατάτμηση του ήπατος στην CT~\cite{christ2016automatic}.
Υπάρχουν επίσης εφαρμογές καρδιολογίας που χρησιμοποίησαν CRF με βαθιά μάθηση για βελτίωση της τμηματοποίησης στα Fundus~\cite{zhou2017improving} και σε LV/RV~\cite{bai2017semi}.
Η πολυτροπική βαθιά μάθηση~\cite{ngiam2011multimodal} μπορεί επίσης να χρησιμοποιηθεί για τη βελτίωση των διαγνωστικών αποτελεσμάτων, π.χ. τη δυνατότητα συνδυασμού δεδομένων fMRI και ECG\@.
Ειδικές βάσεις δεδομένων πρέπει να δημιουργηθούν για να αυξηθεί η έρευνα στον τομέα αυτό, καθώς σύμφωνα με την τρέχουσα ανασκόπηση υπάρχουν μόνο τρεις βάσεις δεδομένων καρδιολογίας με πολυτροπικά δεδομένα.
Εκτός από τις προηγούμενες βάσεις δεδομένων, η MIMIC-III έχει επίσης χρησιμοποιηθεί για πολυτροπική βαθιά μάθηση από~\cite{purushotham2018benchmarking} για την πρόβλεψη ενδονοσοκομειακής, βραχυπρόθεσμης/μακροπρόθεσμης θνησιμότητας και πρόβλεψης του κώδικα ICD-9.

\clearpage
\bibliography{chapter3.bib}
\bibliographystyle{unsrt}

%% file: chapter5.tex
\chapter{Επίπεδα Signal2Image σε βαθιά νευρωνικά δίκτυα για ταξινόμηση EEG}
\label{chapter5}
\graphicspath{{./images/signal2image-modules-in-deep-neural-networkds-for-eeg-classification/}}

\section{Εισαγωγή}
Η βαθιά μάθηση έχει φέρει επανάσταση στην όραση υπολογιστών, χρησιμοποιώντας την αυξημένη διαθεσιμότητα βάσεων δεδομένων μεγάλου όγκου και τη δύναμη των παράλληλων υπολογιστικών μονάδων, όπως είναι οι μονάδες επεξεργασίας γραφικών.
Η συντριπτική πλειοψηφία της έρευνας στην βαθιά μάθηση γίνεται με τη χρήση εικόνων ως δεδομένων εκπαίδευσης, ωστόσο ο βιοϊατρικός τομέας είναι πλούσιος σε σήματα φυσιολογίας που χρησιμοποιούνται για προβλήματα διάγνωσης και πρόβλεψης.
Είναι ακόμα ανοιχτό ερευνητικό ερώτημα, η καλύτερη αξιοποίηση σημάτων για την εκπαίδευση βαθιών νευρωνικών δικτύων.

Οι περισσότερες μέθοδοι για την επίλυση βιοϊατρικών προβλημάτων μέχρι πρόσφατα, περιελάμβαναν χειροποίητη δημιουργία χαρακτηριστικών και προσπάθεια μίμησης ανθρώπινων εμπειρογνωμόνων, τα οποία αποδεικνύονται όλο και περισσότερο ανεπαρκή και επιρρεπή σε σφάλματα.
Η βαθιά μάθηση αναδεικνύεται ως μια ισχυρή λύση για ένα ευρύ φάσμα προβλημάτων στη βιοϊατρική, που επιτυγχάνει καλύτερα αποτελέσματα σε σύγκριση με την παραδοσιακή μηχανική μάθηση.
Το κύριο πλεονέκτημα των μεθόδων που χρησιμοποιούν τη βαθιά μάθηση είναι ότι μαθαίνουν ιεραρχικά χαρακτηριστικά από δεδομένα εκπαίδευσης με αυτόματο τρόπο, κάτι το οποίο τα καθιστά πιο κλιμακωτά και γενικεύσιμα από παραδοσιακές μεθόδους.
Αυτό επιτυγχάνεται με τη χρήση δικτύων πολλαπλών επιπέδων που αποτελούνται από εκατομμύρια παραμέτρους~\cite{krizhevsky2012imagenet}, εκπαιδευμένα με backpropagation~\cite{rumelhart1986learning} σε μεγάλο όγκο δεδομένων.
Παρόλο που η βαθιά μάθηση χρησιμοποιείται κυρίως σε βιοϊατρικές εικόνες υπάρχει επίσης ένα ευρύ φάσμα φυσιολογικών σημάτων, όπως το EEG, τα οποία χρησιμοποιούνται για προβλήματα διάγνωσης και πρόβλεψης.
Το EEG είναι μια μέτρηση του ηλεκτρικού πεδίου που παράγεται από τον εγκέφαλο και χρησιμοποιείται για την ταξινόμηση σταδίων του ύπνου~\cite{aboalayon2016sleep}, τις διεπαφές υπολογιστή-εγκεφάλου~\cite{al2017review}, την συναισθηματική παρακολούθηση~\cite{lotte1999electroencephalography} και την ταξινόμηση της επιληψίας~\cite{acharya2013automated}.

Οι Yannick et al.~\cite{yannick2019deep} εξέτασαν δημοσιεύσεις βαθιάς μάθησης με χρήση του EEG και εντόπισαν μια γενική αύξηση της ακρίβειας όταν χρησιμοποιείται βαθιά μάθηση και συγκεκριμένα όταν χρησιμοποιούνται CNN αντί παραδοσιακών μεθόδων μηχανικής μάθησης.
Ωστόσο δεν αναφέρουν συγκεκριμένα χαρακτηριστικά των αρχιτεκτονικών CNN που ευθύνονται για την αύξηση της απόδοσης.
Είναι ακόμα ανοιχτό ερευνητικό ερώτημα, το πως μπορεί να χρησιμοποιηθεί το EEG για την εκπαίδευση μοντέλων βαθιάς μάθησης.

Μια κοινή προσέγγιση που χρησιμοποίησαν προηγούμενες μελέτες για την ταξινόμηση των σημάτων EEG ήταν η εξαγωγή χαρακτηριστικών από το πεδίο συχνοτήτων και χρόνου--συχνοτήτων, που χρησιμοποιούν τη θεωρία πίσω από τις συχνότητες των ζωνών EEG~\cite{langkvist2012sleep}: δέλτα (0.5--4 Hz), θήτα (4--8 Hz), άλφα (8--13 Hz), βήτα (13--20 Hz) και γάμμα (20--64 Hz).
Οι Truong et al.~\cite{truong2018convolutional} χρησιμοποίησαν STFT σε παράθυρο ολίσθησης 30 δευτερολέπτων, για να εκπαιδεύσουν ένα CNN τριών επιπέδων σε αναπαραστάσεις πεδίου χρόνου-συχνοτήτων, για πρόβλεψη επιληπτικών κρίσεων και αξιολόγησαν τη μέθοδο τους σε τρεις βάσεις δεδομένων EEG\@.
Οι Khan et al.~\cite{khan2018focal} μετασχημάτισαν τα EEG στο πεδίο χρόνου-συχνοτήτων χρησιμοποιώντας κυματίδια πολλαπλών κλιμάκων και έπειτα εκπαίδευσαν ένα CNN έξι επιπέδων, για την πρόβλεψη της εστιακής έναρξης παρουσιάζοντας υποσχόμενα αποτελέσματα.

Η εξαγωγή χαρακτηριστικών από το πεδίο χρόνου-συχνοτήτων έχει επίσης χρησιμοποιηθεί και σε άλλες δημοσιεύσεις που σχετίζονται με το EEG, εκτός από την πρόγνωση επιληπτικών κρίσεων.
Οι Zhang et al.~\cite{zhang2017pattern} εκπαίδευσαν ένα ensemble CNN που περιείχε δύο έως δέκα επίπεδα, χρησιμοποιώντας χαρακτηριστικά STFT που εξήχθησαν από τις EEG συχνότητες για ταξινόμηση ψυχικού φόρτου εργασίας.
Οι Giri et al.~\cite{giri2016ischemic} εξήγαγαν στατιστικά και μέτρα πληροφορίας από το πεδίο συχνοτήτων για την εκπαίδευση ενός 1D CNN με δύο επίπεδα, για τον εντοπισμό ισχαιμικού εγκεφαλικού επεισοδίου.

Σε αυτό το κεφάλαιο και στο πλαίσιο της διδακτορικής διατριβής ορίζουμε τα επίπεδα νευρωνικών δικτύων Signal2Image (S2I) ως κάθε δομικό στοιχείο που τοποθετείται μετά την είσοδο μη-επεξεργασμένου σήματος και πριν από ένα `μοντέλο βάσης' που είναι συνήθως μια καθιερωμένη αρχιτεκτονική για προβλήματα εικόνων.
Μία σημαντική ιδιότητα ενός S2I είναι αν αποτελείται από παραμέτρους που μπορούν να εκπαιδευτούν όπως συνελικτικά και γραμμικά επίπεδα ή είναι μη-εκπαιδεύσιμα όπως οι παραδοσιακές μέθοδοι χρόνου-συχνοτήτων.
Χρησιμοποιώντας αυτόν τον ορισμό μπορούμε επίσης να καταλήξουμε στο συμπέρασμα ότι οι περισσότερες προηγούμενες μέθοδοι για την ταξινόμηση EEG χρησιμοποιούν μη-εκπαιδεύσιμα S2I και ότι καμία προηγούμενη μελέτη δεν έχει συγκρίνει εκπαίδευσιμα με μη-εκπαιδεύσιμα S2I.

Συγκρίνουμε τα εκπαιδεύσιμα και μη-εκπαιδεύσιμα S2Is σε συνδυασμό με τις γνωστές αρχιτεκτονικές νευρωνικών δικτύων `μοντέλων βάσης', μαζί με τις 1D και τις σε βάθος παραλλαγές των τελευταίων.
Μια επισκόπηση υψηλού επιπέδου αυτών των συνδυασμένων μεθόδων παρουσιάζεται στην Εικ.~\ref{fig:highleveloverview}.
Αν και επιλέγουμε το σύνολο δεδομένων αναγνώρισης επιληπτικών ελλείψεων EEG από το Πανεπιστήμιο της Καλιφόρνιας, Irvine (University of California, UCI)~\cite{andrzejak2001indications} για ταξινόμηση EEG, οι συνέπειες αυτής της μελέτης θα μπορούσαν να γενικευτούν σε οποιοδήποτε είδος προβλήματος ταξινόμησης σημάτων.
Εδώ επίσης αναφερόμαστε στο CNN ως ένα νευρωνικό δίκτυο που αποτελείται από εναλλασσόμενα συνελικτικά επίπεδα ακολουθούμενα από ένα ReLU και ένα επίπεδο μέγιστης συγκέντρωσης και ένα πλήρως συνδεδεμένο επίπεδο στο τέλος, ενώ ο όρος `επίπεδο' υποδηλώνει τον αριθμό των συνελικτικών επιπέδων.

\section{Δεδομένα}
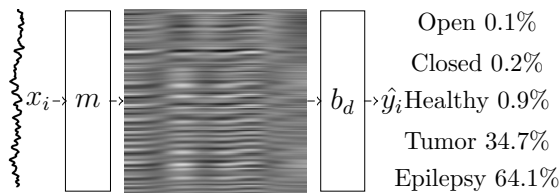
\begin{figure}
	\centering
	\begin{tikzpicture}[]
		\node[] at (-4.5, 0){\includegraphics[scale=0.2,angle=90]{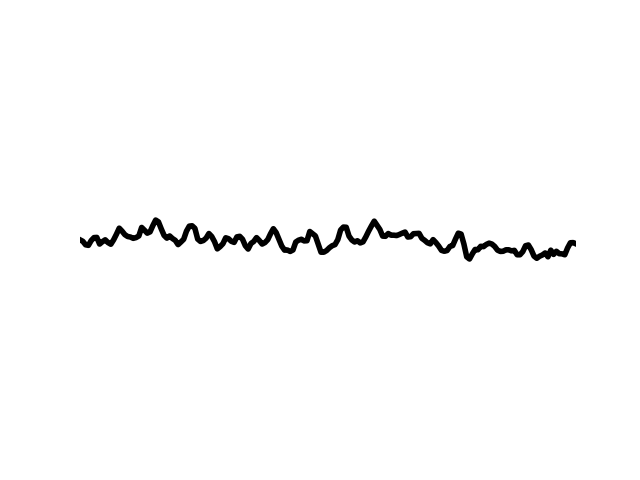}};
		\node[align=left] at (-4.2, 0) {$x_i$};
		\draw[dashed,->] (-4, 0) -- (-3.8, 0);
		\node[draw, minimum height=2.55cm] at (-3.5, 0) {$m$};
		\draw[dashed,->] (-3.2, 0) -- (-3, 0);
		\node[] at (-1.7, 0){\includegraphics[scale=0.41,angle=90]{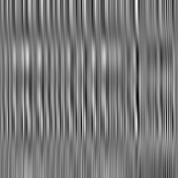}};
		\draw[dashed,->] (-0.4, 0) -- (-0.2, 0);
		\node[draw, minimum height=2.55cm] at (0.1, 0) {$b_d$};
		\draw[dashed,->] (0.4, 0) -- (0.6, 0);
		\node[align=left] at (0.8, 0) {$\hat{y_i}$};
		\node[minimum width=0.5cm, minimum height=0.5cm] at (2, 1.1) {\footnotesize Open 0.1\%};
		\node[minimum width=0.5cm, minimum height=0.5cm] at (2, 0.55) {\footnotesize Closed 0.2\%};
		\node[minimum width=0.5cm, minimum height=0.5cm] at (2, 0) {\footnotesize Healthy 0.9\%};
		\node[minimum width=0.5cm, minimum height=0.5cm] at (2, -0.55) {\footnotesize Tumor 34.7\%};
		\node[minimum width=0.5cm, minimum height=0.5cm] at (2, -1.1) {\footnotesize Epilepsy 64.1\%};
	\end{tikzpicture}
	\caption[Αρχιτεκτονική Signal2Image]{Αρχιτεκτονική Signal2Image.
	$x_i$ είναι η είσοδος, $m$ είναι το Signal2Image, $b_d$ είναι η 1D ή 2D αρχιτεκτονική του `μοντέλου βάσης' για $d=1,2$ αντίστοιχα και $\hat{y_i}$ είναι η προβλεπόμενη έξοδος.
	Τα ονόματα των επισημάνσεων απεικονίζονται δεξιά μαζί με τις προβλέψεις για το απεικονιζόμενο σήμα.
	Η εικόνα μεταξύ του $m$ και του $b_d$ απεικονίζει την έξοδο του CNN Signal2Image ενός επιπέδου, ενώ το `σήμα ως εικόνα' και φασματογράφημα έχουν ενδιάμεσες εικόνες όπως απεικονίζονται στην δεύτερη και τρίτη σειρά του σχήματος~\ref{fig:signal2imageoutputs}.
	Τα βέλη απεικονίζουν την ροή της προς-τα-εμπρός διάδοσης.
	Για την 1D αρχιτεκτονική το $m$ παραλείπεται και δεν δημιουργείται ενδιάμεση εικόνα.}
	\label{fig:highleveloverview}
\end{figure}

Η βάση δεδομένων αναγνώρισης επιληπτικών κρίσεων UCI EEG~\cite{andrzejak2001indications} αποτελείται από $500$ σήματα το καθένα με $4097$ δείγματα (23.5 δευτερόλεπτα).
Αποτελείται από πέντε κατηγορίες με $100$ σήματα για κάθε κλάση (σε παρένθεση τα συντομευμένα ονόματα των επισημάνσεων που χρησιμοποιούνται στα Σχήματα~\ref{fig:highleveloverview} και~\ref{fig:signal2imageoutputs}):
\begin{enumerate}
	\item υγιής ασθενής ενώ έχει τα μάτια του ανοιχτά (Open),
	\item υγιής ασθενής, έχοντας τα μάτια κλειστά (Closed),
	\item ασθενής με όγκο με σήμα που λαμβάνεται από υγιή περιοχή (Healthy),
	\item ασθενής με όγκο με σήμα που λαμβάνεται από την περιοχή του όγκου (Tumor),
	\item ασθενής ενώ έχει επιληπτική δραστηριότητα (Epilepsy)
\end{enumerate}

Για τους σκοπούς αυτού του κεφαλαίου χρησιμοποιούμε μια παραλλαγή της βάσης δεδομένων\footnote{\url{https://archive.ics.uci.edu/ml/datasets/Epileptic+Seizure+Recognition}} στην οποία τα σήματα EEG χωρίζονται σε τμήματα με $178$ δείγματα το καθένα, καταλήγοντας σε ένα ισορροπημένο σύνολο δεδομένων που αποτελείται από $11500$ σήματα EEG\@.

\section{Μέθοδοι}
\subsection{Ορισμοί}
Ορίζουμε το σύνολο δεδομένων $D=\{x_i, y_i\}_{i=1\ldots N}$ όπου $x_i \in \mathbb{Z}^n$ και $y_i \in \{1, 2, 3, 4, 5\}$ υποδηλώνει το $i^{th}$ σήμα εισόδου με διαστάσεις $n=178$ και την $i^{th}$ επισήμανση με πέντε πιθανές κατηγορίες αντίστοιχα.
$N=11500$ είναι ο αριθμός των παρατηρήσεων.

Επίσης, ορίζουμε το σύνολο των S2Is ως $M$ και το μέλος αυτού του συνόλου ως $m$ το οποίο περιλαμβάνει τα ακόλουθα:
\begin{itemize}
	\item 'σήμα ως εικόνα' (μη-εκπαιδεύσιμο)
	\item φασματογράφημα (μη-εκπαιδεύσιμο)
	\item CNN ενός και δύο επιπέδων (εκπαιδεύσιμο)
\end{itemize}

Καθορίζουμε στη συνέχεια το σύνολο των `μοντέλων βάσης' ως $B$ και το μέλος αυτού του συνόλου ως $b_d$ όπου $d=[1,2]$ υποδηλώνει τη διαστασιμότητα των επιπέδων συνέλιξης (convolutional), μέγιστης συγκέντρωσης (max-pooling) και κανονικοποίησης παρτίδας (batch-norm).
Το $B$ περιλαμβάνει τα παρακάτω $b_d$ μαζί με τις παραλλαγές σε βάθος και τις ισοδύναμες 1D αρχιτεκτονικές για $d=1$ (για μια πλήρη λίστα ανατρέξτε στις δύο πρώτες σειρές του Πίνακα.~\ref{table:results}):
\begin{itemize}
	\item LeNet~\cite{lecun1998gradient}
	\item AlexNet~\cite{krizhevsky2012imagenet}
	\item VGGnet~\cite{simonyan2014very}
	\item ResNet~\cite{he2016deep}
	\item DenseNet~\cite{huang2017densely}
\end{itemize}

Τέλος ορίσαμε τους συνδυασμούς των $m$ και $b_d$ ως μέλη $c$ του συνόλου συνδυασμένων μοντέλων $C$.
Χρησιμοποιώντας τους προηγούμενους ορισμούς, ο στόχος αυτού του κεφαλαίου είναι η αξιολόγηση του συνόλου των μοντέλων $C$, όπου $C$ είναι το συνδυασμένο σύνολο των $M$ και $B$ δλδ. $C=M\times B$ σε σχέση με την χρονική απόδοση και την ακρίβεια εκπαιδευμένα στο $D$.

\subsection{Signal2Image}
Σε αυτή την ενότητα περιγράφονται τα εσωτερικά στοιχεία κάθε μονάδας S2I.
Για το S2I `σήμα ως εικόνα' κανονικοποιήσαμε το πλάτος $x_i$ στο εύρος $[1, 178]$.
Τα αποτελέσματα αντιστράφηκαν κατά μήκος του άξονα y, στρογγυλοποιημένα στον πλησιέστερο ακέραιο και στη συνέχεια χρησιμοποιήθηκαν ως y-δείκτες για τα εικονοστοιχεία με πλάτος $255$ σε μια εικόνα $178\times 178$ αρχικοποιημένη με μηδενικά.

Για το S2I φασματογράφου, το οποίο χρησιμοποιείται για την απεικόνιση της μεταβολής της συχνότητας ενός μη-στάσιμου σήματος με την πάροδο του χρόνου~\cite{oppenheim1999discrete}, χρησιμοποιήθηκε παράθυρο Tukey με παράμετρο σχήματος $0.25$, μήκος τμήματος $8$ δειγμάτων, αλληλεπικάλυψη μεταξύ τμημάτων $4$ δειγμάτων και ταχύ Μετασχηματισμό Fourier των $64$ δειγμάτων για την μετατροπή του $x_i$ στο πεδίο χρόνου-συχνοτήτων.
Το προκύπτων φασματογράφημα, το οποίο αντιπροσωπεύει το μέγεθος της φασματικής πυκνότητας ισχύος ($V^2/Hz$) του $x_i$, στη συνέχεια υπερδειγματολήφθηκε σε $178\times 178$ με χρήση διγραμμικής παρεμβολής μεταξύ των εικονοστοιχείων.

Για τα S2I CNN με ένα και δύο επίπεδα, το $x_i$ μετατρέπεται σε εικόνα χρησιμοποιώντας εκπαιδεύσιμες παραμέτρους, αντί χρήσης κάποιας στατικής διαδικασίας.
Το S2I ενός επιπέδου αποτελείται από ένα 1D συνελικτικό επίπεδο (μεγέθη πυρήνα $3$ με $8$ κανάλια).
Το S2I δύο επιπέδων αποτελείται από δύο 1D συνελικτικά επίπεδα (μεγέθη πυρήνα $3$ με $8$ και $16$ κανάλια) με το πρώτο επίπεδο να ακολουθείται από μια συνάρτηση ενεργοποίησης ReLU και ένα επίπεδο 1D max-pooling (μέγεθος πυρήνα $2$).
Οι χάρτες ενεργοποιήσεων του τελευταίου συνελικτικού επιπέδου και για τα δύο S2I συμπτύσσονται σειριακά κατά μήκος του y-άξονα και στη συνέχεια μετατρέπονται σε μέγεθος $178\times 178$ χρησιμοποιώντας τη διγραμμική παρεμβολή.

\begin{figure}
	\centering
	\subfloat{\includegraphics[scale=0.16]{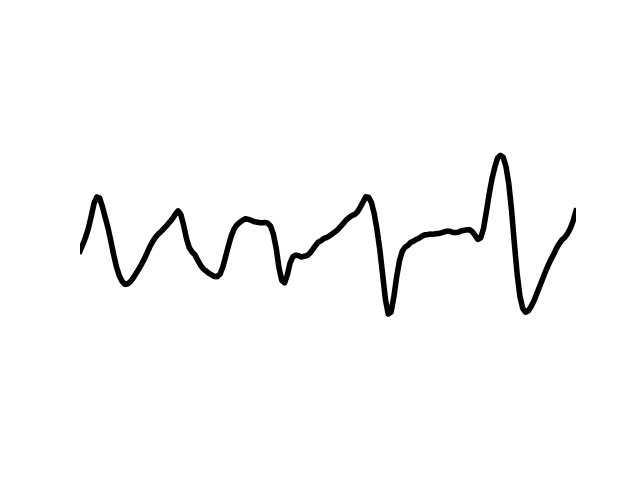}}
	\quad
	\subfloat{\includegraphics[scale=0.16]{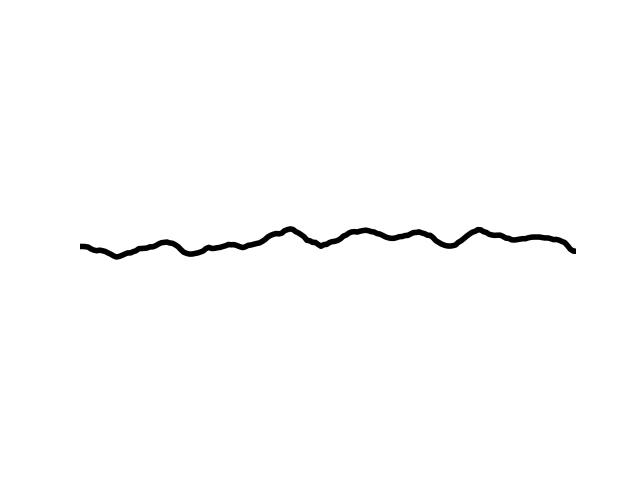}}
	\quad
	\subfloat{\includegraphics[scale=0.16]{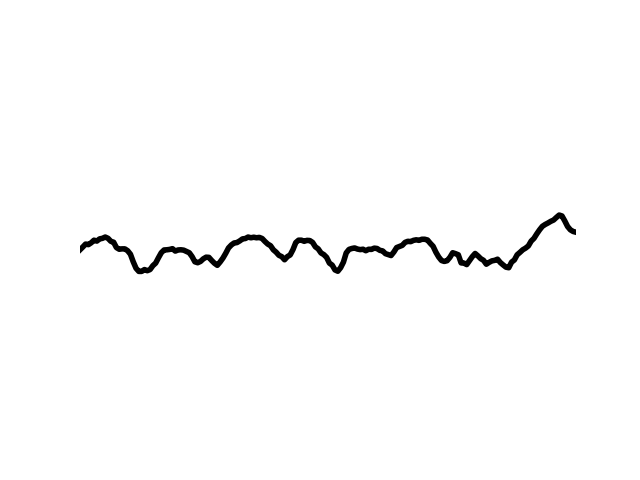}}
	\quad
	\subfloat{\includegraphics[scale=0.16]{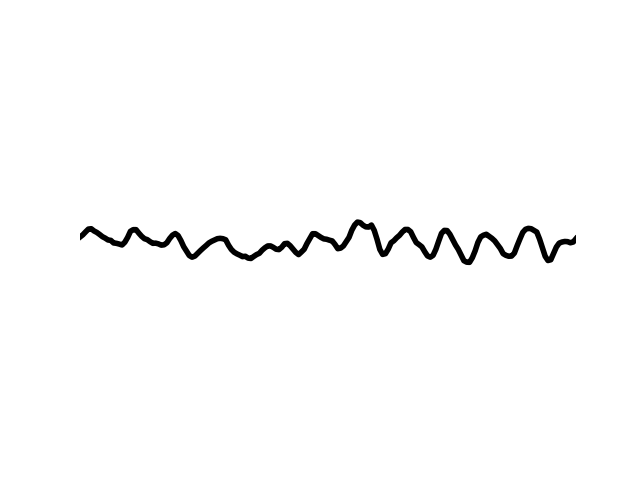}}
	\quad
	\subfloat{\includegraphics[scale=0.16]{signal-epilepsy.png}}
	\\
	\subfloat{\includegraphics[scale=0.42]{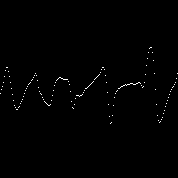}}
	\quad
	\subfloat{\includegraphics[scale=0.42]{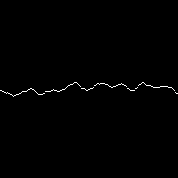}}
	\quad
	\subfloat{\includegraphics[scale=0.42]{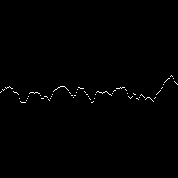}}
	\quad
	\subfloat{\includegraphics[scale=0.42]{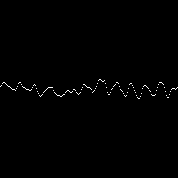}}
	\quad
	\subfloat{\includegraphics[scale=0.42]{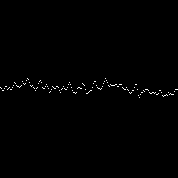}}
	\\
	\subfloat{\includegraphics[scale=0.42]{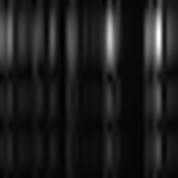}}
	\quad
	\subfloat{\includegraphics[scale=0.42]{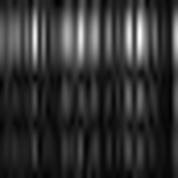}}
	\quad
	\subfloat{\includegraphics[scale=0.42]{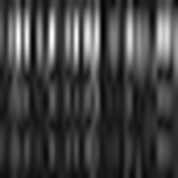}}
	\quad
	\subfloat{\includegraphics[scale=0.42]{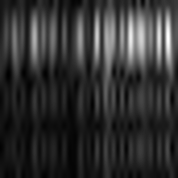}}
	\quad
	\subfloat{\includegraphics[scale=0.42]{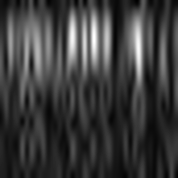}}
	\\
	\setcounter{subfigure}{0}
	\subfloat[Open]{\includegraphics[scale=0.42]{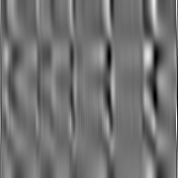}}
	\quad
	\subfloat[Closed]{\includegraphics[scale=0.42]{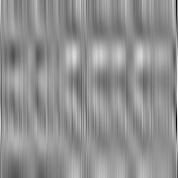}}
	\quad
	\subfloat[Healthy]{\includegraphics[scale=0.42]{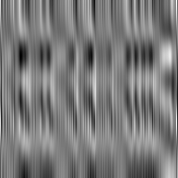}}
	\quad
	\subfloat[Tumor]{\includegraphics[scale=0.42]{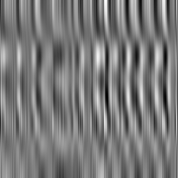}}
	\quad
	\subfloat[Epilepsy]{\includegraphics[scale=0.42]{cnn-epilepsy.png}}
	\caption[Σήματα και έξοδοι των Signal2Image για κάθε κλάση]{Σήματα και έξοδοι των Signal2Image για κάθε κλάση.
	Οι x, y-άξονες της πρώτης σειράς είναι \SI{}{\micro V} και χρονικά δείγματα αντίστοιχα.
	Οι x, y-άξονες των υπόλοιπων εικόνων απεικονίζουν χωρική πληροφορία, επειδή δεν ενημερώνουμε το `μοντέλο βάσης' με την έννοια του χρόνου κατά τον x-άξονα ή την έννοια της συχνότητας κατά τον y-άξονα.
	Υψηλότερη ένταση των εικονοστοιχείων υποδηλώνει υψηλότερο πλάτος.}
	\label{fig:signal2imageoutputs}
\end{figure}

Περιορίζουμε την έξοδο για όλα τα $m$ σε μια εικόνα μεγέθους $178\times 178$ για να επιτρέψουμε οπτική σύγκριση.
Τρία πανομοιότυπα κανάλια στοιβάζονται επίσης για όλες τις εξόδους των $m$ για να ικανοποιήσουν τις απαιτήσεις μεγέθους εισόδου των $b_d$.
Οι αρχιτεκτονικές όλων των $b_d$ παραμένουν οι ίδιες, εκτός του αριθμού των εξόδων του τελευταίου γραμμικού επιπέδου που έχει οριστεί σε πέντε για να αντιστοιχεί στον αριθμό των κατηγοριών του $D$.
Ένα παράδειγμα των αντίστοιχων εξόδων κάποιων από τα $m$ (το CNN ενός/δύο επιπέδων παρήγαγαν παρόμοιες απεικονίσεις) απεικονίζονται στη δεύτερη, τρίτη και τέταρτη σειρά του σχήματος~\ref{fig:signal2imageoutputs}.

\section{Ρύθμιση πειραμάτων}
Τα συνελικτικά επίπεδα αρχικοποιήθηκαν με χρήση ομοιόμορφου θορύβου τύπου Kaiming~\cite{he2015delving}. Οι τιμές λαμβάνονται από την ομοιόμορφη κατανομή $\mathcal{U}(-c, c)$, όπου $c$ είναι:
\begin{equation}
	c = \sqrt{\frac{6}{(1 + a^2) k}}
\end{equation}

\noindent
,$a$ σε αυτήν τη μελέτη τίθεται μηδέν και $k$ είναι το μέγεθος της εισόδου του επιπέδου.
Τα γραμμικά επίπεδα του $m$ αρχικοποιήθηκαν χρησιμοποιώντας $\mathcal{U}(-\frac{1}{\sqrt{k}},\frac{1}{\sqrt{k}})$.
Τα συνελικτικά και γραμμικά επίπεδα όλων των $b_d$ αρχικοποιήθηκαν σύμφωνα με την αρχική τους εφαρμογή.

Χρησιμοποιήθηκε ο Adam~\cite{kingma2014adam} ως βελτιστοποιητής με ρυθμό μάθησης $lr=0.001$, betas $b_1=0.9$, $b_2=0.999$, epsilon $\epsilon=10^{-8}$ χωρίς απώλεια βαρών και τη διασταυρούμενη εντροπία ως συνάρτηση απώλειας.
Το μέγεθος παρτίδας ήταν $20$ και δεν χρησιμοποιήθηκε επιπλέον regularization εκτός από το προϋπάρχον όπως το dropout, το οποίο έχουν κάποια από τα `μοντέλα βάσης' (AlexNet, VGGnet και DenseNet).

Από τα $11500$ σήματα που χρησιμοποιήθηκαν $76\%$ και $12\%$ από τα δεδομένα ($8740,1380,1380$ σήματα) ως δεδομένα εκπαίδευσης, επικύρωσης και δοκιμής αντίστοιχα.
Δεν πραγματοποιήθηκε αφαίρεση θορύβου ή άλλη προεπεξεργασία.
Όλα τα δίκτυα εκπαιδεύτηκαν για $100$ εποχές και η επιλογή των μοντέλων έγινε σύμφωνα με την καλύτερη ακρίβεια επικύρωσης από όλες τις εποχές.
Χρησιμοποιήθηκε το PyTorch~\cite{paszke2017automatic} για την υλοποίηση των αρχιτεκτονικών νευρωνικών δικτύων και η εκπαίδευση/προεπεξεργασία πραγματοποιήθηκε με τη χρήση της NVIDIA Titan X Pascal GPU 12GB RAM και 12 Core Intel i7-8700 CPU @ 3.20GHz σε λειτουργικό σύστημα Linux.

\begin{sidewaystable}
	\caption[Ακρίβειες στα δεδομένα δοκιμής (\%) για τα συνδυασμένα μοντέλα]{Ακρίβειες στα δεδομένα δοκιμής (\%) για τα συνδυασμένα μοντέλα.
	Η δεύτερη σειρά απεικονίζει τον αριθμό των επιπέδων.
	Η έντονη γραμματοσειρά απεικονίζει τις καλύτερες ακρίβειες για κάθε μοντέλο βάσης.}
	\label{table:results}
	\begin{minipage}{\textwidth}
		\centering
		\input{chapter5-results.tex}
	\end{minipage}
\end{sidewaystable}

\section{Αποτελέσματα}
Όπως φαίνεται στον Πίνακα.~\ref{table:results} το CNN ενός επιπέδου DenseNet201 πέτυχε την καλύτερη ακρίβεια $85.3\%$, με διάρκεια εκπαίδευσης 70 δευτερόλεπτα/εποχή κατά μέσο όρο.
Συνολικά, το CNN S2I ενός επιπέδου πέτυχε την καλύτερη ακρίβεια για έντεκα από τα δεκαπέντε `μοντέλα βάσης'.
Το CNN S2I δύο επιπέδων παρουσίασε χειρότερα αποτελέσματα ακόμη και σε σύγκριση με τις 1D παραλλαγές, υποδεικνύοντας ότι η αύξηση του βάθους S2I δεν είναι ευεργετική.
Το `σήμα ως εικόνα' και το φασματογράφημα S2Is παρουσίασαν πολύ χειρότερα αποτελέσματα από τις 1D παραλλαγές και το CNN S2Is.
Τα αποτελέσματα S2I του φασματογράφου είναι αντίθετα με την προσδοκία ότι η ερμηνεύσιμη αναπαράσταση του πεδίου χρόνου-συχνοτήτων θα βοηθούσε στην εύρεση καλών χαρακτηριστικών για ταξινόμηση.
Υποθέτουμε ότι το φασματογράφημα S2I παρεμποδίστηκε από την έλλειψη μη-εκπαιδεύσιμων παραμέτρων.
Ένα άλλο αποτέλεσμα αυτών των πειραμάτων είναι ότι η αύξηση του βάθους των `μοντέλων βάσης' δεν αύξησε την ακρίβεια, κάτι το οποίο είναι σύμφωνο με προηγούμενα αποτελέσματα~\cite{schirrmeister2017deep}.

\section{Συμπεράσματα}
Σε αυτό το κεφάλαιο δείξαμε εμπειρικά αποτελέσματα ότι οι παραλλαγές 1D `μοντέλου βάσης' και τα εκπαιδεύσιμα S2Is (ειδικά το CNN ενός επιπέδου) έχουν καλύτερη απόδοση από τα μη-εκπαιδεύσιμα S2I.
Ωστόσο, πρέπει να καταβληθεί περισσότερη προσπάθεια για την πλήρη αντικατάσταση των μη-εκπαιδεύσιμων S2I, όχι μόνο από την πλευρά επίτευξης υψηλότερων αποτελεσμάτων ακρίβειας, αλλά και για την αύξηση της ερμηνευσιμότητας του μοντέλου.
Ένα άλλο σημείο αναφοράς είναι ότι τα συνδυασμένα μοντέλα εκπαιδεύτηκαν από την αρχή, με βάση την υπόθεση ότι τα προρυθμισμένα χαρακτηριστικά χαμηλού επιπέδου των `μοντέλων βάσης' μπορεί να μην είναι κατάλληλα για εικόνες που μοιάζουν με φασματογράφημα όπως εκείνες που δημιουργούνται από τα S2Is.
Μια μελλοντική εργασία θα μπορούσε να περιλαμβάνει τη δοκιμή αυτής της υπόθεσης με την αρχικοποίηση ενός `μοντέλου βάσης' με τη χρήση μεθόδων μεταφοράς μάθησης ή άλλων μεθόδων αρχικοποίησης.
Επιπλέον, μπορούν να χρησιμοποιηθούν και άλλα εκπαιδεύσιμα S2Is και 1D `μοντέλου βάσης' για άλλα σήματα φυσιολογίας εκτός από το EEG, όπως το ECG, το ηλεκτρομυογράφημα και η γαλβανική απόκριση δέρματος.

\clearpage
\bibliography{chapter5.bib}
\bibliographystyle{unsrt}

%% file: chapter5-results.tex
\begin{tabular}{l | c | c | c c c c | c c c c c | c c c c }
	\toprule
	\multirow{2}{*}{\diagbox{Dim, S2I}{Model}} & LeNet         & AlexNet       & \multicolumn{4}{c|}{VGGnet} & \multicolumn{5}{c|}{ResNet} & \multicolumn{4}{c}{DenseNet}                                                                                                                                                                 \\
	& 2             & 5             & 11                          & 13                          & 16                           & 19            & 18            & 34            & 50            & 101           & 152           & 121           & 161           & 169           & 201           \\
	\midrule
	1D, none                                   & 72.6          & 78.8          & 76.9                        & \textbf{79.0}               & 79.5                         & \textbf{79.3} & 81.5          & 82.5          & 81.4          & 78.8          & 81.4          & 81.8          & \textbf{83.3} & 82.1          & 82.0          \\
	\midrule
	2D, signal as image                        & 67.9          & 68.3          & 74.1                        & 74.7                        & 72.7                         & 72.5          & 73.3          & 71.7          & 74.1          & 72.3          & 74.1          & 74.7          & 72.5          & 75.2          & 75.0          \\
	\midrule
	2D, spectrogram                            & 73.2          & 74.0          & 77.9                        & 76.3                        & 77.5                         & 76.0          & 76.2          & 79.0          & 77.2          & 74.6          & 75.3          & 74.1          & 75.2          & 77.0          & 75.4          \\
	\midrule
	2D, one layer CNN                          & \textbf{75.8} & \textbf{82.0} & \textbf{84.0}               & 77.9                        & 80.7                         & 78.4          & \textbf{85.1} & \textbf{84.6} & \textbf{83.0} & \textbf{85.0} & \textbf{83.3} & \textbf{84.3} & 80.7          & \textbf{85.0} & \textbf{85.3} \\
	\midrule
	2D, two layer CNN                          & 75.0          & 77.9          & 80.7                        & 78.8                        & \textbf{81.1}                & 74.9          & 78.3          & 80.0          & 78.3          & 77.1          & 80.9          & 83.2          & 82.3          & 79.0          & 79.1          \\
	\bottomrule
\end{tabular}

%% file: chapter6.tex
\chapter{Δίκτυα αραιής ενεργοποίησης}
\label{chapter6}
\graphicspath{{./images/sparsely-activated-networks/}}

\newcommand\figscale{0.16}
\tdplotsetmaincoords{100}{-70}
\begin{filecontents*}{keys-values.csv}
key,value
uci_epilepsy_supervised_accuracy,82.14
mnist_supervised_accuracy,92.17
fashionmnist_supervised_accuracy,83.95
\end{filecontents*}
\DTLloaddb{keys_values}{keys-values.csv}

\section{Εισαγωγή}
Τα DNN~\cite{lecun2015deep} χρησιμοποιούν πολλαπλά στοιβαγμένα επίπεδα με βάρη και συναρτήσεις ενεργοποίησης, που μετασχηματίζουν την είσοδο σε ενδιάμεσες αναπαραστάσεις κατά τη διάρκεια της προς-τα-εμπρός διάδοσης.
Χρησιμοποιώντας backpropagation~\cite{rumelhart1986learning} η κλίση του κάθε βάρους σε σχέση με το σφάλμα της εξόδου υπολογίζεται και μεταβιβάζεται σε μια συνάρτηση βελτιστοποίησης όπως η Στοχαστική Κάθοδος Κλίσεων ή Adam~\cite{kingma2014adam} η οποία ενημερώνει τις τιμές των βαρών κάνοντας την έξοδο του δικτύου να συγκλίνει στην επιθυμητή έξοδο.
Τα DNN χρησιμοποιώντας μεγάλο όγκο δεδομένων και ισχυρές μονάδες παράλληλης επεξεργασίας έχουν επιτύχει υψηλού επιπέδου αποτελέσματα σε προβλήματα όπως η αναγνώριση εικόνας~\cite{krizhevsky2012imagenet} και ομιλίας~\cite{graves2013speech}.
Ωστόσο, αυτές οι καινοτομίες έχουν συμβεί σε βάρος της αύξησης του μήκους περιγραφής των αναπαραστάσεων, οι οποίες στα DNN είναι ανάλογες με τον αριθμό των:
\begin{enumerate}
	\item βαρών του μοντέλου και
	\item μη-μηδενικών ενεργοποιήσεων.
\end{enumerate}

Η χρήση μεγάλου αριθμού βαρών ως επιλογή σχεδιασμού σε αρχιτεκτονικές όπως το Inception~\cite{szegedy2016rethinking}, VGGnet~\cite{simonyan2014very} και ResNet~\cite{he2016deep} (συνήθως με την αύξηση του βάθους), ακολουθήθηκε από έρευνα που επέδειξε τον πλεονασμό των βαρών των DNNs.
Αποδείχθηκε ότι τα DNN προσαρμόζονται εύκολα σε τυχαίες επισημάνσεις δεδομένων~\cite{zhang2016understanding} και ότι σε κάθε αρχικοποιημένο DNN υπάρχει ένα υποδίκτυο που μπορεί να επιλύσει το δεδομένο πρόβλημα, με την ίδια ακρίβεια με το αρχικά εκπαιδευμένο~\cite{frankle2018lottery}.

Επιπλέον τα DNN με μεγάλο αριθμό βαρών έχουν υψηλότερες απαιτήσεις αποθήκευσης και είναι πιο αργά κατά τη διάρκεια των συμπερασμών.
Προηγούμενη έρευνα που έγινε πάνω σε αυτό το πρόβλημα, επικεντρώθηκε στο κλάδεμα βαρών από εκπαιδευμένα DNNs~\cite{aghasi2017net} και κλάδεμα βαρών κατά τη διάρκεια της εκπαίδευσης~\cite{lin2017runtime}.
Το κλάδεμα ελαχιστοποιεί την χωρητικότητα του μοντέλου για χρήση σε περιβάλλοντα με χαμηλές υπολογιστικές δυνατότητες ή χαμηλές απαιτήσεις χρόνου συμπερασμών και επίσης βοηθά στην μείωση της προσαρμογής των νευρώνων η οποία επίσης έχει αντιμετωπιστεί από το Dropout~\cite{srivastava2014dropout}.
Ωστόσο οι στρατηγικές κλαδέματος λαμβάνουν υπόψη μόνο τον αριθμό των βαρών του μοντέλου.

Το άλλο στοιχείο που επηρεάζει το μήκος περιγραφής των αναπαραστάσεων των DNN, είναι ο αριθμός των μη-μηδενικών ενεργοποιήσεων στις ενδιάμεσες αναπαραστάσεις που σχετίζονται με την έννοια της αραιότητας.
Στα νευρωνικά δίκτυα η αραιότητα μπορεί να εφαρμοστεί είτε στις συνδέσεις μεταξύ νευρώνων είτε στους χάρτες ενεργοποίησης~\cite{laughlin2003communication}.
Παρόλο που η αραιότητα στους χάρτες ενεργοποίησης συνήθως επιβάλλεται στη συνάρτηση απώλειας με την προσθήκη ενός όρου κανονικοποίησης $L_{1, 2}$ ή απόκλισης Kullback-Leibler~\cite{kingma2013auto}, θα μπορούσαμε επίσης να επιτύχουμε αραιότητα στους χάρτες ενεργοποίησης με τη χρήση μιας κατάλληλης συνάρτησης ενεργοποίησης.

\begin{sidewaysfigure}
	\centering
	\subfloat{\includegraphics[width=0.2\textwidth]{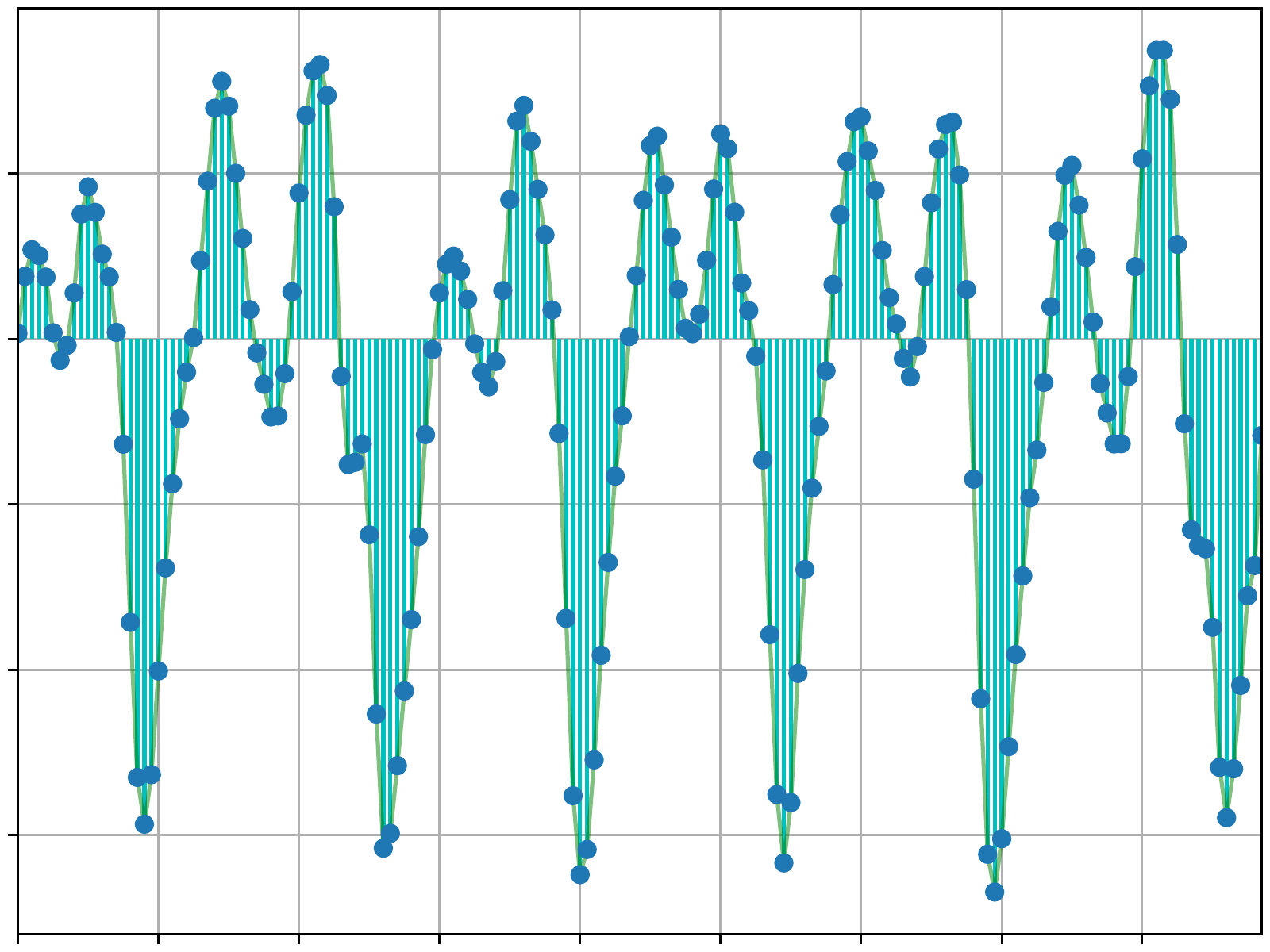}}
	\subfloat{\includegraphics[width=0.2\textwidth]{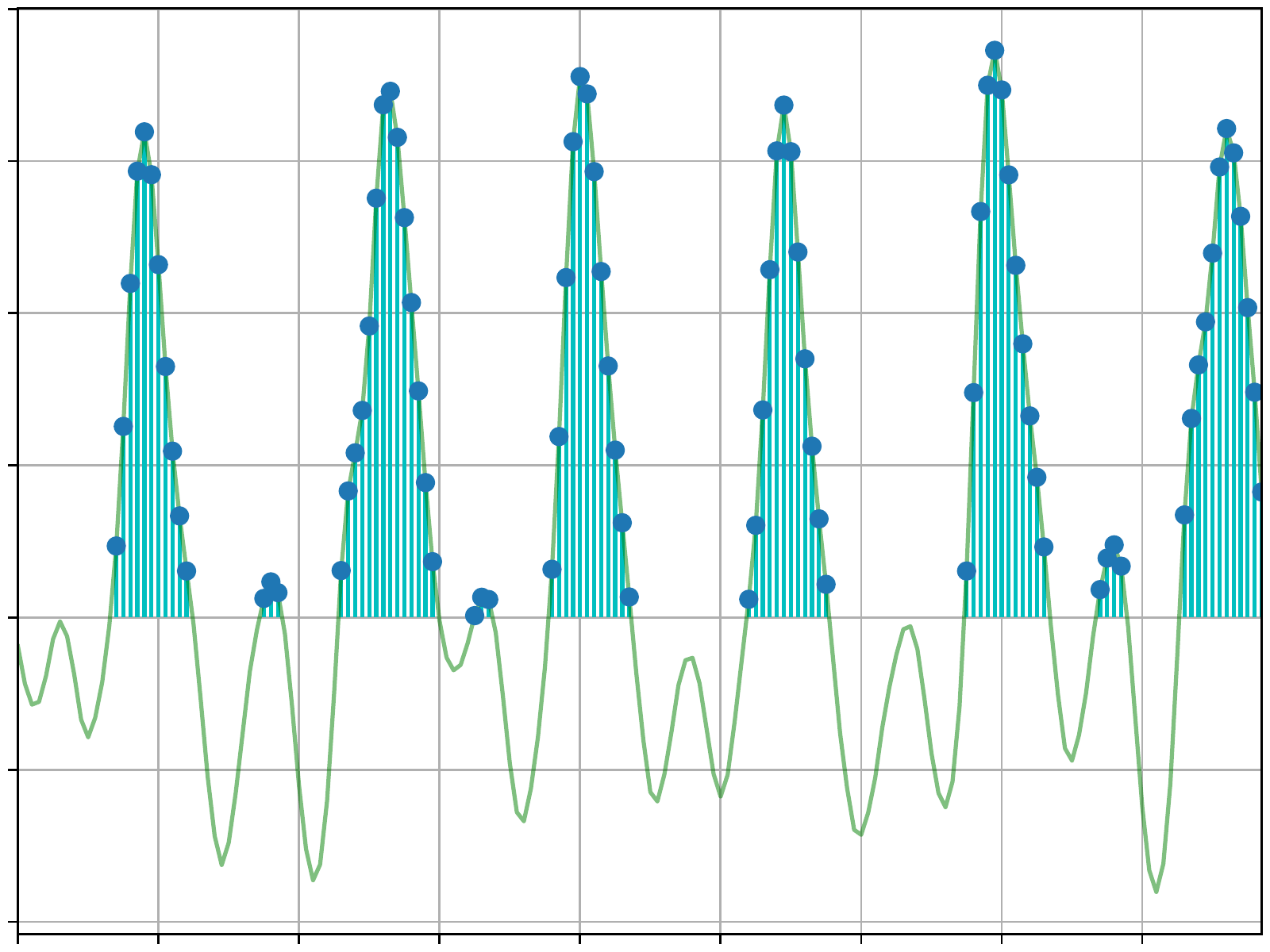}}
	\subfloat{\includegraphics[width=0.2\textwidth]{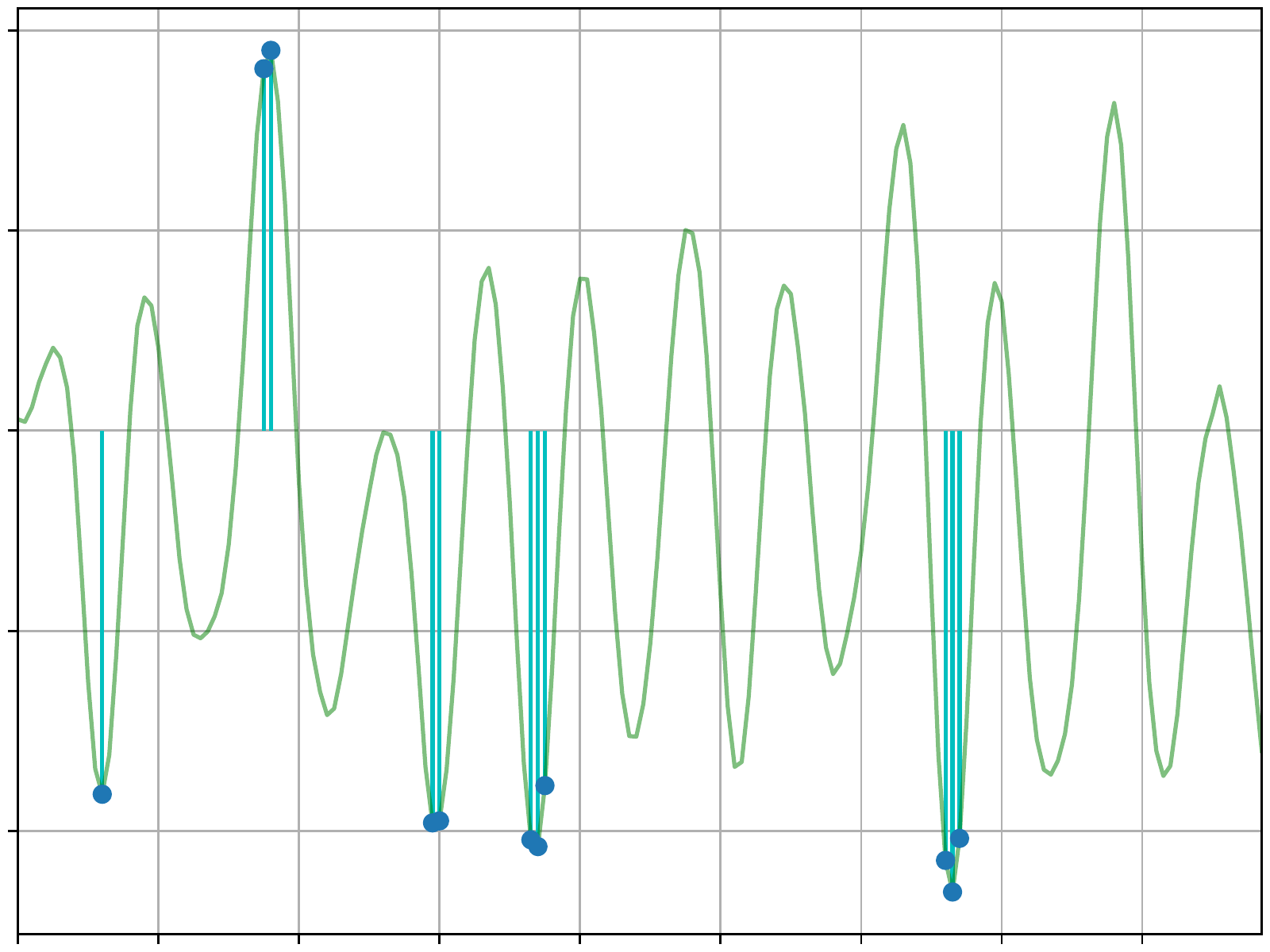}}
	\subfloat{\includegraphics[width=0.2\textwidth]{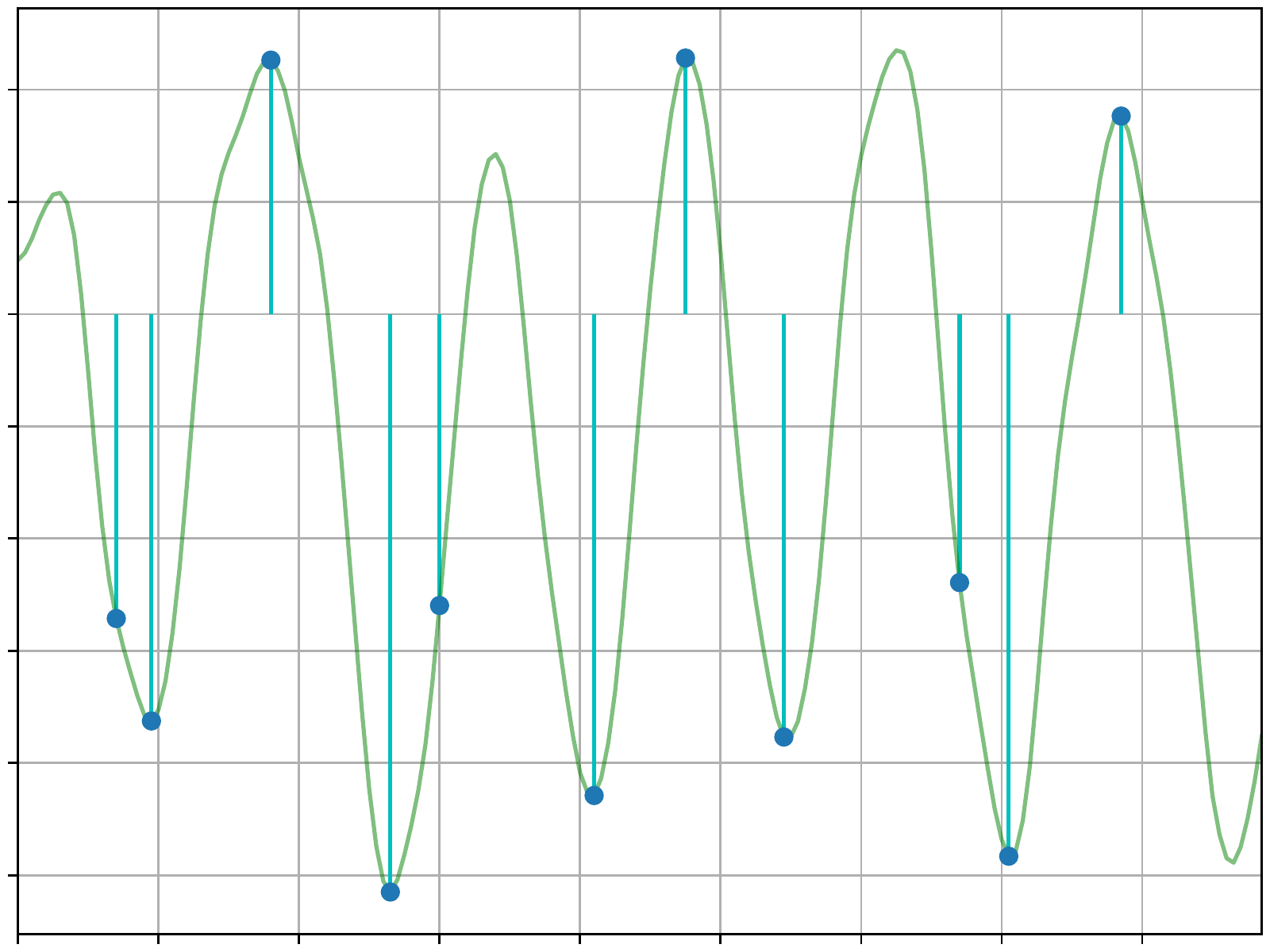}}
	\subfloat{\includegraphics[width=0.2\textwidth]{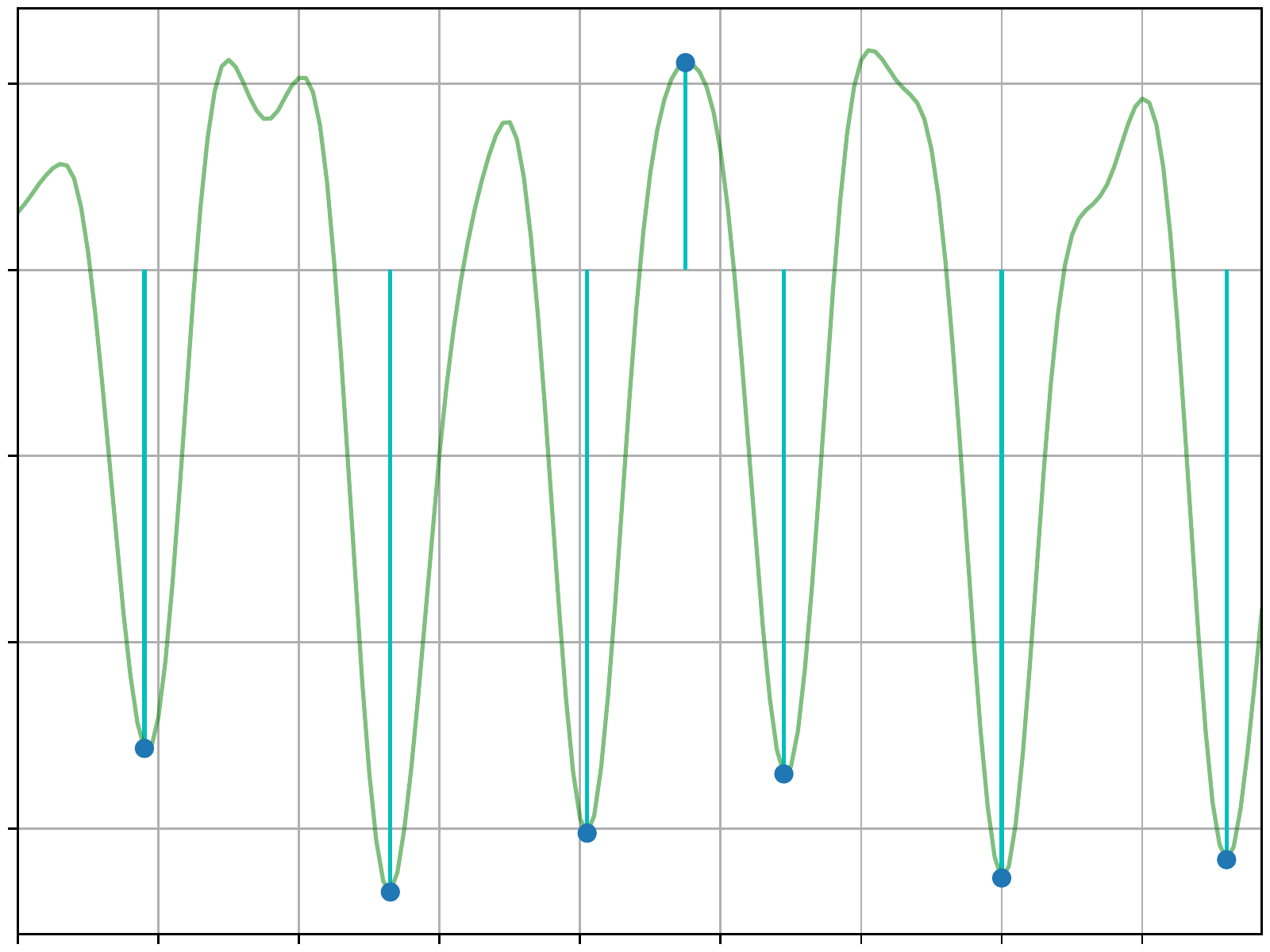}}
	\\
	\setcounter{subfigure}{0}
	\subfloat[Identity]{\includegraphics[width=0.2\textwidth]{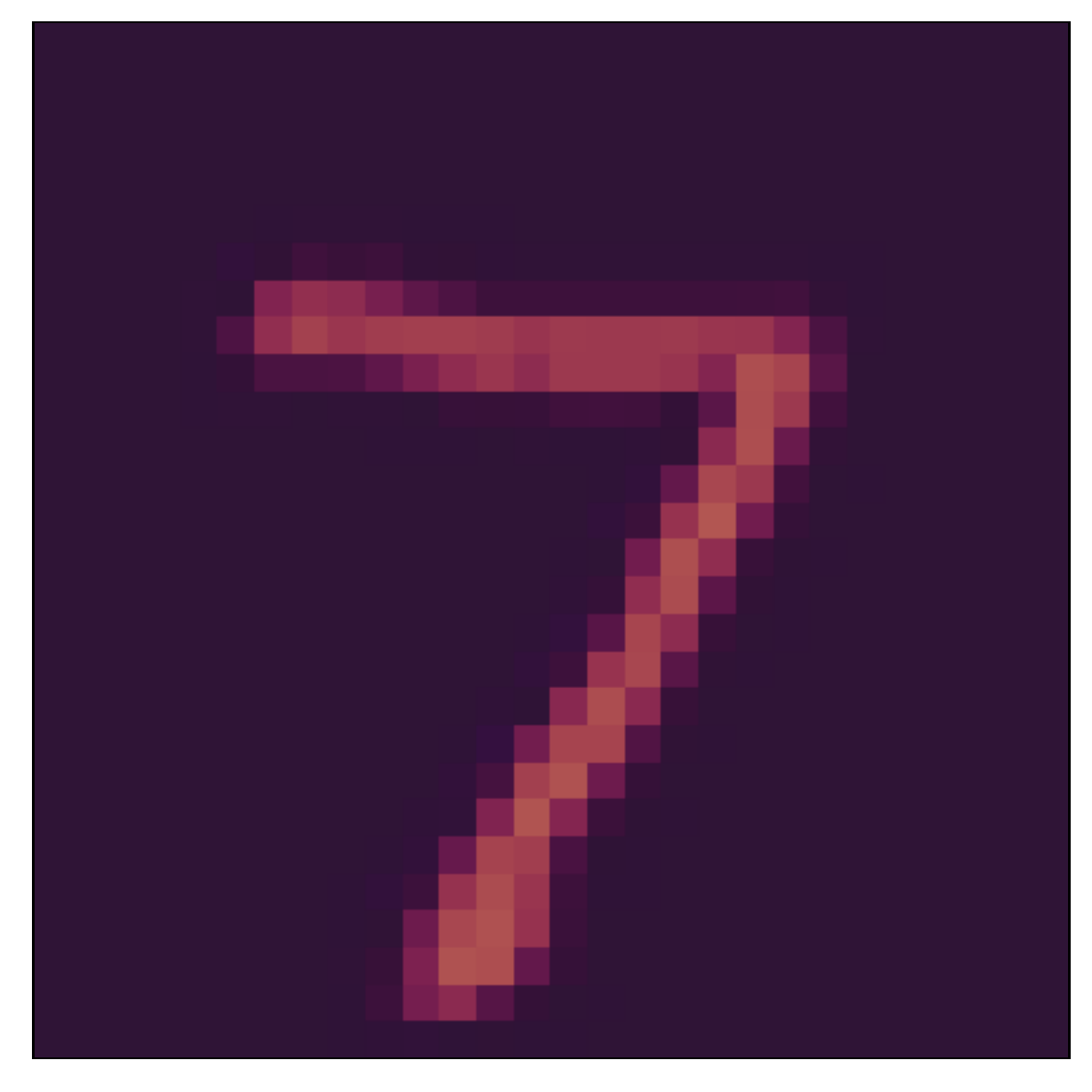}\label{subfig:identity}}
	\subfloat[ReLU]{\includegraphics[width=0.2\textwidth]{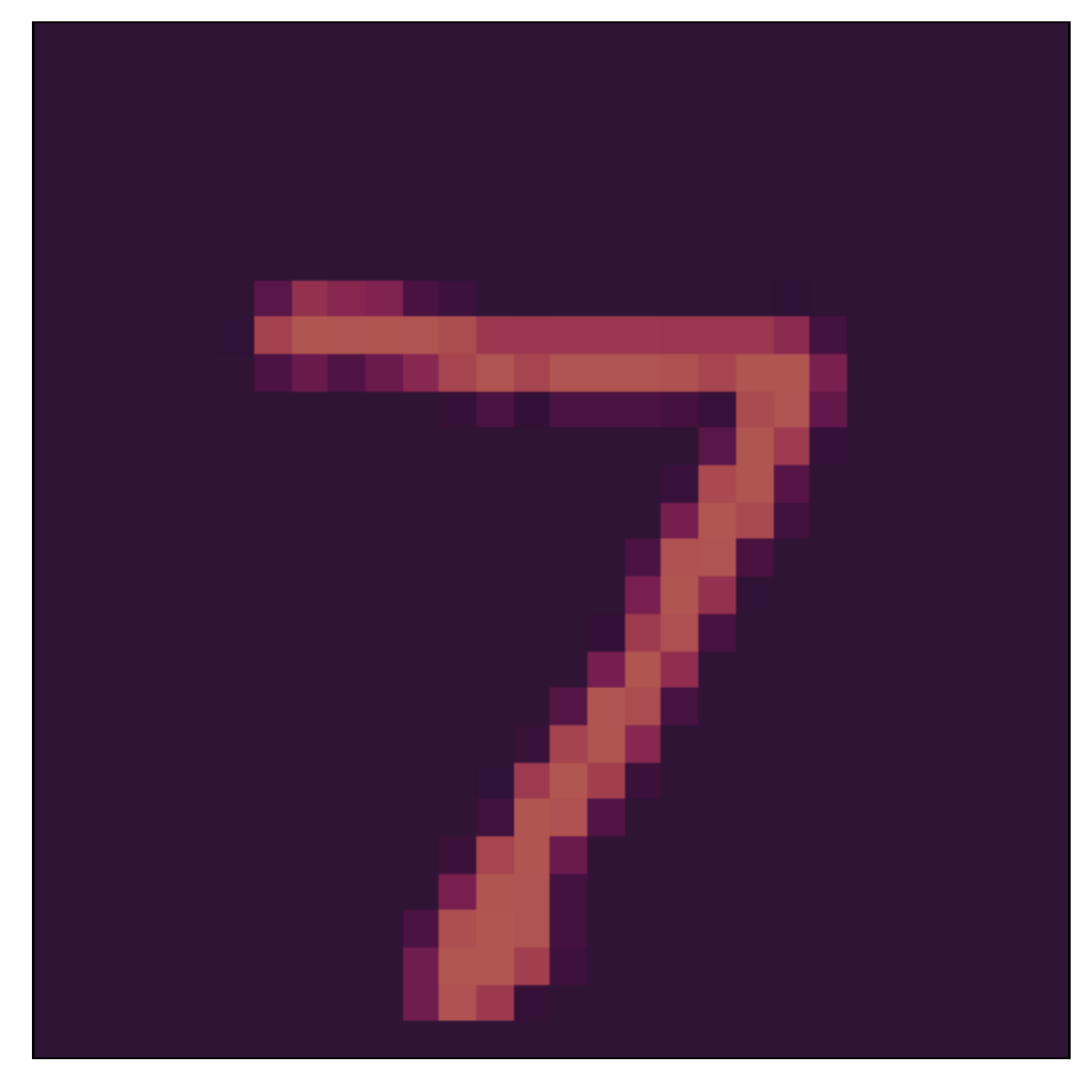}\label{subfig:relu}}
	\subfloat[top-k absolutes]{\includegraphics[width=0.2\textwidth]{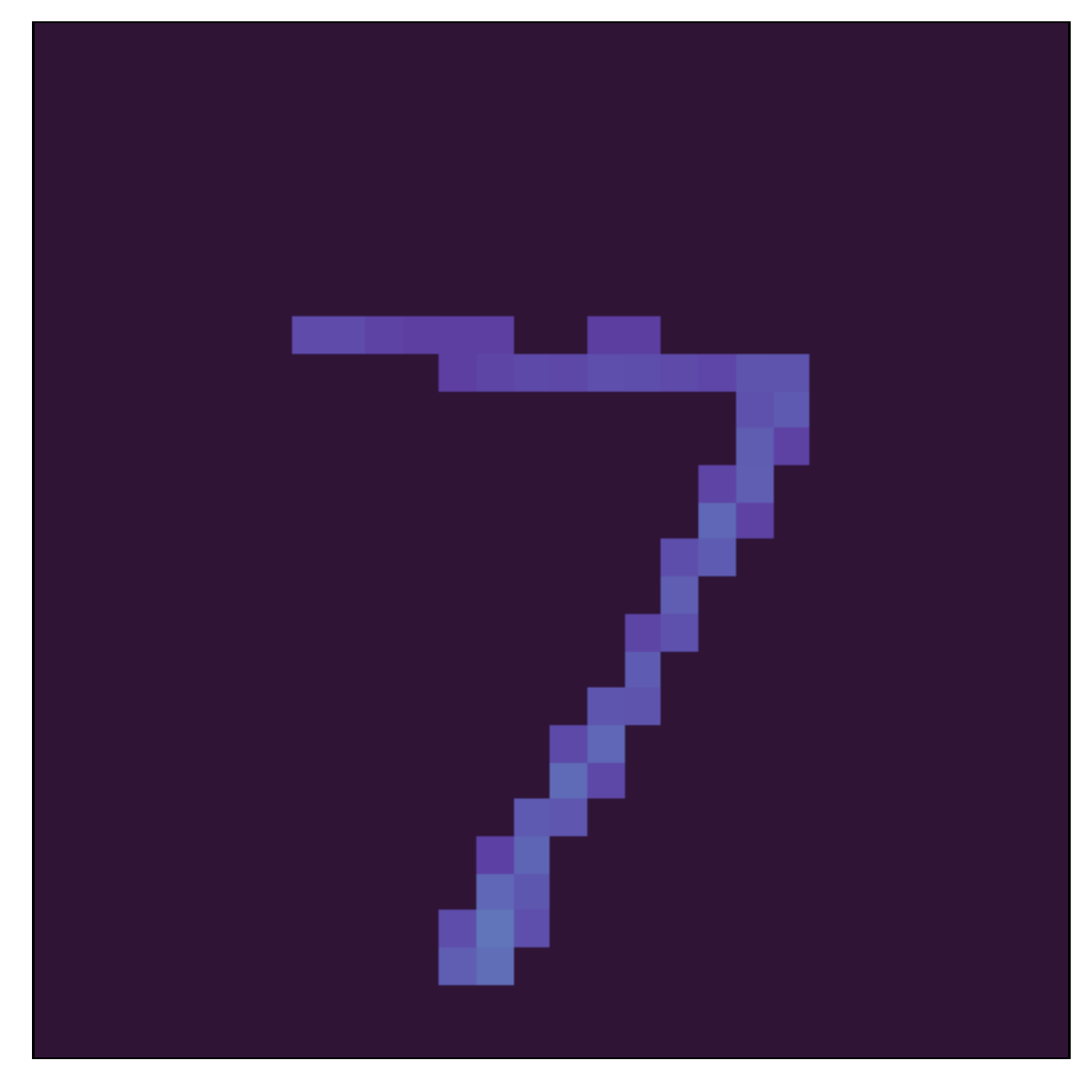}\label{subfig:topk_absolutes}}
	\subfloat[Extrema-Pool indices]{\includegraphics[width=0.2\textwidth]{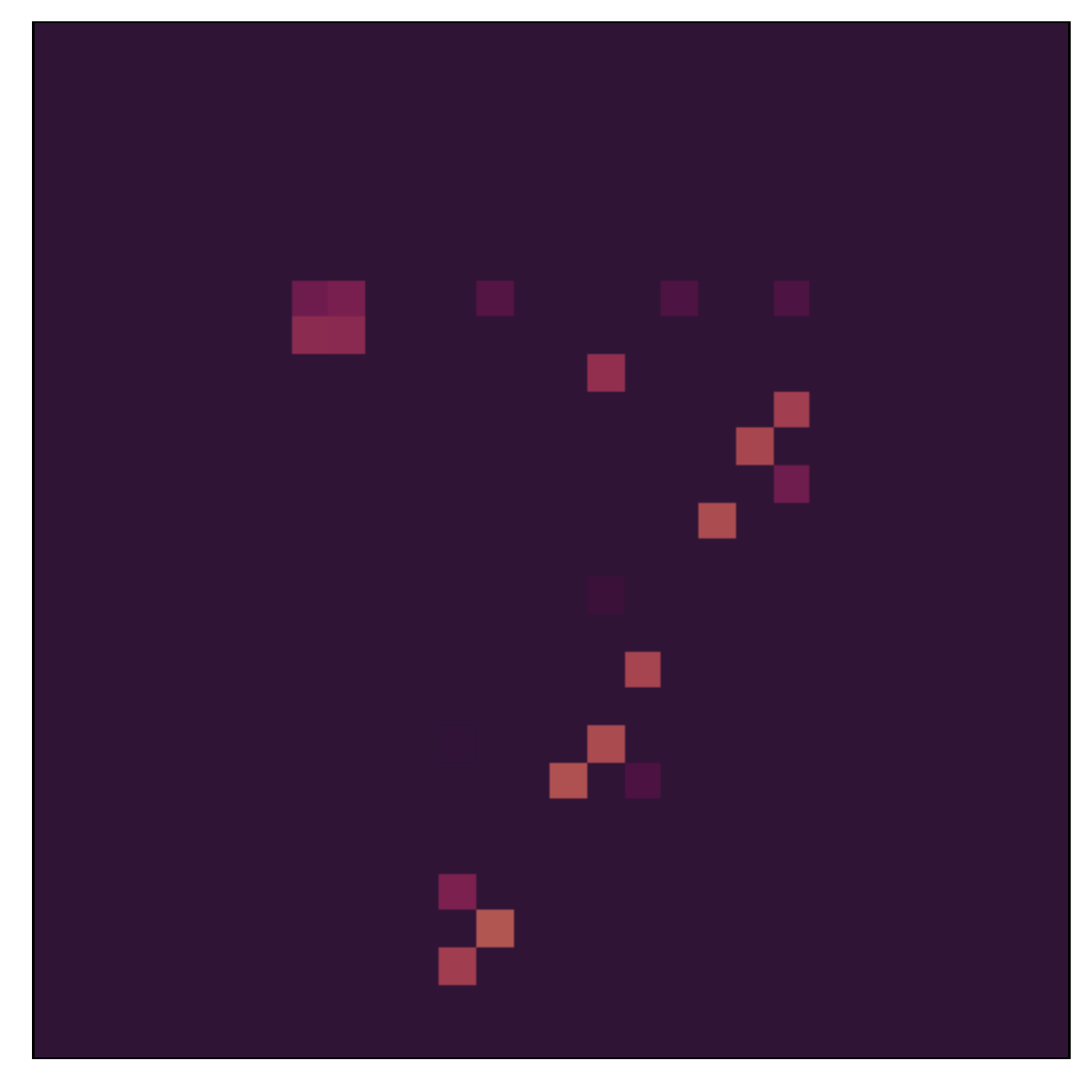}\label{subfig:extremapoolindices}}
	\subfloat[Extrema]{\includegraphics[width=0.2\textwidth]{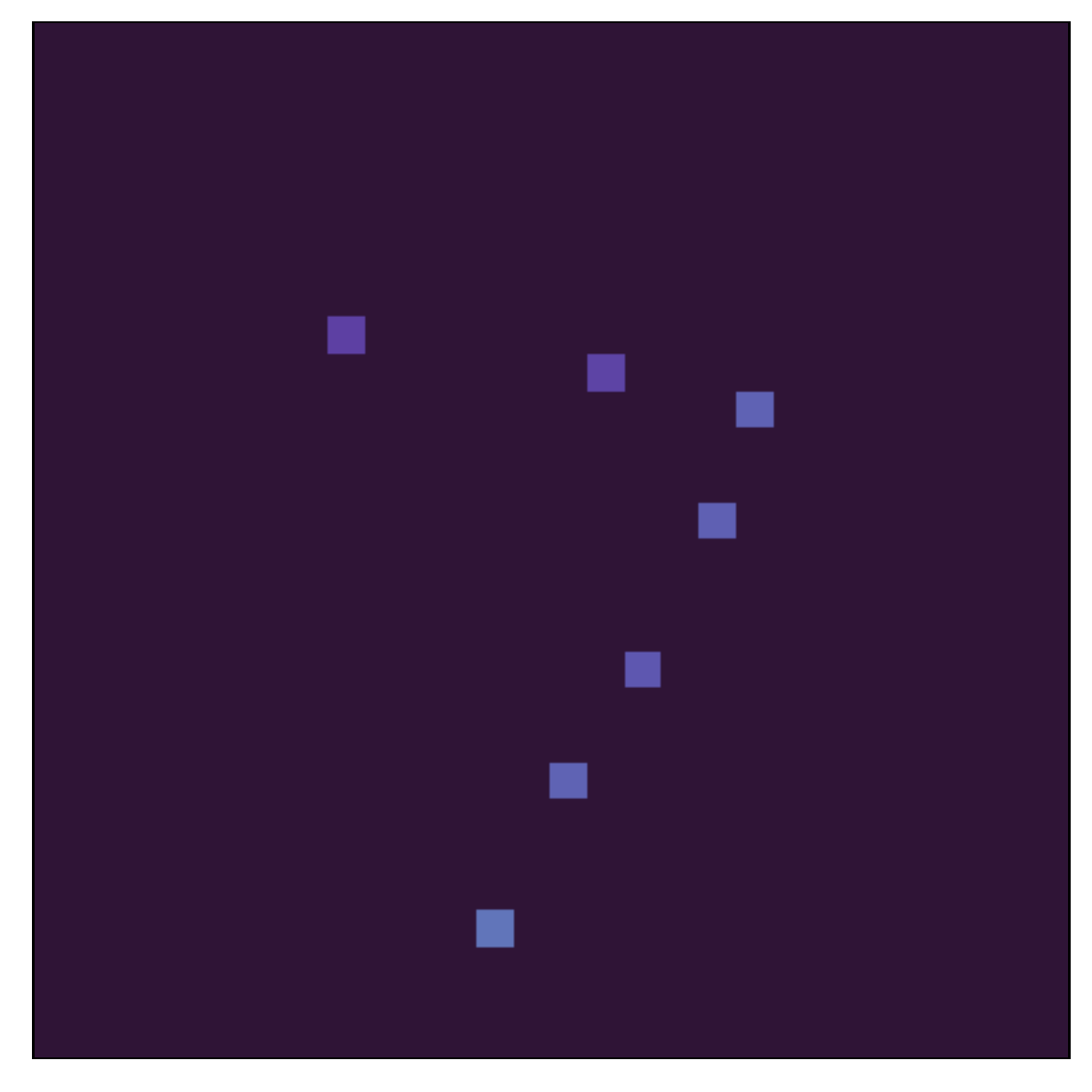}\label{subfig:extrema}}
	\caption[Οπτικοποίηση χαρτών ενεργοποίησης πέντε συναρτήσεων αραιής ενεργοποίησης (Identity, ReLU, top-k absolutes, Extrema-Pool indices και Extrema) για 1D και 2D είσοδο.]{Οπτικοποίηση πέντε συναρτήσεων αραιής ενεργοποίησης (Identity, Max-Activations, Max-Pool indices και Extrema) για 1D και 2D είσοδο στην πρώτη και τη δεύτερη σειρά αντίστοιχα.
	Η 1D είσοδος στις συναρτήσεις ενεργοποίησης υποδηλώνεται με τη συνεχή διαφανή πράσινη γραμμή, χρησιμοποιώντας ένα παράδειγμα από τη βάση δεδομένων επιληψίας του UCI\@.
	Η έξοδος κάθε συνάρτησης ενεργοποίησης απεικονίζεται με τις κυανές γραμμές με τους μπλε δείκτες.
	Το 2D παράδειγμα απεικονίζει μόνο την έξοδο των συναρτήσεων ενεργοποίησης, χρησιμοποιώντας ένα παράδειγμα από τη βάση δεδομένων MNIST\@.
	}
	\label{fig:activationfunctions}
\end{sidewaysfigure}

Αρχικά, χρησιμοποιήθηκαν φραγμένες συναρτήσεις όπως η σιγμοειδής και η υπερβολική εφαπτομένη ($\tanh$), αλλά εκτός από την παραγωγή πυκνών χαρτών ενεργοποίησης παρουσιάζουν επίσης το πρόβλημα του vanishing gradients~\cite{bengio1994learning}.
Το ReLU προτάθηκε αργότερα~\cite{glorot2011deep, nair2010rectified} ως μια συνάρτηση ενεργοποίησης που λύνει το πρόβλημα της διαφυγής των κλίσεων και αυξάνει την αραιότητα των χαρτών ενεργοποίησης.
Αν και το ReLU δημιουργεί μηδενικά (σε αντίθεση με τους προκατόχους του σιγμοειδής και $\tanh$), ο χάρτης ενεργοποίησής του αποτελείται από αραιά χωρισμένες αλλά πυκνές περιοχές (Εικ.~\ref{fig:activationfunctions}\subref{subfig:relu}) αντί για αραιές αιχμές.
Το ίδιο ισχύει για άλλες γενικεύσεις του ReLU, όπως το Παραμετρικό ReLU~\cite{he2015delving} και το Maxout~\cite{goodfellow2013maxout}.
Πρόσφατα, στο $k$-Sparse Autoencoders~\cite{makhzani2013k} χρησιμοποιήθηκε μια συνάρτηση ενεργοποίησης που εφαρμόζει κατωφλίωση μέχρι να παραμείνουν οι $k$ μεγαλύτερες ενεργοποιήσεις, ωστόσο αυτή η μη-γραμμικότητα καλύπτει μια περιορισμένη περιοχή του χάρτη ενεργοποίησης δημιουργώντας αραιά αποσυνδεδεμένες πυκνές περιοχές (Εικ.~\ref{fig:activationfunctions}\subref{subfig:topk_absolutes}), παρόμοιες με την περίπτωση ReLU\@.

Επιπλέον συναρτήσεις ενεργοποίησης που παράγουν συνεχείς χάρτες ενεργοποίησης (όπως το ReLU) είναι λιγότερο βιολογικά εύλογες, επειδή οι βιολογικοί νευρώνες σπάνια βρίσκονται στο μέγιστο επίπεδο κορεσμού τους~\cite{bush1996inhibition} και επίσης χρησιμοποιούν αιχμές για να επικοινωνούν αντί για συνεχείς τιμές~\cite{bengio2015towards}.
Προηγούμενη βιβλιογραφία έχει επίσης καταδείξει την αυξημένη βιολογική ευλογοφάνεια της αραιότητας στα τεχνητά νευρωνικά δίκτυα~\cite{rehn2007network}.
Η αραιότητα επιπέδου αιχμών στους χάρτες ενεργοποίησης έχει ερευνηθεί διεξοδικά στα βιολογικώς πιο αποδεκτά μοντέλα που επικοινωνούν με βάση το ρυθμό (rate-based)~\cite{heiberg2018firing}, αλλά δεν έχει διερευνηθεί διεξοδικά ως επιλογή σχεδιασμού για συναρτήσεις ενεργοποίησης σε συνδυασμό με συνελικτικά φίλτρα.

Ο αυξημένος αριθμός βαρών και των μη-μηδενικών ενεργοποιήσεων καθιστούν τα DNNs  πολύπλοκα και έτσι είναι δυσκολότερο να χρησιμοποιηθούν σε προβλήματα που απαιτούν την αντιστοίχηση αιτιότητας της εξόδου με ένα συγκεκριμένο σύνολο νευρώνων.
Η πλειοψηφία των τομέων που κάνουν χρήση της μηχανικής μάθησης συμπεριλαμβανομένων τομέων όπως η υγειονομική περίθαλψη~\cite{bizopoulos2019deep} απαιτούν τα μοντέλα να είναι ερμηνεύσιμα και εξηγήσιμα πριν θεωρηθούν ως πιθανή λύση.
Αν και αυτές οι ιδιότητες μπορούν να αυξηθούν χρησιμοποιώντας ανάλυση ευαισθησίας~\cite{simonyan2013deep}, μεθόδους αποσυνέλιξης~\cite{zeiler2014visualizing}, Layerwise-Relevance Propagation~\cite{bach2015pixel} και Local-Interpretable Model agnostic Explanations~\cite{ribeiro2016should} θα ήταν προτιμότερο να έχουμε αυτο-ερμηνεύσιμα μοντέλα.

Επιπλέον λαμβάνοντας υπόψη ότι τα DNNs μαθαίνουν να αναπαριστούν δεδομένα χρησιμοποιώντας το συνδυαστικό σύνολο των βαρών και των μη-μηδενικών ενεργοποιήσεων κατά τη διάρκεια της προς-τα-εμπρός διάδοσης, προκύπτει ένα ενδιαφέρον ερώτημα:
\\\\
\indent\textit{Ποιες είναι οι επιπτώσεις του συμβιβασμού μεγαλύτερου σφάλματος ανακατασκευής των αναπαραστάσεων και του λόγου συμπίεσης των αναπαραστάσεων σε σχέση με στα αρχικά δεδομένα;}
\\

Προηγούμενη εργασία από τους Blier et al.~\cite{blier2018description} έδειξε την ικανότητα των DNN να συμπιέζουν χωρίς απώλειες τα δεδομένα εισόδου και τα βάρη, αλλά χωρίς να παίρνουν υπόψη τον αριθμό των μη-μηδενικών ενεργοποιήσεων.
Σε αυτό το έργο, χαλαρώνουμε την απαίτηση της μη-απώλειας και επίσης θεωρούμε τα νευρωνικά δίκτυα καθαρά ως προσεγγιστές συναρτήσεων, αντί για πιθανοτικά μοντέλα.
Οι συνεισφορές του παρόντος είναι οι ακόλουθες προτάσεις:
\begin{itemize}
	\item Το μέτρο $\varphi$ που αξιολογεί μη-επιβλεπώμενα μοντέλα με βάση το πόσο συμπιεσμένες είναι οι αναπαραστάσεις σε σχέση με τα αρχικά δεδομένα και κατα πόσο ακριβή είναι στις ανακατασκευές τους.
	\item Δίκτυα Αραιής Ενεργοποίησης (SANs) (Εικ.~\ref{fig:sans}), στα οποία επιβάλλεται αραιότητα αιχμών στον χάρτη ενεργοποίησης μέσω μιας συνάρτησης αραιής ενεργοποίησης (Εικ.~\ref{fig:activationfunctions}\subref{subfig:extremapoolindices} και \subref{subfig:extrema}).
\end{itemize}

\section{Μέτρο $\varphi$}
\label{sec6:flithos}
Έστω $M$ ένα μοντέλο με $q$ πυρήνες κάθε ένας από τους οποίους με $m^{(i)}$ δείγματα και έστω $\mathcal{L}$ η συνάρτηση απώλειας ανακατασκευής:
\begin{equation}
	\label{eq:model}
	M: \bm{x} \longmapsto \hat{\bm{x}}
\end{equation}

\noindent
, όπου $\bm{x} \in \mathbb{R}^n$ είναι το διάνυσμα εισόδου και $\hat{\bm{x}}$ είναι η ανακατασκευή του $\bm{x}$.
Για τον ορισμό του μέτρου $\varphi$ χρησιμοποιούμε ένα νευρωνικό δίκτυο το οποίο περιέχει συνελικτικά φίλτρα, παρ'όλ'αυτά ο συγκεκριμένος ορισμός μπορεί να γενικευτεί και σε άλλες αρχιτεκτονικές.
Το μέτρο $\varphi$ αξιολογεί ένα μοντέλο με βάση δύο έννοιες: την `φλυαρία' (verbosity) και την ακρίβειά του.

Η `φλυαρία' στα νευρωνικά δίκτυα μπορεί να γίνει αντιληπτή ως αντιστρόφως ανάλογη με το λόγο συμπίεσης των αναπαραστάσεων.
Πρώτον, υπολογίζουμε τον αριθμό των βαρών $W$ ενός μοντέλου $M$ ως εξής:
\begin{equation}
	\label{eq:numberofweights}
	W = \sum\limits_{i=1}^q m^{(i)}
\end{equation}

Επίσης υπολογίζουμε τον αριθμό των μη-μηδενικών ενεργοποιήσεων $A$ ενός μοντέλου $M$ για είσοδο $\bm{x}$ ως:
\begin{equation}
	\label{eq:numberofactivations}
	A_{\bm{x}} = \sum\limits_{i=1}^q \Big\lVert\bm{\alpha}^{(i)}\Big\lVert_0
\end{equation}

\noindent
, όπου $\lVert \cdot \rVert_0$ δηλώνει την ψευδο-νόρμα $\ell_0$ και $\bm{\alpha}^{(i)}$ τον χάρτη ενεργοποίησης του $i^{th}$ πυρήνα.
Στη συνέχεια, χρησιμοποιώντας τις Εξισώσεις~\ref{eq:numberofweights} και~\ref{eq:numberofactivations} ορίζουμε τη σχέση συμπίεσης $CR$ του $\bm{x}$ σε σχέση με $M$ ως:
\begin{equation}
	\label{eq:compressionratio}
	CR = \frac{n}{W + (\dim(\bm{x}) + 1)A_{\bm{x}}}
\end{equation}

\noindent
, όπου $\dim$ ορίζει την διαστασιμότητα.
Ο λόγος που πολλαπλασιάζουμε τη διαστασιμότητα του $\bm{x}$ με τον αριθμό των ενεργοποιήσεων $A_{\bm{x}}$, είναι ότι πρέπει να εξετάσουμε τη χωρική θέση κάθε μη-μηδενικής ενεργοποίησης εκτός από το πλάτος του για την ανακατασκευή $\bm{x}$.
Επιπλέον, χρησιμοποιώντας αυτόν τον ορισμό του $CR$ δημιουργείται ένας επιθυμητός συμβιβασμός μεταξύ της χρήσης ενός μεγαλύτερου πυρήνα με λιγότερες εμφανίσεις και ενός μικρότερου πυρήνα με περισσότερες εμφανίσεις, βάσει του οποίου αποφασίζεται το μέγεθος του πυρήνα που ελαχιστοποιεί το $CR$.

Όσον αφορά την ακρίβεια ορίζουμε την κανονικοποιημένη απώλεια ανακατασκευής ως εξής:
\begin{equation}
	\label{eq:normalizedreconstrucionloss}
	\tilde{\mathcal{L}}(\hat{\bm{x}},\bm{x}) = \frac{\mathcal{L}(\hat{\bm{x}},\bm{x})}{\mathcal{L}(0,\bm{x})}
\end{equation}

Τέλος, χρησιμοποιώντας τις Εξισώσεις~\ref{eq:compressionratio} και\ref{eq:normalizedreconstrucionloss} ορίζουμε το μέτρο $\varphi$\footnote{Η χρήση του συμβόλου $\varphi$ προέρχεται από τον πρώτο χαρακτήρα του σύνθετου ελληνικού ουσιαστικού `φλύθος' = φλύ + θος. Αποτελείται από το πρώτο μέρος της λέξης φλύ-αρος και το δεύτερο μέρος της λέξης λά-θος. Το φλύθος ορίζεται κυριολεκτικά ως:\textit{Παροχή ανακριβών πληροφοριών χρησιμοποιώντας πολλές λέξεις. Η κατάσταση του να είναι κανείς λάθος και φλύαρος ταυτόχρονα}.} του $\bm{x}$ σε σχέση με $M$ ως εξής:
\begin{equation}
	\label{eq:flithos}
	\varphi = \lVert(CR^{-1}, \tilde{\mathcal{L}}(\hat{\bm{x}},\bm{x}))\rVert_2
\end{equation}

\noindent
, όπου $\lVert\cdot\rVert_2$ ορίζει την ευκλείδια απόσταση.

Όσον αφορά την επιλογή υπερπαραμέτρων, ορίζουμε επίσης το μέτρο $\varphi$ ενός συνόλου δεδομένων ή μια μίνι-παρτίδας σε σχέση με το $M$ ως:
\begin{equation}
	\label{eq:meanflithos}
	\bar\varphi = \frac{1}{l} \sum \limits_{j=1}^l \varphi^{(j)}
\end{equation}

\noindent
, όπου $l$ είναι ο αριθμός παρατηρήσεων της βάσης δεδομένων ή το μέγεθος της παρτίδας.

Το $\bar\varphi$ είναι μη διαφοροποιήσιμο λόγω της παρουσίας της ψευδο-νόρμας $\ell_0$ στην Εξ.~\ref{eq:numberofactivations}.
Ένας τρόπος για να ξεπεραστεί αυτό το πρόβλημα, είναι η χρήση του $\mathcal{L}$ ως τη διαφοροποιήσιμη συνάρτηση βελτιστοποίησης κατά τη διάρκεια της εκπαίδευσης και η χρήση του $\bar\varphi$ ως το μέτρο για την επιλογή μοντέλου κατά την επικύρωση, κατά την διάρκεια της οποίας παίρνονται οι αποφάσεις για τις επιλογές των τιμών των υπερπαραμέτρων (όπως το μέγεθος του πυρήνα).

\section{Δίκτυα αραιής ενεργοποίησης}
\label{sec6:sans}

\subsection{Συναρτήσεις αραιής ενεργοποίησης}
\label{sec6:safs}
Σε αυτή την υποενότητα ορίζουμε πέντε συναρτήσεις ενεργοποίησης $\phi$ και την αντίστοιχη παράμετρο αραιότητας πυκνότητας $d^{(i)}$ για τα οποία έχουμε:
\begin{equation}
	\label{eq:phi}
	\phi: s \longmapsto \alpha
\end{equation}

Επιλέγουμε τιμές για το $d^{(i)}$ για κάθε συνάρτηση αραιής ενεργοποίησης έτσι ώστε να έχουμε περίπου τον ίδιο αριθμό ενεργοποιήσεων για να έχουμε δίκαιη σύγκριση μεταξύ των συναρτήσεων αραιής ενεργοποίησης (εκτός από την Identity, η οποία δεν έχει παράμετρο αραιότητας).

\subsubsection{Ταυτότητα}
\label{sec6:identity}
$\phi = \mathds{1}$.
Η συνάρτηση ενεργοποίησης ταυτότητας (Identity) χρησιμεύει ως βάση αναφοράς και διατηρεί ακέραια την είσοδό του, όπως φαίνεται στην Εικ.~\ref{fig:activationfunctions}\subref{subfig:identity}.
Για αυτήν την περίπτωση, δεν εφαρμόζεται παράμετρος αραιότητας $d^{(i)}$.

\subsubsection{ReLU}
\label{sec6:relu}
$\phi = ReLU(s)$.
Η συνάρτηση ενεργοποίησης ReLU παράγει αραιά συνδεδεμένες αλλά πυκνές περιοχές, όπως φαίνεται στην Εικ.~\ref{fig:activationfunctions}\subref{subfig:relu}.
Για αυτήν την περίπτωση, δεν εφαρμόζεται παράμετρος αραιότητας $d^{(i)}$.

\subsubsection{κ-μέγιστα απολύτων}
\label{sec6:topk_absolutes}
Η συνάρτηση κ-μεγίστων απολύτων (top-k absolutes) (που ορίζεται στον αλγόριθμο~\ref{alg:topk_absolutes}) διατηρεί τους δείκτες των $k$ ενεργοποιήσεων με τη μεγαλύτερη απόλυτη τιμή και μηδενίζει τις υπόλοιπες, όπου $1 \le k <n \in \mathbb{N}$.
Ορίσαμε $d^{(i)} = k$, όπου $k = \lfloor n / m \rfloor^{\dim(\bm{x})}$.
Η συνάρτηση ενεργοποίησης απόλυτων κ-μεγίστων είναι πιο αραιή από το ReLU αλλά κάποια ακρότατα ενεργοποιούνται πολλαπλά σε σχέση με άλλα ακρότατα που δεν ενεργοποιούνται καθόλου, όπως φαίνεται στην Εικ.~\ref{fig:activationfunctions}\subref{subfig:topk_absolutes}.

\begin{algorithm}[H]
	\caption{top-k absolutes}
	\label{alg:topk_absolutes}
	\input{chapter6-topk-absolutes.tex}
\end{algorithm}

\subsubsection{Δείκτες Συγκέντρωσης Ακρότατων}
\label{sec6:extremapoolindices}
Η συνάρτηση ενεργοποίησης των δεικτών συγκέντρωσης ακρότατων (Extrema-Pool indices) (που ορίζεται στον Αλγόριθμο~\ref{alg:extremapoolindices}) διατηρεί μόνο τον δείκτη της ενεργοποίησης με το μέγιστο απόλυτο πλάτος από κάθε περιοχή που περιγράφεται από ένα πλέγμα με την ίδια αναλυτικότητα όπως το μέγεθος πυρήνα $m^{(i)}$ και μηδενίζει τις υπόλοιπες.
Αποτελείται από ένα στρώμα μέγιστης συγκέντρωσης ακολουθούμενο από ένα στρώμα μέγιστης αποσυγκέντρωσης με τις ίδιες παραμέτρους, ενώ η παράμετρος αραιότητας $d^{(i)}$ σε αυτή την περίπτωση ορίζεται ως $d^{(i)} = m^{(i)} <n \in \mathbb{N}$.
Αυτή η συνάρτηση ενεργοποίησης δημιουργεί πιο αραιούς χάρτες ενεργοποίησης από τα απόλυτα κ-μέγιστα, παρ'όλ'αυτά σε περιπτώσεις που το πλέγμα της συγκέντρωσης είναι κοντά σε ακρότατο υπάρχει το ενδεχόμενο να ενεργοποιηθεί διπλά (όπως φαίνεται στην Εικ.~\ref{fig:activationfunctions}\subref{subfig:extremapoolindices}).

\begin{algorithm}[H]
	\caption{Δείκτες συγκέντρωσης ακρότατων}
	\label{alg:extremapoolindices}
	\input{chapter6-extremapoolindices.tex}
\end{algorithm}

\subsubsection{Ακρότατα}
\label{sec6:extrema}
Η συνάρτηση ενεργοποίησης ακρότατων (Extrema) (που ορίζεται στον Αλγόριθμο~\ref{alg:extrema}) ανιχνεύει τα υποψήφια ακρότατα χρησιμοποιώντας τη μηδενική διέλευση της πρώτης παραγώγου και στη συνέχεια τις ταξινομεί σε φθίνουσα σειρά και σταδιακά εξαλείφει εκείνες τα ακρότατα που έχουν μικρότερο πλάτος από ένα γειτονικό ακρότατο σύμφωνα με μια προκαθορισμένη απόσταση $med$.
Η επιβολή ελάχιστης απόστασης ακρότατων στον αλγόριθμο ανίχνευσης ακρότατων καθιστά την $\bm{\alpha}$ πιο αραιή από τις προηγούμενες περιπτώσεις και λύνει το πρόβλημα της διπλής ενεργοποίησης ακρότατων που εμφανίζουν οι δείκτες συγκέντρωσης ακρότατων (όπως φαίνεται στην Εικ.~\ref{fig:activationfunctions}\subref{subfig:extrema}).
Η παράμετρος αραιότητας σε αυτή την περίπτωση ορίζεται $d^{(i)} = med$, όπου $1 \le med <n \in \mathbb{N}$ είναι η ελάχιστη απόσταση ακρότατων.
Ορίσαμε $med = m^{(i)}$ για τη δίκαιη σύγκριση μεταξύ των συναρτήσεων αραιής ενεργοποίησης.
Ειδικά για την συνάρτηση ενεργοποίησης ακρότατων εισάγουμε μια παράμετρο `συνοριακής ανοχής' για να επιτρέψουμε μια πιο έγκαιρη ενεργοποίηση του νευρώνα.

\begin{algorithm}[H]
	\caption{Ανίχνευση ακρότατων με ελάχιστη απόσταση $med$}
	\label{alg:extrema}
	\scalebox{0.9}{
		\begin{minipage}{\linewidth}
			\input{chapter6-extrema.tex}
		\end{minipage}
		}
\end{algorithm}

\subsection{Αρχιτεκτονική και εκπαίδευση των SANs}

\begin{figure}
	\subfloat[1D SAN]{\begin{tikzpicture}[]
		\input{chapter6-san-1d.tex}
	\end{tikzpicture}
	}
	\qquad
	\subfloat[2D SAN]{\begin{tikzpicture}[]
		\input{chapter6-san-2d.tex}
	\end{tikzpicture}
	}
	\caption[Διαγράμματα της προς-τα-εμπρός διάδοσης ενός 1D και 2D SAN με δύο πυρήνες για παραδείγματα από την βάση δεδομένων επιληψίας UCI και MNIST αντίστοιχα.]{Διαγράμματα της προς-τα-εμπρός διάδοσης 1D και 2D SAN με δύο πυρήνες για παραδείγματα από την βάση δεδομένων επιληψίας UCI και MNIST αντίστοιχα.
	Οι εικόνες απεικονίζουν τις ενδιάμεσες αναπαραστάσεις; το $\bm{x}$ δηλώνει το σήμα εισόδου (μπλε γραμμή), το $\bm{w}^{(i)}$ τους πυρήνες (κόκκινη γραμμή), το $\bm{s}^{(i)}$ τους πίνακες ομοιότητας (πράσινη γραμμή), το $\bm{\alpha}{(i)}$ τους χάρτες ενεργοποίησης (κυανές γραμμές με μπλε δείκτες), το $\bm{r}^{(i)}$ την μερική ανακατασκευή από κάθε $\bm{w}^{(i)}$ και το $\hat{\bm{x}}$ την ανακατασκευασμένη είσοδο (κόκκινη γραμμή).
	Για λόγους σύγκρισης, η διαφανής πράσινη γραμμή στο $\bm{\alpha}^{(i)}$ δηλώνει το αντίστοιχο $\bm{s}^{(i)}$ και η διαφανή μπλε γραμμή στο $\hat{\bm{x}}$ δηλώνει την είσοδο $\bm{x}$.
	Ο εκθέτης $i = 0,1$ αντιστοιχεί στον πρώτο και στον δεύτερο πυρήνα και στις ενδιάμεσες αναπαραστάσεις.
	Οι κύκλοι δηλώνουν συναρτήσεις. Το $\mathcal{L}$ δηλώνει τη συνάρτηση απώλειας, το $\phi$ τη συνάρτηση αραιής ενεργοποίησης, το $\ast$ τη συνάρτηση συνέλιξης και το $+$ τον τελεστή πρόσθεσης.
	Όλες οι συναρτήσεις εκτελούνται ξεχωριστά για κάθε $\bm{w}^{(i)}$, ωστόσο για οπτική διαύγεια απεικονίζουμε μόνο μία συνάρτηση για κάθε βήμα.
	Αποχρώσεις του κόκκινου και του μπλε απεικονίζουν θετικές και αρνητικές τιμές αντίστοιχα.
	Η Peak συνάρτηση ενεργοποίησης χρησιμοποιήθηκε και για τα δύο δίκτυα.
	}
	\label{fig:sans}
\end{figure}

Έστω $\bm{x} \in \mathbb{R}^n$ είναι ένα δεδομένο εισόδου, ωστόσο τα παρακάτω μπορεί να γενικευτούν σε εισόδους παρτίδων με διαφορετικά μήκη.
Έστω $\bm{w}^{(i)} \in \mathbb{R}^{m^{(i)}}$ ο πίνακας βάρους του $i^{th}$ πυρήνα ο οποίος αρχικοποιείται χρησιμοποιώντας μια κανονική κατανομή με μέση τιμή $\mu$ και τυπική απόκλιση $\sigma$:
\begin{equation}
	\label{eq:weightinitialization}
	\bm{w}^{(i)} \sim \mathcal{N}(\mu, \sigma)
\end{equation}

\noindent
, όπου $0 \le i <q \in \mathbb{N}$ είναι ο αριθμός των πυρήνων.

Αρχικά υπολογίζουμε τους πίνακες ομοιότητας\footnote{Η προηγούμενη βιβλιογραφία αναφέρεται σε αυτό ως `κρυφή μεταβλητή', εδώ όμως χρησιμοποιούμε μια πιο άμεση ονομασία που ταιριάζει στο πλαίσιο του παρόντος κειμένου.} Το $\bm{s}^{(i)}$ για καθένα από τους πίνακες βάρους $\bm{w}^{(i)}$ είναι:
\begin{equation}
	\label{eq:similarity}
	\bm{s}^{(i)} = \bm{x} * \bm{w}^{(i)}
\end{equation}

\noindent
, όπου $*$ είναι η συνέλιξη\footnote{Χρησιμοποιούμε την συνέλιξη αντί της αλληλοσυσχέτισης μόνο για λόγους συμβατότητας με την προηγούμενη βιβλιογραφία και τα υπολογιστικά πλαίσια. Η χρήση αλληλοσυσχέτισης θα παρήγαγε τα ίδια αποτελέσματα και επιπλέον δεν θα απαιτούσε την περιστροφή των πυρήνων κατά τη διάρκεια της οπτικοποίησης.}
Για την διατήρηση του αρχικού μεγέθους του διανύσματος εισόδου συμπληρώνουμε με μηδενικά.
Δεν χρειαζόμαστε όρο μεροληψίας επειδή θα εφαρμοζόταν συνολικά στο $\bm{s}^{(i)}$ κάτι το οποίο είναι σχεδόν ισοδύναμο με την μάθηση της γραμμής βάσης $\bm{x}$.

Έπειτα, περνάμε την $\bm{s}^{(i)}$ και μια παράμετρο αραιότητας $d^{(i)}$ στην συνάρτηση αραιής ενεργοποίησης $\phi$ με αποτέλεσμα τον χάρτη ενεργοποίησης $\bm{\alpha}^{(i)}$:
\begin{equation}
	\label{eq:extrema}
	\bm{\alpha}^{(i)} = \phi(\bm{s}^{(i)}, d^{(i)})
\end{equation}

\noindent
, όπου $\bm{\alpha}^{(i)}$ είναι ένας αραιός πίνακας, του οποίου τα μη-μηδενικά στοιχεία υποδηλώνουν τις χωρικές θέσεις των στιγμιότυπων του $i^{th}$ πυρήνα.
Η ακριβής μορφή του $\phi$ και του $d^{(i)}$ εξαρτώνται από την επιλογή της συνάρτησης αραιής ενεργοποίησης, οι οποίες παρουσιάζονται στην ενότητα~\ref{sec6:safs}.

Συνελλίσουμε κάθε $\bm{\alpha}^{(i)}$ με το αντίστοιχο $\bm{w}^{(i)}$, έχοντας ως αποτέλεσμα ένα σύνολο ατομικών ανακατασκευών $\bm{r}^{(i)}$ της εισόδου:
\begin{equation}
	\label{eq:reconstructions}
	\bm{r}^{(i)} = \bm{\alpha}^{(i)} * \bm{w}^{(i)}
\end{equation}

\noindent
, που αποτελείται από αραιά επαναλαμβανόμενα πρότυπα $\bm{w}^{(i)}$ με μεταβλητό πλάτος.
Τέλος, μπορούμε να ανακατασκευάσουμε την είσοδο ως το άθροισμα των μεμονωμένων ανακατασκευών $\bm{r}^{(i)}$ ως εξής:
\begin{equation}
	\label{eq:output1}
	\hat{\bm{x}} = \sum \limits_{i=1}^q \bm{r}^{(i)}
\end{equation}

Το μέσο απόλυτο σφάλμα (Mean Absolute Error, MAE) της εισόδου $\bm{x}$ και η πρόβλεψη $\hat{\bm{x}}$ υπολογίζεται ως εξής:
\begin{equation}
	\label{eq:lossfunction}
	\mathcal{L}\left( {\bm{x},\hat{\bm{x}}} \right) = \frac{1}{n}\sum\limits_{t=1}^n \left|\hat{\bm{x}}_t - \bm{x}_t \right|
\end{equation}

\noindent
, όπου ο δείκτης $t$ δηλώνει το $t^{th}$ δείγμα.
Η επιλογή της ΜΑΕ βασίζεται στην ανάγκη να αντιμετωπιστούν οι απομακρυσμένες τιμές των δεδομένων με το ίδιο βάρος με τις κανονικές τιμές.
Ωστόσο τα SAN δεν περιορίζονται στη χρήση του MAE αλλά μπορούν να χρησιμοποιήσουν και άλλες συναρτήσεις απώλειας, όπως το μέσο τετραγωνικό σφάλμα.

Χρησιμοποιώντας backpropagation υπολογίζονται οι κλίσεις του σφάλματος απώλειας $\mathcal{L}$ σε σχέση με το $\bm{w}^{(i)}$:
\begin{equation}
	\label{eq:backpropagation1}
	\nabla\mathcal{L} = \left( \frac{\partial\mathcal{L}}{\partial\bm{w}^{(1)}},\ldots,\frac{\partial\mathcal{L}}{\partial\bm{w}^{(q)}}\right)
\end{equation}

Τέλος, το $\bm{w}^{(i)}$ ενημερώνεται χρησιμοποιώντας τον ακόλουθο κανόνα μάθησης:
\begin{equation}
	\label{eq:backpropagation2}
	\Delta\bm{w}^{(i)} = -\lambda\frac{\partial\mathcal{L}}{\partial\bm{w}^{(i)}}
\end{equation}

\noindent
, όπου $\lambda$ είναι ο ρυθμός μάθησης.

Μετά την εκπαίδευση, θεωρούμε τα $\bm{\alpha}^{(i)}$ (τα οποία υπολογίζονται κατά τη διάρκεια της προς-τα-εμπρός διάδοσης από την Εξ.~\ref{eq:extrema}) και τα $\bm{w}^{(i)}$ (τα οποία υπολογίζονται με τη χρήση του backpropagation από την Εξ.~\ref{eq:backpropagation2}) ως τη συμπιεσμένη αναπαράσταση $\bm{x}$, η οποία μπορεί να ανακατασκευαστεί από τις Εξισώσεις~\ref{eq:reconstructions} και~\ref{eq:output1}:
\begin{equation}
	\label{eq:output2}
	\hat{\bm{x}} = \sum\limits_{i=1}^q \left(\bm{\alpha}^{(i)} * \bm{w}^{(i)}\right)
\end{equation}

Όσον αφορά το μέτρο $\varphi$ και παίρνοντας υπόψη την Εξ.~\ref{eq:output2} στόχος μας είναι να υπολογίσουμε μια όσο το δυνατόν ακριβή αναπαράσταση του $\bm{x}$ μέσω των $\bm{\alpha}^{(i)}$ και $\bm{w}^{(i)}$ με τον μικρότερο αριθμό μη-μηδενικών ενεργοποιήσεων και βαρών.

Η γενική διαδικασία εκπαίδευσης των SAN για πολλαπλές εποχές χρησιμοποιώντας παρτίδες (αντί για ένα παράδειγμα όπως παρουσιάστηκε προηγουμένως) παρουσιάζεται στον Αλγόριθμο~\ref{alg:training}.

\begin{algorithm}[H]
	\caption{Εκπαίδευση δικτύων αραιής ενεργοποίησης}
	\label{alg:training}
	\input{chapter6-san-train.tex}
\end{algorithm}

\section{Πειράματα}
\label{sec6:experiments}
Για όλα τα πειράματα τα βάρη των πυρήνων SAN αρχικοποιούνται χρησιμοποιώντας την κανονική κατανομή $\mathcal{N} (\mu, \sigma)$ με $\mu = 0$ και $\sigma = 0.1$.
Χρησιμοποιήσαμε τον Adam~\cite{kingma2014adam} ως βελτιστοποιητή με ρυθμό μάθησης $\lambda = 0.01$, $b_1 = 0.9$, $b_2 = 0.999$, epsilon $\epsilon = 10^{-8}$ χωρίς σταδιακή απομείωση βαρών.
Για την υλοποίηση και την εκπαίδευση των SANs χρησιμοποιήσαμε το Pytorch~\cite{paszke2017automatic}, με NVIDIA Titan X Pascal GPU 12GB RAM και 12 Core Intel i7-8700 CPU @ 3.20GHz σε λειτουργικό σύστημα Linux.

\subsection{Σύγκριση του μέτρου $\varphi$ για συναρτήσεις αραιής ενεργοποίησης και διάφορα μεγέθη πυρήνα στην Physionet}
Εδώ μελετάμε την επίδραση στο $\bar\varphi$, της επιλογής του μεγέθους πυρήνα $m$ και των συναρτήσεων αραιής ενεργοποίησης που ορίστηκαν στην ενότητα~\ref{sec6:safs}.

\subsubsection{Βάσεις δεδομένων}
Χρησιμοποιούμε ένα σήμα από καθεμία από τις 15 βάσεις δεδομένων της Physionet που αναφέρονται στην πρώτη στήλη του Πίνακα~\ref{table:crrl}.
Κάθε σήμα αποτελείται από $12000$ δείγματα τα οποία με τη σειρά τους κατανέμονται σε $12$ σήματα με $1000$ δείγμα το καθένα, για τη δημιουργία των σημάτων εκπαίδευσης ($6$), επικύρωσης ($2$ σήματα) και δοκιμής ($4$ σήματα).
Η μόνη προεπεξεργασία που γίνεται είναι η αφαίρεση του μέσου όρου και διαίρεση με την τυπική απόκλιση στα σήματα των $1000$ δειγμάτων.

\subsubsection{Ρύθμιση πειράματος}
Εκπαιδεύουμε τέσσερα SAN (ένα για κάθε συνάρτηση αραιής ενεργοποίησης) για καθεμία από τις $15$ βάσεις δεδομένων της Physionet για $30$ εποχές με μέγεθος παρτίδας $2$ και μέγεθος πυρήνων που κυμαίνονται στην περιοχή $[1, 250]$.
Κατά τη διάρκεια της επικύρωσης επιλέξαμε τα μοντέλα με μέγεθος πυρήνα που πέτυχαν το καλύτερο $\bar\varphi$ από όλες τις εποχές.
Κατά τη διάρκεια των δοκιμών τροφοδοτούμε τα δεδομένα δοκιμής στο επιλεγμένο μοντέλο και υπολογίζουμε τα $CR^{-1}$, $\tilde{\mathcal{L}}$ και $\bar\varphi$ για αυτό το σύνολο υπερπαραμέτρων όπως φαίνεται στον Πίνακα~\ref{table:crrl}.
Για τη συνάρτηση ενεργοποίησης ακρότατων θέτουμε `συνοριακή ανοχή' τριών δειγμάτων.

\subsubsection{Αποτελέσματα}
\begin{figure}
	\input{chapter6-crrl.tex}
	\caption[Αντίστροφος λόγος συμπίεσης ($CR^{-1}$) έναντι κανονικοποιημένης απώλειας ανακατασκευής ($\tilde{\mathcal{L}}$) για $15$ βάσεις δεδομένων της Physionet και για διάφορα μεγέθη πυρήνα.]{Αντίστροφος λόγος συμπίεσης ($CR^{-1}$) έναντι κανονικοποιημένης απώλειας ανακατασκευής ($\tilde{\mathcal{L}}$) για $15$ βάσεις δεδομένων της Physionet και για διάφορα μεγέθη πυρήνα.
	Οι πέντε μικρές γραφικές παραστάσεις με το κίτρινο φόντο στα δεξιά της κάθε εικόνας, απεικονίζουν τον αντίστοιχο πυρήνα για το μέγεθος του πυρήνα που πέτυχε το καλύτερο $\varphi$.}
	\label{fig:crrl}
\end{figure}
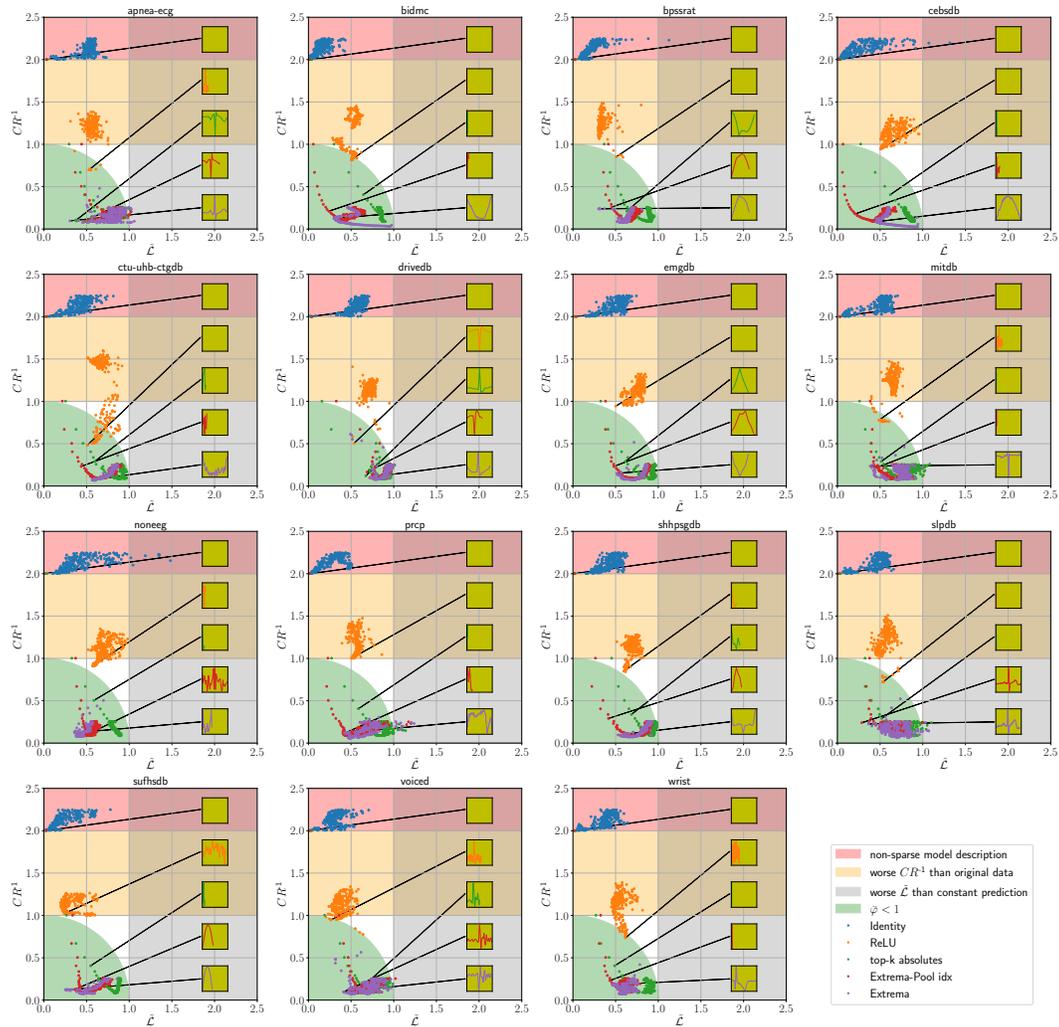

\begin{sidewaysfigure}
	\centering
	\subfloat[Διάγραμμα πυκνότητας $CR^{-1}$ vs. $\tilde{\mathcal{L}}$]{\includegraphics[width=0.32\textwidth]{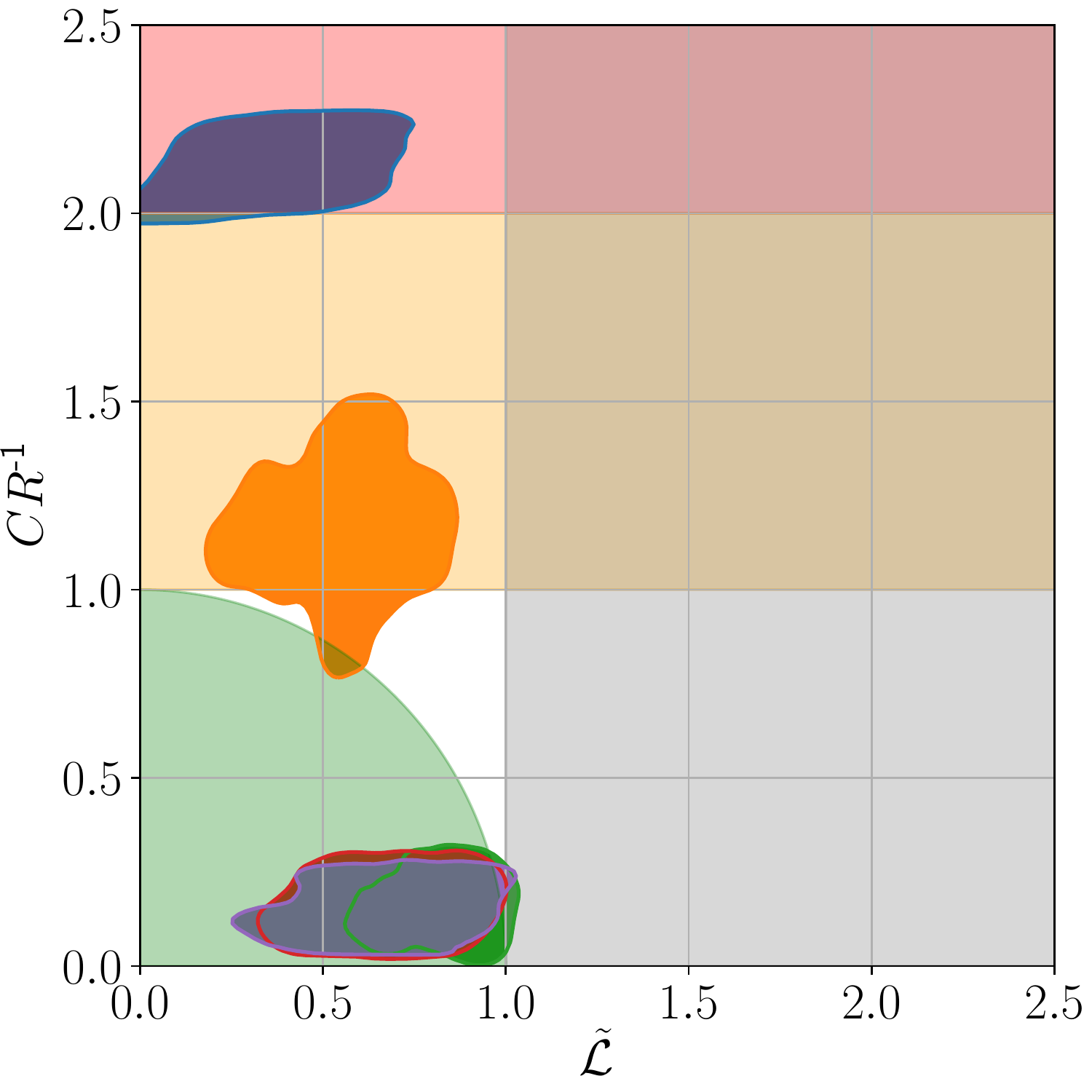}\label{subfig:crrl_density_plot}}
	\subfloat[Διαστήματα εμπιστοσύνης $\bar\varphi$ vs. $epochs$]{\includegraphics[width=0.32\textwidth]{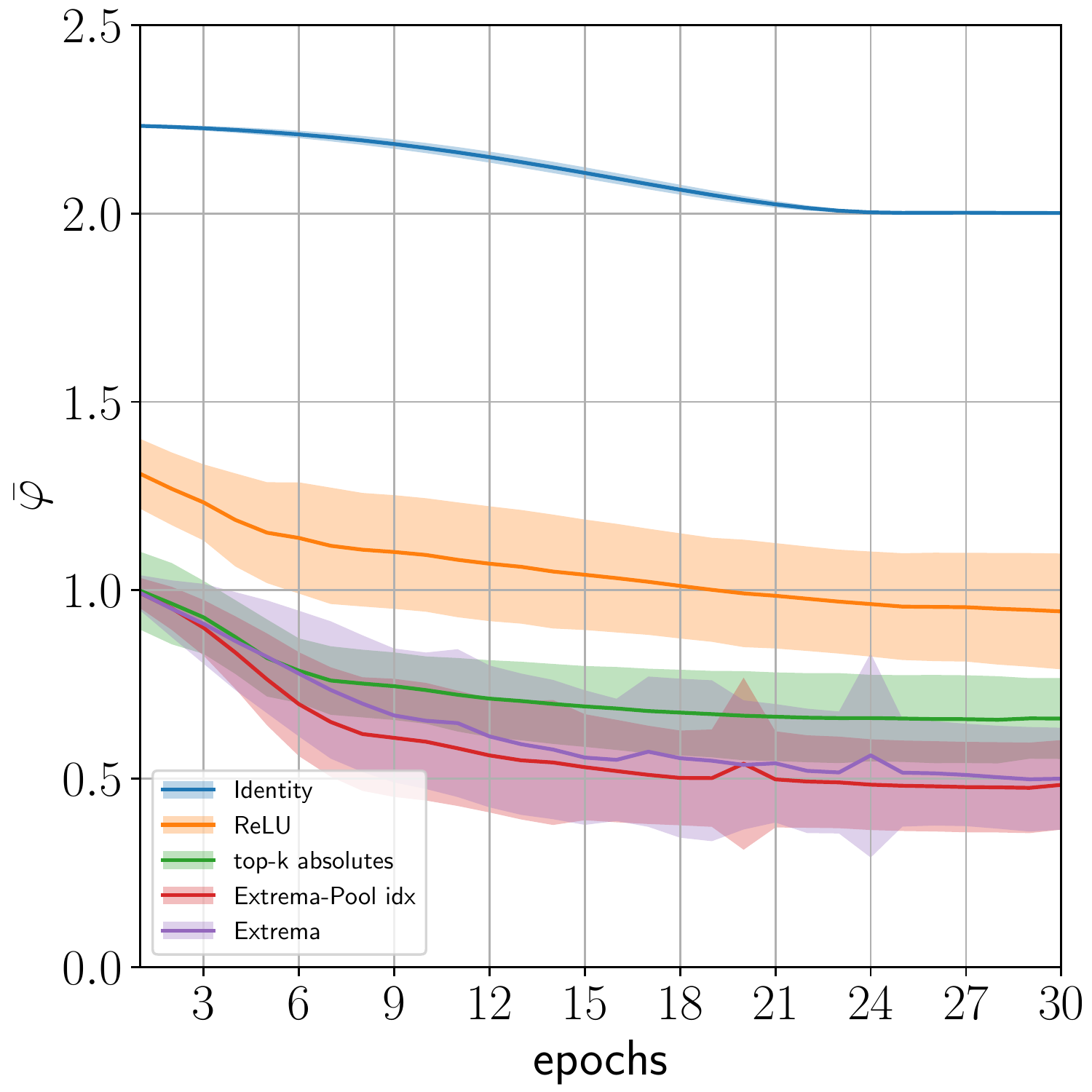}\label{subfig:flithos_epochs}}
	\subfloat[Διαστήματα εμπιστοσύνης $\bar\varphi$ vs. $m$]{\includegraphics[width=0.32\textwidth]{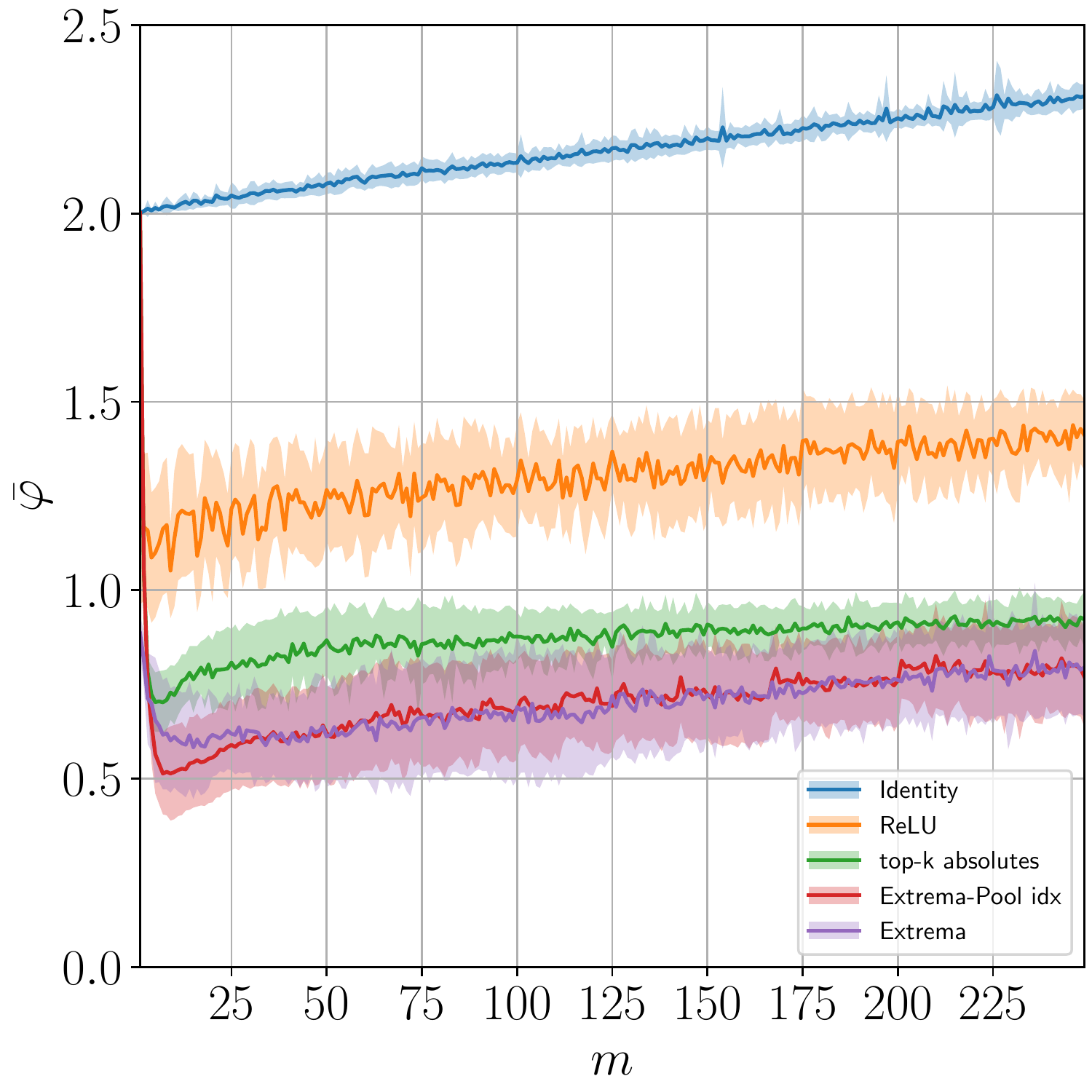}\label{subfig:flithos_m}}
	\caption{Συνολικά αποτελέσματα της αξιολόγησης των βάσεων δεδομένων της Physionet με τη χρήση του μέτρου $\varphi$.
	Το διάγραμμα πυκνότητας δημιουργήθηκε χρησιμοποιώντας εκτίμηση πυκνότητας γκαουσιανών πυρήνων και τα διαστήματα εμπιστοσύνης απεικονίζουν μια τυπική απόκλιση.}
	\label{fig:flithos}
\end{sidewaysfigure}

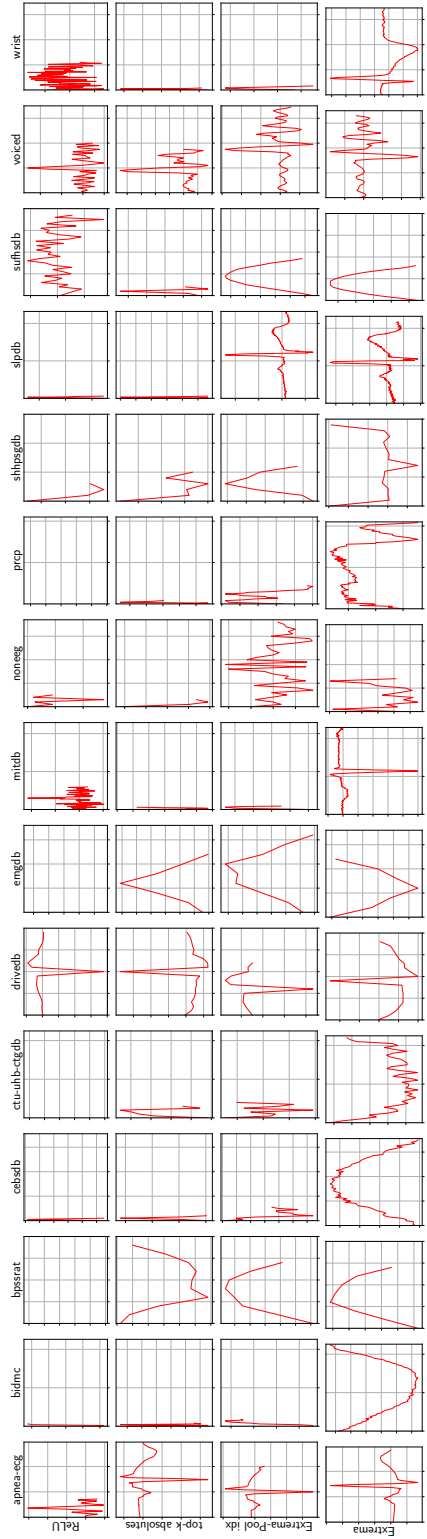
\begin{sidewaysfigure}
	\centering
	\input{chapter6-kernelvisualization.tex}
	\caption{Οπτικοποίηση των πυρήνων για κάθε αραιή συνάρτηση ενεργοποίησης (γραμμή) και για κάθε βάση δεδομένων της Physionet (στήλη).
	}
	\label{fig:kernelvisualization}
\end{sidewaysfigure}

Η ύπαρξη των τριών ξεχωριστών ομάδων που απεικονίζονται στην Εικ.~\ref{fig:crrl} και Εικ.~\ref{fig:flithos}\subref{subfig:crrl_density_plot} μεταξύ της Identity, του ReLU και των υπολοίπων είναι το αποτέλεσμα της επίδρασης της αραιότητας των συναρτήσεων ενεργοποίησης στις αναπαραστάσεις.
Όσο αραιότερη η συνάρτηση ενεργοποίησης είναι τόσο περισσότερο συμπιέζεται η αναπαράσταση, μερικές φορές σε βάρος του σφάλματος ανακατασκευής.
Ωστόσο, με οπτική επιθεώρηση της Εικ.\ref{fig:kernelvisualization} μπορούμε να επιβεβαιώσουμε ότι τα βάρη των SANs με αραιότερους χάρτες ενεργοποίησης (Extrema-Pool indices και Extrema) αντιστοιχούν σε επαναλαμβανόμενα μοτίβα των βάσεων δεδομένων, επιτυγχάνοντας έτσι υψηλή ερμηνευσιμότητα.
Αυτά τα αποτελέσματα δείχνουν ότι το σφάλμα ανακατασκευής από μόνο του δεν είναι επαρκές μέτρο για την αποσύνθεση δεδομένων σε ερμηνεύσιμα στοιχεία.
Προσπαθώντας να επιτύχουμε αποκλειστικά μικρότερο σφάλμα ανακατασκευής (όπως στην περίπτωση της Identity) έχουμε ως αποτέλεσμα θορυβώδεις πυρήνες, ενώ το συνδυασμένο μέτρο του σφάλματος ανακατασκευής και του λόγου συμπίεσης (μικρότερο $\bar\varphi$) έχει ως αποτέλεσμα ερμηνεύσιμους πυρήνες.
Συγκρίνοντας τις διαφορές στο $\bar\varphi$ μεταξύ του Identity, του ReLU και των υπόλοιπων συναρτήσεων αραιής ενεργοποίησης στην Εικ.~\ref{fig:flithos}\subref{subfig:flithos_m} παρατηρούμε ότι οι τελευταίες παράγουν μια περιοχή ελαχίστων στην οποία έχουμε ως αποτέλεσμα ερμηνεύσιμους πυρήνες.

\begin{sidewaystable}
	\centering
	\caption{Μέγεθος πυρήνα $m$ με το καλύτερο $\varphi$ για κάθε συνάρτηση αραιής ενεργοποίησης για κάθε βάση δεδομένων της Physionet}
	\label{table:crrl}
	\input{table-mean-inverse-compression-ratio-mean-reconstruction-loss-variable-kernel-size.tex}
\end{sidewaystable}

\subsection{Αξιολόγηση της ανακατασκευής των SANs με χρήση επιβλεπώμενου CNN για αναγνώριση επιληψίας στην UCI}
Εδώ μελετάμε την ποιότητα των ανακατασκευών των SAN εκπαιδεύοντας ένα επιτηρούμενο 1D Συνελικτικό Νευρωνικό Δίκτυο (CNN) στις εξόδους του κάθε SAN\@.
Επίσης, μελετάμε την επίδραση που έχει το $m$ στο $\bar\varphi$ και την ακρίβεια του ταξινομητή για τις πέντε συναρτήσεις αραιής ενεργοποίησης.

\subsubsection{Βάση δεδομένων}
Χρησιμοποιούμε τη βάση δεδομένων αναγνώρισης επιληψίας από την UCI που αποτελείται από $500$ σήματα των $4097$ δειγμάτων (23.5 δευτερόλεπτα) το καθένα.
Η βάση δεδομένων απαριθμείται σε πέντε κατηγορίες με $100$ σήματα για κάθε κατηγορία.
Για τους σκοπούς αυτής της εργασίας χρησιμοποιούμε μια παραλλαγή της βάσης δεδομένων\footnote{\url{https://archive.ics.uci.edu/ml/datasets/Epileptic+Seizure+Recognition}} στην οποία τα EEG σήματα χωρίζονται σε τμήματα των $178$ δειγμάτων το καθένα, με αποτέλεσμα μια ισορροπημένη βάση δεδομένων που αποτελείται από $11500$ σήματα EEG συνολικά.

\subsubsection{Ρύθμιση πειράματος}
Αρχικά, συγχωνεύουμε τις κατηγορίες όγκων ($2$ και $3$) και των ματιών ($4$ και $5$) με αποτέλεσμα μια τροποποιημένη βάση δεδομένων τριών κατηγοριών (όγκος, μάτια, επιληψία).
Στη συνέχεια διαιρούμε τα $11500$ σήματα σε $76\%$, $12\%$ και $12\%$ ($8740,1380,1380$) ως δεδομένα εκπαίδευσης, επικύρωσης και δοκιμής αντίστοιχα και κανονικοποιούμε στο εύρος $[0, 1]$ χρησιμοποιώντας το ολικό μέγιστο και ελάχιστο.
Για τα SAN χρησιμοποιήσαμε δύο πυρήνες $q = 2$ με μεταβλητό μήκος εύρους $[15, 22]$ και εκπαιδεύσαμε για $5$ εποχές με μέγεθος παρτίδας $32$.
Μετά την εκπαίδευση, επιλέγουμε το μοντέλο που εμφάνισε το χαμηλότερο $\bar\varphi$ από όλες τις εποχές.

Κατά τη διάρκεια της επιβλεπώμενης μάθησης, τα βάρη των πυρήνων παγώνουν και ένα CNN στοιβάζεται πάνω από τις ανακατασκευές των SAN\@.
Ο εξαγωγέας χαρακτηριστικών του CNN αποτελείται από δύο συνελικτικά επίπεδα με φίλτρα $3$ και $16$ και μέγεθος πυρήνα $5$, το καθένα από τα οποία ακολουθείται από ένα ReLU και Extrema-Pool με μέγεθος συγκέντρωσης $2$.
Ο ταξινομητής αποτελείται από τρία πλήρως συνδεδεμένα επίπεδα με μονάδες $656$, $120$ και $84$.
Τα πρώτα δύο από τα πλήρως συνδεδεμένα επίπεδα ακολουθούνται από ένα ReLU ενώ το τελευταίο περνάει από ένα log-softmax το οποίο παράγει τις προβλέψεις.
Το CNN εκπαιδεύεται για $5$ πρόσθετες εποχές με το ίδιο μέγεθος παρτίδας και τη διαδικασία επιλογής μοντέλου όπως και με τα SAN και την αρνητική λογαριθμική πιθανοφάνεια ως συνάρτηση απώλειας.
Για τη συνάρτηση ενεργοποίησης ακρότατων θέτουμε `συνοριακή ανοχή' δύο δειγμάτων.

\subsubsection{Αποτελέσματα}
Όπως φαίνεται στον Πίνακα.~\ref{table:uciepilepsysupervised}, αν και χρησιμοποιούμε ένα σημαντικά μειωμένο μέγεθος αναπαράστασης, η ακρίβεια δεν πέφτει ανάλογα κάτι το οποίο δείχνει ότι τα SANs επιλέγουν τα πιο σημαντικά χαρακτηριστικά για να αναπαραστήσουν τα δεδομένα.
Για παράδειγμα, για $m = 15$ για την Peak συνάρτηση ενεργοποίησης, υπάρχει μια πτώση ακρίβειας $1.44\%$ (το CNN βάση αναφοράς με τα αρχικά δεδομένα πέτυχε $\DTLfetch{keys_values}{key}{uci_epilepsy_supervised_accuracy}{value}\%$) παρόλο που χρησιμοποιήθηκε μια μειωμένη αναπαράσταση με μόλις $34\%$ μέγεθος σε σχέση με τα αρχικά δεδομένα.

\begin{sidewaystable}
	\centering
	\caption{SANs με επιβλεπώμενο στοιβαγμένο CNN για αναγνώριση επιληψίας στην UCI}
	\label{table:uciepilepsysupervised}
	\input{table-uci-epilepsy-supervised.tex}
\end{sidewaystable}

\subsection{Αξιολόγηση της ανακατασκευής των SANs με χρήση επιβλεπώμενου FNN στην MNIST και FMNIST}
\subsubsection{Βάση δεδομένων}
Για την ίδια εργασία με την προηγούμενη, αλλά για 2D, χρησιμοποιούμε την MNIST~\cite{lecun1998gradient} η οποία αποτελείται από μια βάση δεδομένων εκπαίδευσης $60000$ χειρόγραφων ψηφίων στην κλίμακα του γκρι και μια βάση δεδομένων δοκιμής με $10000$ εικόνες καθεμία με μέγεθος $28\times 28$.
Η ίδια διαδικασία ακολουθείται και για την FMNIST~\cite{xiao2017fashion}.

\subsubsection{Ρύθμιση πειράματος}
Τα μοντέλα αποτελούνται από δύο πυρήνες $q = 2$ με μεταβλητό μήκος στο εύρος $[1, 6]$.
Χρησιμοποιούμε $10000$ εικόνες από τη βάση δεδομένων εκπαίδευσης για επικύρωση και εκπαιδεύουμε με τα υπόλοιπα $50000$ για $5$ εποχές και μέγεθος παρτίδας $64$.
Δεν εφαρμόζουμε προεπεξεργασία στις εικόνες.

Κατά τη διάρκεια της επιβλεπώμενης μάθησης, τα βάρη των πυρήνων παγώνουν και ένα μονοστρωματικό πλήρως συνδεδεμένο δίκτυο (FNN) στοιβάζεται πάνω από τις τελικές ανακατασκευές των SAN\@.
Το FNN εκπαιδεύεται για $5$ πρόσθετες εποχές με την ίδια διαδικασία επιλογής μοντέλου και μέγεθος παρτίδας όπως και με τα SANs και την αρνητική λογαριθμική πιθανοφάνεια ως συνάρτηση απώλειας.
Για τη συνάρτηση ενεργοποίησης ακρότατων θέτουμε `συνοριακή ανοχή' δύο δειγμάτων.

\subsubsection{Αποτελέσματα}
Όπως φαίνεται στον Πίνακα~\ref{table:mnistsupervised}, η ακρίβεια που επιτυγχάνεται με τις ανακατασκευές ορισμένων SANs είναι συγκρίσιμη με εκείνες ενός FNN που έχει εκπαιδευτεί στα αρχικά δεδομένα ($\DTLfetch{keys_values}{key}{mnist_supervised_accuracy}{value}\%$), αν και έχουν συμπιεστεί σε μεγάλο βαθμό.
Είναι ενδιαφέρον να τονιστεί ότι σε μερικές περιπτώσεις οι ανακατασκευές των SANs, όπως των Extrema-Pool indices, πέτυχαν καλύτερη ακρίβεια από τα αρχικά δεδομένα.
Αυτό υποδηλώνει τη συντριπτική παρουσία περιττής πληροφορίας που βρίσκεται στις αρχικές εικόνες των αρχικών δεδομένων και την δυνατότητα των SANs να εξάγουν τα πιο σημαντικά χαρακτηριστικά από τα δεδομένα.

\begin{sidewaystable}
	\centering
	\caption{SANs με επιβλεπώμενο στοιβαγμένο FNN στην MNIST}
	\label{table:mnistsupervised}
	\input{table-mnist-supervised.tex}
\end{sidewaystable}

\section{Συζήτηση}
\label{sec6:discussion}
Τα SANs σε συνδυασμό με το μέτρο $\varphi$ συμπιέζουν την περιγραφή των δεδομένων σε $\bm{w}^{(i)}$ και $\bm{\alpha}^{(i)}$ κατά παρόμοιο τρόπο με ένα πλαίσιο γλώσσας ελάχιστης περιγραφής (Minimum Description Language).
Τα πειράματα που έγιναν στην ενότητα~\ref{sec6:experiments} δείχνουν ότι η χρήση της ταυτότητας, του ReLU και (σε μικρότερο βαθμό) των Μέγιστων-Ενεργοποιήσεων παράγουν θορυβώδη χαρακτηριστικά, ενώ από την άλλη πλευρά οι δείκτες συγκέντρωσης ακρότατων και τα ακρότατα παράγουν σταθερά χαρακτηριστικά και μπορούν να προσαρμοστούν χρησιμοποιώντας παραμέτρους (μήκος πυρήνα, και $med$) των οποίων οι τιμές μπορούν να καθοριστούν με απλή επισκόπηση των δεδομένων.

Από την πλευρά της μάθησης αραιών λεξιλογίων (Sparse Dictionary Learning), οι πυρήνες των SANs θα μπορούσαν να θεωρηθούν ως άτομα ενός λεξικού που ειδικεύεται στην ερμηνευτική αντιστοίχιση μοτίβων (e.g.\ για είσοδο ECG οι πυρήνες των SAN είναι ECG beats) και ο χάρτης αραιής ενεργοποίησης ως η αναπαράσταση.
Το γεγονός ότι τα SANs είναι ευρύ με λιγότερα και μεγαλύτερα μεγέθη πυρήνα αντί για βαθιά με μικρότερα και περισσότερα μεγέθη πυρήνα, τα καθιστούν πιο ερμηνεύσιμα από τα DNN και σε ορισμένες περιπτώσεις χωρίς να θυσιάζουν σημαντική ακρίβεια.

Ένα πλεονέκτημα των SANs σε σχέση με τους Αραιούς Αυτοκωδικοποιητές (Sparse Autoencoders)~\cite{ng2011sparse} είναι ότι ο περιορισμός της εγγύτητας των ενεργοποιήσεων μπορεί να εφαρμοστεί ατομικά για κάθε διάνυσμα εισόδου σε αντίθεση με τον υπολογισμό της προς-τα-εμπρός διάδοσης όλων των διανυσμάτων εισόδου.
Επιπλέον, τα SANs δημιουργούν ακριβή μηδενικά αντί για περίπου-μηδενικά, κάτι το οποίο μειώνει την προσαρμογή μεταξύ των ενεργοποιήσεων των νευρώνων.

Το $\varphi$ θα μπορούσε να θεωρηθεί ως μια εναλλακτική τυποποίηση του ξυραφιού του Occam~\cite{soklakov2002occam}, όπως η θεωρία του Solomonov για την επαγωγική εξαγωγή~\cite{solomonoff1964formal}, αλλά με μια αιτιοκρατική ερμηνεία αντί για πιθανοτική.
Το κόστος της περιγραφής των δεδομένων μπορεί να θεωρηθεί ότι είναι ανάλογο του αριθμού των βαρών και του αριθμού των μη-μηδενικών ενεργοποιήσεων, ενώ η ποιότητα της περιγραφής είναι ανάλογη με την απώλεια ανακατασκευής.
Το μέτρο $\varphi$ σχετίζεται επίσης με τη θεωρία του rate-distortion~\cite{burger1971rate}, όπου η μέγιστη παραμόρφωση ορίζεται σύμφωνα με την ανθρώπινη αντίληψη, η οποία όμως αναπόφευκτα εισάγει προκατάληψη.
Υπάρχει επίσης σχέση με τον τομέα της συμπιεσμένης ανίχνευσης (Compressed Sensing)~\cite{donoho2006compressed}, στην οποία εκμεταλλευόμαστε την αραιότητα των δεδομένων, επιτρέποντάς μας να τα ανακατασκευάσουμε με λιγότερα δείγματα από αυτό που απαιτείται από το θεώρημα Nyquist-Shannon και τον τομέα της εξαγωγής σταθερών χαρακτηριστικών (Robust Feature Extraction)~\cite{kim2013deep} όπου τα χαρακτηριστικά χρησιμοποιούνται για αντιπροσώπευση των αρχικών δεδομένων.
Οι Olshausen et al.~\cite{olshausen1996emergence} παρουσίασαν μια συνάρτηση βελτιστοποίησης που εξετάζει υποκειμενικά μέτρα της αραιότητας των χαρτών ενεργοποίησης, ωστόσο σε αυτό το έργο χρησιμοποιούμε το άμεσο μέτρο του λόγου συμπίεσης.
Προηγούμενες εργασίες όπως~\cite{zhang2017ecg} χρησιμοποίησαν ένα σταθμισμένο συνδυασμό του αριθμού των νευρώνων, της διαφοράς μεταξύ των διαστημάτων root-mean-squared και ενός συντελεστή συσχέτισης για τη συνάρτηση βελτιστοποίησης ενός FNN ως μέτρο αλλά χωρίς να ληφθεί υπόψη ο αριθμός των μη-μηδενικών ενεργοποιήσεων.

Ένας περιορισμός των SAN είναι η χρήση μεταβλητών πυρήνων μόνο στο πλάτος, το οποίο δεν επαρκεί για πιο περίπλοκα δεδομένα και επίσης δεν αξιοποιείται πλήρως η συμπιεστότητα των δεδομένων.
Μια πιθανή λύση θα ήταν να χρησιμοποιηθεί ένας δειγματολήπτης πλέγματος~\cite{jaderberg2015spatial} για τον πυρήνα, ο οποίος θα του επέτρεπε να μάθει πιο γενικές μεταβολές (όπως κλίμακα) από την απλή μεταβλητότητα πλάτους.
Ωστόσο, οι πρόσθετες ιδιότητες του πυρήνα θα πρέπει να επιλέγονται παίρνοντας υπόψη τις επιπτώσεις στο μέτρο $\varphi$; το μοντέλο θα πρέπει να συμπιεστεί περισσότερο με μειωμένη απώλεια ανακατασκευής.

\clearpage
\bibliography{chapter6.bib}
\bibliographystyle{unsrt}

%% file: chapter6-topk-absolutes.tex
\begin{algorithmic}[1]
	\renewcommand{\algorithmicrequire}{\textbf{Input:}}
	\renewcommand{\algorithmicensure}{\textbf{Output:}}
	\REQUIRE $s$, $k$
	\ENSURE $\alpha$
	\STATE $\alpha_i \leftarrow 0, i=1\ldots card(s)$
	\STATE $p \leftarrow topk(\lvert s\rvert, k)$
	\FOR {$i$ = 0 to $card(s)$}
	\STATE $\alpha_i(p_i) = s_i(p_i)$
	\ENDFOR
	\RETURN $\alpha$
\end{algorithmic}

%% file: chapter6-extremapoolindices.tex
\begin{algorithmic}[1]
	\renewcommand{\algorithmicrequire}{\textbf{Input:}}
	\renewcommand{\algorithmicensure}{\textbf{Output:}}
	\REQUIRE $s$, $m$
	\ENSURE $\alpha$
	\STATE $p \leftarrow maxpool(\lvert s\rvert, m)$
	\STATE $\alpha \leftarrow maxunpool(s(p), p, m)$
	\RETURN $\alpha$
\end{algorithmic}

%% file: chapter6-extrema.tex
\begin{algorithmic}[1]
	\renewcommand{\algorithmicrequire}{\textbf{Input:}}
	\renewcommand{\algorithmicensure}{\textbf{Output:}}
	\REQUIRE $s$, $med$
	\ENSURE $\alpha$
	\STATE $peaks \leftarrow \left(\frac{d s}{d t}^+ \geq 0\right) \land \left(\frac{d s}{d t}^- < 0\right)$
	\STATE $valleys \leftarrow \left(\frac{d s}{d t}^+ < 0\right) \land \left(\frac{d s}{d t}^- \geq 0\right)$ \\
	\textit{\scriptsize \# + and - denote one sample padding to the right and left respectively}
	\STATE $z = peaks \lor valleys$
	\STATE $p_i \leftarrow z > 0$
	\STATE $p_{i_i} \leftarrow sort(z)$
	\STATE $p_{i_{sorted}} \leftarrow p_i(p_{i_i})$
	\STATE $q_i \leftarrow 0, i=1\ldots card(s)$
	\FOR {$i$ = 0 to $card(s)$}
	\IF {$\lnot q_i$}
	\STATE $p_{i_r} \leftarrow p_i \geq p_{i_i} - med$
	\STATE $p_{i_l} \leftarrow p_i \leq p_{i_i} + med$
	\STATE $p_{i_m} \leftarrow p_{i_r} \land p_{i_l}$
	\STATE $q \leftarrow q \lor p_{i_m}$
	\STATE $q_i \leftarrow 0$
	\ENDIF
	\ENDFOR
	\STATE $\alpha_{ind} \leftarrow p_{i_{sorted}}(\lnot q)$
	\STATE $\alpha_i \leftarrow 0, i=1\ldots card(s)$
	\STATE $\alpha(\alpha_i) \leftarrow s(\alpha_{ind})$
	\RETURN $\alpha$
\end{algorithmic}

%% file: chapter6-san-1d.tex
\begin{scope}[tdplot_main_coords, canvas is yz plane at x=-0.5,xscale=-1, transform shape]
	\node[opacity=0] at (0, 0)(input){\includegraphics[scale=\figscale]{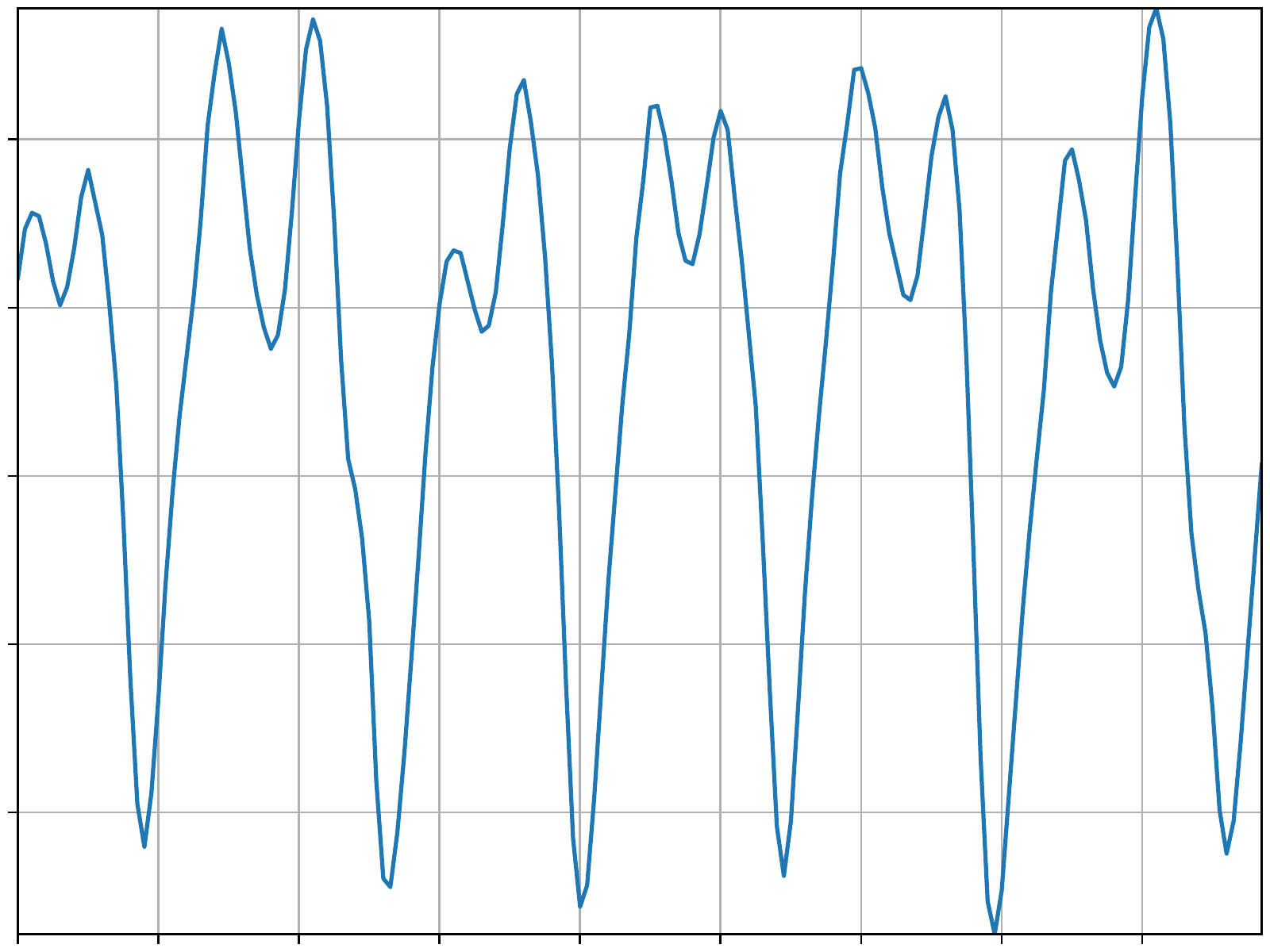}};
	\node[opacity=0, draw, right=0.5cm of input, circle] (loss){$\mathcal{L}$};
	\node[opacity=0, right=0.5cm of loss] (reconstructed){\includegraphics[scale=\figscale]{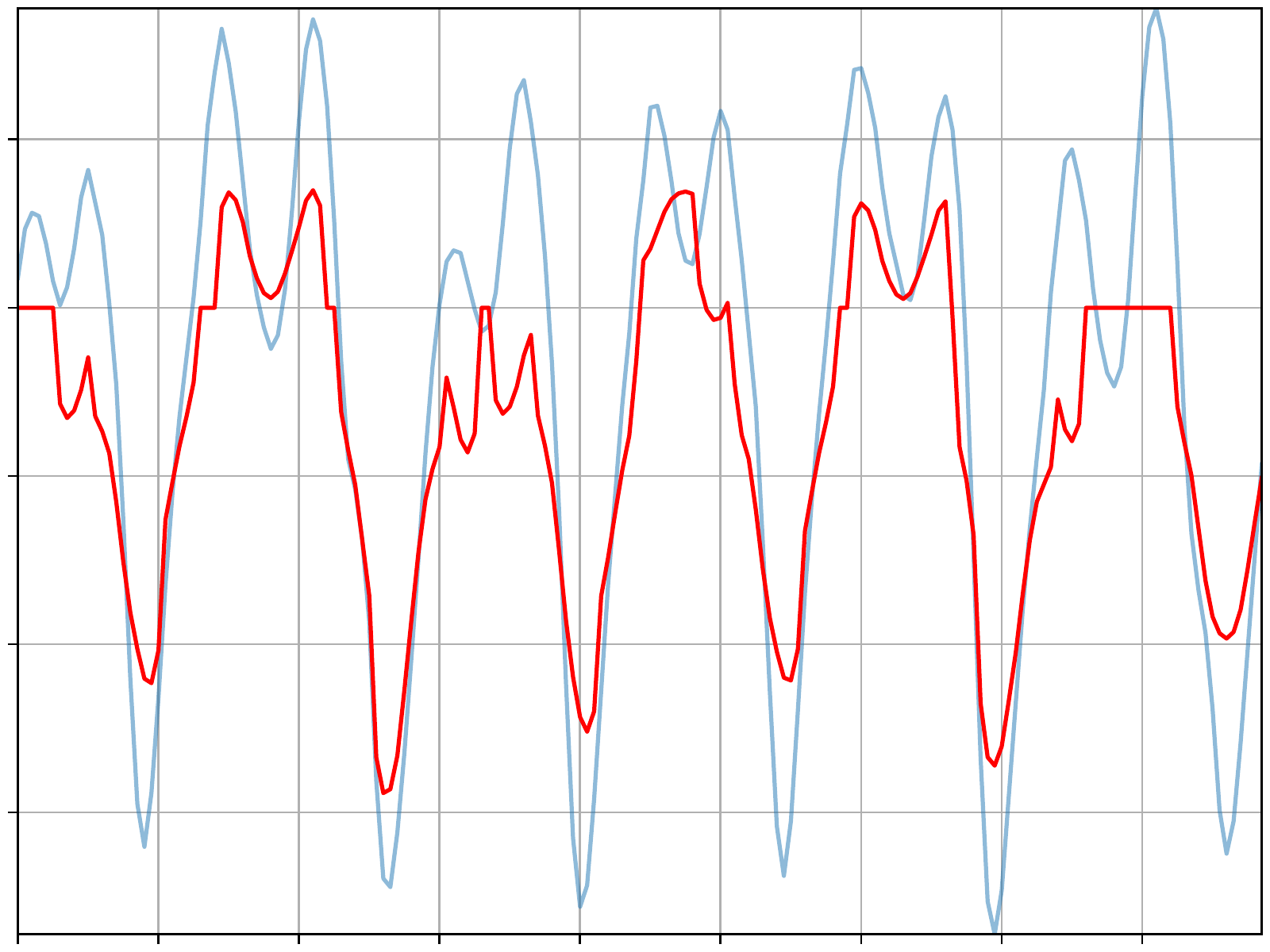}};
	\node[opacity=0, draw, below=0.5cm of reconstructed, circle] (plus){$+$};
	\node[opacity=0.8, below=0.5cm of plus] (reconstruction){\includegraphics[scale=\figscale]{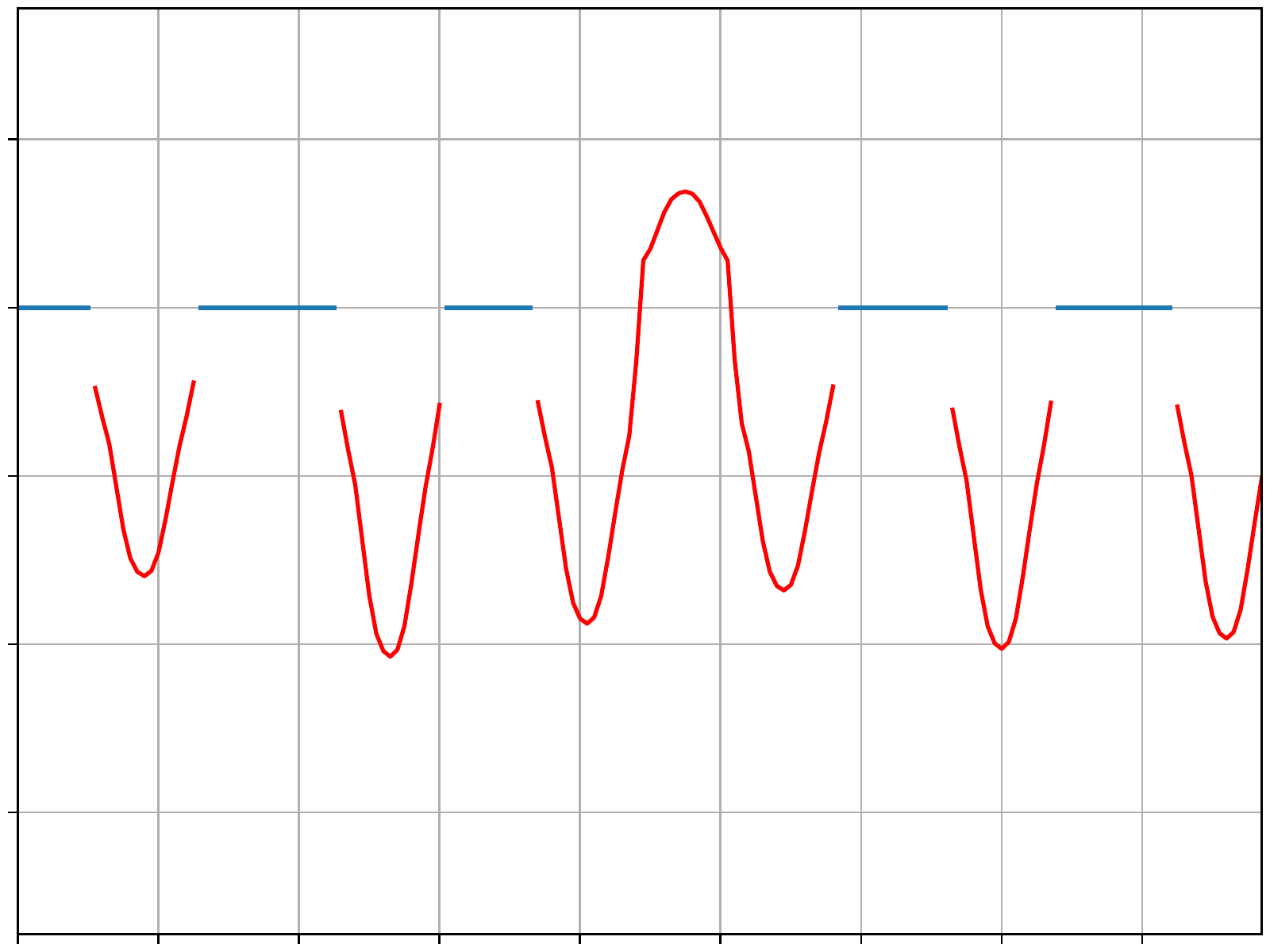}};
	\node[opacity=0, draw, below=0.5cm of reconstruction, circle] (conv2){$\ast$};
	\node[] at (4.8, -3.4){$\bm{r}^{(0)}$};
	\node[opacity=0.8, below=0.5cm of conv2] (extrema){\includegraphics[scale=\figscale]{"images-1d/UCI-epilepsy-extrema-1d-2-activations-0"}};
	\node[opacity=0, draw, left=0.55cm of extrema, circle] (phi){$\phi$};
	\node[] at (4.8, -7.4){$\bm{\alpha}^{(0)}$};
	\node[opacity=0.8, left=0.55cm of phi] (similarity){\includegraphics[scale=\figscale]{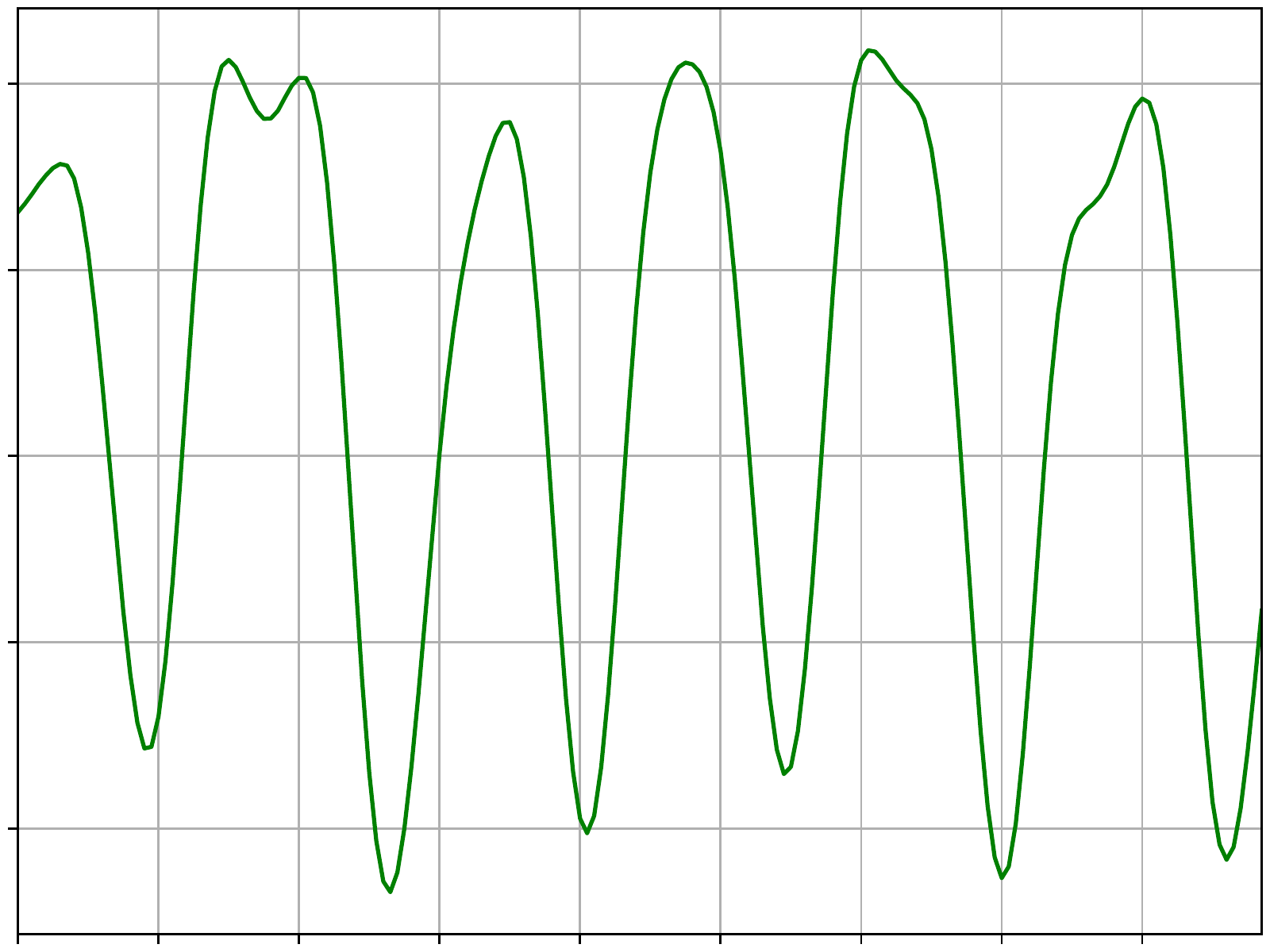}};
	\node[] at (0.1, -7.4){$\bm{s}^{(0)}$};
	\node[left=1.5cm of conv2] (kernel){\includegraphics[scale=0.1]{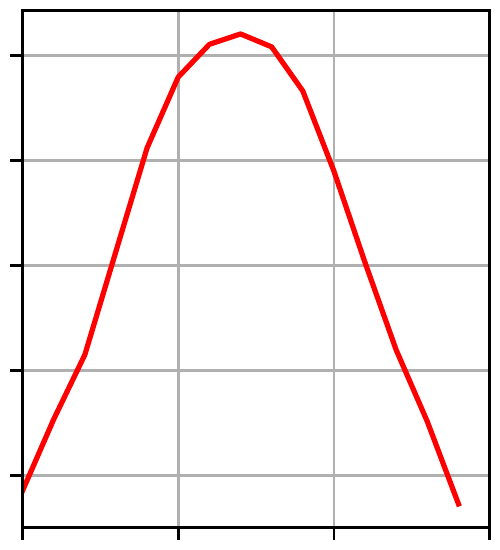}};
	\node[] at (2.6, -5.6){\tiny $\bm{w}^{(0)}$};
	\node[opacity=0, draw, above=0.5cm of similarity, circle] (conv1){$\ast$};
\end{scope}
\begin{scope}[tdplot_main_coords, canvas is yz plane at x=0,xscale=-1, transform shape]
	\node[] at (0, 0)(input){\includegraphics[scale=\figscale]{"images-1d/UCI-epilepsy-extrema-1d-2-signal"}};
	\node[] at (0.1, 0.7){$\bm{x}$};
	\node[draw, right=0.5cm of input, circle] (loss){$\mathcal{L}$};
	\node[right=0.5cm of loss] (reconstructed){\includegraphics[scale=\figscale]{"images-1d/UCI-epilepsy-extrema-1d-2-reconstructed"}};
	\node[] at (4.8, 0.7){$\hat{\bm{x}}$};
	\node[draw, below=0.5cm of reconstructed, circle] (plus){$+$};
	\node[opacity=0, below=0.5cm of plus] (reconstruction){\includegraphics[scale=\figscale]{"images-1d/UCI-epilepsy-extrema-1d-2-reconstruction-0"}};
	\node[draw, below=0.5cm of reconstruction, circle] (conv2){$\ast$};
	\node[opacity=0, below=0.5cm of conv2] (extrema){\includegraphics[scale=\figscale]{"images-1d/UCI-epilepsy-extrema-1d-2-activations-0"}};
	\node[draw, left=0.55cm of extrema, circle, inner sep=2pt] (phi){$\phi$};
	\node[opacity=0, left=0.55cm of phi] (similarity){\includegraphics[scale=\figscale]{"images-1d/UCI-epilepsy-extrema-1d-2-similarity-0"}};
	\node[opacity=0, left=1.5cm of conv2] (kernel){\includegraphics[scale=0.1]{"images-1d/UCI-epilepsy-extrema-1d-2-kernel-0"}};
	\node[draw, above=0.5cm of similarity, circle] (conv1){$\ast$};
	\draw[->](input) -- node{} (conv1);
	\draw[->](conv1) -- node{} (similarity);
	\draw[->](similarity) -- node{} (phi);
	\draw[->](phi) -- node{} (extrema);
	\draw[->](extrema) -- node{} (conv2);
	\draw[->](conv2) -- node{} (reconstruction);
	\draw[->](reconstruction) -- node{} (plus);
	\draw[->](plus) -- node{} (reconstructed);
	\draw[->](kernel) -- node{} (conv1);
	\draw[->](kernel) -- node{} (conv2);
	\draw[->](input) -- node{} (loss);
	\draw[->](reconstructed) -- node{} (loss);
\end{scope}
\begin{scope}[tdplot_main_coords, canvas is yz plane at x=0.5,xscale=-1, transform shape]
	\node[opacity=0] at (0, 0)(input){\includegraphics[scale=\figscale]{"images-1d/UCI-epilepsy-extrema-1d-2-signal"}};
	\node[opacity=0, draw, right=0.5cm of input, circle] (loss){$\mathcal{L}$};
	\node[opacity=0, right=0.5cm of loss] (reconstructed){\includegraphics[scale=\figscale]{"images-1d/UCI-epilepsy-extrema-1d-2-reconstructed"}};
	\node[opacity=0, draw, below=0.5cm of reconstructed, circle] (plus){$+$};
	\node[opacity=0.8, below=0.5cm of plus] (reconstruction){\includegraphics[scale=\figscale]{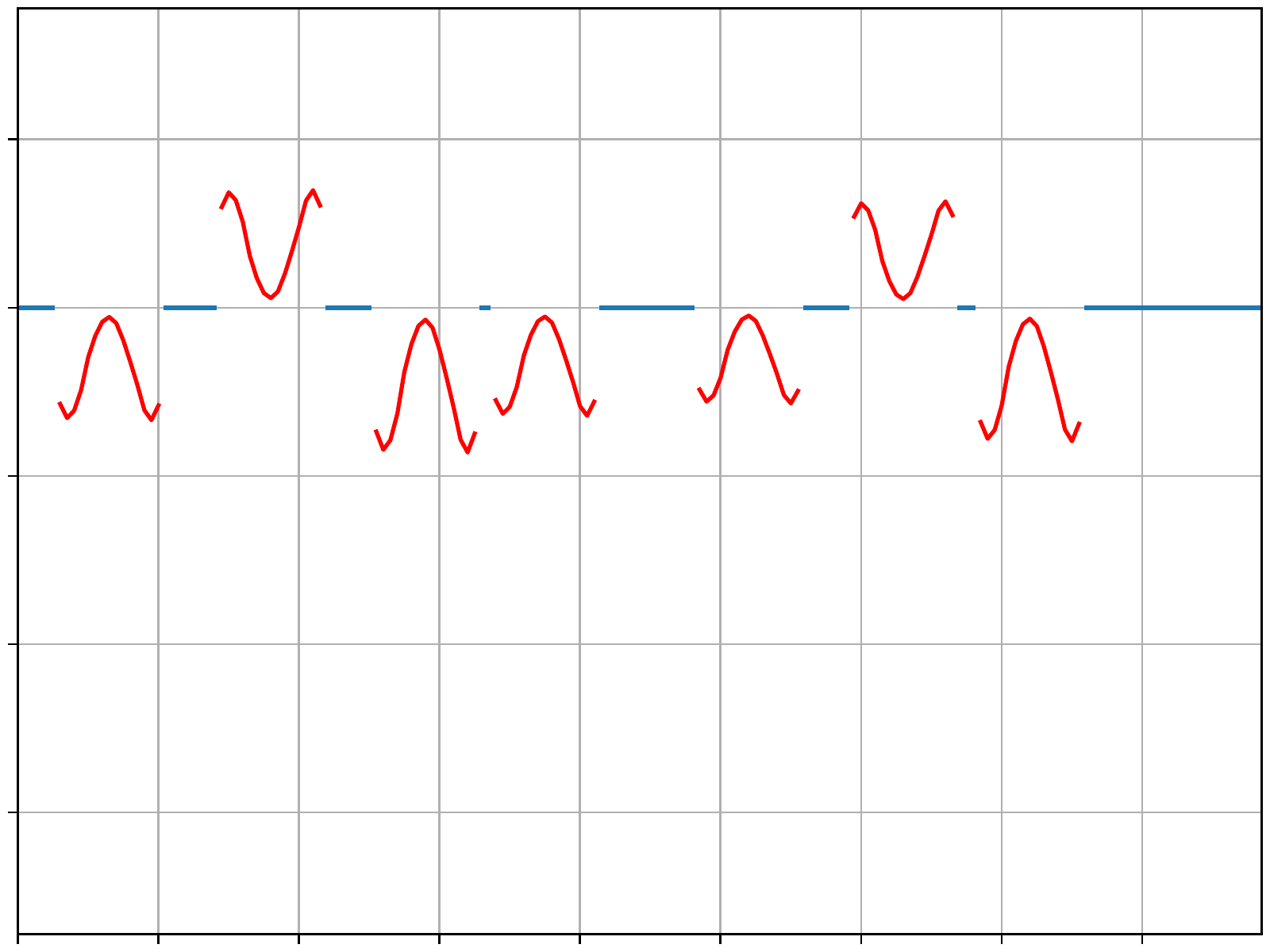}};
	\node[] at (4.8, -3.4){$\bm{r}^{(1)}$};
	\node[opacity=0, draw, below=0.5cm of reconstruction, circle] (conv2){$\ast$};
	\node[opacity=0.8, below=0.5cm of conv2] (extrema){\includegraphics[scale=\figscale]{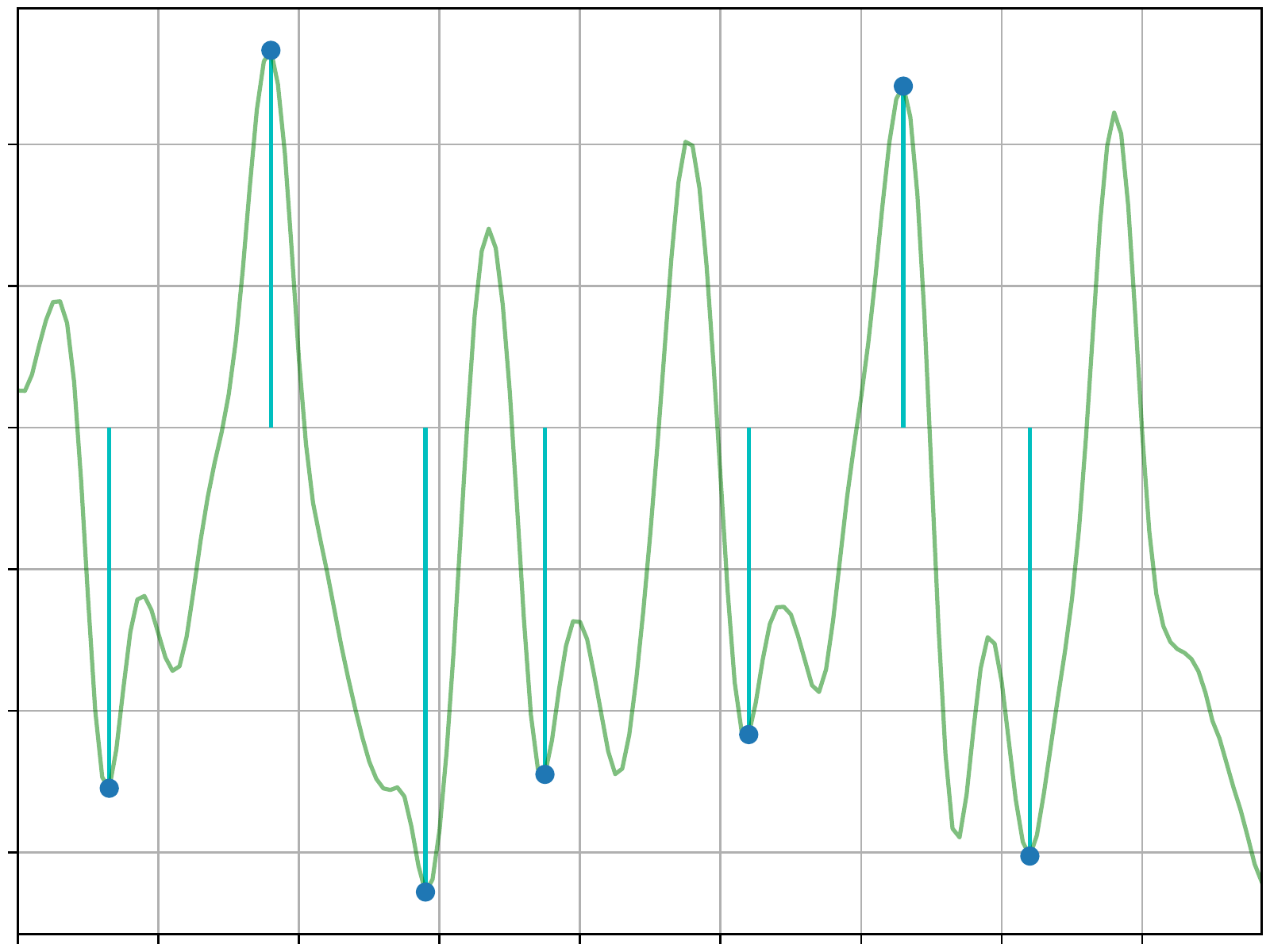}};
	\node[] at (4.8, -7.4){$\bm{\alpha}^{(1)}$};
	\node[opacity=0, draw, left=0.55cm of extrema, circle] (phi){$\phi$};
	\node[opacity=0.8, left=0.55cm of phi] (similarity){\includegraphics[scale=\figscale]{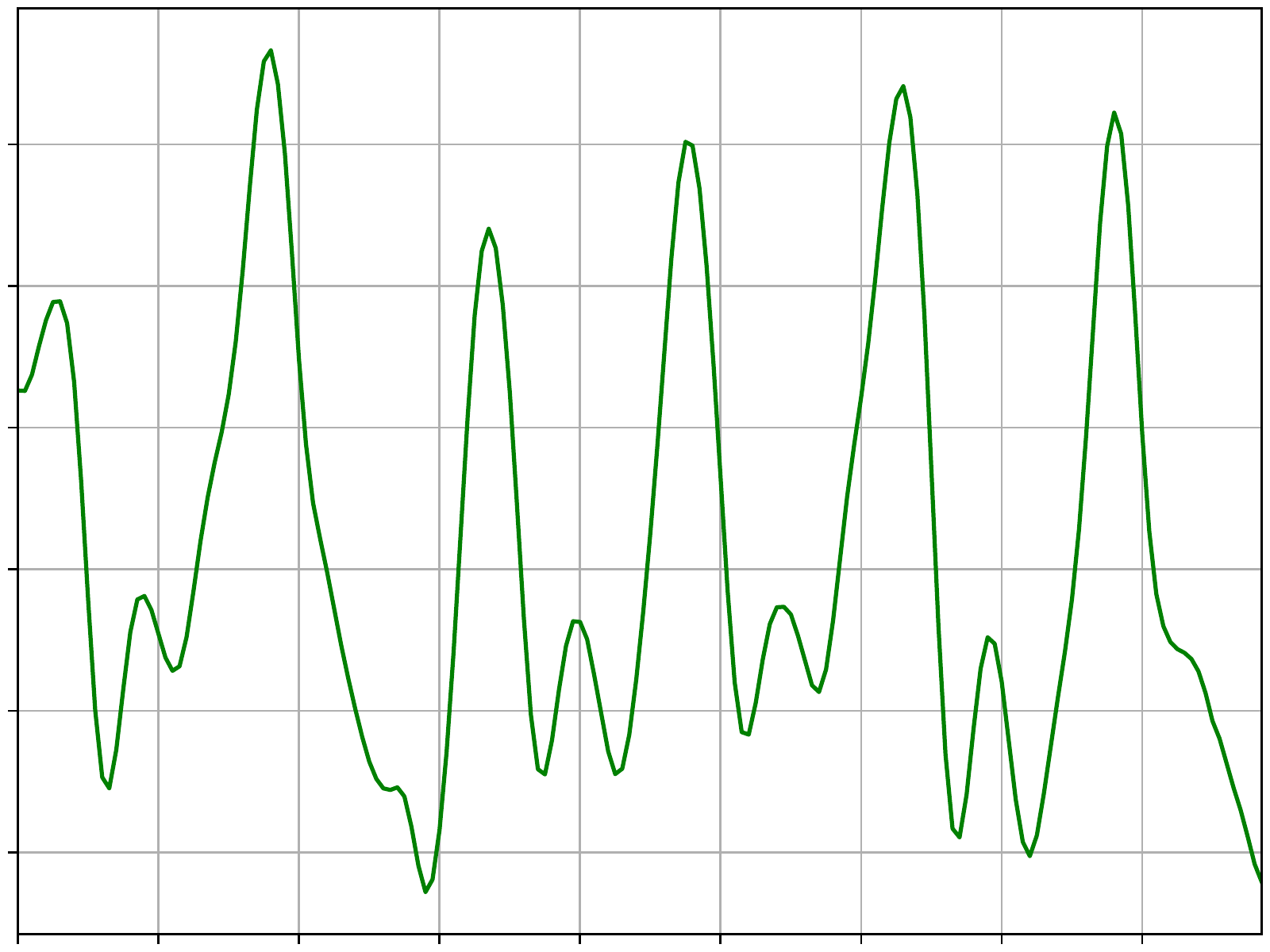}};
	\node[] at (0.1, -7.4){$\bm{s}^{(1)}$};
	\node[left=1.5cm of conv2] (kernel){\includegraphics[scale=0.1]{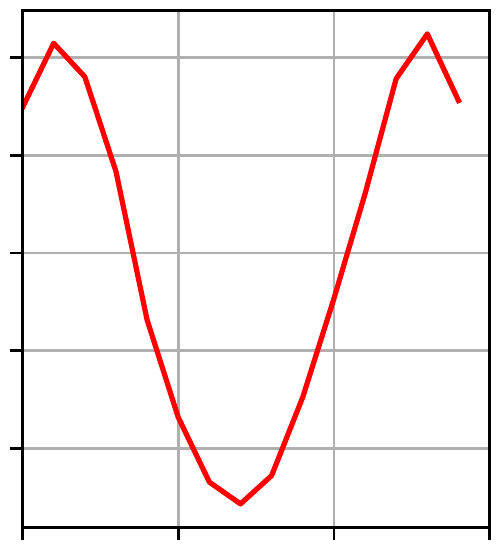}};
	\node[] at (2.6, -5.6){\tiny $\bm{w}^{(1)}$};
	\node[opacity=0, draw, above=0.5cm of similarity, circle] (conv1){$\ast$};
\end{scope}

%% file: chapter6-san-2d.tex
\begin{scope}[tdplot_main_coords, canvas is yz plane at x=-0.5,xscale=-1, transform shape]
	\node[opacity=0] at (0, 0)(input){\includegraphics[scale=\figscale]{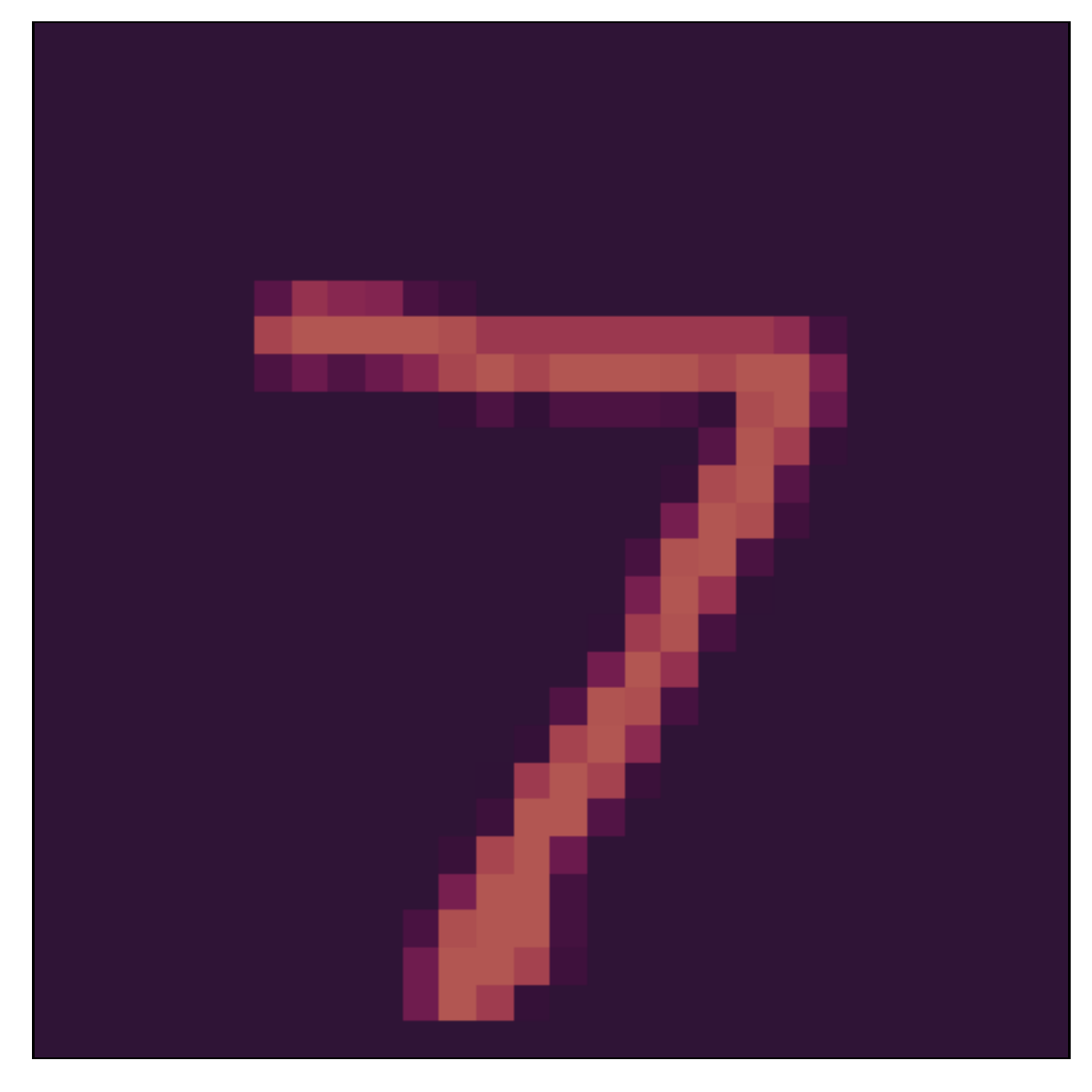}};
	\node[opacity=0, draw, right=0.5cm of input, circle] (loss){$\mathcal{L}$};
	\node[opacity=0, right=0.5cm of loss] (reconstructed){\includegraphics[scale=\figscale]{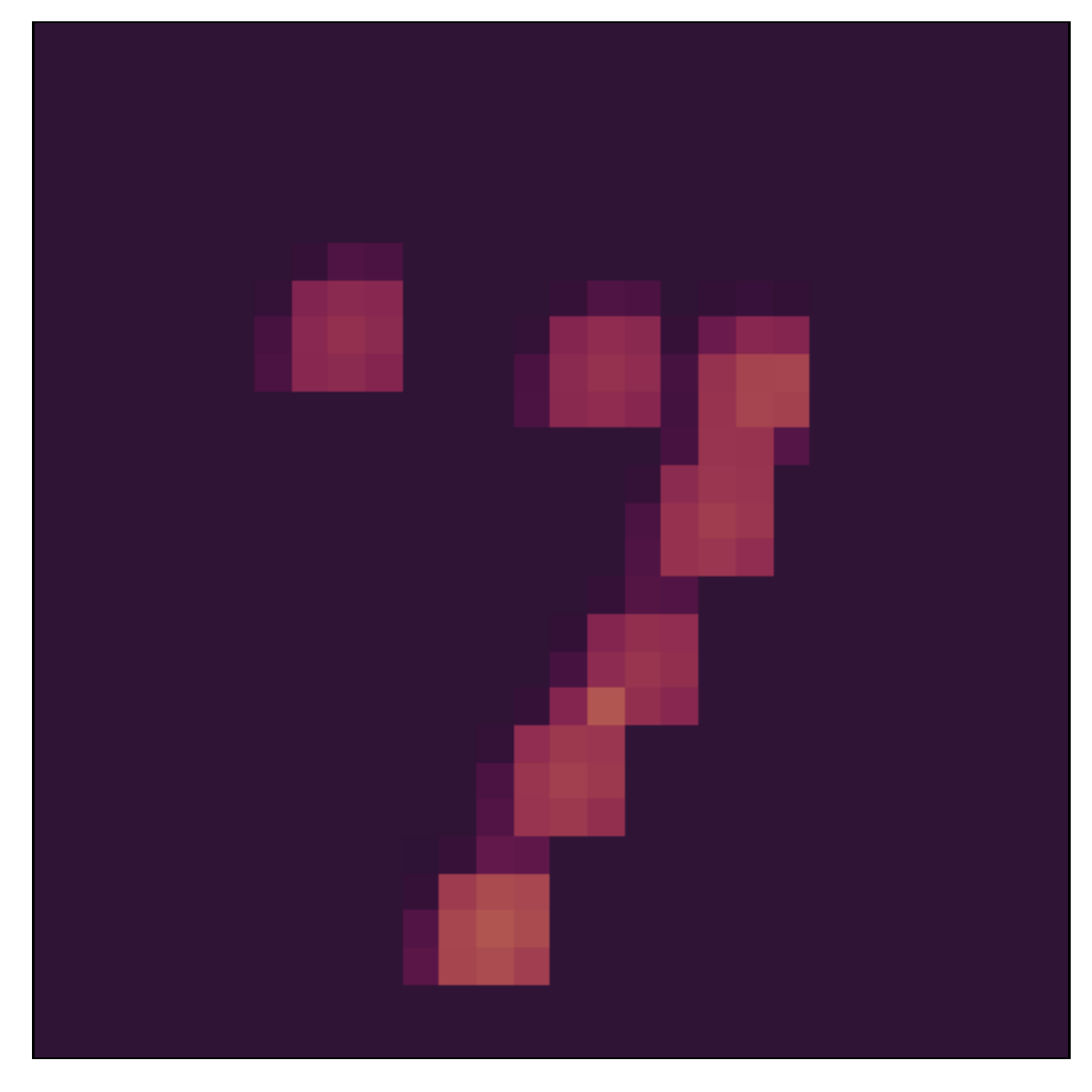}};
	\node[opacity=0, draw, below=0.5cm of reconstructed, circle] (plus){$+$};
	\node[opacity=0.8, below=0.5cm of plus] (reconstruction){\includegraphics[scale=\figscale]{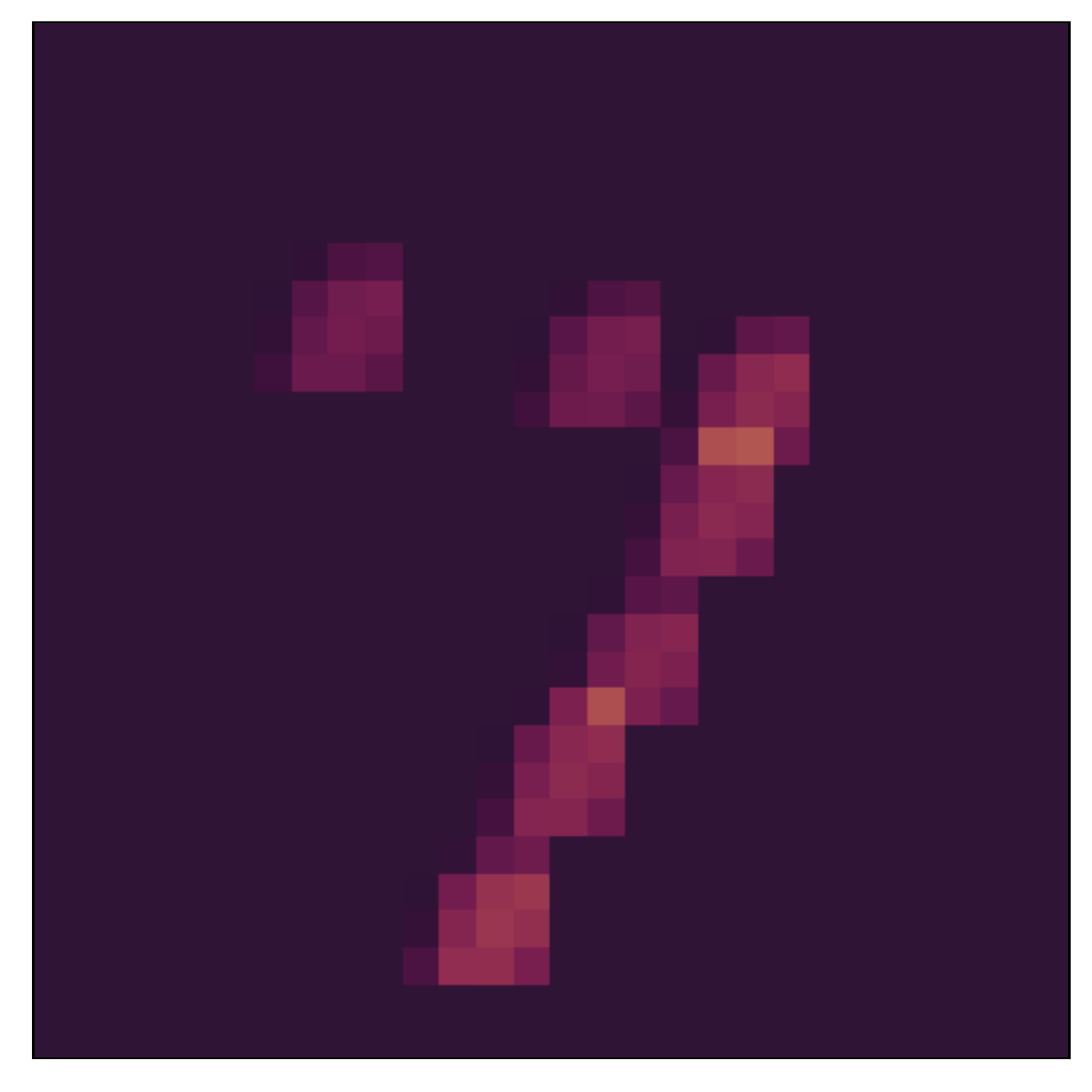}};
	\node[opacity=0, draw, below=0.5cm of reconstruction, circle] (conv2){$\ast$};
	\node[opacity=0.8, below=0.5cm of conv2] (extrema){\includegraphics[scale=\figscale]{"images-2d/MNIST-extrema-2d-2-activations-0"}};
	\node[opacity=0, draw, left=0.55cm of extrema, circle] (phi){$\phi$};
	\node[opacity=0.8, left=0.55cm of phi] (similarity){\includegraphics[scale=\figscale]{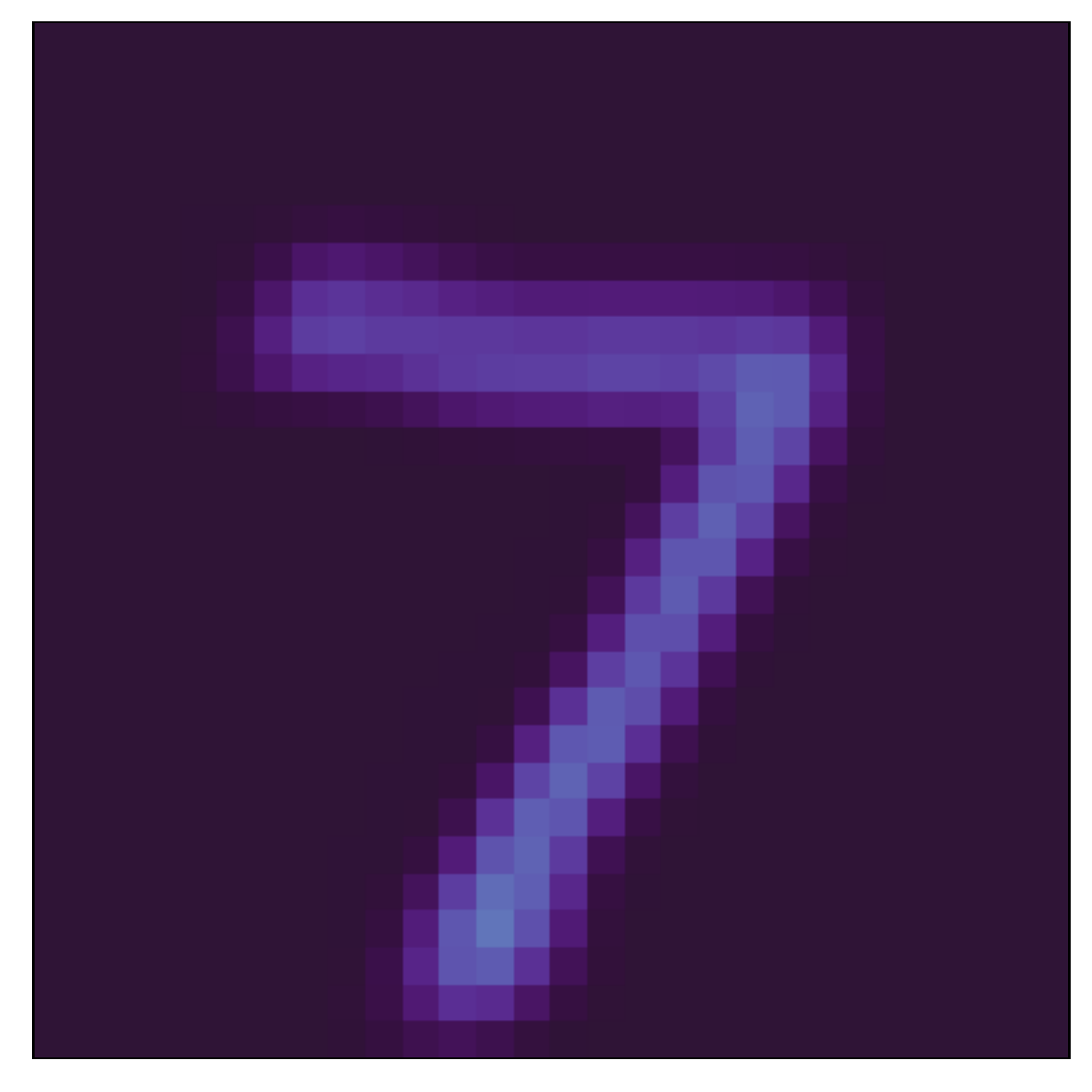}};
	\node[opacity=0.8, left=1.2cm of conv2] (kernel){\includegraphics[scale=0.1]{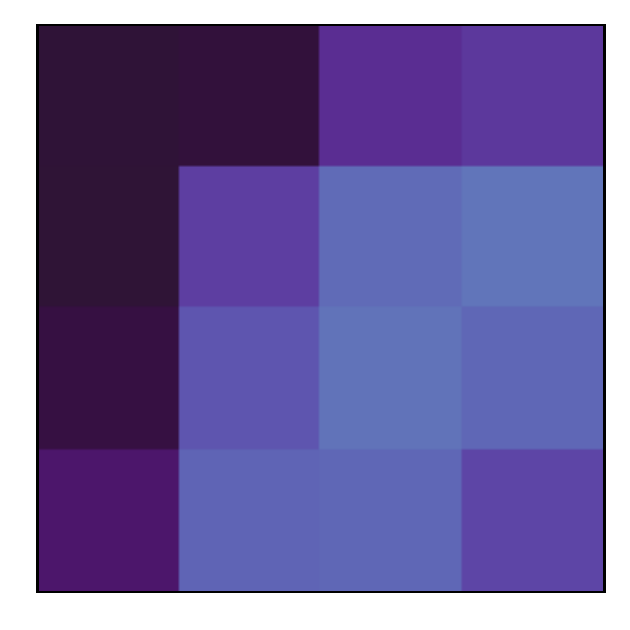}};
	\node[opacity=0, draw, above=0.5cm of similarity, circle] (conv1){$\ast$};
\end{scope}
\begin{scope}[tdplot_main_coords, canvas is yz plane at x=0,xscale=-1, transform shape]
	\node[] at (0, 0)(input){\includegraphics[scale=\figscale]{"images-2d/MNIST-extrema-2d-2-signal"}};
	\node[draw, right=0.5cm of input, circle] (loss){$\mathcal{L}$};
	\node[right=0.5cm of loss] (reconstructed){\includegraphics[scale=\figscale]{"images-2d/MNIST-extrema-2d-2-reconstructed"}};
	\node[draw, below=0.5cm of reconstructed, circle] (plus){$+$};
	\node[opacity=0, below=0.5cm of plus] (reconstruction){\includegraphics[scale=\figscale]{"images-2d/MNIST-extrema-2d-2-reconstruction-0"}};
	\node[draw, below=0.5cm of reconstruction, circle] (conv2){$\ast$};
	\node[opacity=0, below=0.5cm of conv2] (extrema){\includegraphics[scale=\figscale]{"images-2d/MNIST-extrema-2d-2-activations-0"}};
	\node[draw, left=0.55cm of extrema, circle, inner sep=2pt] (phi){$\phi$};
	\node[opacity=0, left=0.55cm of phi] (similarity){\includegraphics[scale=\figscale]{"images-2d/MNIST-extrema-2d-2-similarity-0"}};
	\node[opacity=0, left=1.2cm of conv2] (kernel){\includegraphics[scale=0.1]{"images-2d/MNIST-extrema-2d-2-kernel-0"}};
	\node[draw, above=0.5cm of similarity, circle] (conv1){$\ast$};
	\draw[->](input) -- node{} (conv1);
	\draw[->](conv1) -- node{} (similarity);
	\draw[->](similarity) -- node{} (phi);
	\draw[->](phi) -- node{} (extrema);
	\draw[->](extrema) -- node{} (conv2);
	\draw[->](conv2) -- node{} (reconstruction);
	\draw[->](reconstruction) -- node{} (plus);
	\draw[->](plus) -- node{} (reconstructed);
	\draw[->](kernel) -- node{} (conv1);
	\draw[->](kernel) -- node{} (conv2);
	\draw[->](input) -- node{} (loss);
	\draw[->](reconstructed) -- node{} (loss);
\end{scope}
\begin{scope}[tdplot_main_coords, canvas is yz plane at x=0.5,xscale=-1, transform shape]
	\node[opacity=0] at (0, 0)(input){\includegraphics[scale=\figscale]{"images-2d/MNIST-extrema-2d-2-signal"}};
	\node[opacity=0, draw, right=0.5cm of input, circle] (loss){$\mathcal{L}$};
	\node[opacity=0, right=0.5cm of loss] (reconstructed){\includegraphics[scale=\figscale]{"images-2d/MNIST-extrema-2d-2-reconstructed"}};
	\node[opacity=0, draw, below=0.5cm of reconstructed, circle] (plus){$+$};
	\node[opacity=0.8, below=0.5cm of plus] (reconstruction){\includegraphics[scale=\figscale]{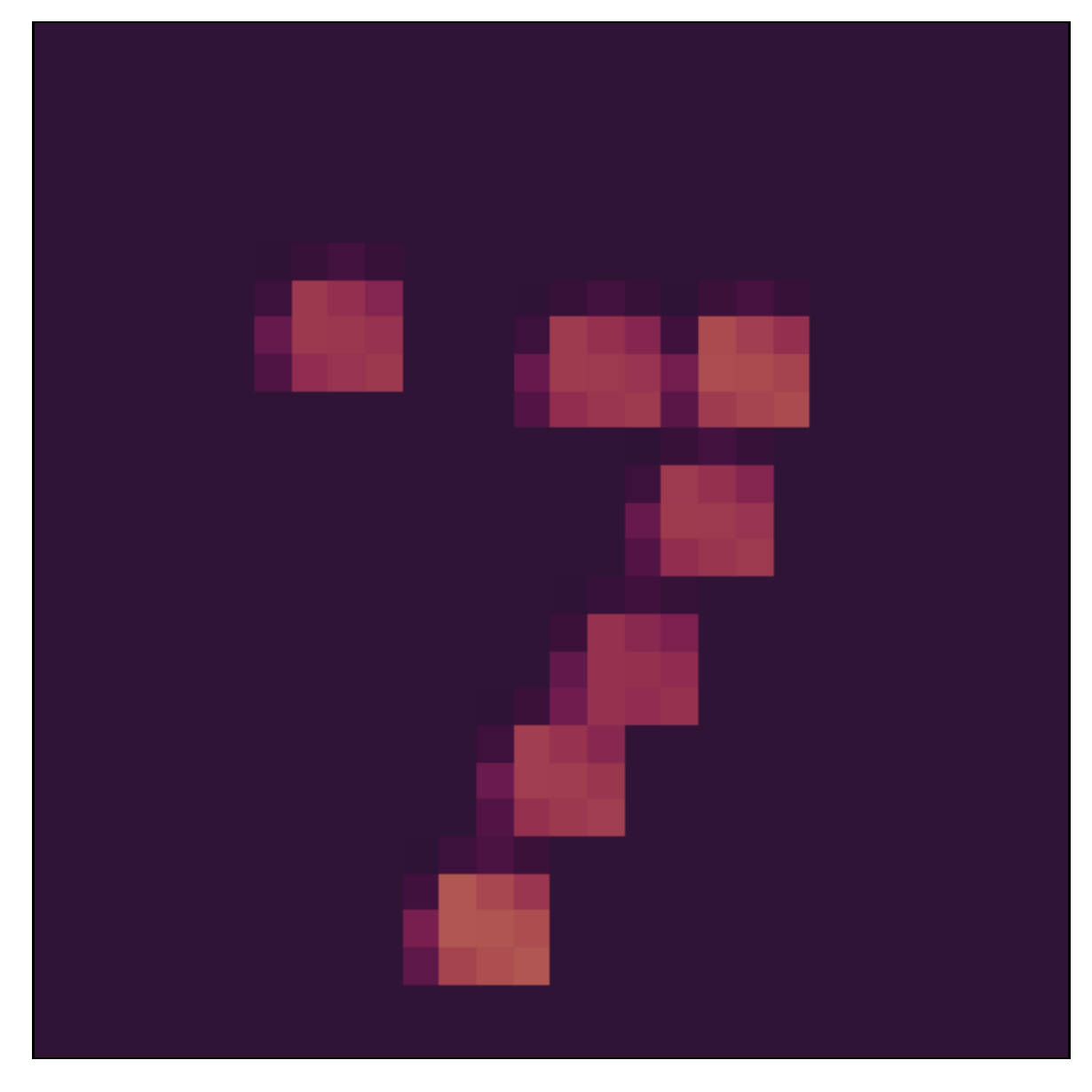}};
	\node[opacity=0, draw, below=0.5cm of reconstruction, circle] (conv2){$\ast$};
	\node[opacity=0.8, below=0.5cm of conv2] (extrema){\includegraphics[scale=\figscale]{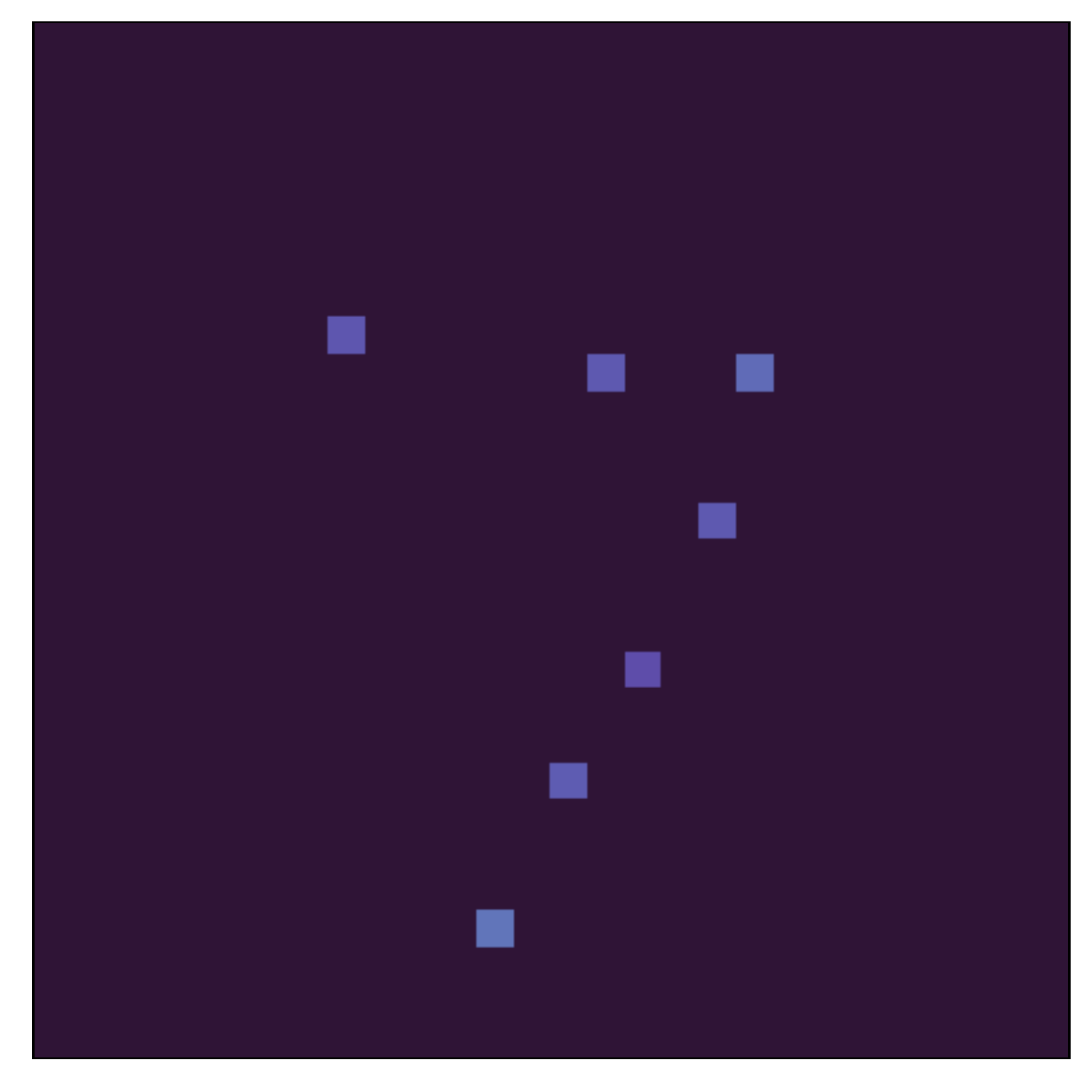}};
	\node[opacity=0, draw, left=0.55cm of extrema, circle] (phi){$\phi$};
	\node[opacity=0.8, left=0.55cm of phi] (similarity){\includegraphics[scale=\figscale]{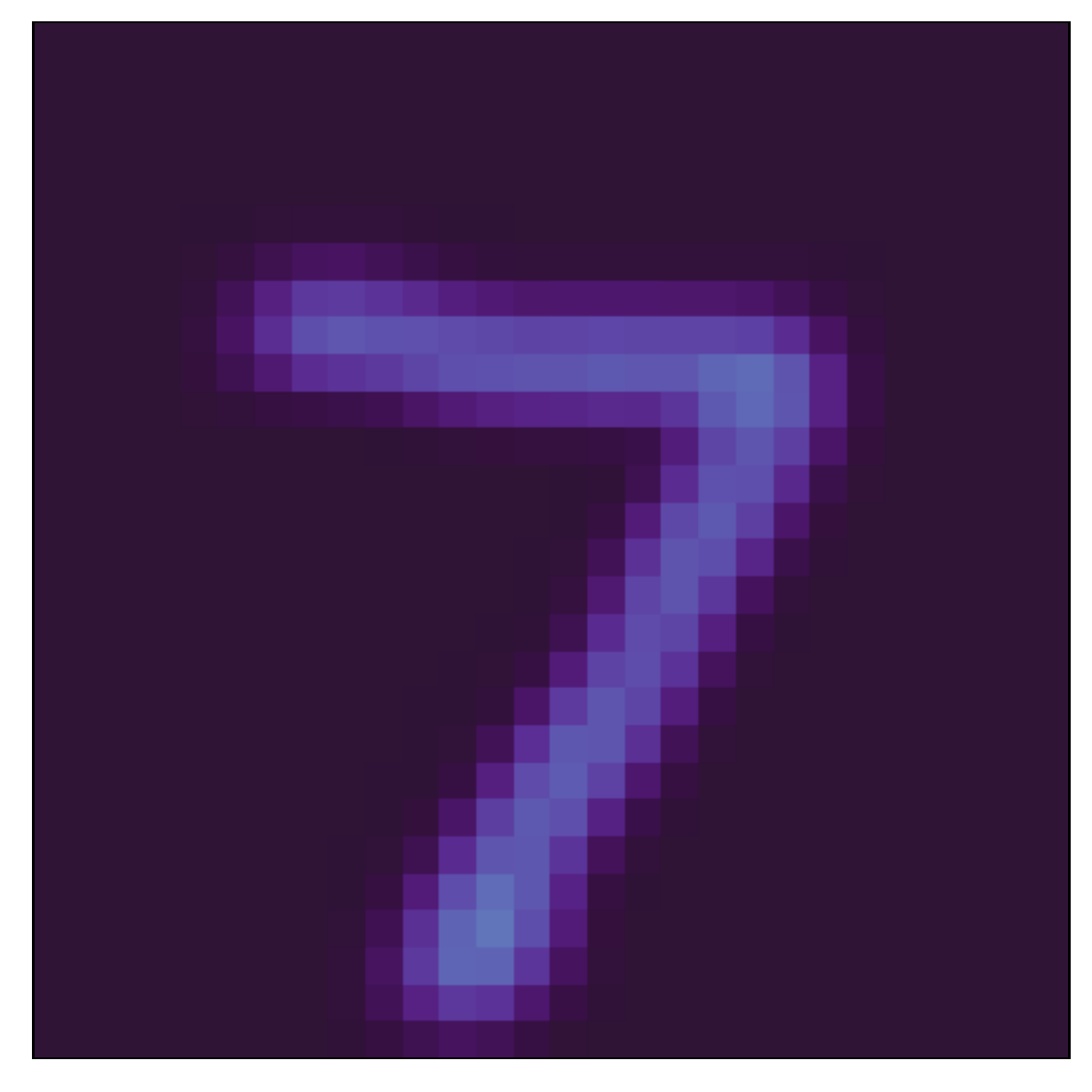}};
	\node[opacity=0.8, left=1.2cm of conv2] (kernel){\includegraphics[scale=0.1]{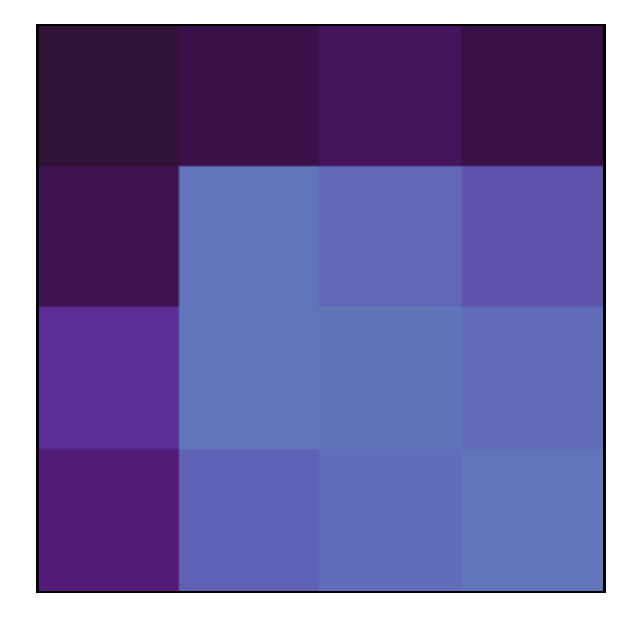}};
	\node[opacity=0, draw, above=0.5cm of similarity, circle] (conv1){$\ast$};
\end{scope}

%% file: chapter6-san-train.tex
\begin{algorithmic}[1]
	\renewcommand{\algorithmicrequire}{\textbf{Input:}}
	\renewcommand{\algorithmicensure}{\textbf{Output:}}
	\REQUIRE $\bm{x}$
	\ENSURE $\bm{\alpha}$, $\bm{w}$
	\\ \textit{Hyperparameters:  $m$, $\mu$, $\sigma$, $q$, $\phi$, $d$, $\lambda$, $epochs$, $batches$}
	\FOR {$i$ = 1 to $q$}
	\STATE $\bm{w}^{(i)} \sim \mathcal{N}(\mu, \sigma)$
	\ENDFOR
	\FOR {$e$ = 1 to $epochs$}
	\FOR {$b$ = 1 to $batches$}
	\STATE $\bm{x}^{(b)} \sim \bm{x}$
	\FOR {$i$ = 1 to $q$}
	\STATE $\bm{s}^{(i)} \leftarrow \bm{x}^{(b)} * \bm{w}^{(i)}$
	\STATE $\bm{\alpha}^{(i)} \leftarrow \phi(\bm{s}^{(i)}, d^{(i)})$
	\STATE $r^{(i)} \leftarrow \bm{\alpha}^{(i)} * \bm{w}^{(i)}$
	\ENDFOR
	\STATE $\hat{\bm{x}}^{(b)} \leftarrow \sum\limits_{i=1}^q r^{(i)}$
	\STATE $\mathcal{L} \leftarrow \frac{1}{n}\sum\limits_{t=1}^n \lvert\hat{\bm{x}}^{(b)}_t - \bm{x}^{(b)}_t \rvert$
	\STATE $\nabla\mathcal{L} = \left( \frac{\partial\mathcal{L}}{\partial\bm{w}^{(1)}},\ldots\frac{\partial\mathcal{L}}{\partial\bm{w}^{(q)}}\right)$
	\STATE $\Delta\bm{w}^{(i)} = -\lambda\frac{\partial\mathcal{L}}{\partial\bm{w}^{(i)}}$
	\ENDFOR
	\ENDFOR
	\RETURN $\bm{\alpha}$, $\bm{w}$
\end{algorithmic}

%% file: chapter6-crrl.tex
\includegraphics[width=0.24\textwidth]{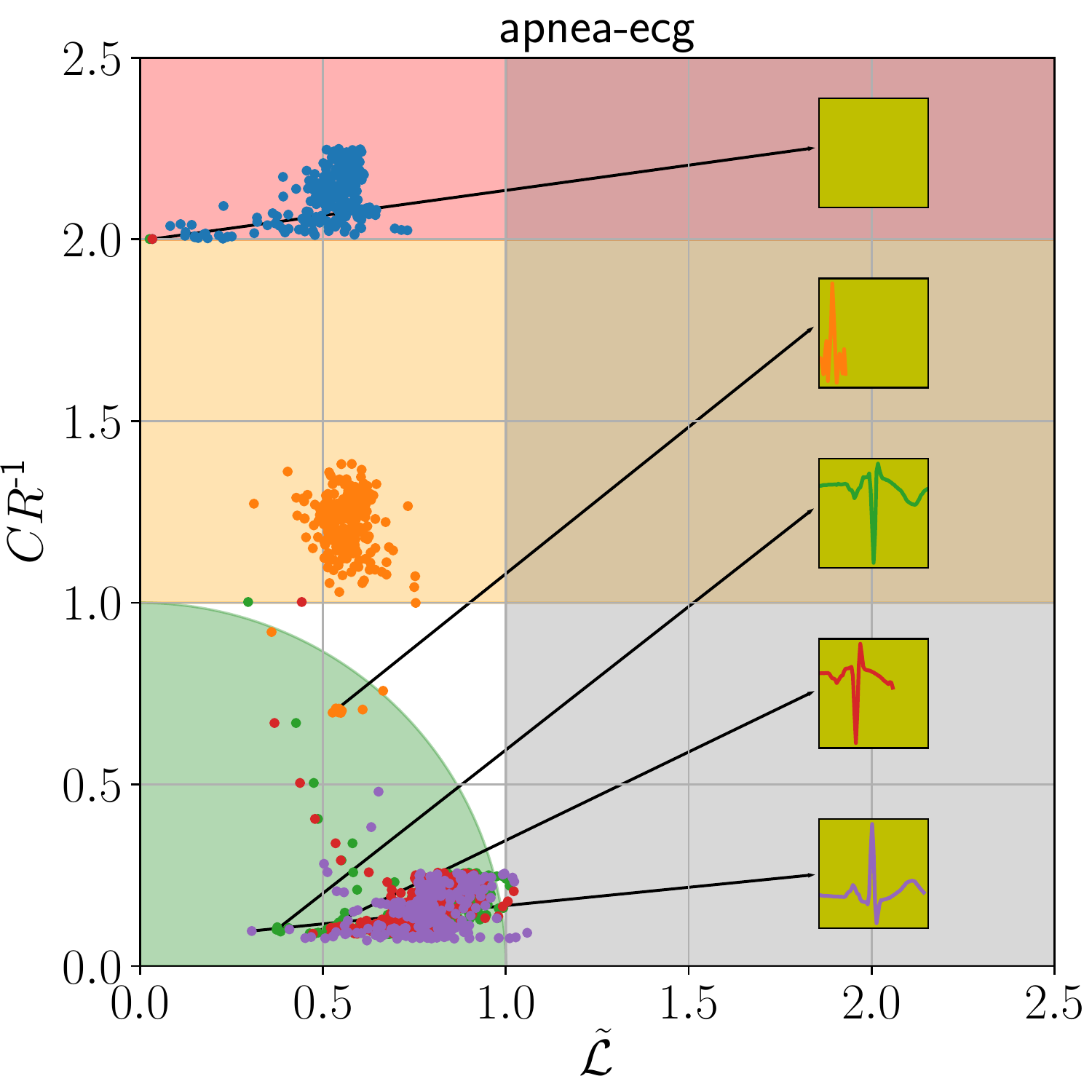}
\includegraphics[width=0.24\textwidth]{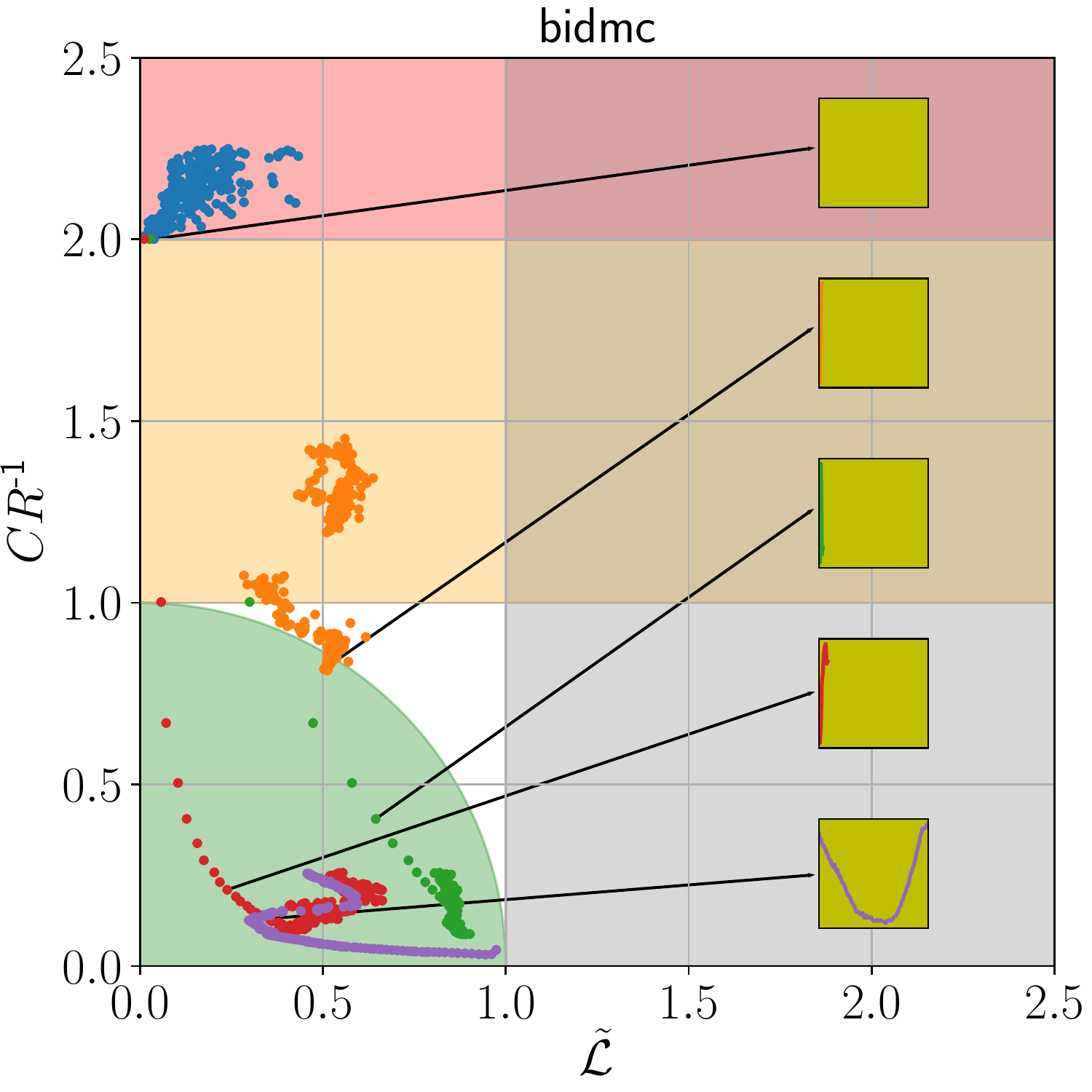}
\includegraphics[width=0.24\textwidth]{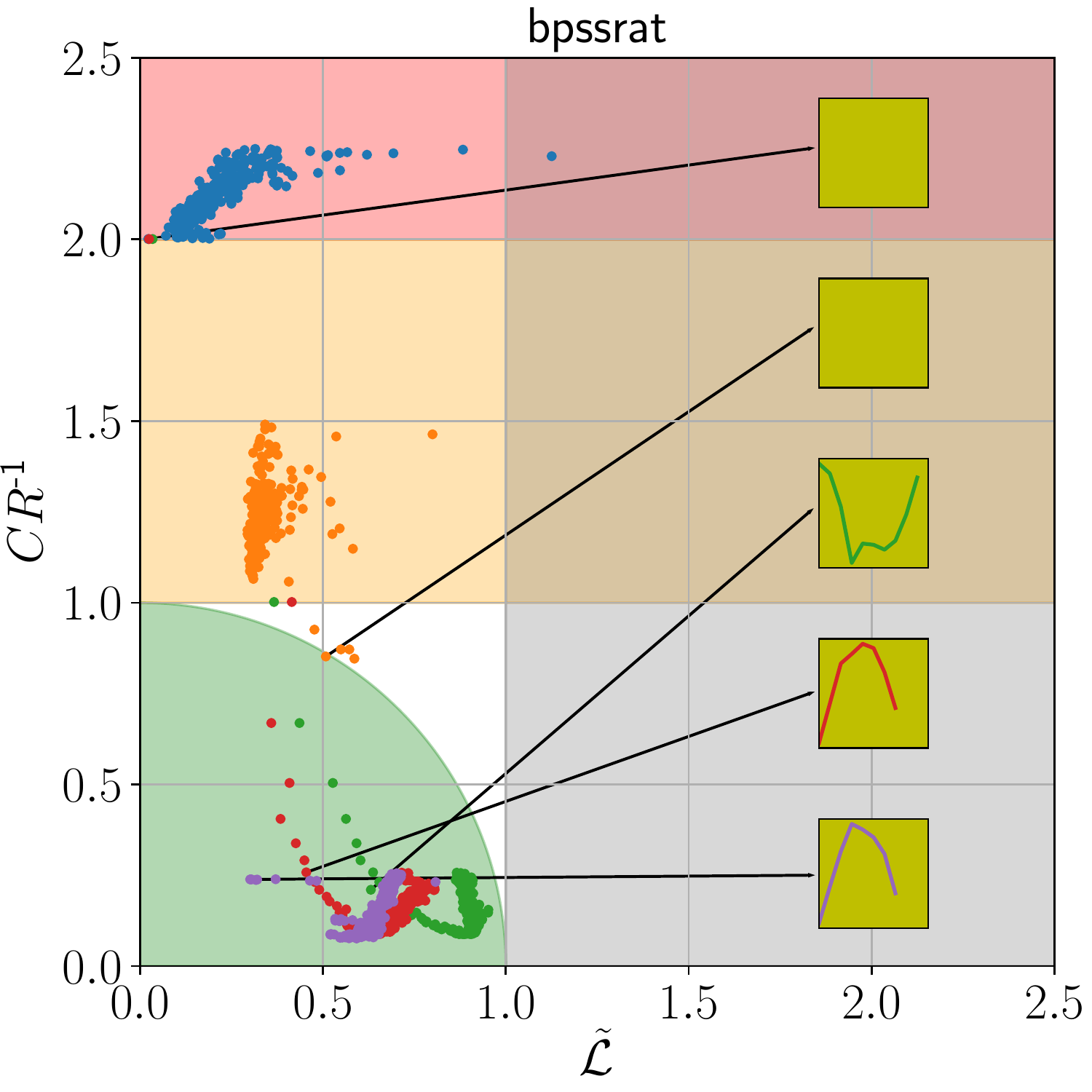}
\includegraphics[width=0.24\textwidth]{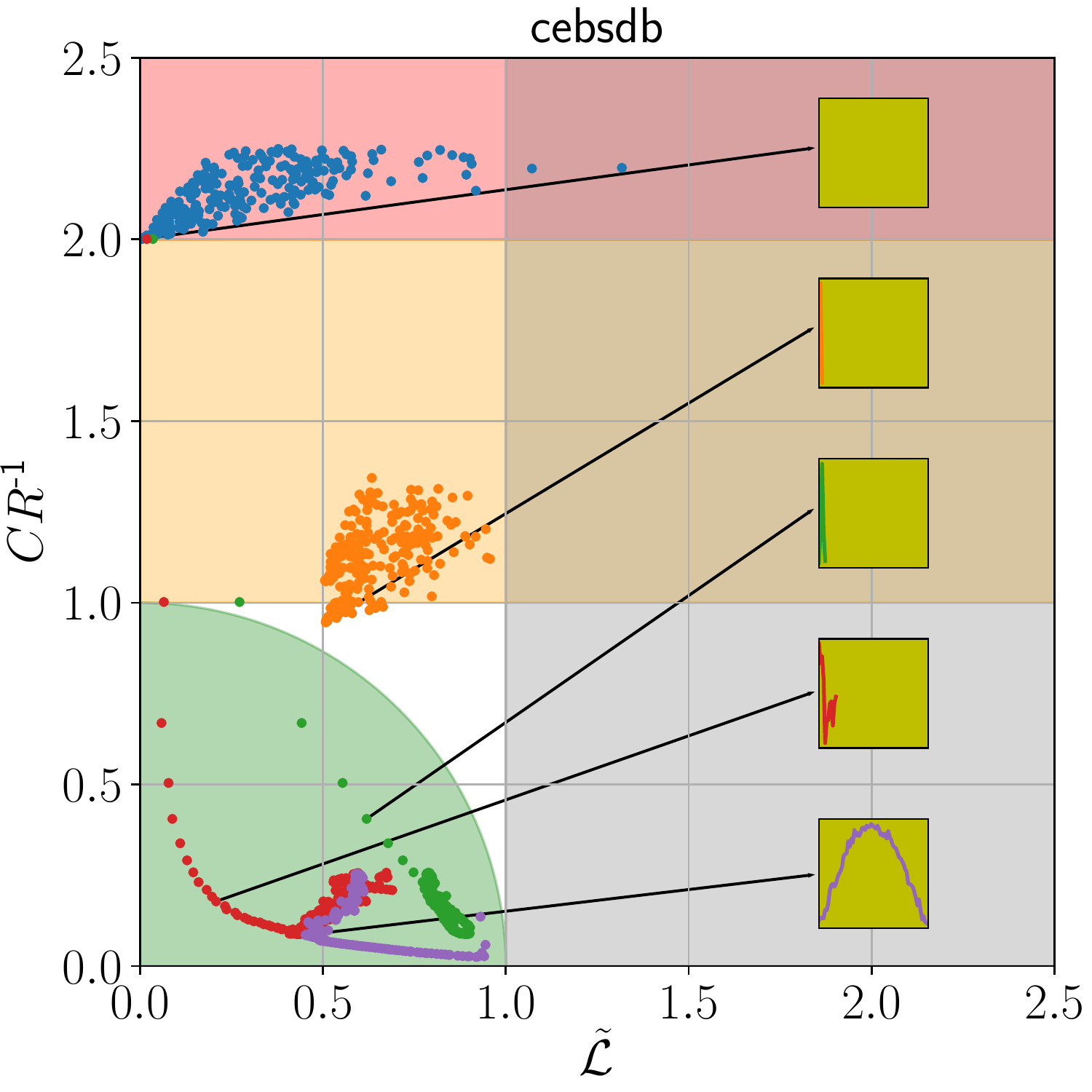}
\\
\includegraphics[width=0.24\textwidth]{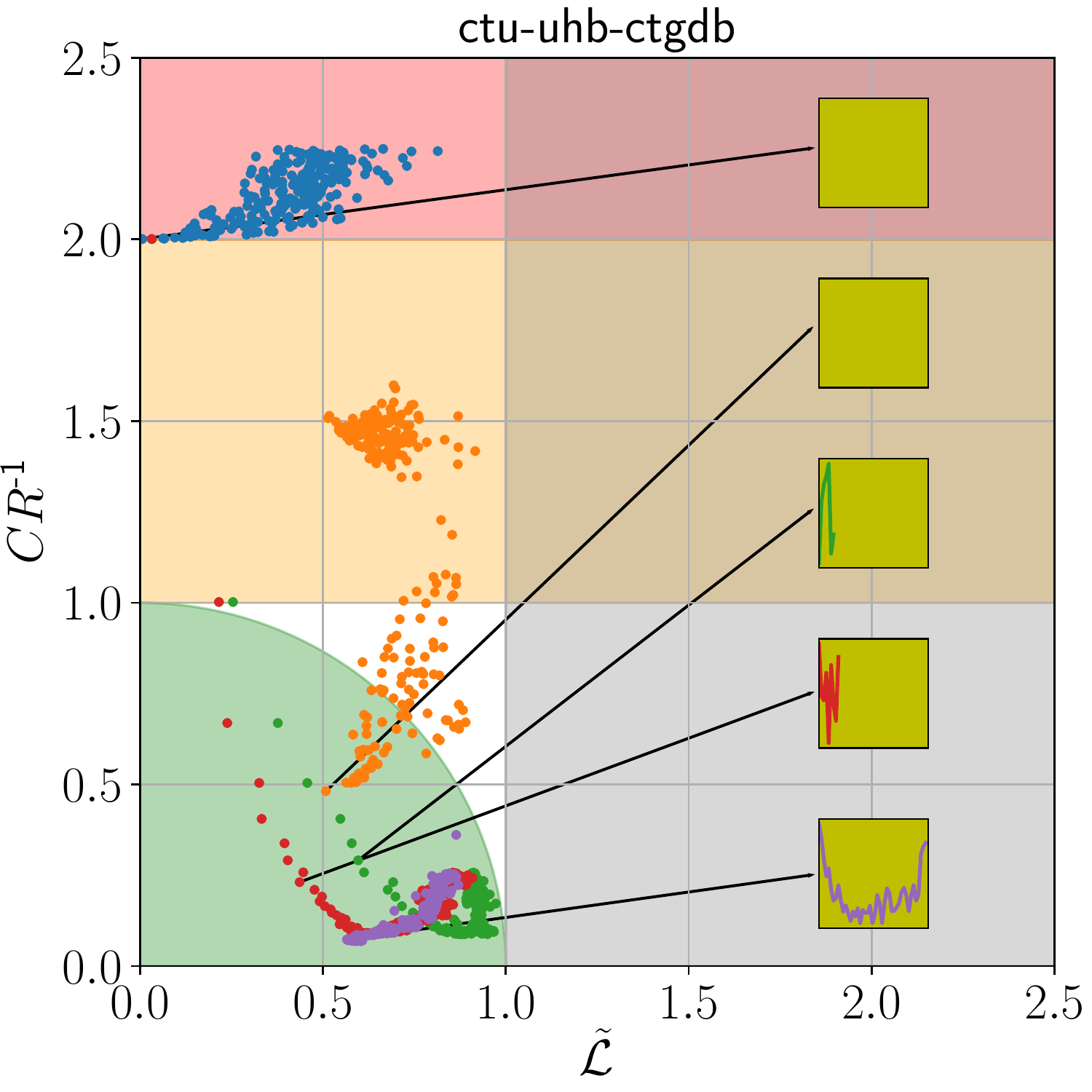}
\includegraphics[width=0.24\textwidth]{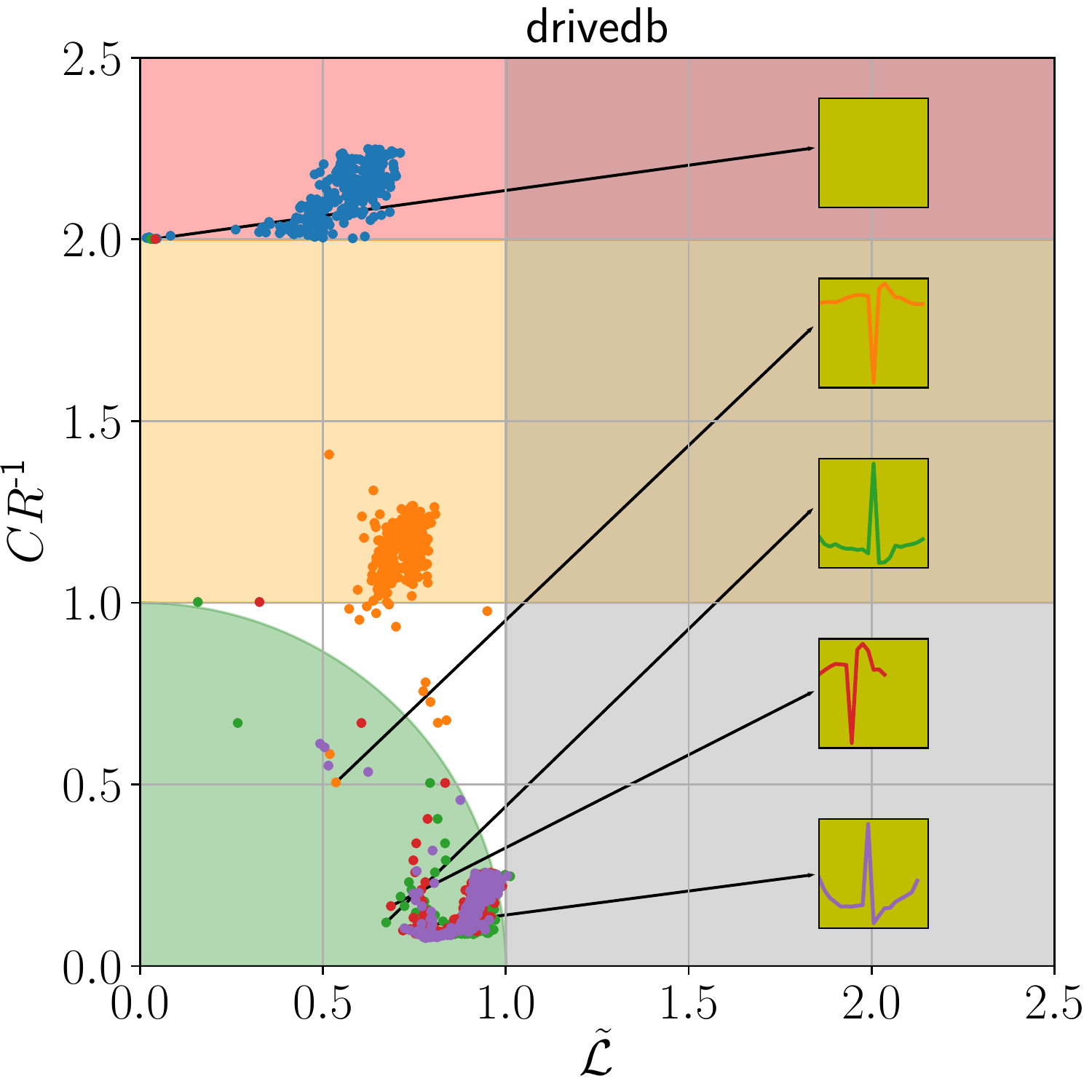}
\includegraphics[width=0.24\textwidth]{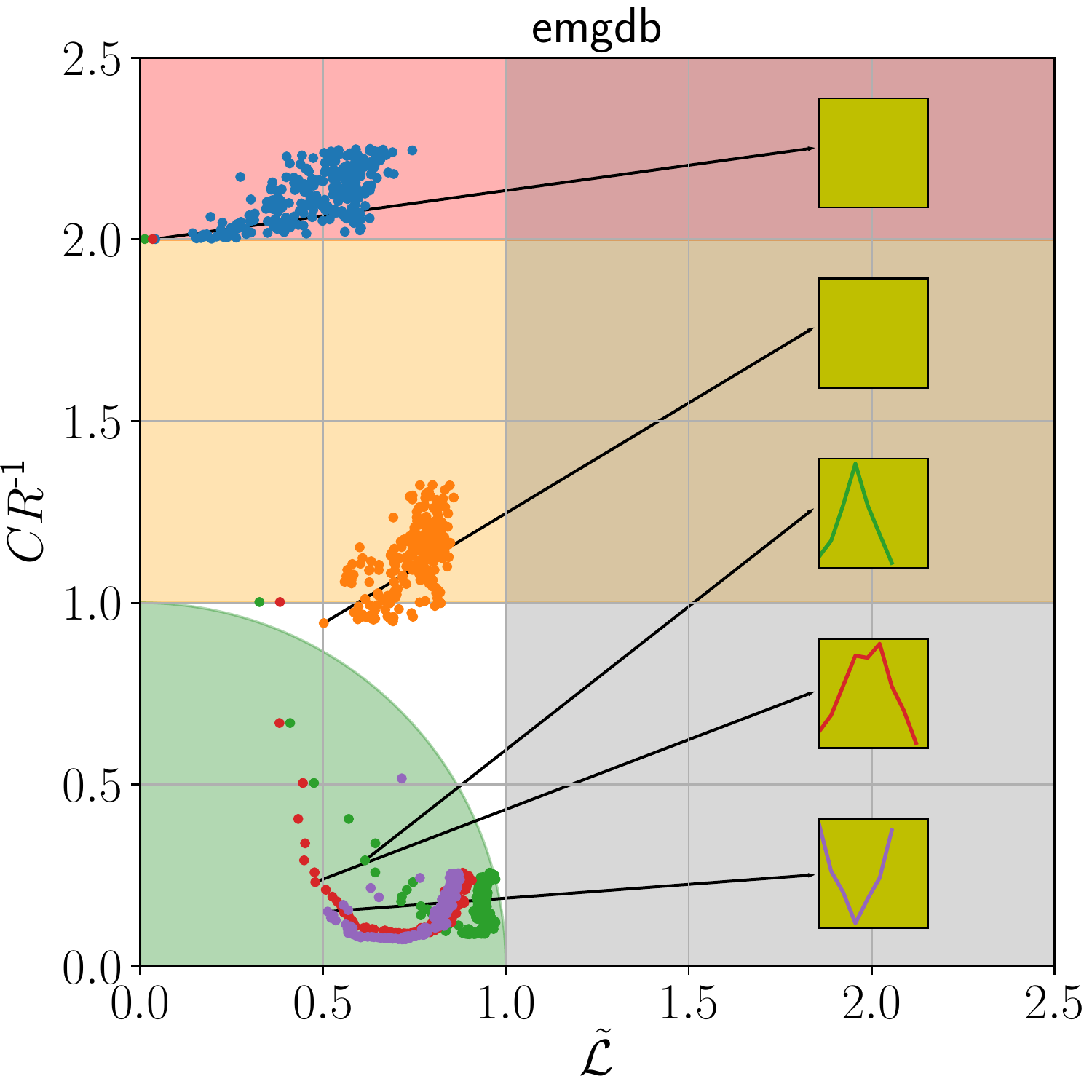}
\includegraphics[width=0.24\textwidth]{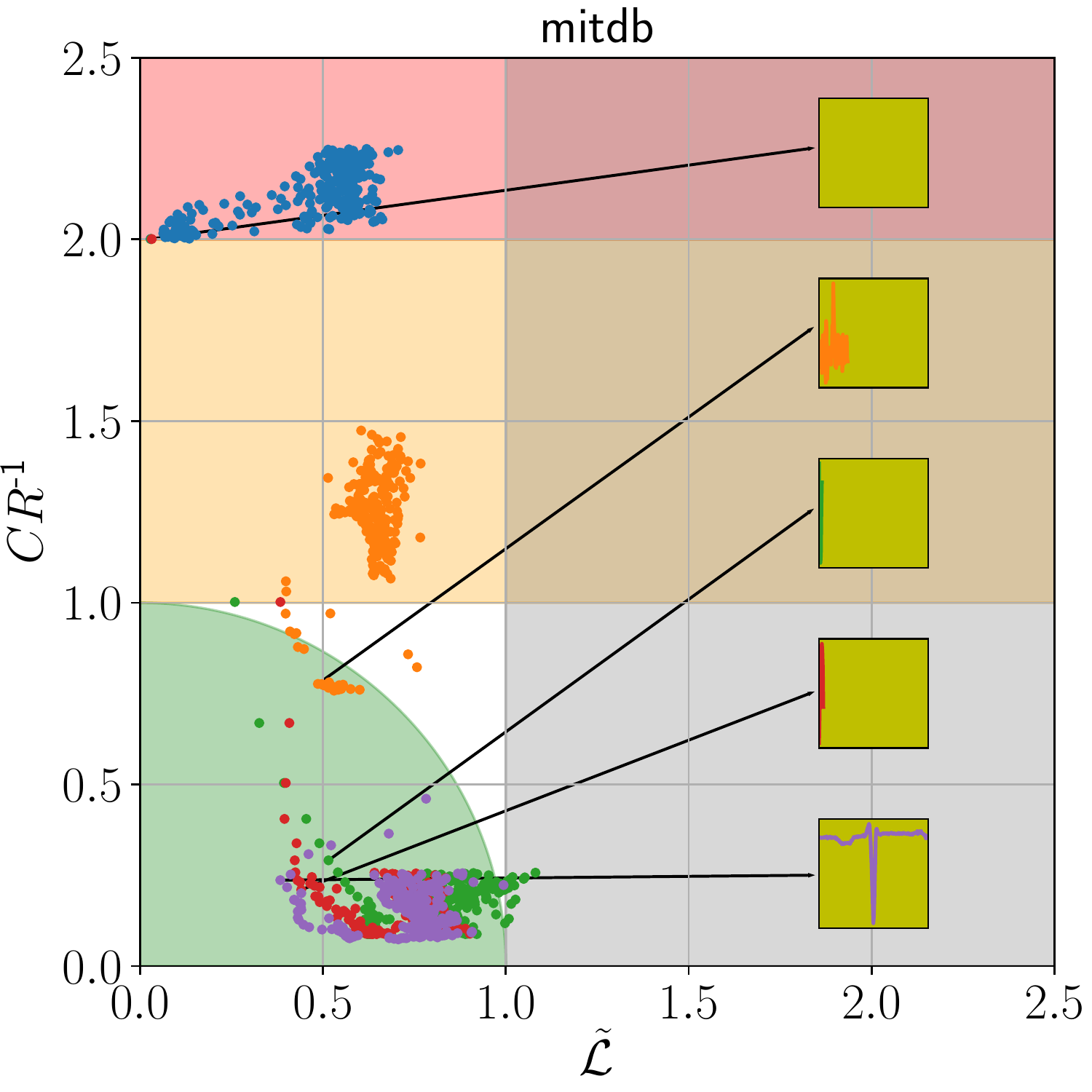}
\\
\includegraphics[width=0.24\textwidth]{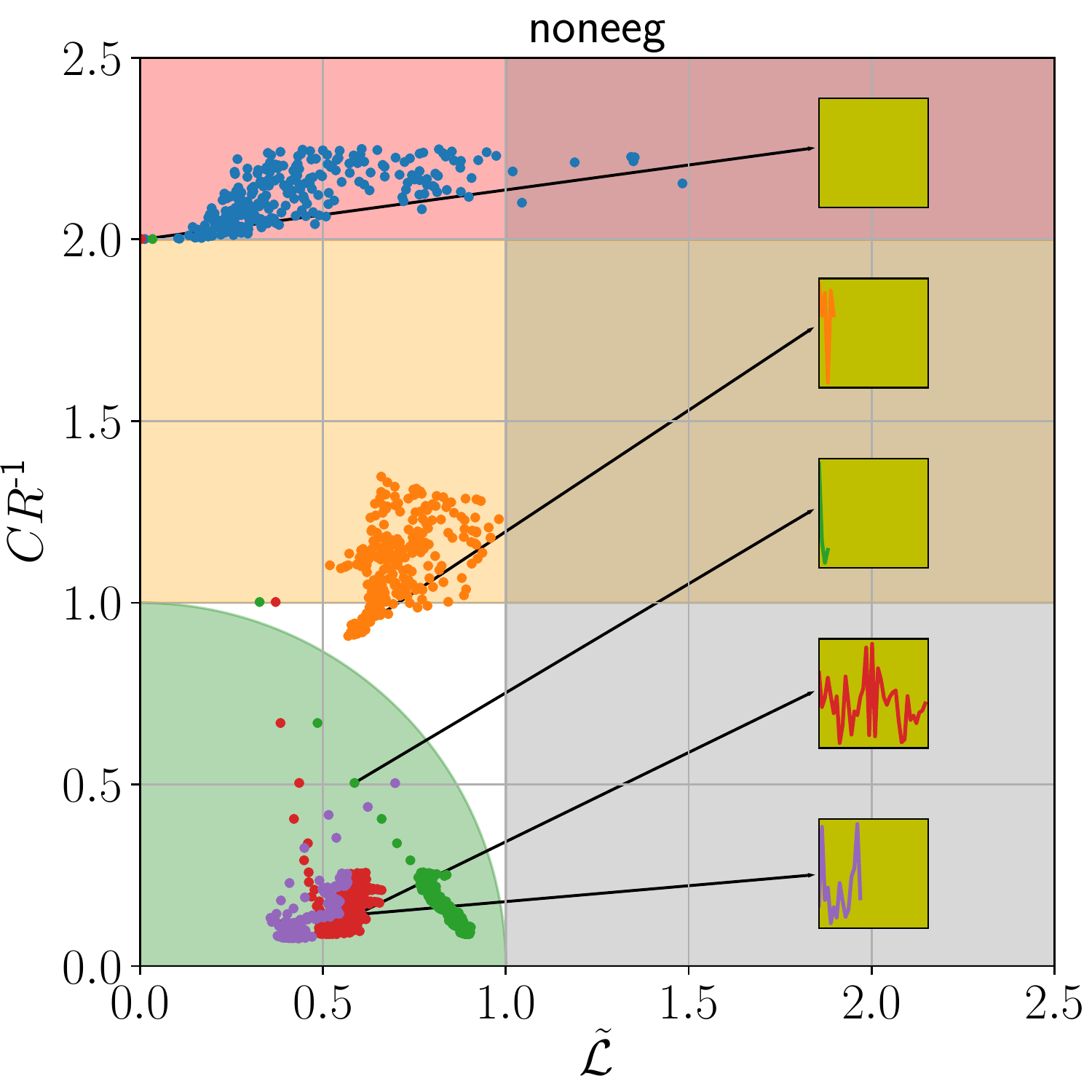}
\includegraphics[width=0.24\textwidth]{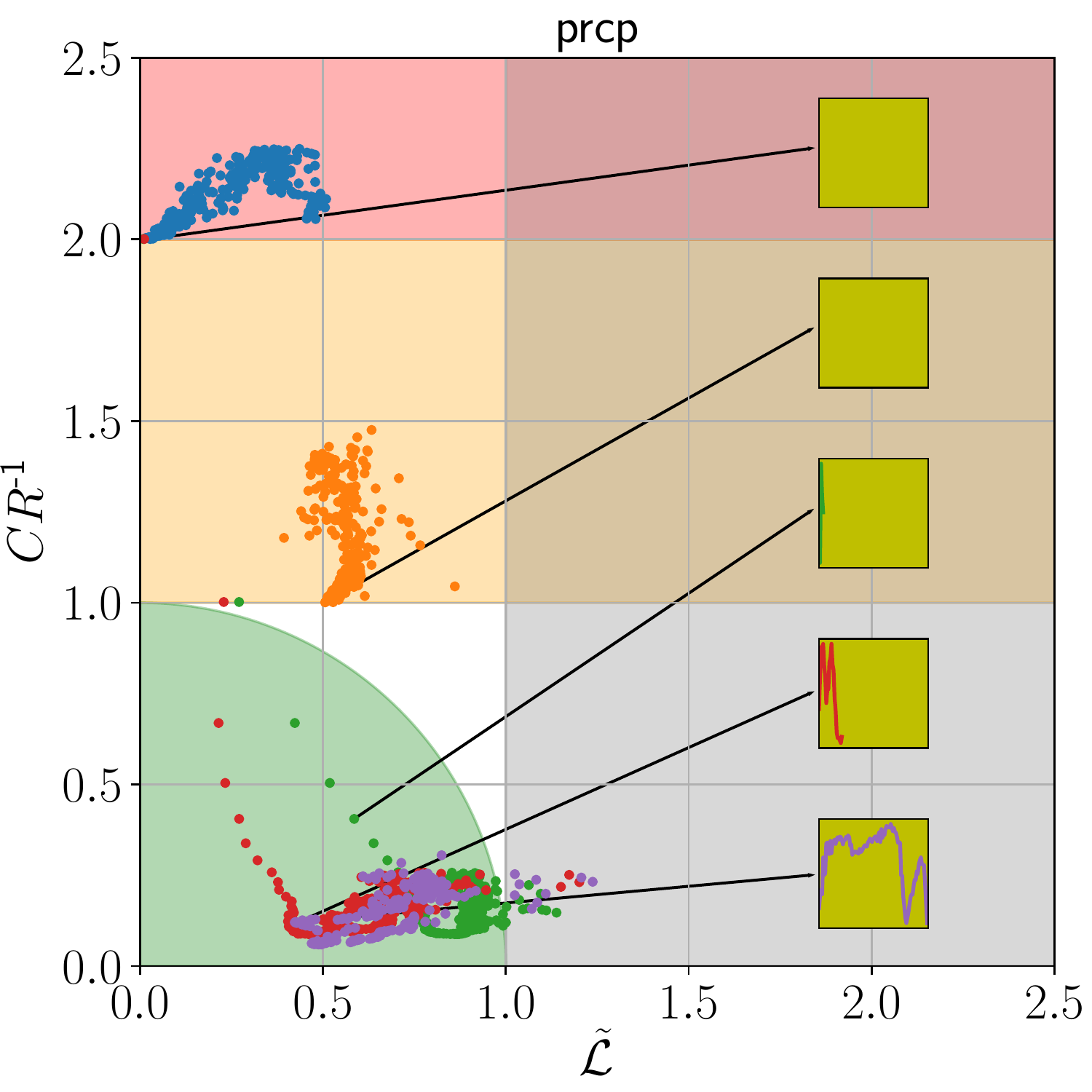}
\includegraphics[width=0.24\textwidth]{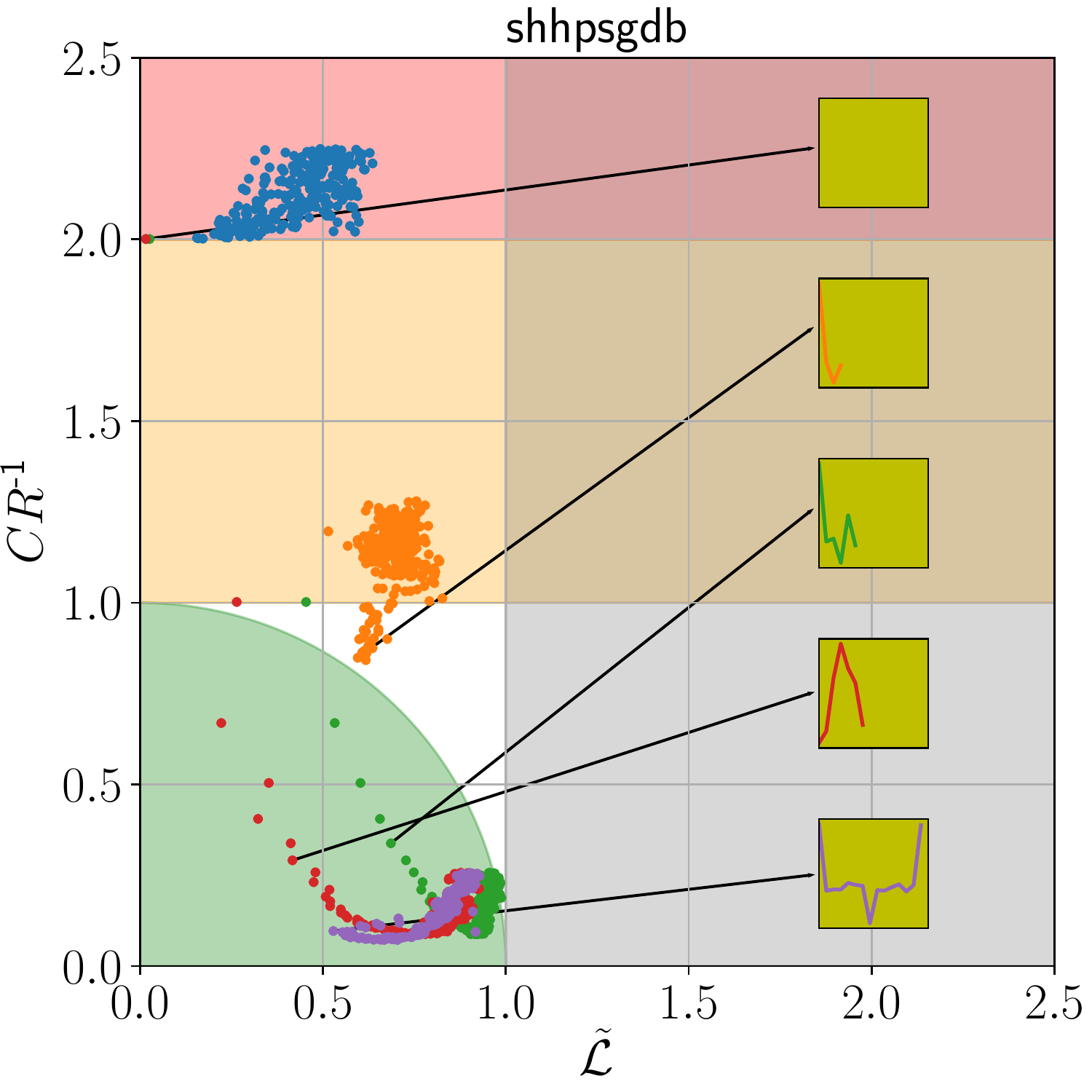}
\includegraphics[width=0.24\textwidth]{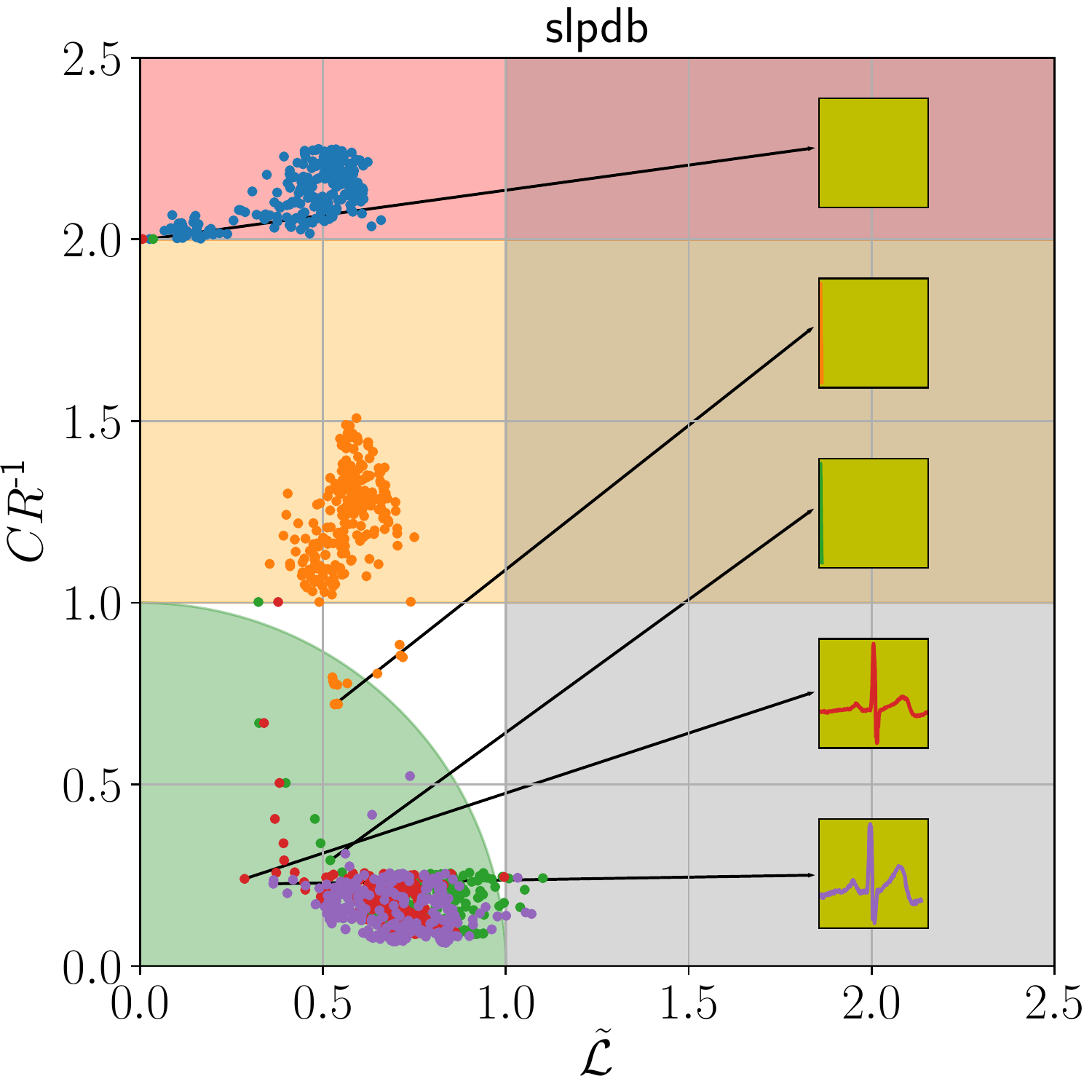}
\\
\includegraphics[width=0.24\textwidth]{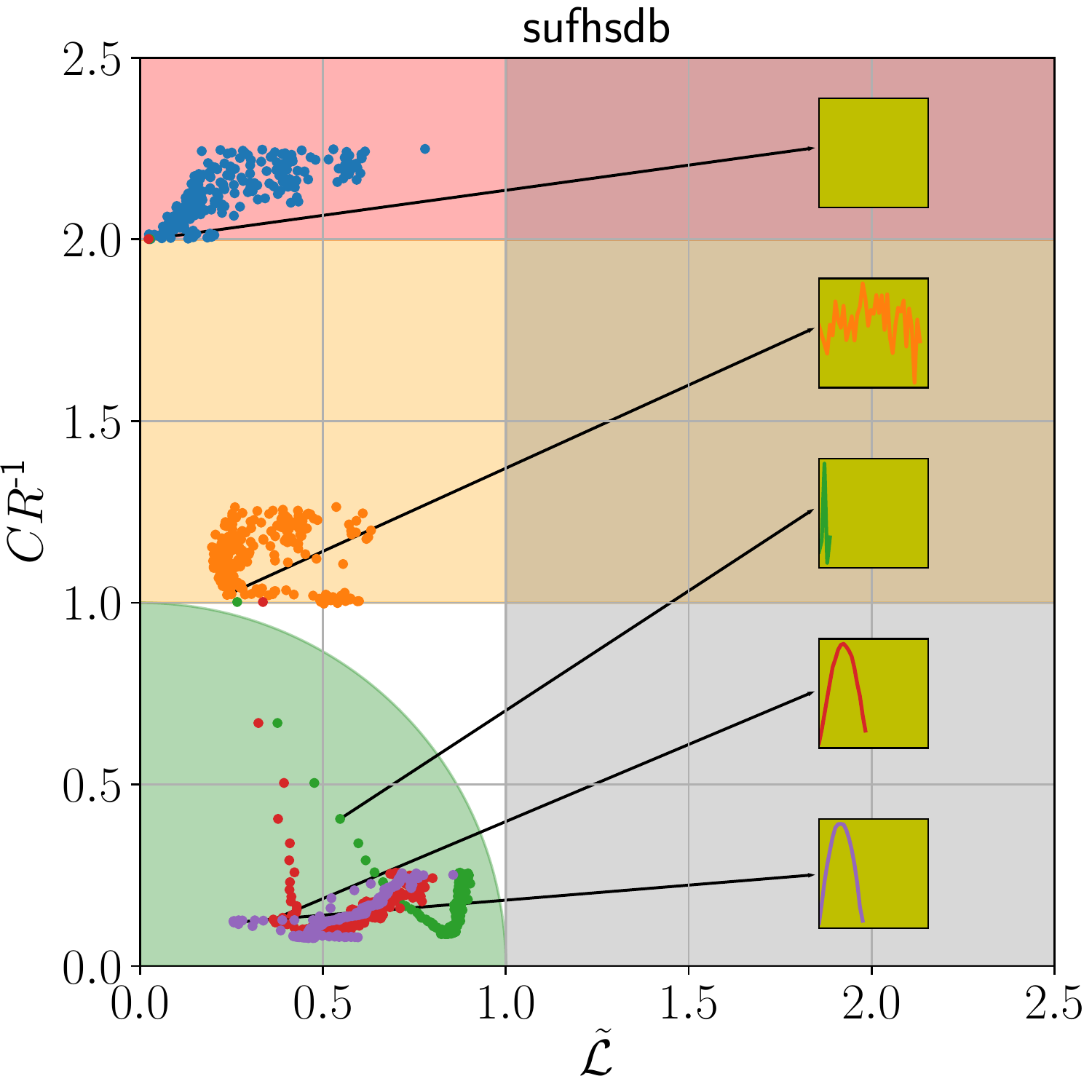}
\includegraphics[width=0.24\textwidth]{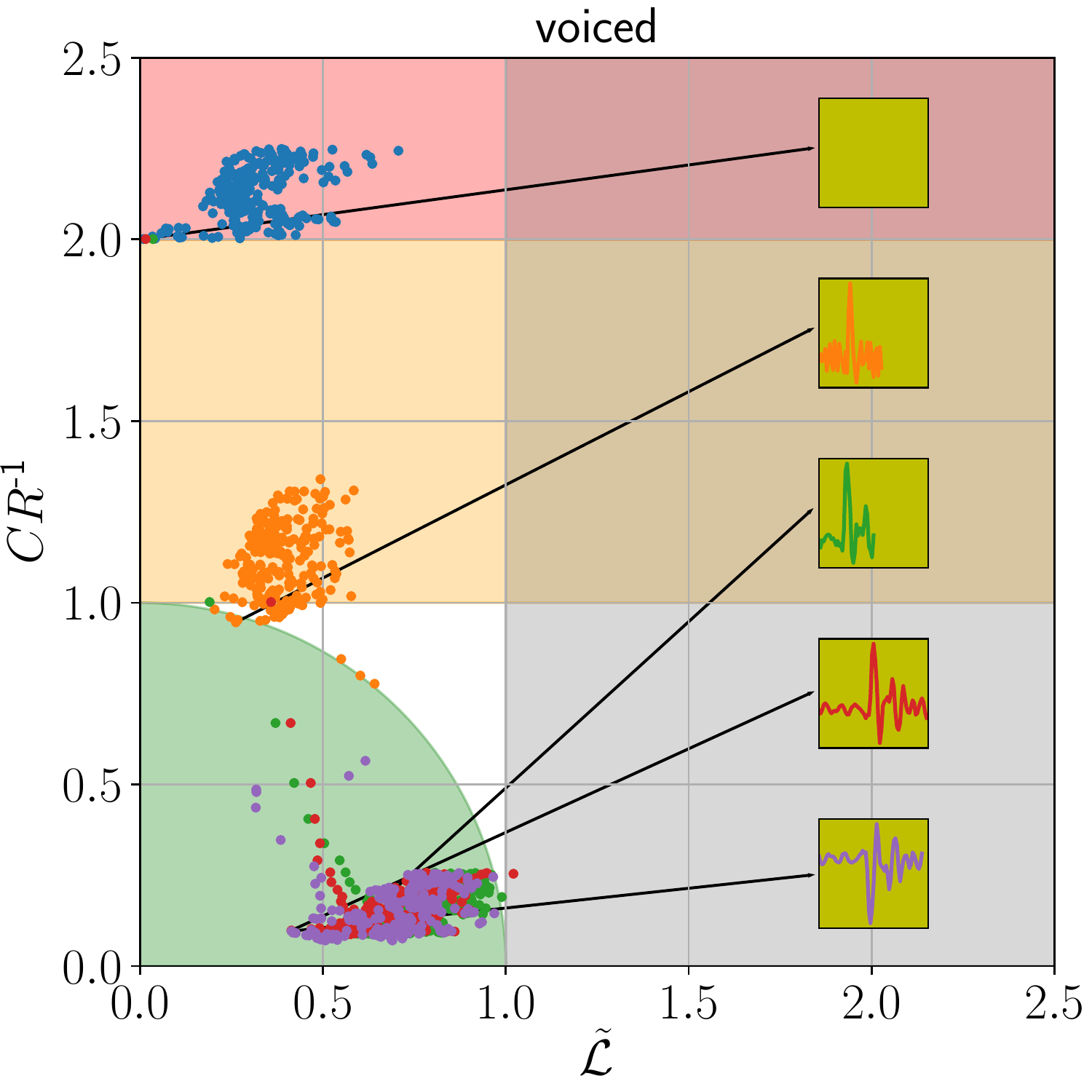}
\includegraphics[width=0.24\textwidth]{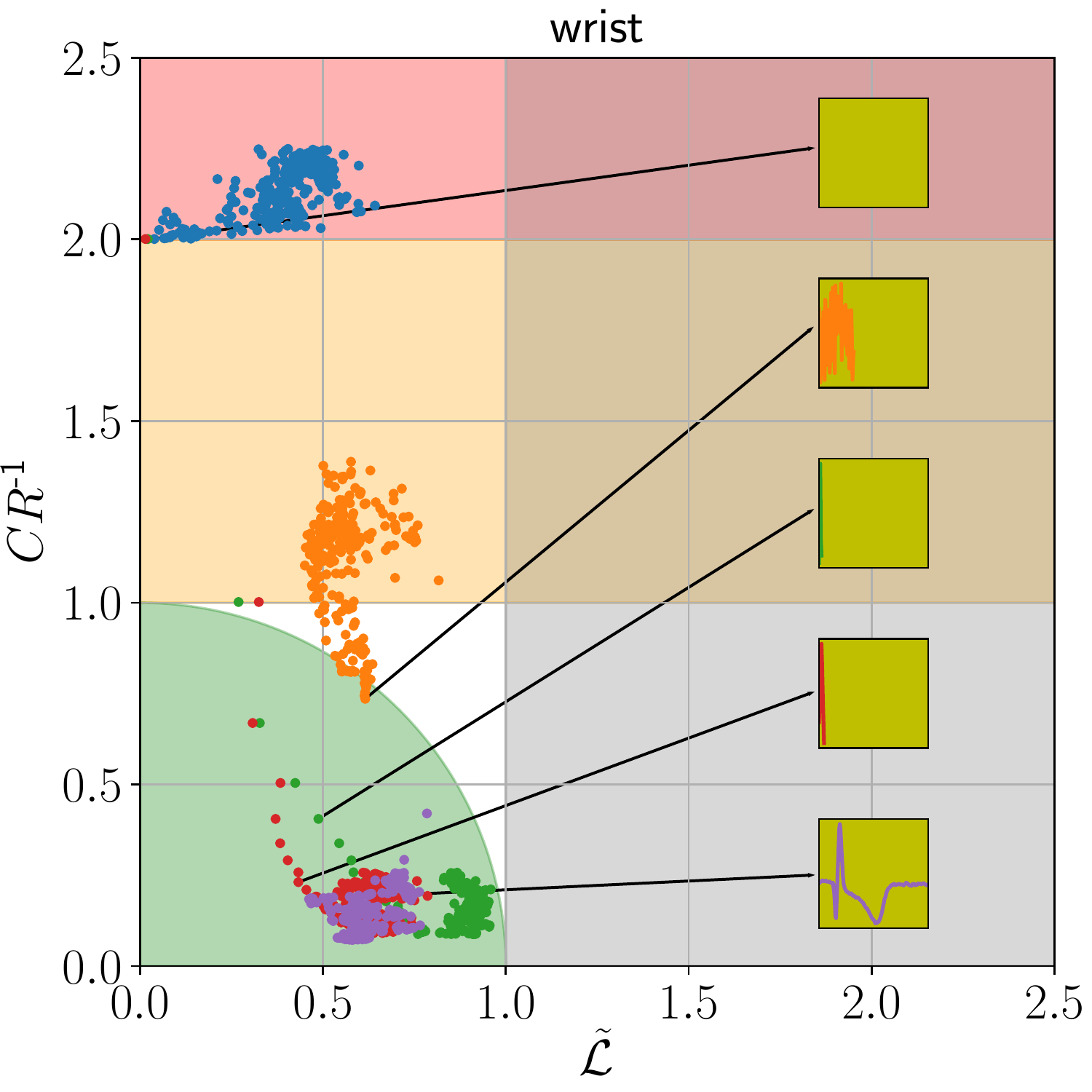}
\includegraphics[width=0.24\textwidth]{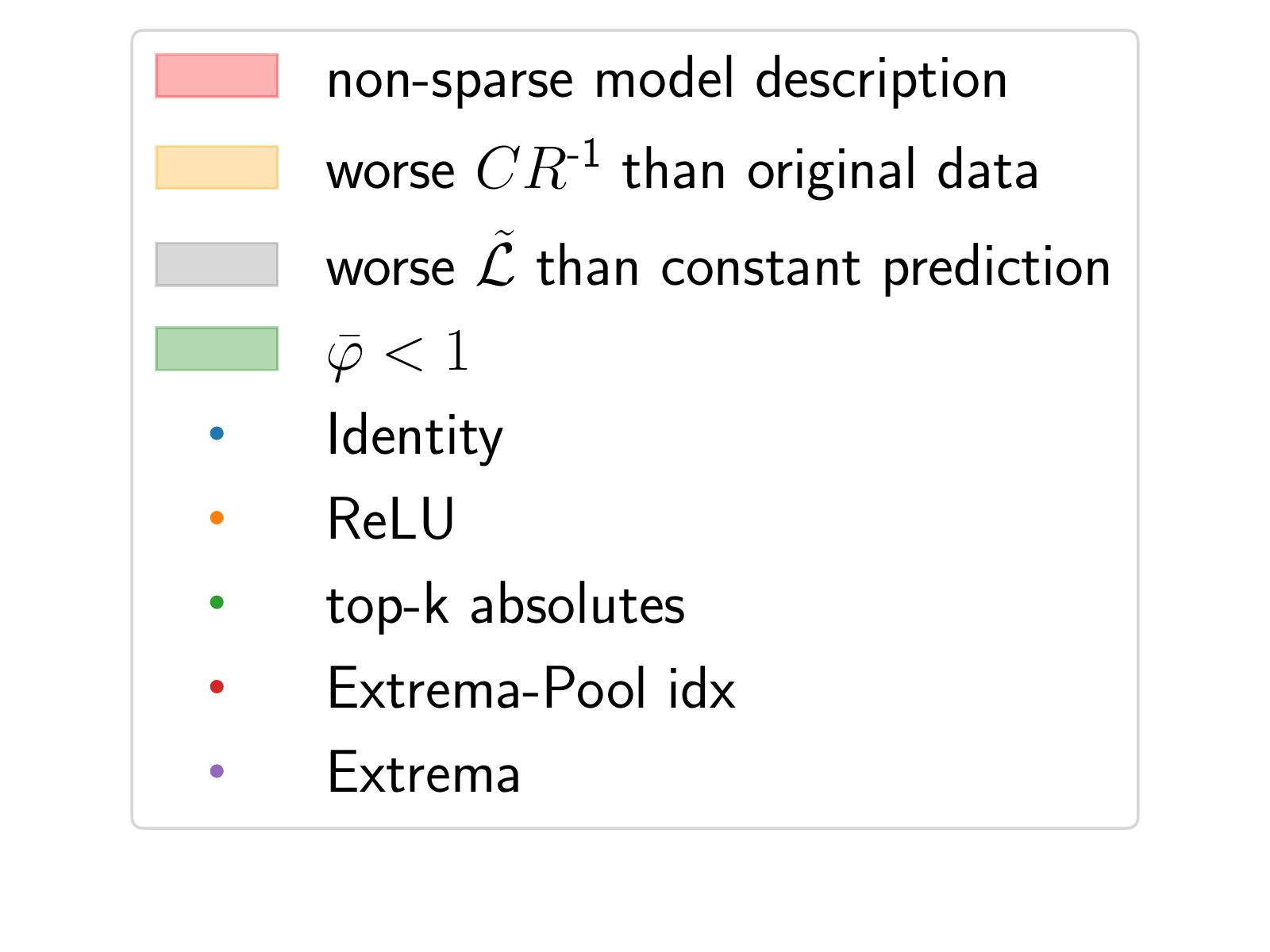}

%% file: chapter6-kernelvisualization.tex
\includegraphics[width=0.056\textwidth]{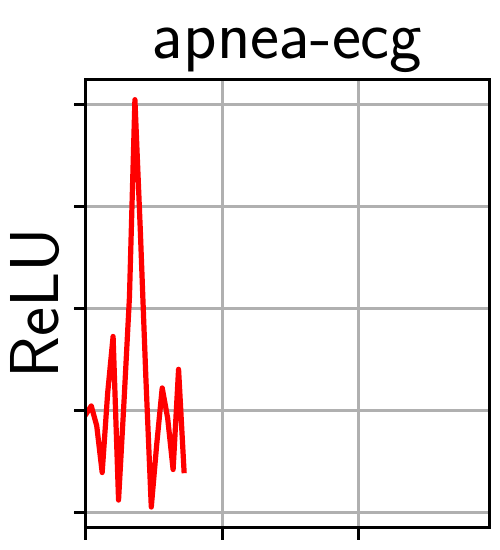}
\includegraphics[width=0.056\textwidth]{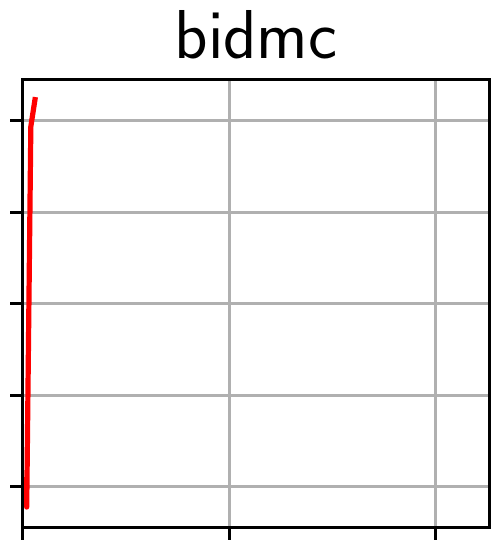}
\includegraphics[width=0.056\textwidth]{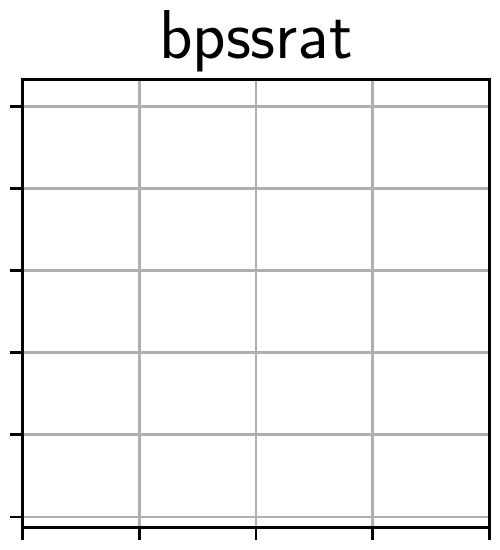}
\includegraphics[width=0.056\textwidth]{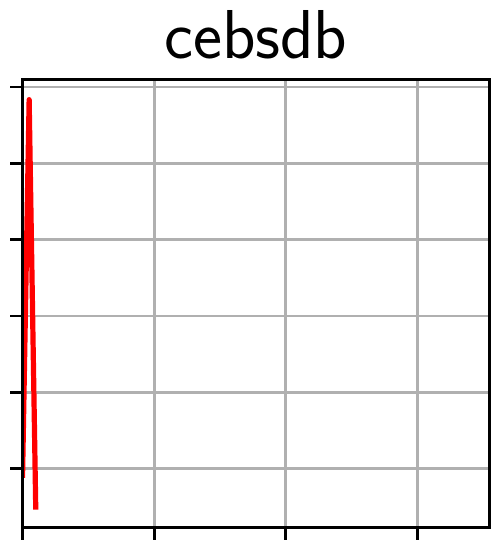}
\includegraphics[width=0.056\textwidth]{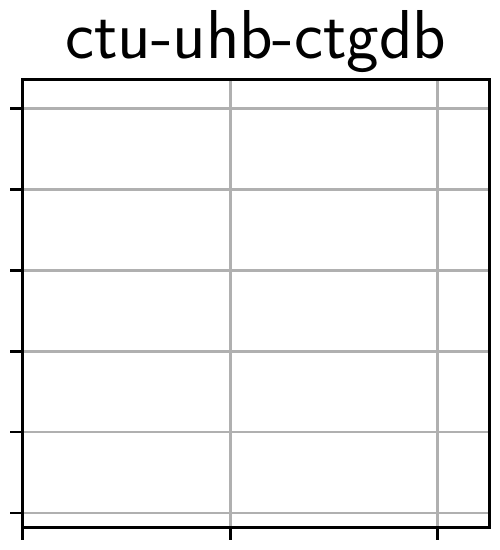}
\includegraphics[width=0.056\textwidth]{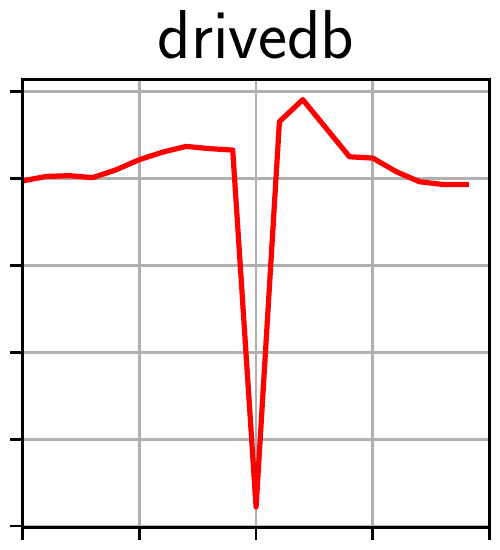}
\includegraphics[width=0.056\textwidth]{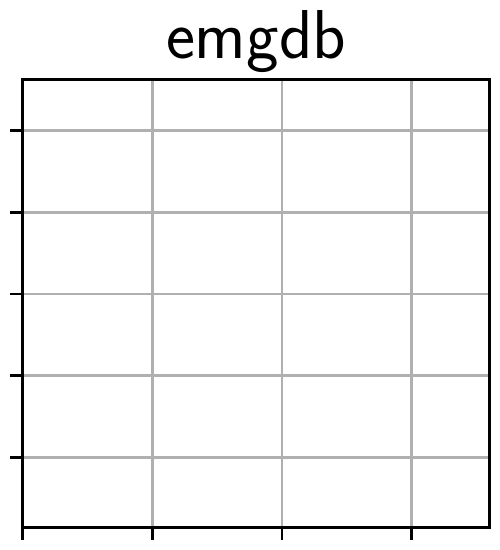}
\includegraphics[width=0.056\textwidth]{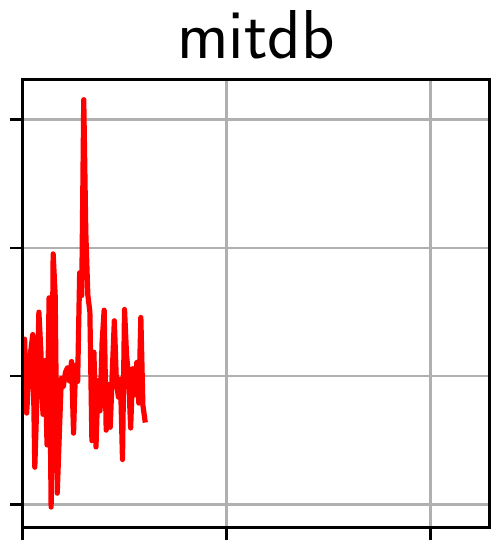}
\includegraphics[width=0.056\textwidth]{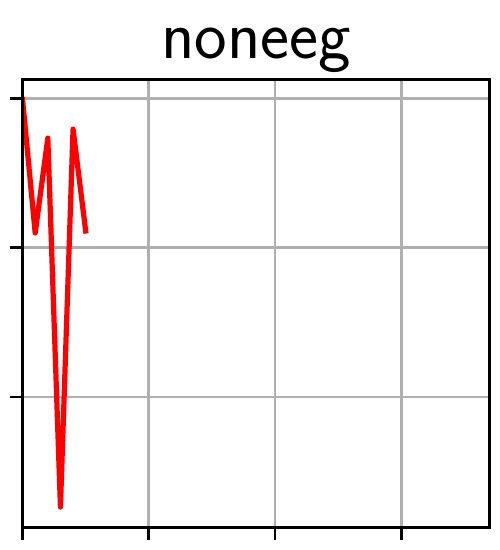}
\includegraphics[width=0.056\textwidth]{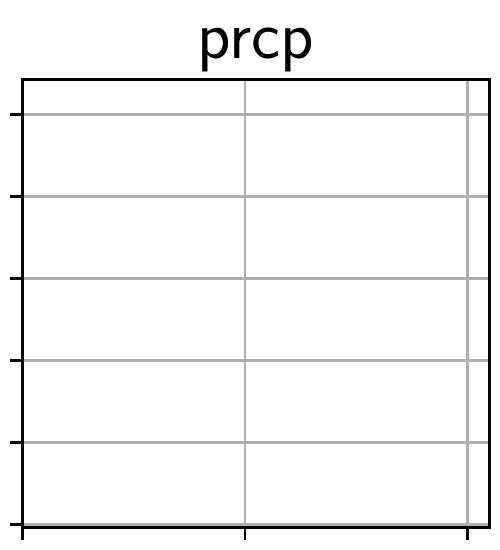}
\includegraphics[width=0.056\textwidth]{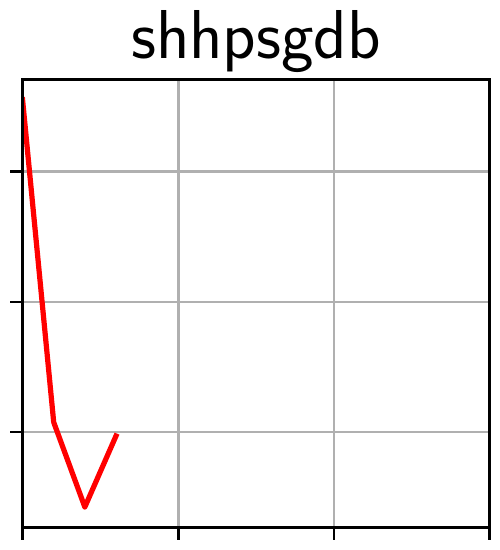}
\includegraphics[width=0.056\textwidth]{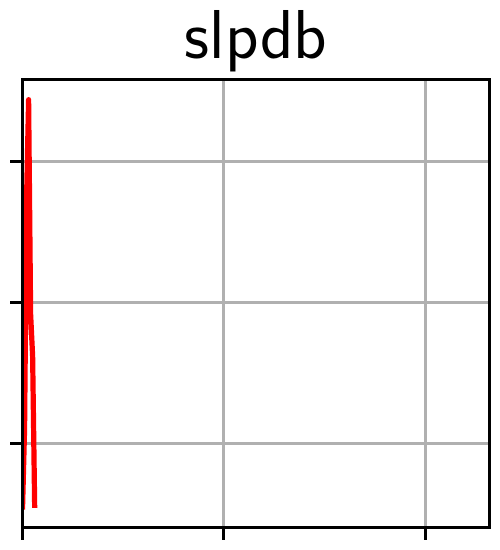}
\includegraphics[width=0.056\textwidth]{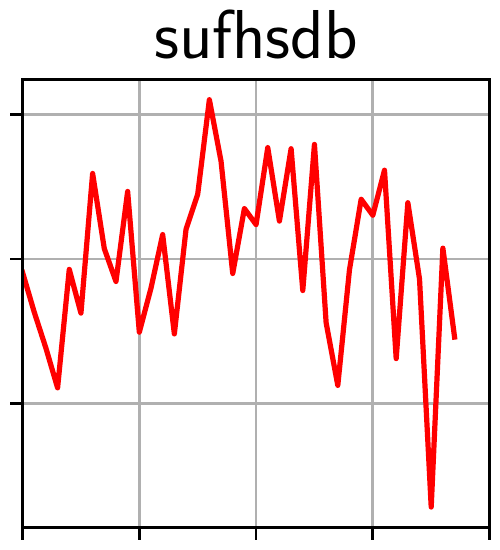}
\includegraphics[width=0.056\textwidth]{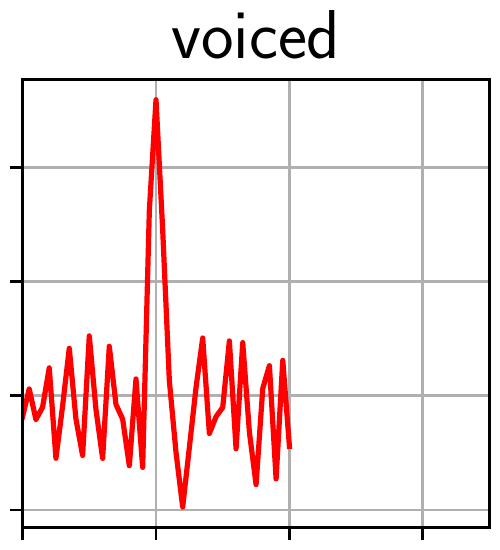}
\includegraphics[width=0.056\textwidth]{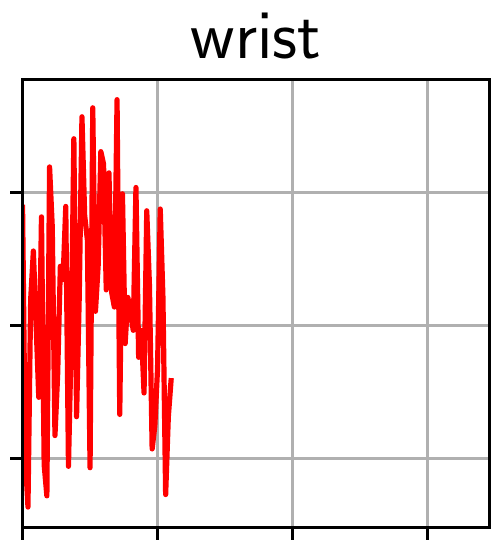}
\\
\includegraphics[width=0.056\textwidth]{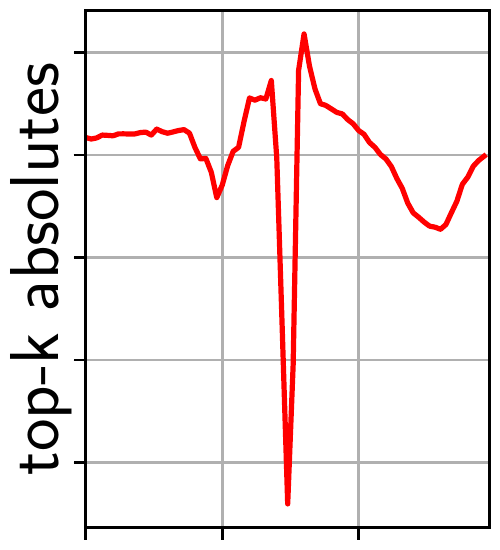}
\includegraphics[width=0.056\textwidth]{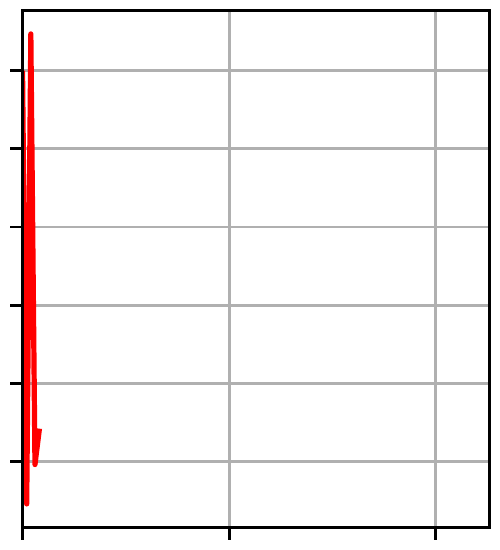}
\includegraphics[width=0.056\textwidth]{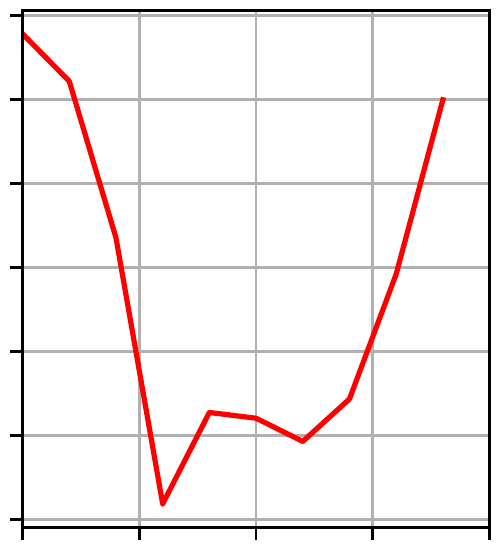}
\includegraphics[width=0.056\textwidth]{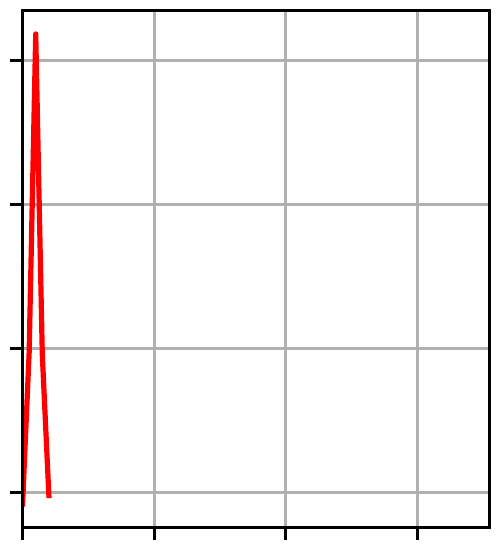}
\includegraphics[width=0.056\textwidth]{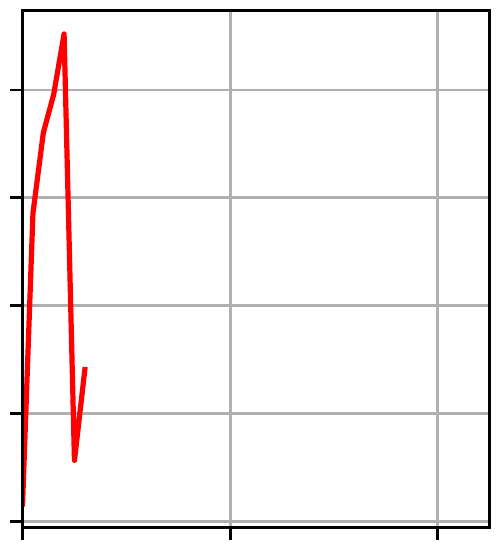}
\includegraphics[width=0.056\textwidth]{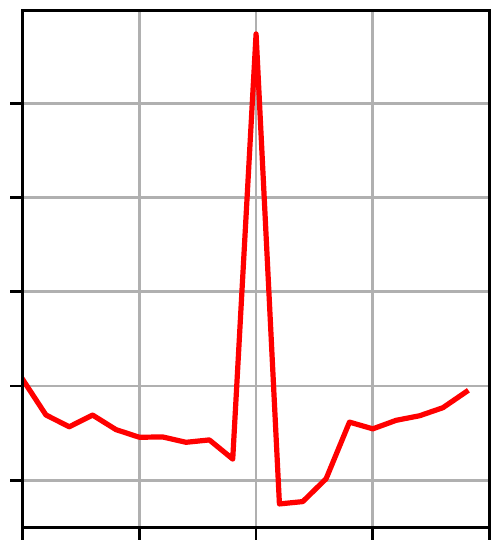}
\includegraphics[width=0.056\textwidth]{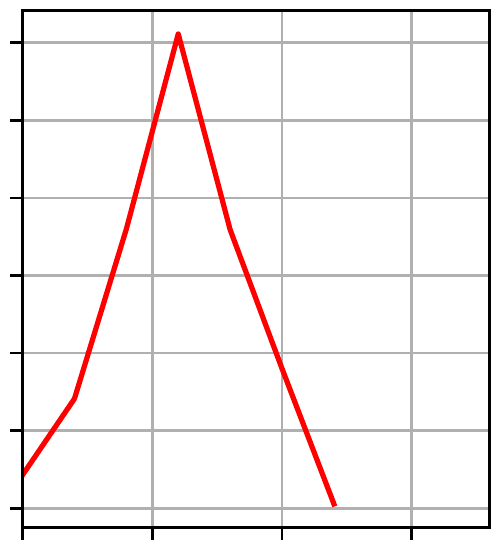}
\includegraphics[width=0.056\textwidth]{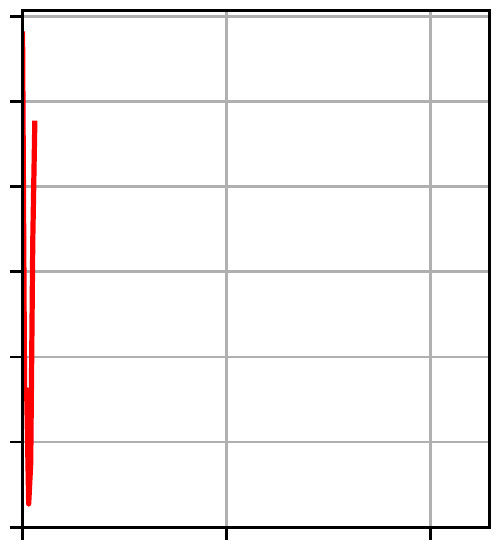}
\includegraphics[width=0.056\textwidth]{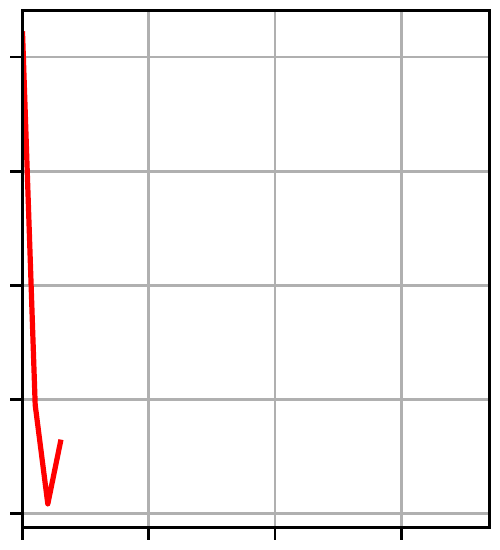}
\includegraphics[width=0.056\textwidth]{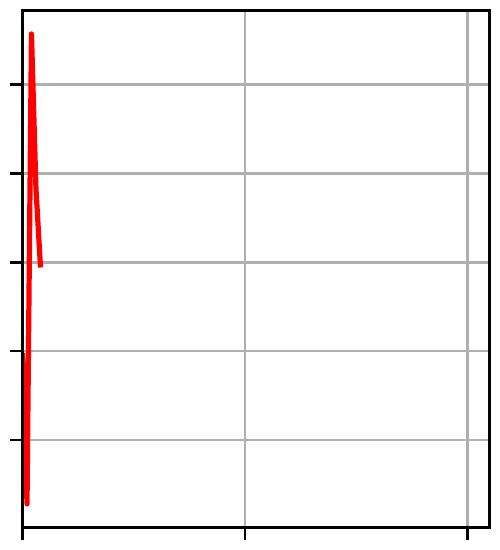}
\includegraphics[width=0.056\textwidth]{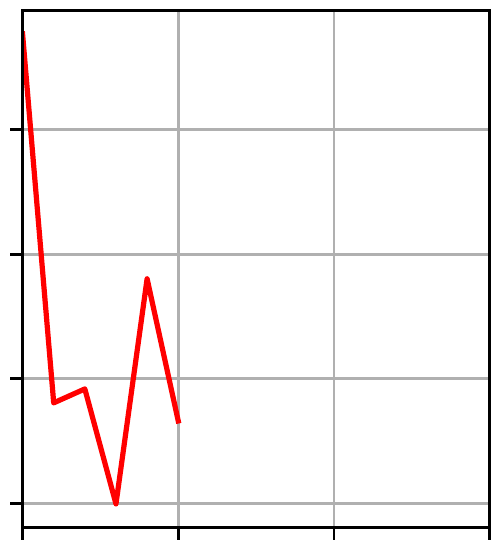}
\includegraphics[width=0.056\textwidth]{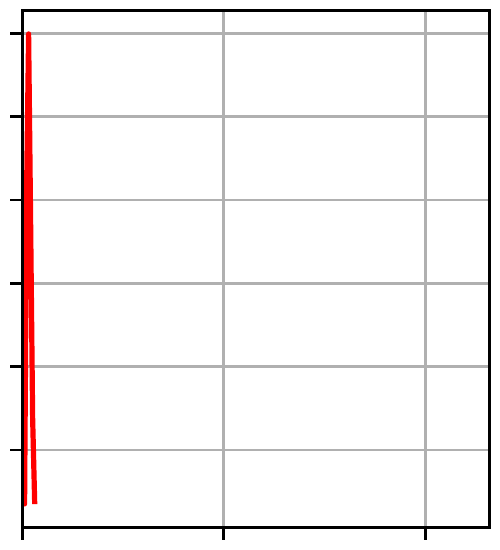}
\includegraphics[width=0.056\textwidth]{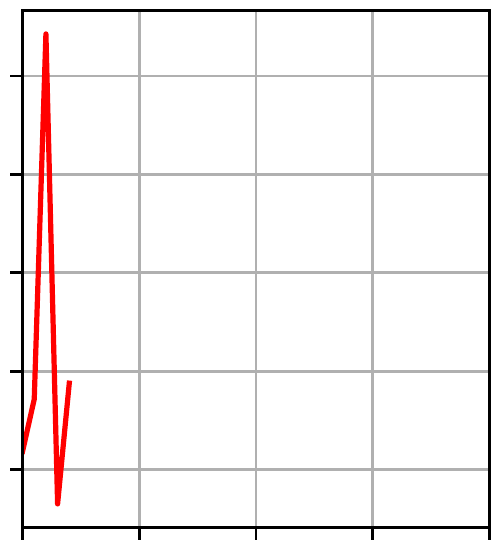}
\includegraphics[width=0.056\textwidth]{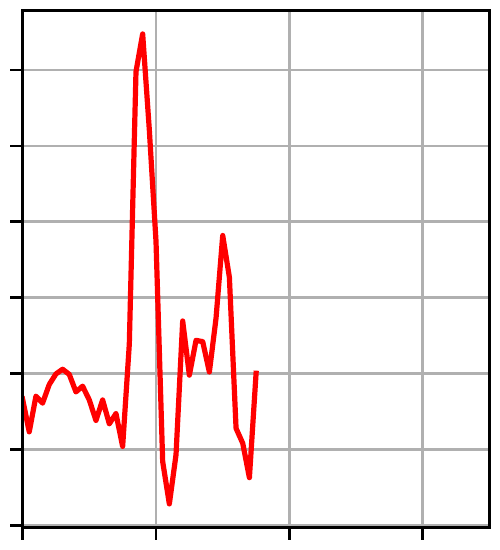}
\includegraphics[width=0.056\textwidth]{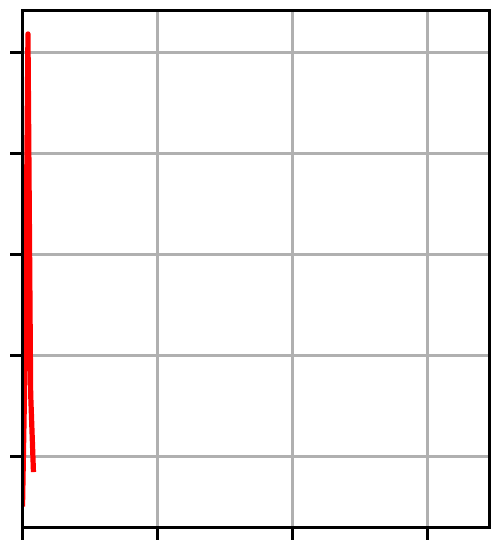}
\\
\includegraphics[width=0.056\textwidth]{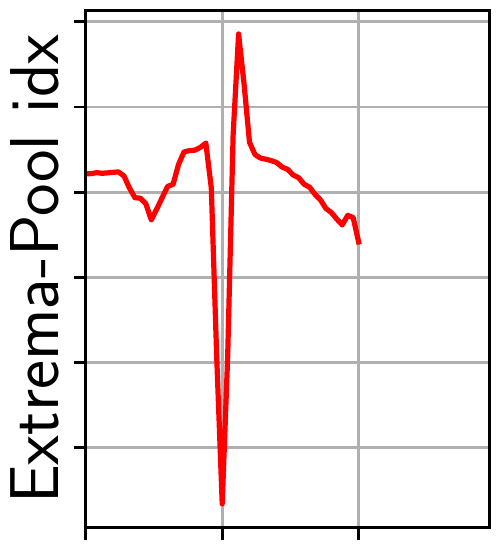}
\includegraphics[width=0.056\textwidth]{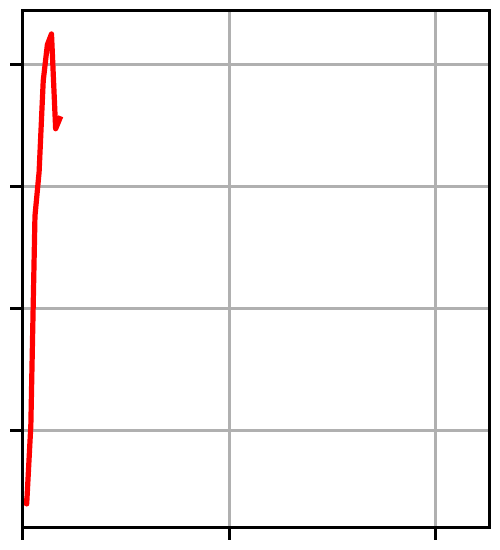}
\includegraphics[width=0.056\textwidth]{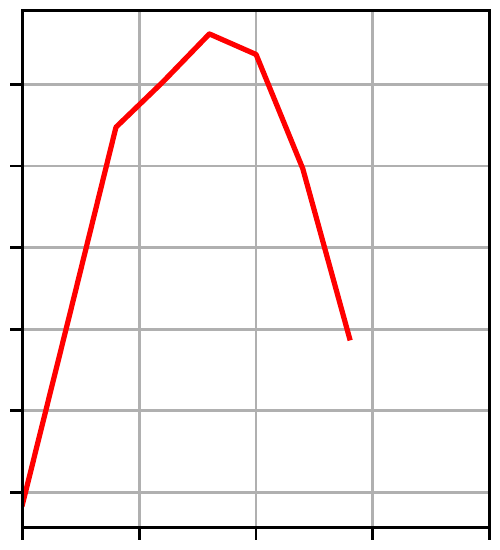}
\includegraphics[width=0.056\textwidth]{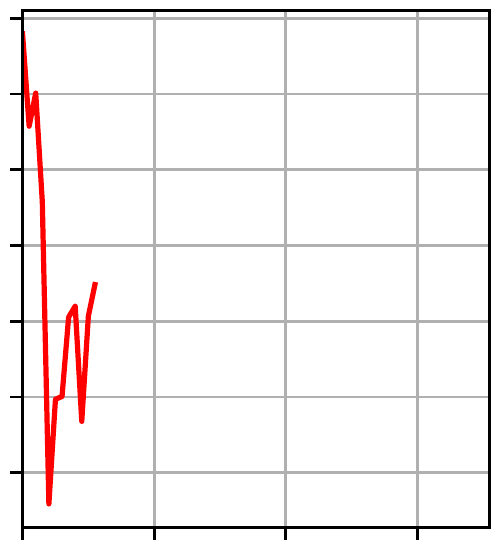}
\includegraphics[width=0.056\textwidth]{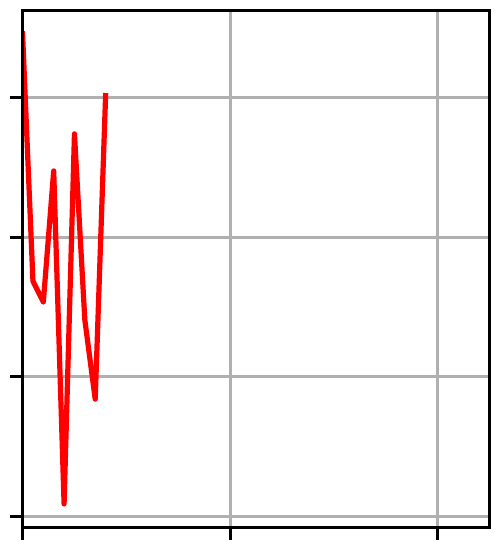}
\includegraphics[width=0.056\textwidth]{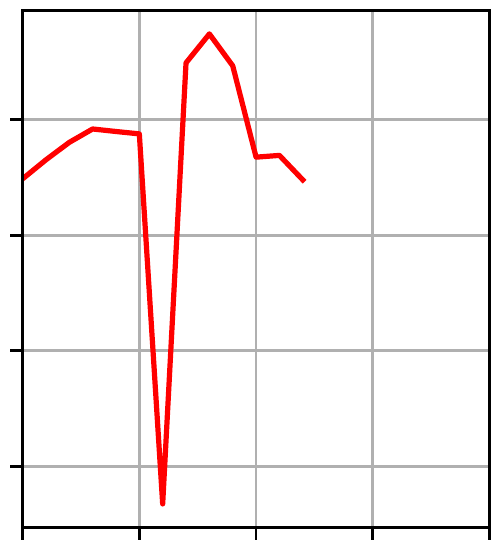}
\includegraphics[width=0.056\textwidth]{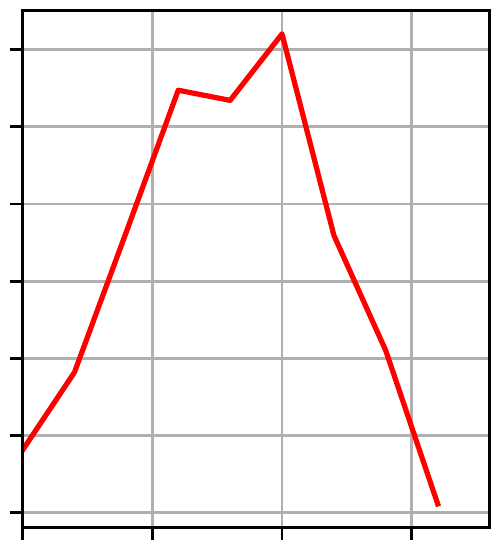}
\includegraphics[width=0.056\textwidth]{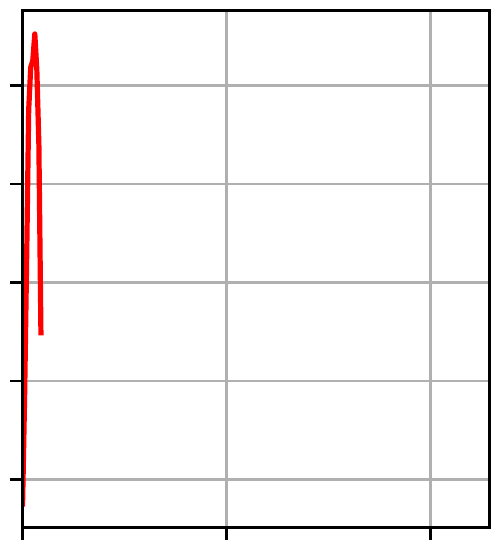}
\includegraphics[width=0.056\textwidth]{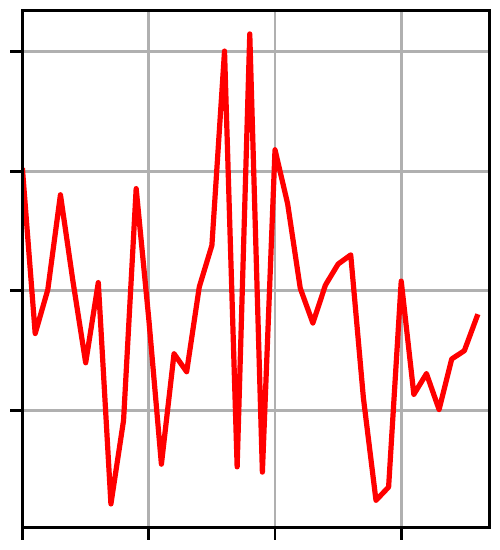}
\includegraphics[width=0.056\textwidth]{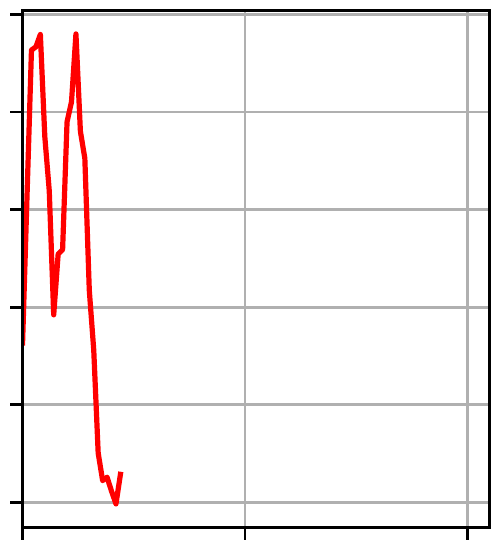}
\includegraphics[width=0.056\textwidth]{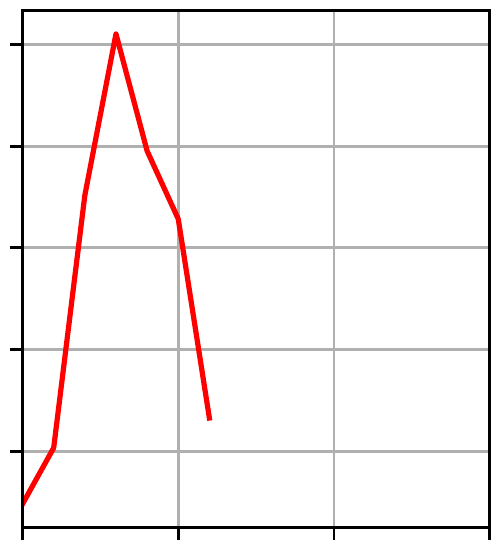}
\includegraphics[width=0.056\textwidth]{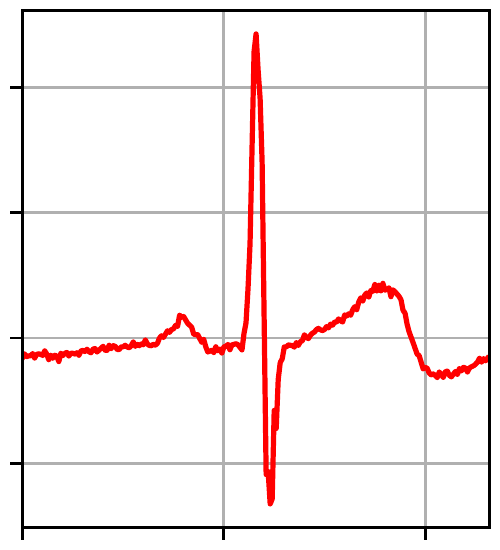}
\includegraphics[width=0.056\textwidth]{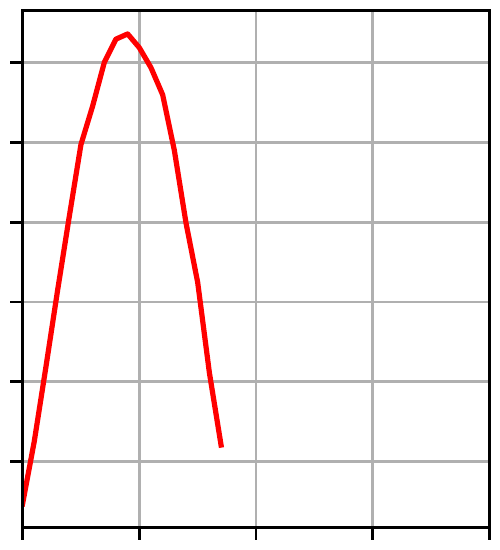}
\includegraphics[width=0.056\textwidth]{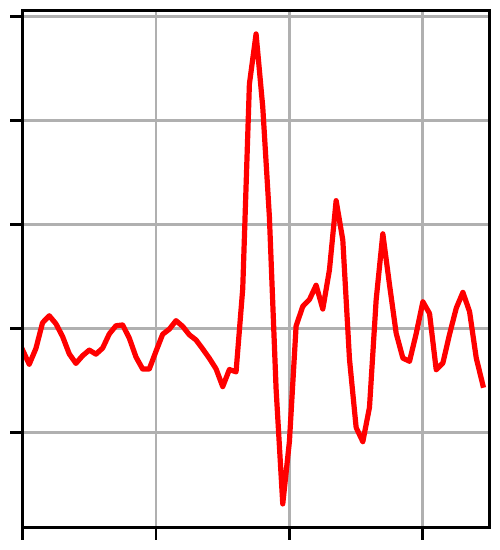}
\includegraphics[width=0.056\textwidth]{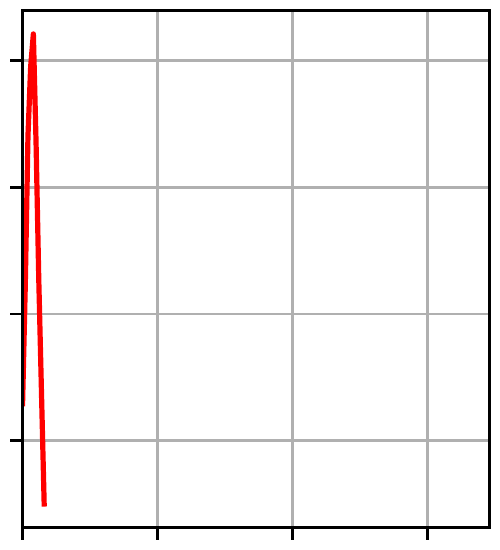}
\\
\includegraphics[width=0.056\textwidth]{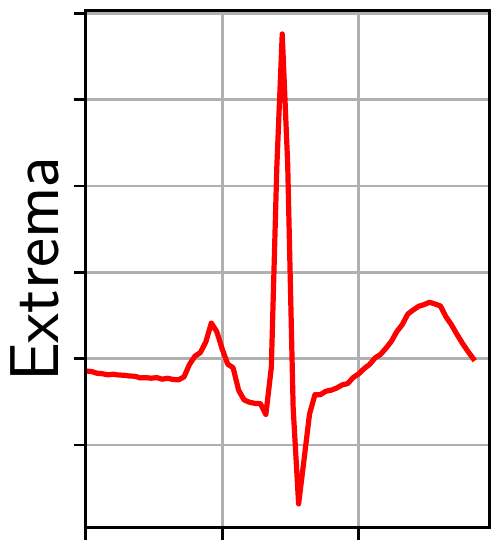}
\includegraphics[width=0.056\textwidth]{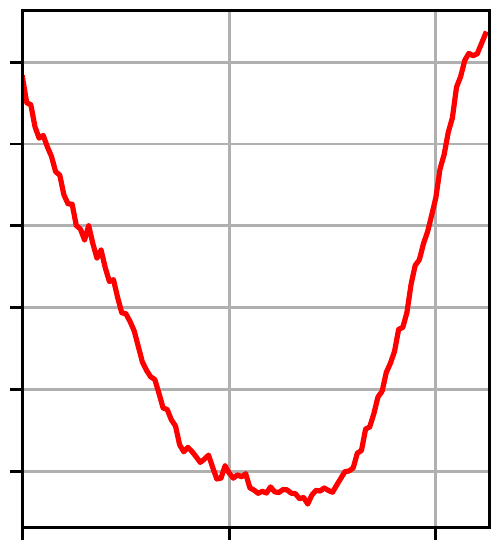}
\includegraphics[width=0.056\textwidth]{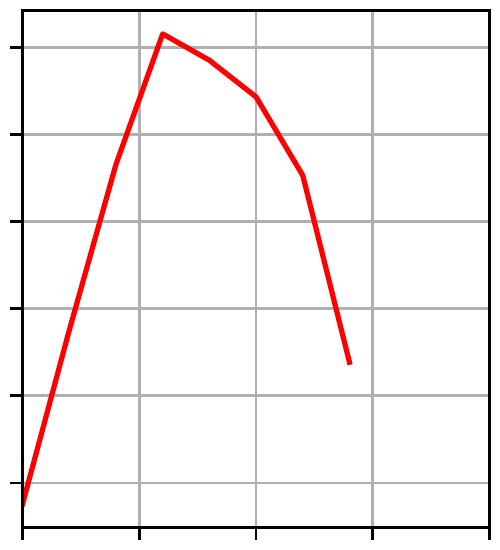}
\includegraphics[width=0.056\textwidth]{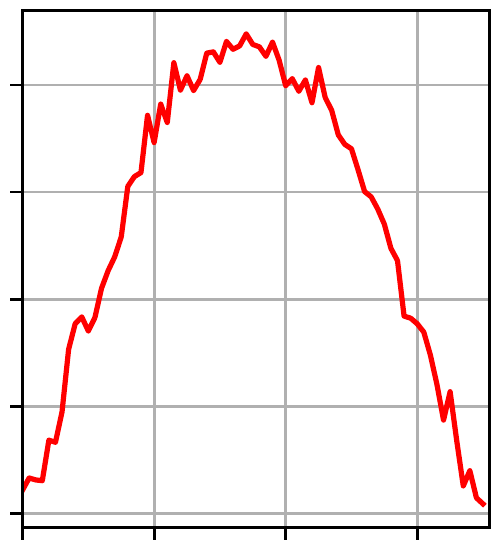}
\includegraphics[width=0.056\textwidth]{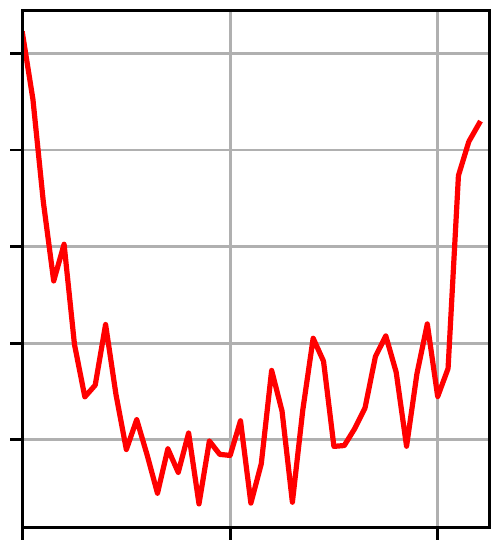}
\includegraphics[width=0.056\textwidth]{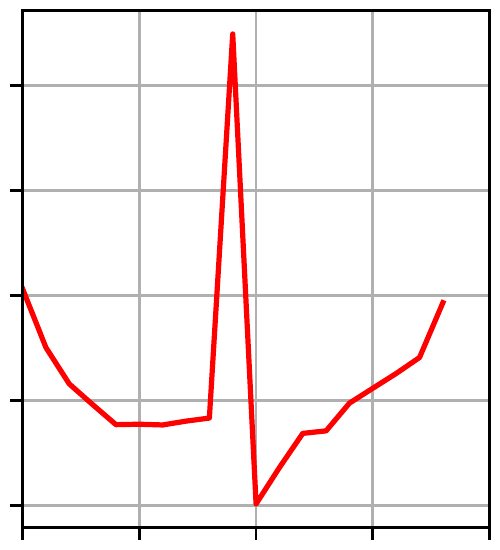}
\includegraphics[width=0.056\textwidth]{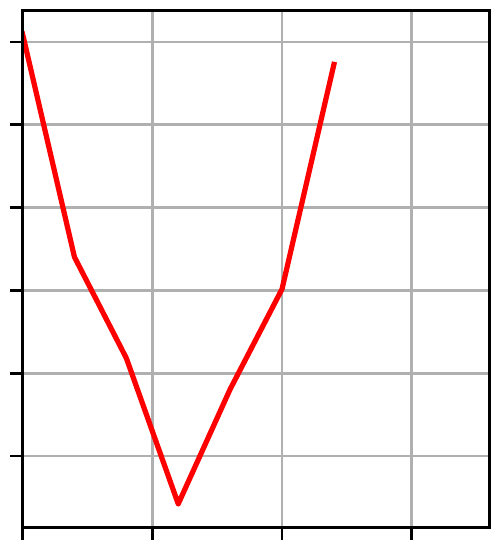}
\includegraphics[width=0.056\textwidth]{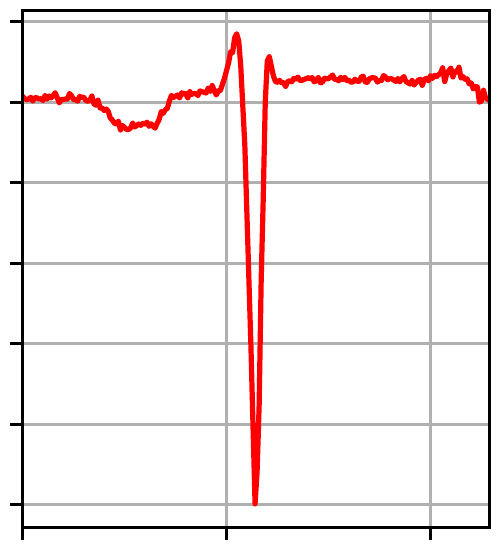}
\includegraphics[width=0.056\textwidth]{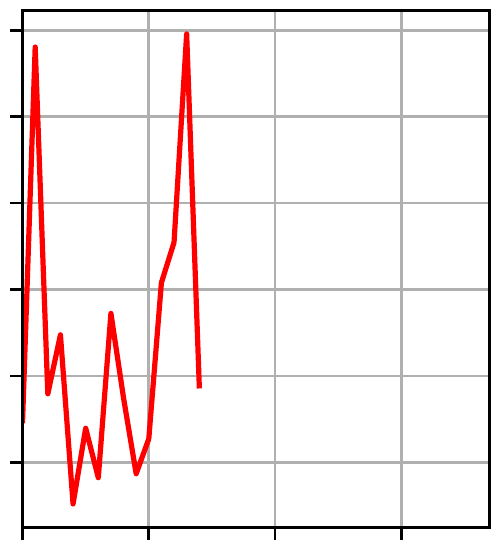}
\includegraphics[width=0.056\textwidth]{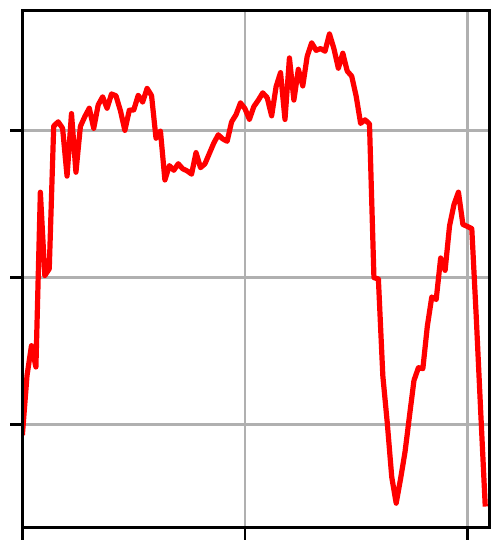}
\includegraphics[width=0.056\textwidth]{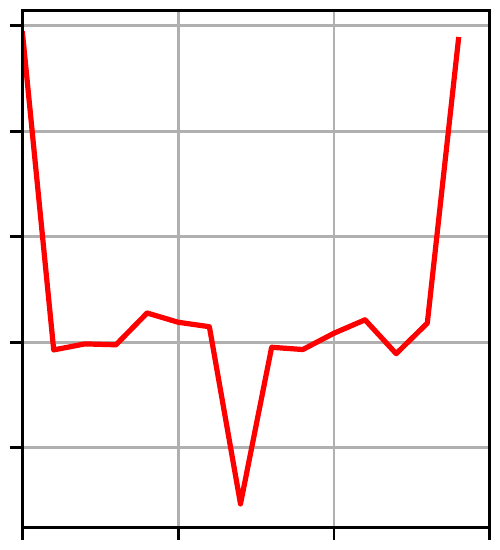}
\includegraphics[width=0.056\textwidth]{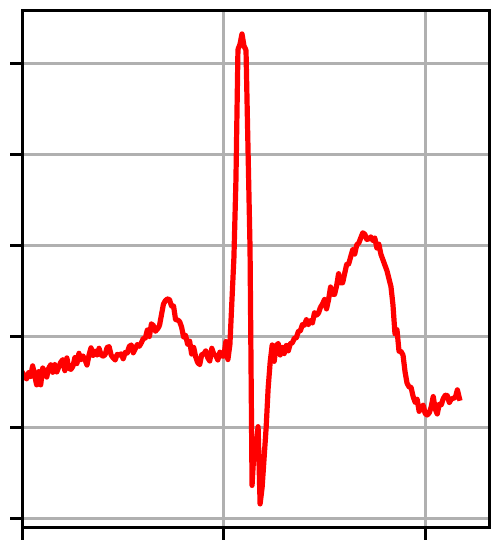}
\includegraphics[width=0.056\textwidth]{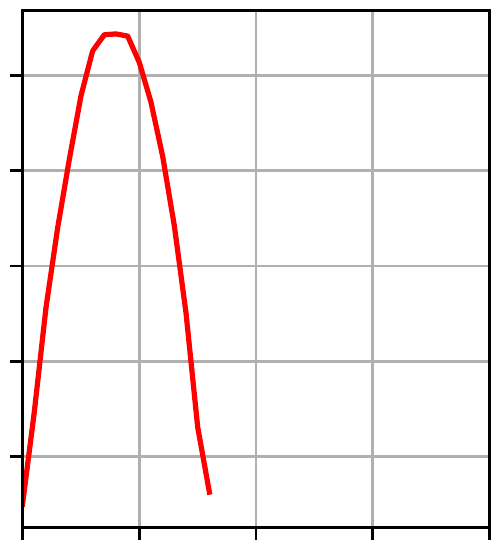}
\includegraphics[width=0.056\textwidth]{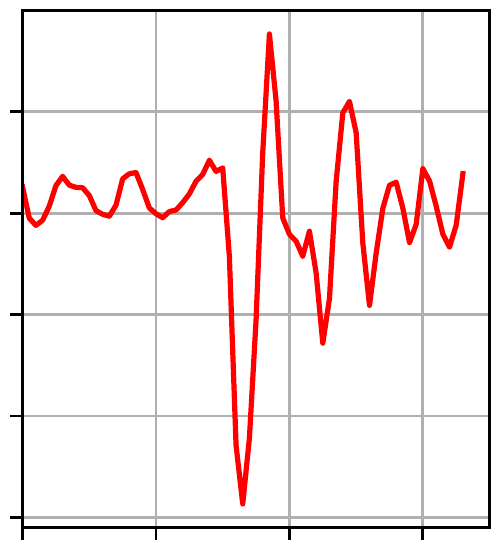}
\includegraphics[width=0.056\textwidth]{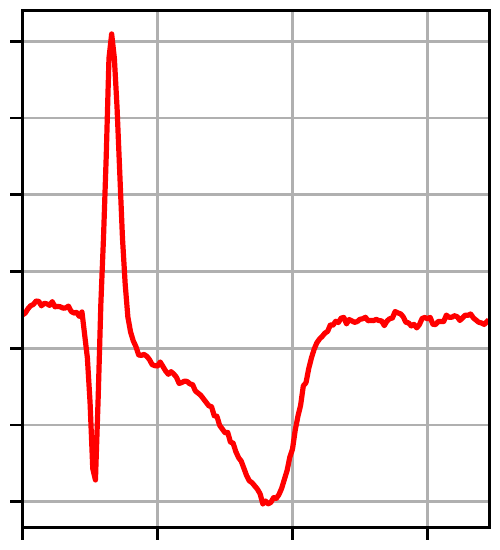}

%% file: table-mean-inverse-compression-ratio-mean-reconstruction-loss-variable-kernel-size.tex
\begin{tabular}{l|rrrr|rrrr|rrrr|rrrr|rrrr}
	\toprule
	{} & \multicolumn{4}{c}{Identity} & \multicolumn{4}{c}{ReLU} & \multicolumn{4}{c}{top-k absolutes} & \multicolumn{4}{c}{Extrema-Pool idx} & \multicolumn{4}{c}{Extrema} \\
	{} &      $m$ & $CR\textsuperscript{-1}$ & $\tilde{\mathcal{L}}$ & $\bar\varphi$ &  $m$ & $CR\textsuperscript{-1}$ & $\tilde{\mathcal{L}}$ & $\bar\varphi$ &             $m$ & $CR\textsuperscript{-1}$ & $\tilde{\mathcal{L}}$ & $\bar\varphi$ &              $m$ & $CR\textsuperscript{-1}$ & $\tilde{\mathcal{L}}$ & $\bar\varphi$ &     $m$ & $CR\textsuperscript{-1}$ & $\tilde{\mathcal{L}}$ & $\bar\varphi$ \\
	\textbf{Datasets     } &          &                          &                       &               &      &                          &                       &               &                 &                          &                       &               &                  &                          &                       &               &         &                          &                       &               \\
	\midrule
	\textbf{apnea-ecg    } &        1 &                     2.00 &                  0.03 &          2.00 &   19 &                     0.70 &                  0.53 &          0.87 &              74 &                     0.10 &                  0.37 &          0.39 &               51 &                     0.09 &                  0.47 &          0.48 &      72 &                     0.10 &                  0.31 &          0.32 \\
	\textbf{bidmc        } &        1 &                     2.00 &                  0.04 &          2.00 &    4 &                     0.82 &                  0.50 &          0.96 &               5 &                     0.41 &                  0.64 &          0.76 &               10 &                     0.21 &                  0.24 &          0.32 &     113 &                     0.13 &                  0.30 &          0.32 \\
	\textbf{bpssrat      } &        1 &                     2.00 &                  0.02 &          2.00 &    1 &                     0.85 &                  0.51 &          0.99 &              10 &                     0.21 &                  0.63 &          0.67 &                8 &                     0.26 &                  0.45 &          0.52 &       8 &                     0.24 &                  0.30 &          0.38 \\
	\textbf{cebsdb       } &        1 &                     2.00 &                  0.01 &          2.00 &    3 &                     0.95 &                  0.51 &          1.07 &               5 &                     0.41 &                  0.62 &          0.74 &               12 &                     0.18 &                  0.21 &          0.28 &      71 &                     0.09 &                  0.45 &          0.46 \\
	\textbf{ctu-uhb-ctgdb} &        1 &                     2.00 &                  0.01 &          2.00 &    1 &                     0.48 &                  0.51 &          0.71 &               7 &                     0.29 &                  0.60 &          0.66 &                9 &                     0.23 &                  0.44 &          0.49 &      45 &                     0.07 &                  0.57 &          0.57 \\
	\textbf{drivedb      } &        1 &                     2.00 &                  0.04 &          2.00 &   20 &                     0.51 &                  0.54 &          0.74 &              20 &                     0.12 &                  0.67 &          0.68 &               13 &                     0.17 &                  0.69 &          0.71 &      19 &                     0.10 &                  0.72 &          0.73 \\
	\textbf{emgdb        } &        1 &                     2.00 &                  0.04 &          2.00 &    1 &                     0.94 &                  0.50 &          1.07 &               7 &                     0.29 &                  0.62 &          0.68 &                9 &                     0.23 &                  0.48 &          0.53 &       7 &                     0.15 &                  0.51 &          0.53 \\
	\textbf{mitdb        } &        1 &                     2.00 &                  0.03 &          2.00 &   61 &                     0.78 &                  0.49 &          0.92 &               7 &                     0.29 &                  0.52 &          0.59 &               10 &                     0.21 &                  0.44 &          0.49 &     229 &                     0.24 &                  0.38 &          0.45 \\
	\textbf{noneeg       } &        1 &                     2.00 &                  0.01 &          2.00 &    6 &                     0.91 &                  0.57 &          1.08 &               4 &                     0.50 &                  0.59 &          0.77 &               37 &                     0.09 &                  0.49 &          0.50 &      15 &                     0.12 &                  0.36 &          0.38 \\
	\textbf{prcp         } &        1 &                     2.00 &                  0.03 &          2.00 &    1 &                     1.00 &                  0.51 &          1.12 &               5 &                     0.41 &                  0.59 &          0.71 &               23 &                     0.11 &                  0.41 &          0.42 &     105 &                     0.12 &                  0.42 &          0.44 \\
	\textbf{shhpsgdb     } &        1 &                     2.00 &                  0.02 &          2.00 &    4 &                     0.85 &                  0.60 &          1.05 &               6 &                     0.34 &                  0.69 &          0.77 &                7 &                     0.29 &                  0.42 &          0.51 &      15 &                     0.10 &                  0.53 &          0.54 \\
	\textbf{slpdb        } &        1 &                     2.00 &                  0.03 &          2.00 &    7 &                     0.72 &                  0.53 &          0.90 &               7 &                     0.29 &                  0.52 &          0.60 &              232 &                     0.24 &                  0.29 &          0.37 &     218 &                     0.23 &                  0.36 &          0.43 \\
	\textbf{sufhsdb      } &        1 &                     2.00 &                  0.03 &          2.00 &   38 &                     1.02 &                  0.24 &          1.05 &               5 &                     0.41 &                  0.55 &          0.68 &               18 &                     0.13 &                  0.36 &          0.39 &      17 &                     0.12 &                  0.26 &          0.28 \\
	\textbf{voiced       } &        1 &                     2.00 &                  0.01 &          2.00 &   41 &                     0.95 &                  0.26 &          0.98 &              36 &                     0.09 &                  0.56 &          0.57 &               70 &                     0.10 &                  0.41 &          0.43 &      67 &                     0.10 &                  0.41 &          0.43 \\
	\textbf{wrist        } &        1 &                     2.00 &                  0.04 &          2.00 &   56 &                     0.74 &                  0.62 &          0.96 &               5 &                     0.41 &                  0.49 &          0.63 &                9 &                     0.23 &                  0.43 &          0.49 &     173 &                     0.18 &                  0.46 &          0.50 \\
	\bottomrule
\end{tabular}

%% file: table-uci-epilepsy-supervised.tex
\begin{tabular}{l|rrrr|rrrr|rrrr|rrrr|rrrr}
	\toprule
	{} & \multicolumn{4}{c}{Identity} & \multicolumn{4}{c}{ReLU} & \multicolumn{4}{c}{top-k absolutes} & \multicolumn{4}{c}{Extrema-Pool idx} & \multicolumn{4}{c}{Extrema} \\
	{} & $CR\textsuperscript{-1}$ & $\tilde{\mathcal{L}}$ & $\bar\varphi$ & Acc\textsubscript{\%} & $CR\textsuperscript{-1}$ & $\tilde{\mathcal{L}}$ & $\bar\varphi$ & Acc\textsubscript{\%} & $CR\textsuperscript{-1}$ & $\tilde{\mathcal{L}}$ & $\bar\varphi$ & Acc\textsubscript{\%} & $CR\textsuperscript{-1}$ & $\tilde{\mathcal{L}}$ & $\bar\varphi$ & Acc\textsubscript{\%} & $CR\textsuperscript{-1}$ & $\tilde{\mathcal{L}}$ & $\bar\varphi$ & Acc\textsubscript{\%} \\
	\textbf{$m$} &                          &                       &               &                       &                          &                       &               &                       &                          &                       &               &                       &                          &                       &               &                       &                          &                       &               &                       \\
	\midrule
	\textbf{15 } &                     4.17 &                  0.02 &          4.17 &                  40.8 &                     2.17 &                  0.03 &          2.17 &                  82.6 &                     0.42 &                  0.79 &          0.89 &                  74.1 &                     0.42 &                  0.46 &          0.62 &                  73.4 &                     0.34 &                  0.43 &          0.55 &                  80.7 \\
	\textbf{16 } &                     4.18 &                  0.02 &          4.18 &                  91.2 &                     2.18 &                  0.04 &          2.18 &                  87.0 &                     0.43 &                  0.78 &          0.89 &                  70.5 &                     0.43 &                  0.44 &          0.61 &                  78.8 &                     0.34 &                  0.44 &          0.56 &                  74.5 \\
	\textbf{17 } &                     4.19 &                  0.02 &          4.19 &                  89.0 &                     2.20 &                  0.02 &          2.20 &                  90.1 &                     0.42 &                  0.78 &          0.89 &                  73.0 &                     0.42 &                  0.45 &          0.61 &                  78.9 &                     0.34 &                  0.44 &          0.56 &                  75.1 \\
	\textbf{18 } &                     4.20 &                  0.04 &          4.20 &                  91.2 &                     2.21 &                  0.05 &          2.21 &                  87.1 &                     0.40 &                  0.79 &          0.88 &                  73.9 &                     0.40 &                  0.48 &          0.63 &                  79.7 &                     0.34 &                  0.45 &          0.57 &                  67.2 \\
	\textbf{19 } &                     4.21 &                  0.02 &          4.21 &                  88.6 &                     2.21 &                  0.03 &          2.21 &                  89.6 &                     0.42 &                  0.78 &          0.88 &                  76.1 &                     0.42 &                  0.45 &          0.61 &                  78.8 &                     0.36 &                  0.44 &          0.57 &                  78.3 \\
	\textbf{20 } &                     4.22 &                  0.03 &          4.22 &                  88.6 &                     2.22 &                  0.03 &          2.22 &                  87.4 &                     0.40 &                  0.79 &          0.89 &                  69.6 &                     0.40 &                  0.49 &          0.64 &                  77.5 &                     0.35 &                  0.45 &          0.58 &                  73.8 \\
	\textbf{21 } &                     4.24 &                  0.02 &          4.24 &                  89.4 &                     2.23 &                  0.03 &          2.24 &                  87.0 &                     0.42 &                  0.79 &          0.89 &                  73.0 &                     0.42 &                  0.47 &          0.63 &                  70.7 &                     0.37 &                  0.44 &          0.58 &                  73.8 \\
	\textbf{22 } &                     4.25 &                  0.03 &          4.25 &                  89.3 &                     2.26 &                  0.04 &          2.26 &                  89.4 &                     0.43 &                  0.78 &          0.89 &                  73.6 &                     0.43 &                  0.45 &          0.62 &                  78.7 &                     0.38 &                  0.44 &          0.59 &                  76.5 \\
	\bottomrule
\end{tabular}

%% file: table-mnist-supervised.tex
\begin{tabular}{l|rrrr|rrrr|rrrr|rrrr|rrrr}
	\toprule
	{} & \multicolumn{4}{c}{Identity} & \multicolumn{4}{c}{ReLU} & \multicolumn{4}{c}{top-k absolutes} & \multicolumn{4}{c}{Extrema-Pool idx} & \multicolumn{4}{c}{Extrema} \\
	{} & $CR\textsuperscript{-1}$ & $\tilde{\mathcal{L}}$ & $\bar\varphi$ & Acc\textsubscript{\%} & $CR\textsuperscript{-1}$ & $\tilde{\mathcal{L}}$ & $\bar\varphi$ & Acc\textsubscript{\%} & $CR\textsuperscript{-1}$ & $\tilde{\mathcal{L}}$ & $\bar\varphi$ & Acc\textsubscript{\%} & $CR\textsuperscript{-1}$ & $\tilde{\mathcal{L}}$ & $\bar\varphi$ & Acc\textsubscript{\%} & $CR\textsuperscript{-1}$ & $\tilde{\mathcal{L}}$ & $\bar\varphi$ & Acc\textsubscript{\%} \\
	\textbf{$m$} &                          &                       &               &                       &                          &                       &               &                       &                          &                       &               &                       &                          &                       &               &                       &                          &                       &               &                       \\
	\midrule
	\textbf{1  } &                     1.16 &                  0.01 &          1.16 &                  91.7 &                     0.58 &                  0.00 &          0.58 &                  92.0 &                     1.16 &                  0.00 &          1.16 &                  91.1 &                     1.16 &                  0.00 &          1.16 &                  91.1 &                     0.08 &                  0.88 &          0.88 &                  83.3 \\
	\textbf{2  } &                     1.55 &                  0.02 &          1.55 &                  92.1 &                     0.58 &                  0.01 &          0.58 &                  90.4 &                     1.37 &                  0.01 &          1.37 &                  92.0 &                     0.48 &                  0.62 &          0.79 &                  93.2 &                     0.09 &                  0.83 &          0.83 &                  84.5 \\
	\textbf{3  } &                     1.93 &                  0.02 &          1.93 &                  91.5 &                     0.65 &                  0.00 &          0.65 &                  91.5 &                     0.63 &                  0.26 &          0.68 &                  90.3 &                     0.30 &                  0.51 &          0.59 &                  92.1 &                     0.08 &                  0.50 &          0.51 &                  84.0 \\
	\textbf{4  } &                     2.30 &                  0.05 &          2.30 &                  91.7 &                     1.57 &                  0.02 &          1.57 &                  91.3 &                     0.39 &                  0.41 &          0.57 &                  87.9 &                     0.22 &                  0.59 &          0.63 &                  91.9 &                     0.06 &                  0.57 &          0.57 &                  82.8 \\
	\textbf{5  } &                     2.66 &                  0.07 &          2.66 &                  90.0 &                     0.65 &                  0.03 &          0.65 &                  91.3 &                     0.20 &                  0.54 &          0.58 &                  87.3 &                     0.16 &                  0.60 &          0.62 &                  91.5 &                     0.08 &                  0.57 &          0.58 &                  82.4 \\
	\textbf{6  } &                     3.02 &                  0.13 &          3.02 &                  91.8 &                     1.58 &                  0.02 &          1.58 &                  91.7 &                     0.14 &                  0.61 &          0.63 &                  84.2 &                     0.12 &                  0.63 &          0.65 &                  90.8 &                     0.06 &                  0.60 &          0.61 &                  81.2 \\
	\bottomrule
\end{tabular}

%% file: chapter7.tex
\chapter{Επίλογος}
\label{chapter7}
Με κάθε τεχνολογική πρόοδο η ιατρική, η οποία μέχρι τώρα είναι βαθιά εξαρτώμενη απο τον ανθρώπινο παράγοντα, έρχεται πιο κοντά σε ένα αυτοματοποιημένο τομέα κατευθυνόμενο από την τεχνητή νοημοσύνη.
Το AI όχι μόνο θα φτάσει στο σημείο που θα ανιχνεύει ασθένειες σε πραγματικό χρόνο, αλλά θα ερμηνεύει επίσης αμφιλεγόμενες καταστάσεις, φαινοτυπικές πολύπλοκες ασθένειες και θα παίρνει ιατρικές αποφάσεις.
Ωστόσο, η πλήρης θεωρητική κατανόηση της βαθιάς μάθησης δεν είναι ακόμη διαθέσιμη και η κριτική κατανόηση των πλεονεκτημάτων και των περιορισμών της εσωτερικής λειτουργίας της είναι ζωτικής σημασίας για να κερδίσει τη θέση της στην καθημερινή κλινική χρήση.
Η επιτυχής εφαρμογή του AI στον ιατρικό τομέα βασίζεται στην επίτευξη ερμηνεύσιμων μοντέλων και δημιουργία μεγάλων βάσεων δεδομένων.

Στο πλαίσιο της διδακτορικής διατριβής προτείναμε το μέτρο $\varphi$ για να αξιολογήσουμε πόσο καλά τα μοντέλα μηχανικής μάθησης ανταλλάσσουν απώλεια ανακατασκευής με συμπίεση.
Έπειτα προτείναμε μια νέα αρχιτεκτονική νευρωνικών δικτύων τα \textbf{SANs} τα οποία έχουν ελάχιστη δομή και με την χρήση των συναρτήσεων αραιής ενεργοποίησης μαθαίνουν να συμπιέζουν δεδομένα χωρίς να χάνουν σημαντικές πληροφορίες.
Χρησιμοποιώντας τις βάσεις δεδομένων Physionet και MNIST αποδείξαμε ότι τα SANs είναι σε θέση να δημιουργούν αναπαραστάσεις υψηλής ποιότητας με ερμηνεύσιμους πυρήνες.

Η ελάχιστη δομή των SANs καθιστά εύκολη τη χρήση τους για την εξαγωγή χαρακτηριστικών, την ομαδοποίηση και την πρόβλεψη χρονοσειρών.
Άλλες μελλοντικές εργασίες σχετικά με τα SANs περιλαμβάνουν:
\begin{itemize}
	\item Εφαρμογή αλγορίθμων σταδιακής μείωσης των ελάχιστων αποστάσεων των ακρότατων, για την αύξηση του βαθμού ελευθερίας των πυρήνων.
	\item Επιβολή ελάχιστης απόστασης ακρότατων κατά μήκος όλων των πινάκων ομοιότητας για πολλαπλούς πυρήνες, κάνοντας έτσι τους πυρήνες να ανταγωνίζονται για περιοχές.
	\item Εφαρμογή του dropout στις ενεργοποιήσεις για να διορθωθούν τα βάρη που έχουν υπερβεί, ειδικά όταν αρχικοποιούνται με υψηλές τιμές.
		Ωστόσο, η επίδραση του dropout στα SAN θα ήταν γενικά αρνητική, καθώς τα SANs έχουν πολύ μικρότερο αριθμό βαρών από τα DNNs και επομένως δεν χρειάζονται ισχυρή κανονικοποίηση.
	\item Χρήση των SAN με δυναμικά δημιουργούμενους πυρήνες οι οποίοι θα μπορούσαν να μάθουν πολυτροπικά δεδομένα από μεταβλητές πηγές (e.g.\ από ECG σε αναπνευστικά) χωρίς να καταστρέψουν προηγούμενα βάρη.
\end{itemize}

%% file: acronyms.tex
\chapter*{Ακρωνύμια}
\label{sec:acronyms}

\addcontentsline{toc}{chapter}{Ακρωνύμια}

\begin{tabbing}
	aaaaaaaaaaaaaaaaaa \= aaaa\kill
	\Large\textbf{Ακρωνύμιο} \> \Large\textbf{Επεξήγηση} \\
	ACDC \> Automated Cardiac Diagnosis Challenge \\
	ACS \> Acute Coronary Syndrome \\
	AE \> Autoencoder \\
	AF \> Atrial Fibrillation \\
	AI \> Artificial Intelligence \\
	BIH \> Beth Israel Hospital \\
	BP \> Blood Pressure \\
	CAC \> Coronary Artery Calcification \\
	CAD \> Coronary Artery Disease \\
	CHF \> Congestive Heart Failure \\
	CNN \> Convolutional Neural Network \\
	CRF \> Conditional Random Field \\
	CT \> Computerized Tomography \\
	CVD \> Cardiovascular Disease \\
	DBN \> Deep Belief Networks \\
	DBP \> Diastolic Blood Pressure \\
	DNN \> Deep Neural Networks \\
	DWI \> Diffusion Weighted Imaging \\
	ECG \> Electrocardiogram \\
	EEG \> Electroencephalogram \\
	EHR \> Electronic Health Record \\
	FCN \> Fully Convolutional Network \\
	FECG \> Fetal ECG \\
	FNN \> Fully Connected Networks \\
	GAN \> Generative Adversarial Network \\
	GRU \> Gated Recurrent Unit \\
	HF \> Heart Failure \\
	HT \> Hemorrhagic Transformation \\
	HVSMR \> Heart \& Vessel Segmentation from 3D MRI \\
	ICD \> International Classification of Diseases \\
	IVUS \> Intravascular Ultrasound \\
	LSTM \> Long-Short Term Memory \\
	LV \> Left Ventricle \\
	MA \> Microaneurysm \\
	MFCC \> Mel-Frequency Cepstral Coefficient \\
	MICCAI \> Medical Image Computing \& Computer-Assisted Intervention \\
	MI \> Myocardial Infarction \\
	MMWHS \> Multi-Modality Whole Heart Segmentation Challenge \\
	MRA \> Magnetic Resonance Angiography \\
	MRI \> Magnetic Resonance Imaging \\
	MRP \> Magnetic Resonance Perfusion \\
	OCT \> Optical Coherence Tomography \\
	PCA \> Principal Component Analysis \\
	PCG \> Phonocardiogram \\
	PPG \> Pulsatile Photoplethysmography \\
	RBM \> Restricted Boltzmann Machine \\
	RF \> Random Forest \\
	RNN \> Recurrent Neural Network \\
	ROI \> Region of Interest \\
	RV \> Right Ventricle \\
	SAE \> Stacked Autoencoder \\
	SATA \> Segmentation Algorithms, Theory and Applications \\
	SBP \> Systolic Blood Pressure \\
	SDAE \> Stacked Denoised Autoencoder \\
	SLO \> Scanning Laser Ophthalmoscopy \\
	SSAE \> Stacked Sparse Autoencoder \\
	STACOM \> Statistical Atlases \& Computational Modeling of the Heart \\
	SVM \> Support Vector Machine \\
	VGG \> Visual Geometry Group \\
	WT \> Wavelet Transform \\
\end{tabbing}

%% file: publications.tex
\chapter{Κατάλογος Δημοσιεύσεων}
\section*{Άρθρα σε Επιστημονικά Περιοδικά}
\begin{enumerate}
	\item \textlatin{\textbf{Paschalis Bizopoulos}, Dimitris D.G Koutsouris, Sparsely Activated Networks, \textit{Under Review/Considered for publication, 2019.}}
	\item \textlatin{\textbf{Paschalis Bizopoulos}, Dimitris D.G Koutsouris, Deep Learning in Cardiology, \textit{IEEE Reviews in Biomedical Engineering, 12:168--193, 2019.}}
	\item \textlatin{Bourantas, Christos V and Papafaklis, Michail I and Lakkas, Lampros and Sakellarios, Antonis and Onuma, Yoshinobu and Zhang, Yao-Jun and Muramatsu, Takashi and Diletti, Roberto and \textbf{Bizopoulos, Paschalis} and Kalatzis, Fanis and Naka, Katerina and Fotiadis, Dimitrios and Wang, Jin and Garcia Garcia, Hector and Kimura, Takeshi and Michalis, Lampros and Serruys Patrick, Fusion of optical coherence tomographic and angiographic data for more accurate evaluation of the endothelial shear stress patterns and neointimal distribution after bioresorbable scaffold implantation: comparison with intravascular ultrasound-derived reconstructions, \textit{The international journal of cardiovascular imaging, 30, 3, 485--494, 2014, Springer Netherlands}}
	\item \textlatin{Bourantas, Christos V and Papadopoulou, Stella-Lida and Serruys, Patrick W and Sakellarios, Antonis and Kitslaar, Pieter H and \textbf{Bizopoulos, Paschalis} and Girasis, Chrysafios and Zhang, Yao-Jun and de Vries, Ton and Boersma, Eric and Papafaklis, Michail and Naka, Katerina and Fotiadis, Dimitrios and Stone, Gregg and Reiber, Johan and Michalis, Lampros and Feyter, Pim and Garcia-Garcia, Hector, Noninvasive Prediction of Atherosclerotic Progression: The PROSPECT-MSCT Study, \textit{JACC\@. Cardiovascular imaging, 9, 8, 1009, 2016}}
	\item \textlatin{Sakellarios, Antonis I and \textbf{Bizopoulos, Paschalis} and Papafaklis, Michail I and Athanasiou, Lambros and Exarchos, Themis and Bourantas, Christos V and Naka, Katerina K and Patterson, Andrew J and Young, Victoria EL and Gillard, Jonathan H and Parodi, Oberdan and Michalis, Lampros and Fotiadis, Dimitrios, Natural history of carotid atherosclerosis in relation to the hemodynamic environment: a low-density lipoprotein transport modeling study with serial magnetic resonance imaging in humans, \textit{Angiology, 68, 2, 109--118, 2017,SAGE Publications Sage CA\@: Los Angeles, CA}}
\end{enumerate}

\section*{Ανακοινώσεις σε Συνέδρια}
\begin{enumerate}
	\item \textlatin{\textbf{Bizopoulos, Paschalis} and Koutsouris, Dimitrios D, Signal2Image Modules in Deep Neural Networks for EEG Classification, \textit{Engineering in Medicine and Biology Society (EMBC), 2019 41th Annual International Conference of the IEEE, 2019, IEEE}}
	\item \textlatin{\textbf{Bizopoulos, Paschalis} A and Tsalikakis, Dimitrios G and Tzallas, Alexandros T and Koutsouris, Dimitrios D and Fotiadis, Dimitrios I, EEG epileptic seizure detection using k-means clustering and marginal spectrum based on ensemble empirical mode decomposition, \textit{Bioinformatics and Bioengineering (BIBE), 2013 IEEE 13th International Conference on, 1--4, 2013, IEEE}}
	\item \textlatin{\textbf{Bizopoulos, Paschalis} A and Al-Ani, Tarik and Tsalikakis, Dimitrios G and Tzallas, Alexandros T and Koutsouris, Dimitrios D and Fotiadis, Dimitrios I, An automatic electroencephalography blinking artefact detection and removal method based on template matching and ensemble empirical mode decomposition, \textit{Engineering in Medicine and Biology Society (EMBC), 2013 35th Annual International Conference of the IEEE, 5853--5856, 2013, IEEE}}
	\item \textlatin{\textbf{Bizopoulos, Paschalis} A and Sakellarios, Antonis I and Koutsouris, Dimitrios D and Iliopoulou, Dimitra and Michalis, Lampros K and Fotiadis, Dimitrios I, Randomly generated realistic vessel geometry using spline interpolation and 2D Perlin noise, \textit{Biomedical and Health Informatics (BHI), 2014 IEEE-EMBS International Conference on, 157--160, 2014, IEEE}}
	\item \textlatin{\textbf{Bizopoulos, Paschalis} A and Sakellarios, Antonis I and Koutsouris, Dimitrios D and Kountouras, Jannis and Kostretzis, Lazaros and Karagergou, Stella and Michalis, Lampros K and Fotiadis, Dimitrios I, Prediction of atheromatic plaque evolution in carotids using features extracted from the arterial geometry, \textit{Engineering in Medicine and Biology Society (EMBC), 2015 37th Annual International Conference of the IEEE, 6556--6559, 2015, IEEE}}
	\item \textlatin{\textbf{Bizopoulos, Paschalis} A and Vavuranakis, Manolis and Papaioannou, Theodoros G and Vrachatis, Dimitrios A and Sakellarios, Antonis I and Iliopoulou, Dimitra and Tousoulis, Dimitris and Koutsouris, Dimitrios D and Fotiadis, Dimitrios I, A Preliminary Study on In-Vivo 3-D Imaging of Bioprosthetic Aortic Valve Deformation, \textit{XIV Mediterranean Conference on Medical and Biological Engineering and Computing 2016, 332--336, 2016, Springer Cham}}
	\item \textlatin{\textbf{Bizopoulos, Paschalis} A and Sakellarios, Antonis and Michalis, Lampros K and Koutsouris, Dimitrios D and Fotiadis, Dimitrios I, 3-D Registration on Carotid Artery imaging data: MRI for different timesteps, \textit{Engineering in Medicine and Biology Society (EMBC), 2016 IEEE 38th Annual International Conference of the, 1159--1162, 2016, IEEE}}
	\item \textlatin{Sakellarios, Antonis I and \textbf{Bizopoulos, Paschalis} and Stefanou, Kostas and Athanasiou, Lambros S and Papafaklis, Michail I and Bourantas, Christos V and Naka, Katerina K and Michalis, Lampros K and Fotiadis, Dimitrios I, A proof-of-concept study for predicting the region of atherosclerotic plaque development based on plaque growth modeling in carotid arteries, \textit{Engineering in Medicine and Biology Society (EMBC), 2015 37th Annual International Conference of the IEEE, 6552--6555, 2015, IEEE}}
	\item \textlatin{Salis, Christos I and Malissovas, Anastasios E and \textbf{Bizopoulos, Paschalis} A and Tzallas, Alexandros T and Angelidis, Pantelis A and Tsalikakis, Dimitrios G, Denoising simulated EEG signals: A comparative study of EMD, wavelet transform and kalman filter, \textit{Bioinformatics and Bioengineering (BIBE), 2013 IEEE 13th International Conference on, 1--4, 2013, IEEE}}
\end{enumerate}